\documentclass[10pt,twocolumn,letterpaper]{article}

\usepackage[pagenumbers]{cvpr} 

\usepackage{graphicx}
\usepackage{amsmath}
\usepackage{amssymb}
\usepackage{booktabs}
\usepackage{caption}
\usepackage{subcaption}
\usepackage{multirow}
\usepackage[pagebackref,breaklinks,colorlinks=true,citecolor=blue]{hyperref}
\usepackage{enumitem}
\usepackage{lipsum}
\usepackage{nicefrac}
\usepackage{cuted}
\AfterEndEnvironment{strip}{\leavevmode}
\usepackage{listings}
\usepackage{xcolor,colortbl}
\usepackage{algpseudocode}
\usepackage{lscape}

\usepackage[capitalize]{cleveref}
\crefname{section}{Sec.}{Secs.}
\Crefname{section}{Section}{Sections}
\Crefname{table}{Table}{Tables}
\crefname{table}{Tab.}{Tabs.}

\DeclareMathOperator*{\argmin}{arg\,min}

\def\cvprPaperID{*****} 
\def\confName{CVPR}
\def\confYear{2023}

\begin{document}

\definecolor{codegreen}{rgb}{0,0.6,0}
\definecolor{codegray}{rgb}{0.5,0.5,0.5}
\definecolor{codepurple}{rgb}{0.58,0,0.82}
\definecolor{backcolour}{rgb}{0.95,0.95,0.92}

\lstdefinestyle{mystyle}{
    backgroundcolor=\color{backcolour},   
    commentstyle=\color{codegreen},
    keywordstyle=\color{magenta},
    numberstyle=\tiny\color{codegray},
    stringstyle=\color{codepurple},
    basicstyle=\ttfamily\footnotesize,
    breakatwhitespace=false,         
    breaklines=true,                 
    captionpos=b,                    
    keepspaces=true,                 
    numbers=left,                    
    numbersep=5pt,                  
    showspaces=false,                
    showstringspaces=false,
    showtabs=false,                  
    tabsize=2
}

\lstset{style=mystyle,linewidth=\linewidth,xleftmargin=0.15\textwidth,xrightmargin=0.15\textwidth}

\title{Visual Prompt Tuning for Generative Transfer Learning} 

\author{
Kihyuk Sohn, Yuan Hao, Jos\'e Lezama, Luisa Polania, \\
Huiwen Chang, Han Zhang, Irfan Essa, Lu Jiang
\\
Google Research
}
\maketitle

\begin{abstract}
Transferring knowledge from an image synthesis model trained on a large dataset is a promising direction for learning generative image models from various domains efficiently. While previous works have studied GAN models, we present a recipe for learning vision transformers by generative knowledge transfer. We base our framework on state-of-the-art generative vision transformers that represent an image as a sequence of visual tokens to the autoregressive or non-autoregressive transformers. To adapt to a new domain, we employ prompt tuning, which prepends learnable tokens called prompt to the image token sequence, and introduce a new prompt design for our task. We study on a variety of visual domains, including visual task adaptation benchmark~\cite{zhai2019large}, with varying amount of training images, and show effectiveness of knowledge transfer and a significantly better image generation quality over existing works. 
\end{abstract}

\begin{figure*}
    \centering
    \includegraphics[width=0.99\textwidth]{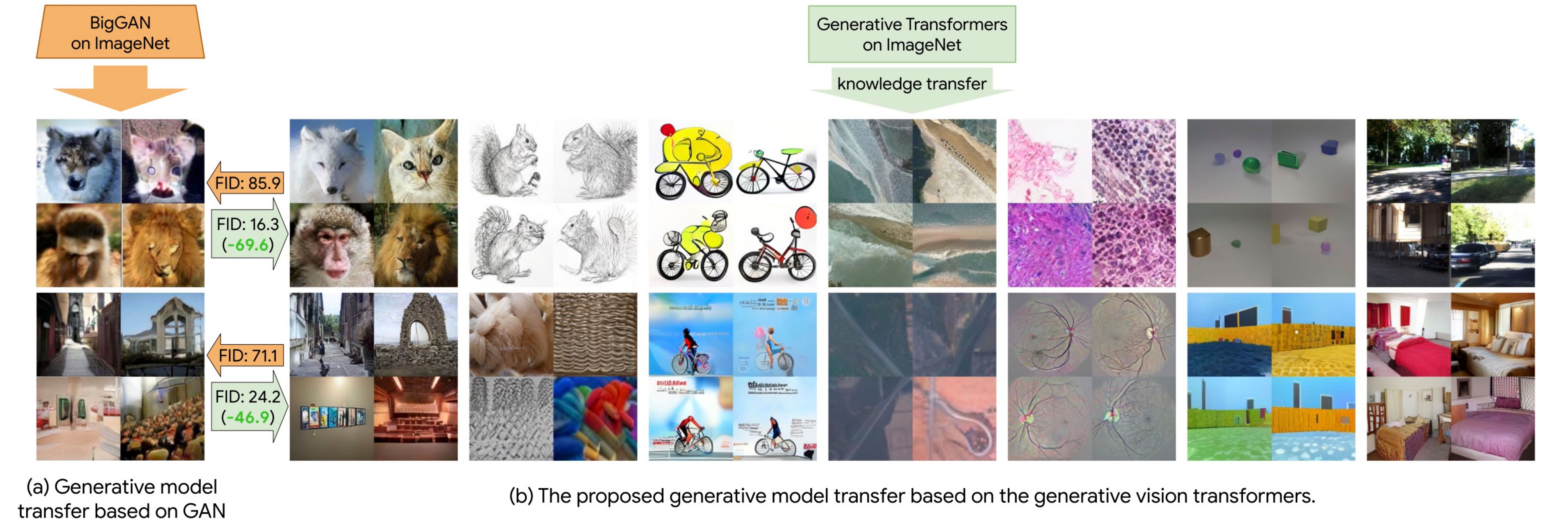}
    \vspace{-0.05in}
    \caption{Image synthesis by knowledge transfer. Unlike previous works using GANs as source model and test transfer on relatively narrow visual domains, we transfer knowledge of generative vision transformers~\cite{esser2021taming,chang2022maskgit} to a comprehensive list of visual domains, including natural (\eg, scene, flower), specialized (\eg, satellite, medical), and structured (\eg, road scenes, synthetic, infograph, sketch), as defined by the visual task adaptation benchmark~\cite{zhai2019large}, with a few training images (\eg, as low as 2 images per class).}
    \label{fig:teaser}
    \vspace{-0.1in}
\end{figure*}

\vspace{-0.3in}
\section{Introduction} 
\label{sec:intro}

Image synthesis has achieved tremendous progress recently with the improvement of deep generative models~\cite{goodfellow2014generative,van2016pixel,Oord17vqvae,brock2018large,dhariwal2021diffusion}. The goal of image synthesis is to generate diverse and plausible scenes resembling the training images. A good image synthesis system can capture the appearance of objects and model their interactions to generalize and create novel scenes. However, the generalization ability is usually determined by the amount of training images. Without sufficient data, the synthesis results are often unsatisfactory. 

Transfer learning, a cornerstone invention in deep learning, has been proving its indispensable role across a broad array of computer vision tasks, including classification~\cite{kolesnikov2020big}, object detection~\cite{girshick2014rich,girshick2015fast}, image segmentation~\cite{he2016deep,he2020momentum}, \etc.
However, transfer learning is not yet a \emph{de facto} technique for image synthesis. While recent efforts have shown success in transferring knowledge from pretrained Generative Adversarial Network (GAN) models~\cite{wang2020minegan,shahbazi2021efficient,ojha2021few,yang2021one}, these demonstrations are limited by narrow visual domains, \eg, faces or cars~\cite{ojha2021few,yang2021one}, as illustrated in Fig.~\ref{fig:teaser}, or requiring a non-trivial amount of training data~\cite{wang2020minegan,shahbazi2021efficient} to transfer to an off-manifold distribution.

In fact, some recent works~\cite{tseng2021regularizing,zhao2020differentiable} have found that, even when the training data is limited in quantity, learning GANs from scratch with advanced techniques outperforms GAN transfer approaches, implying that the transfer learning may not even be necessary for generative modeling.
Such observation is in direct contrast to the essential role of transfer learning for discriminative models,\footnote{Here discriminative models refer to a board of machine learning models that directly model the conditional distribution of the target variables.} which suggests transfer learning for image synthesis remains under-exploited.

In this work, we approach the transfer learning for image synthesis using generative vision transformers, an emerging class of image synthesis models, such as DALL·E~\cite{Ramesh21dalle}, Taming Transformer~\cite{esser2021taming}, MaskGIT~\cite{chang2022maskgit}, CogView~\cite{ding2021cogview}, NÜWA~\cite{wu2021n}, or Parti~\cite{yu2022scaling}, that excel in several image synthesis tasks. We closely follow the recipe of transfer learning for image classification~\cite{kolesnikov2020big}, in which a source model is first trained on a large dataset (\eg, ImageNet) and then transferred to a diverse collection of downstream tasks, except in our setting the input and output are reversed and the model generates images from a class label. Our study employs the visual task adaptation benchmark (or VTAB)~\cite{zhai2019large}, a standard and challenging benchmark for studying transfer learning. VTAB consists of 19 visual recognition tasks and compiles images from diverse and distinctly different visual domains, such as natural (\eg, flowers, scenes), specialized (\eg, satellite, medical), or structured (\eg, road scenes).

We present a transfer learning framework using \emph{prompt tuning}~\cite{lester2021power,li2021prefix}. While the technique has been used for transfer learning of discriminative models for vision tasks~\cite{jia2022visual,bahng2022visual}, to our knowledge, this work appears to be the first to adopt  prompt tuning for transfer learning of image synthesis.
Moreover, we propose two technical innovations. First, a parameter efficient design of prompt token generator that admits condition variables (\eg, class, instance), a key for controllable image synthesis yet often neglected in prompt tuning for discriminative transfer~\cite{lester2021power,jia2022visual}. Second, a marquee header prompt that engineers (\eg, composes and interpolates) learned prompts to enhance generation diversity.

We conduct a large-scale empirical study to understand the mechanics of generative transfer learning for autoregressive~\cite{Ramesh21dalle,esser2021taming} and non-autoregressive~\cite{chang2022maskgit} generative transformers. To this end, we show that generative vision transformers with prompt tuning outperforms the prior state-of-the-art held by GANs~\cite{wang2020minegan,shahbazi2021efficient} through a vast margin. Moreover, in contrast to prior works~\cite{wang2020minegan,shahbazi2021efficient} limited to show transfer to a few visual domains, we show the efficacy of knowledge transfer from pretrained ImageNet models to 19 downstream tasks of diverse visual distributions and varying amounts of training data in VTAB. \cref{fig:teaser} compares visual domains, showing the great expansion on the varieties of downstream tasks to what is achieved by previous works. On the on-manifold distributions on which previous studies mainly focused, our method slashes the prior state-of-the-art in FID from 71 to 24 on Places~\cite{zhou2014learning} and 86 to 16 on Animal Face~\cite{si2011learning} datasets. Moreover, the proposed method is used to demonstrate the few-shot generative transfer capabilities (\cref{sec:exp_fewshot}), showing extreme data efficiency while being able to generate images that are realistic and diverse, while following the target distribution.

In summary, our contributions are as follows:
\begin{itemize}[noitemsep]
    \item We present a generative visual transfer learning framework for vision transformers with prompt tuning~\cite{lester2021power}, proposing a novel prompt token generator design and a prompt engineering method for image synthesis.
    \item We conduct a large-scale empirical study for generative transfer learning to validate our method on a variety of visual domains (\eg, VTAB~\cite{zhai2019large}) and scenarios (\eg, few-shot). To this end, we show state-of-the-art image synthesis performance.
    \item To our knowledge, we are the first to employ prompt tuning for transfer learning of image synthesis, and provide one-of-the-first substantial empirical evidence on the necessity of knowledge transfer for data and compute efficient generative image modeling using the standard visual transfer learning benchmark.
\end{itemize}

\section{Preliminary}
\label{sec:prelim}

\subsection{Generative Vision Transformers}
\label{sec:prelim_gvt}
This paper uses generative vision transformers to denote the vision transformer models for image synthesis. Generally, there are two types of generative vision transformers: \emph{AutoRegressive (AR)} and \emph{Non-AutoRegressive (NAR)} transformers, both consisting of two stages~\cite{Ramesh21dalle,esser2021taming}: image quantization and decoding. The first stage is the same between the two types of models in which the image is quantized into a grid of discrete tokens by a Vector-Quantized (VQ) auto-encoder~\cite{Oord17vqvae,razavi2019generating,esser2021taming,vim2021}. The VQ encoder quantizes image patches into integer indices (or tokens) in a codebook. The 2D image is then flattened into a 1D sequence to which a special token indicating its class label is prepended.

\begin{figure}[t]
    \centering
    \includegraphics[width=0.95\linewidth]{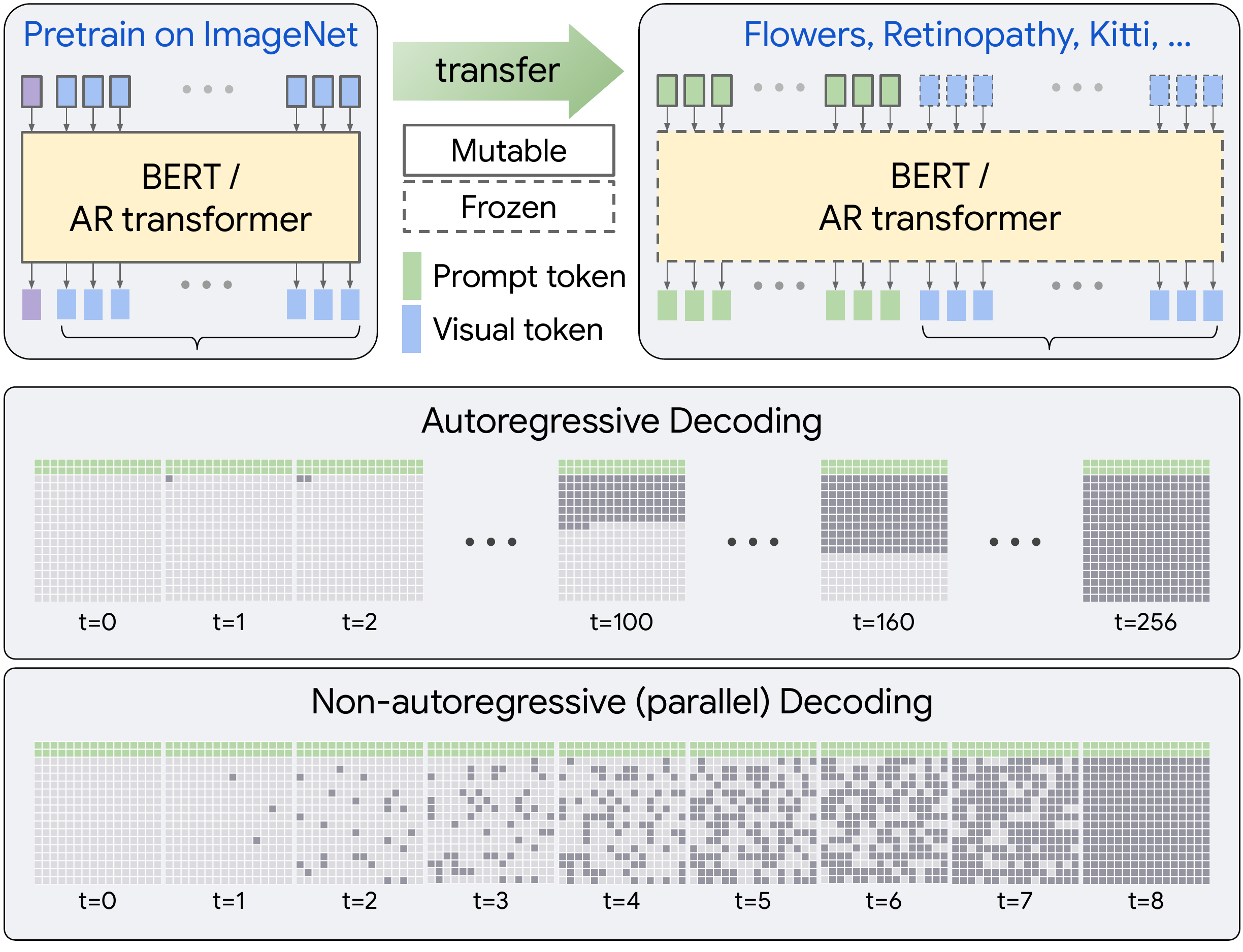}
    \caption{Our method transfers knowledge from generative vision transformers (\eg, autoregressive~\cite{esser2021taming} or non-autoregressive~\cite{chang2022maskgit}) trained on a large dataset to various visual domains by prepending learnable prompt tokens (green) to visual tokens (blue). 
    }
    \label{fig:method}
\end{figure}

AR and NAR transformers differ in the second stage of decoding. AR transformers~\cite{chen2020imagegpt}, including DALL·E~\cite{Ramesh21dalle}, Taming Transformer~\cite{esser2021taming}, NÜWA~\cite{wu2021n}, CogView~\cite{ding2021cogview}, and Parti~\cite{yu2022scaling}, are inspired by the AR language model~\cite{mikolov2010recurrent,gpt3}. They learn an AR decoder on the flattened token sequence to generate image tokens sequentially based on the previously generated tokens. As illustrated in \cref{fig:method}, the generation follows a raster scan ordering, generating tokens from left to right, line-by-line. Finally, the generated tokens are mapped to the pixel space using the VQ decoder.

On the other hand, NAR transformers~\cite{ghazvininejad2019mask,gu-kong-2021-fully,kong2020incorporating}, which are originally proposed for machine translation, are recently extended to improve the AR image decoding~\cite{chang2022maskgit,zhang2021m6,lezama2022improved}. Unlike their AR counterpart, NAR transformers (\eg, MaskGIT~\cite{chang2022maskgit}, Token-Critic~\cite{lezama2022improved}, BLT~\cite{kong2021blt}) are bidirectional and are trained on the masked modeling proxy task of BERT~\cite{devlin2018bert}. During inference, the model adopts a non-autoregressive decoding method to synthesize an image in a few steps~\cite{gu-kong-2021-fully,chang2022maskgit,kong2021blt,lezama2022improved}. As shown in \cref{fig:method}, the NAR transformer starts from a blank canvas with all tokens masked out, and generate an image in 8 steps or so. In each step, it predicts all tokens in parallel while retaining ones with the highest prediction scores. The remaining tokens are masked out and will be predicted in the next iteration until all tokens are generated. NAR transformers~\cite{chang2022maskgit,lezama2022improved} show faster inference than their AR counterparts (\eg, \cite{esser2021taming}) while offering on-par or superior fidelity and diversity.

\subsection{Prompt Tuning}
\label{sec:method_prompt_tuning}

Prompt tuning~\cite{lester2021power,li2021prefix} has been introduced recently in natural language processing as a way of efficiently adapting pretrained large language models to downstream tasks. 
Here, prompt is a sequence of additional tokens prepended to a token sequence. In prompt engineering~\cite{brown2020language}, their values are often chosen by heuristic. On the other hand, in prompt tuning~\cite{lester2021power,li2021prefix}, tokens are parameterized by learnable parameters and their parameters are updated via a gradient descent to adopt transformers to the downstream tasks.

Due to its simplicity and as transformers getting popular, prompt tuning has been also applied to some vision tasks for knowledge transfer, \eg, image classification~\cite{bahng2022visual,jia2022visual}, detection and segmentation~\cite{nie2022protuning}. To our knowledge, we appear to be the first to use prompt tuning for image synthesis. 

\begin{figure*}
    \centering
    \begin{subfigure}[b]{0.14\linewidth}
        \centering
        \includegraphics[height=1.3in]{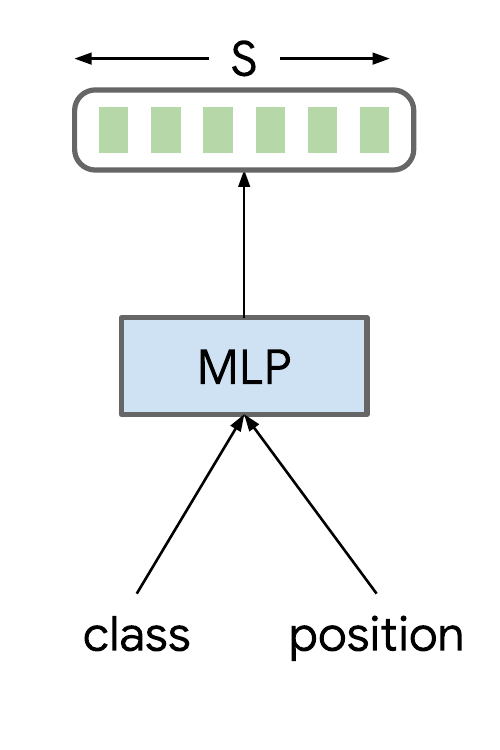}
        \caption{Baseline prompt token generators of length $S$ conditioned on class.}
        \label{fig:prompt_design_baseline}
    \end{subfigure}
    \hspace{0.05in}
    \begin{subfigure}[b]{0.53\linewidth}
        \centering
        \includegraphics[height=1.3in]{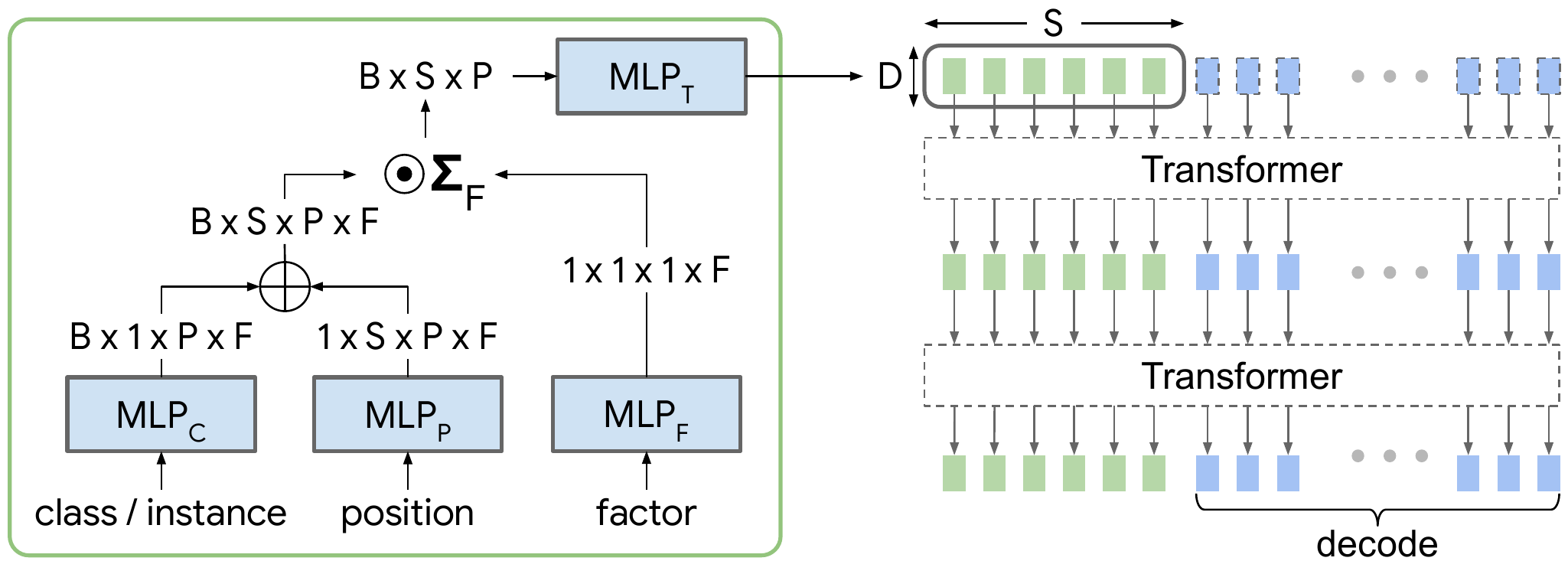}
        \caption{The proposed parameter efficient prompt token generator via factorization of class / instance and position. $\oplus$ is an element-wise sum, $\odot$ is an element-wise product, $\Sigma_{F}$ is a sum over $F$ dimension. $S$: sequence length, $B$: batch size, $P$: feature dimension, $D$: token dimension. }
        \label{fig:prompt_design_proposed}
    \end{subfigure}
    \hspace{0.05in}
    \begin{subfigure}[b]{0.25\linewidth}
        \centering
        \includegraphics[height=1.3in]{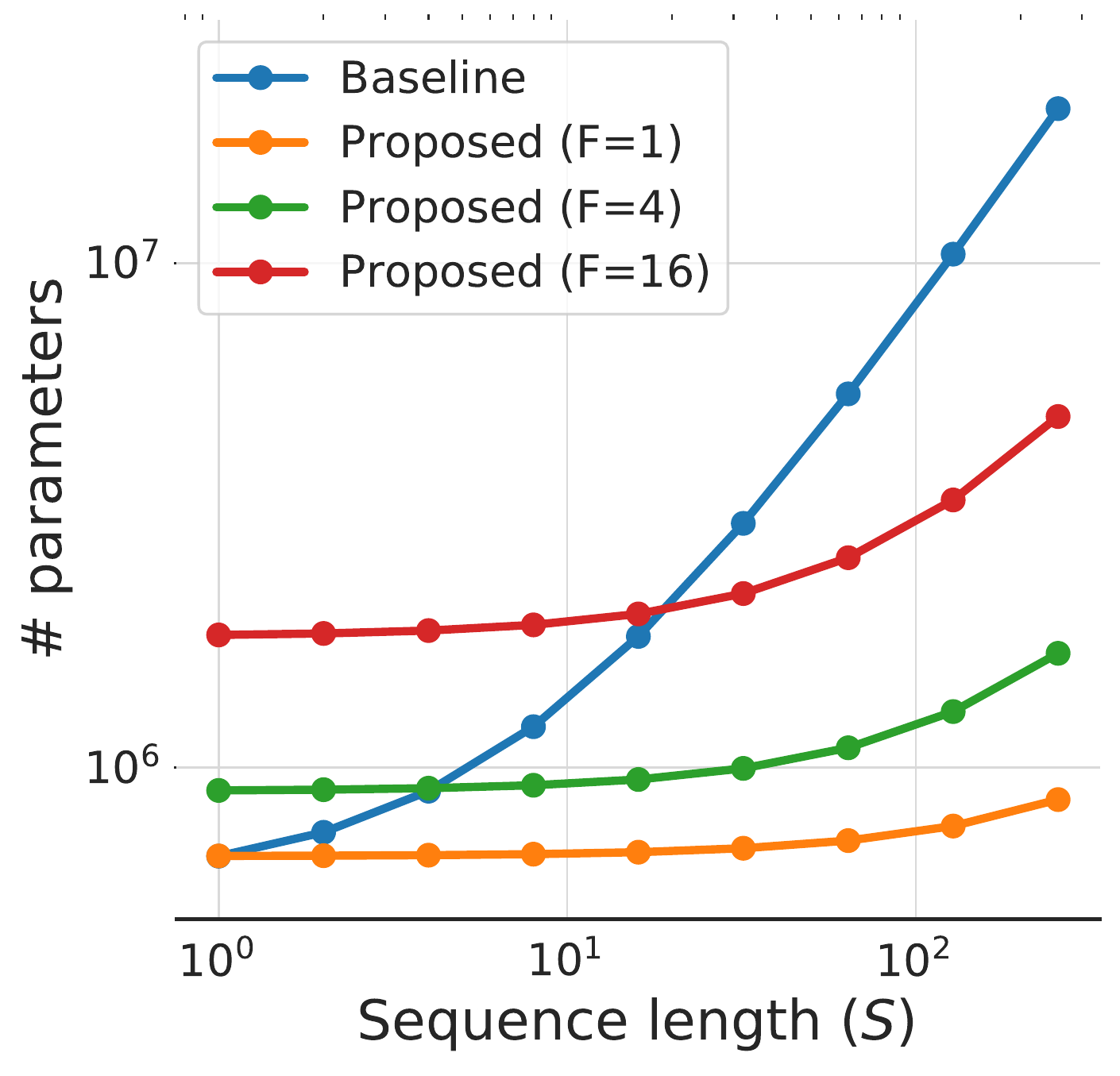}
        \caption{Number of parameters with respect to the sequence length and different number of factors $F$.}
        \label{fig:prompt_num_params}
    \end{subfigure}
    \caption{Prompt token generators and their use in transformer. (a) a straightforward extension of baseline prompt token generators~\cite{li2021prefix,lester2021power,jia2022visual} with a class condition. When using an MLP with a single dense layer of $P$ units, the number of trainable parameters is $P{\cdot}(C{\cdot}S{+}D)$. (b) The proposed parameter efficient prompt token generators that factorizes data dependent conditions (e.g., class, instance) and token position. Under a similar design choice as baseline models, the number of trainable parameters is $P{\cdot}(F{\cdot}(C{+}S){+}D)$, which could be significantly fewer when $F{\ll}\min(C, S)$. (c) Number of parameters for prompt token generators with respect to the sequence length (S), while setting $P\,{=}\,768$, $D\,{=}\,768$, and $C\,{=}\,100$ with different number of factors $F$.
    }
    \label{fig:prompt_design}
\end{figure*}

\section{Visual Prompt for Generative Transfer}
\label{sec:method}

Our goal is to design a transfer learning framework for image synthesis using vision transformers. Starting from a generative vision transformer pretrained on a large dataset (\eg, ImageNet), we discuss a method to adapt transformers on various target domains (\eg, VTAB). \cref{sec:method_prompt_design_and_learning} presents a visual prompt tuning for AR and NAR transformers. Then, in \cref{sec:method_prompt_composition}, we propose a novel prompt, named marquee header prompt, tailored to NAR transformers to trade-off generation fidelity and diversity.

\subsection{Building and Learning Visual Prompt}
\label{sec:method_prompt_design_and_learning}

\cref{fig:method} overviews the proposed generative transfer learning framework. We aim at transferring a generative prior, parameterized by generative vision transformers, while utilizing the same VQ encoder and decoder trained from the large source dataset. We employ a prompt tuning~\cite{lester2021power,li2021prefix,jia2022visual} that uses a sequence of learnable tokens (\eg, green blocks with a solid line in \cref{fig:method}) to adapt to target distributions, while leaving transformer parameters frozen.
In the following sections, we discuss how to learn (\cref{sec:method_prompt_learning}) a prompt token generator designed for a conditional image generation (\cref{sec:method_prompt_token_generator}) and use them for image synthesis (\cref{sec:method_prompt_composition}). 

\subsubsection{Learning Visual Prompt}
\label{sec:method_prompt_learning}

A sequence of prompt tokens is prepended to visual tokens to guide the pretrained transformer models to the target distribution. Prompt tuning, learning parameters of token generator, is done by gradient descent with respective loss functions, while fixing parameters of pretrained transformers.
%
{To be specific, let $\mathcal{Z}\,{=}\,\{z_{i}\}_{i=1}^{H{\times}W}$ be a sequence of visual tokens (\ie, an output of VQ encoder followed by the vectorization) and $\mathcal{P}_{\phi}\,{=}\,\{p_{s;\phi}\}_{s=1}^{S}$ be a sequence of prompt tokens.}
For AR transformer, the loss is given as follows:
\begin{eqnarray}
    \mathcal{L}_{\mathrm{AR}} = \mathbb{E}_{x{\sim}P_{\mathcal{X}}}\big[-\log P_{\theta}(\mathcal{Z}|\mathcal{P}_{\phi})\big]\\
    P_{\theta}(\mathcal{Z}|\mathcal{P}_{\phi}) = \prod\nolimits_{i=1}^{H{\times}W} P_{\theta}(z_{i}|z_{<i}, \mathcal{P}_{\phi} )
\end{eqnarray}
For NAR transformer, we follow the loss of MaskGIT~\cite{chang2022maskgit}:
\begin{eqnarray}
    \mathcal{L}_{\mathrm{NAR}} = \mathbb{E}_{x{\sim}P_{\mathcal{X}},M{\sim}P_{\mathcal{M}}}\big[-\log P_{\theta}(\mathcal{Z}_{M}|\mathcal{Z}_{\overline{M}}, \mathcal{P}_{\phi})\big]\\
    P_{\theta}(\mathcal{Z}_{M}|\mathcal{Z}_{\overline{M}}, \mathcal{P}_{\phi}) = \prod\nolimits_{i\,{\in}\,M} P_{\theta}(z_{i}|\mathcal{Z}_{\overline{M}}, \mathcal{P}_{\phi} )
\end{eqnarray}
where $M\,{\subset}\,\{1,...,H{\times}W\}$ is a set of visual token indices sampled from a masking schedule distribution $P_{\mathcal{M}}$, $\overline{M}$ is its complement, and $\mathcal{Z}_{M}\,{=}\,\{z_{i}\}_{i\,{\in}\,M}$. Prompt tuning proceeds by minimizing respective loss functions with respect to the prompt parameters $\phi$ while fixing transformer parameters $\theta$:
\begin{equation}
    \phi^{\ast} = \argmin_{\phi}\mathcal{L}_{\mathrm{AR/NAR}}
\end{equation}

While our focus is at the prompt tuning due to its effectiveness and compute-efficiency for large source models, we note that the proposed learning framework is amenable with other transfer learning methods, such as adapter~\cite{houlsby2019parameter} or fine-tuning~\cite{kolesnikov2020big}, with learnable prompts, as shown in \cref{sec:abl_beyond_prompt_tunig}.

After prompt tuning, we generate visual tokens for image synthesis by iterative decoding. For AR transformer, 
\begin{algorithmic}[1]
{\small
\For {$i \gets 1$ to $H\,{\times}\,W$}
\State $\hat{z}_{i}\,{\sim}\,P_{\theta}(z_{i}|\hat{z}_{<i}, \mathcal{P}_{\phi})$
\EndFor
}
\end{algorithmic}
For NAR model, scheduled parallel decoding~\cite{chang2022maskgit} is used:
\begin{algorithmic}[1]
{\small
\Require{$\overline{M}\,{=}\,\{\}$, $T$, $\{n_{1},...,n_{T}\}, \sum_{t=1}^{T}n_{t}\,{=}\,H\,{\times}\,W$} 
\For {$t \gets 1$ to $T$}
\State $\hat{z}_{i}\,{\sim}\,P_{\theta}(z_{i}|\widehat{\mathcal{Z}}_{\overline{M}}, \mathcal{P}_{\phi})$, $\forall i\,{\in}\,{M}$
\State $\overline{M} \gets \overline{M}\cup \{\arg\mathrm{topk}_{i\,{\in}\,M}\big(P_{\theta}(z_{i}|\widehat{\mathcal{Z}}_{\overline{M}}, \mathcal{P}_{\phi}), k\,{=}\,n_{t}\big)\}$
\EndFor
}
\end{algorithmic}
where $\{n_{1},...,n_{T}\}$ is a masking schedule that decides the number of tokens to decode at each decoding step. We refer to \cite{chang2022maskgit} for details on decoding for NAR transformer. Illustrations of decoding steps for both models are in \cref{fig:method}.

\subsubsection{Prompt Token Generator Design}
\label{sec:method_prompt_token_generator}

For discriminative transfer learning, prompts are designed without condition variables~\cite{jia2022visual}. For generation, it is beneficial to have condition variables (\eg, class, attribute) for better control in generation. We accomplish this with rather a straightforward extension of existing prompt designs using a class-condition, $\mathcal{P}_{\phi}(c)$, as in \cref{fig:prompt_design_baseline}.

One caveat of the baseline token generator design is that the number of learnable parameters increases as the product of three factors: the number of classes $C$, the prompt sequence length $S$ and the feature dimension $P$. For example, when using a prompt of length $S{=}128$, hidden $P{=}768$ and embedding dimension $D{=}768$, the token generator would introduce $10.4$M parameters for $C{=}100$ class conditions, as in \cref{fig:prompt_num_params}. 
The bottleneck occurs at the 3d weight tensor of size $C{\times}S{\times}P$. To make it parameter efficient, we propose a factorized token generator, as in \cref{fig:prompt_design_proposed}. Specifically, we encode class and sequence position index via $\textsc{MLP}_{\textsc{C}}$ and $\textsc{MLP}_{\textsc{P}}$ with $F$ factors, respectively. The MLP outputs are element-wise summed, multiplied by an 1d factor vector from $\textsc{MLP}_{\textsc{F}}$, and reduced along the factor dimension. The output is then fed to $\textsc{MLP}_{\textsc{T}}$ to produce a prompt of length $S$. As in \cref{fig:prompt_num_params}, the number of parameters of the proposed architecture is greatly reduced, requiring only $0.76$M parameters, down from $10.4$M, for a prompt of length $128$ when $F\,{=}\,1$.\footnote{The proposed factorization can be extended to incorporate the ``depth'' position of deep visual prompt~\cite{jia2022visual} to reduce the number of parameters.} An implementation of the proposed token generator in Flax~\cite{flax2020github} is in \cref{fig:prompt_generator} of Appendix.
We empirically find that $F\,{=}\,1$ is sufficient for NAR transformers, demonstrating extreme parameter efficiency. For AR transformers, we need extra capacity and use $F\,{=}\,16$.

Moreover, we build a new type of prompt tokens conditioned on individual data instances, inspired by the instance-conditioned GAN~\cite{casanova2021instance}.
We assign each data a unique index and map it into a distinct embedding via $\textsc{MLP}_{\textsc{C}}$. When both class label and instance index are used, instance index is simply treated as an extra class, indexed from $C$. To train the model, we sample between class label and instance index.
As we explain below in \cref{sec:method_prompt_composition}, instance conditioned prompts add more fine-grained control on generation.

\begin{figure*}
    \centering
    \begin{subfigure}[b]{\textwidth}
        \centering
        \includegraphics[width=0.95\textwidth]{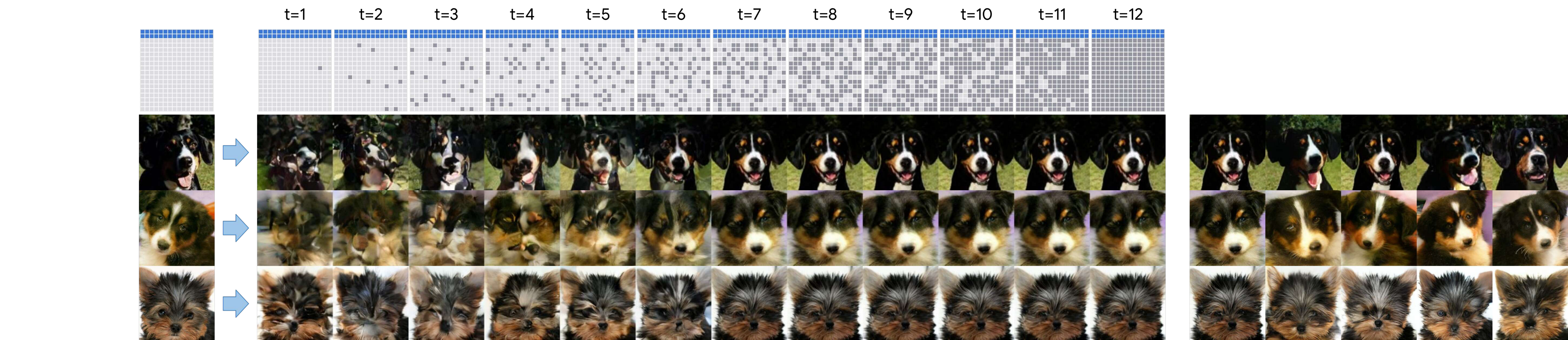}
        \caption{Image synthesis using instance-conditioned prompts.}
        \label{fig:autoreg_gen_prompt}
    \end{subfigure}
    \begin{subfigure}[b]{\textwidth}
        \centering
        \includegraphics[width=0.95\textwidth]{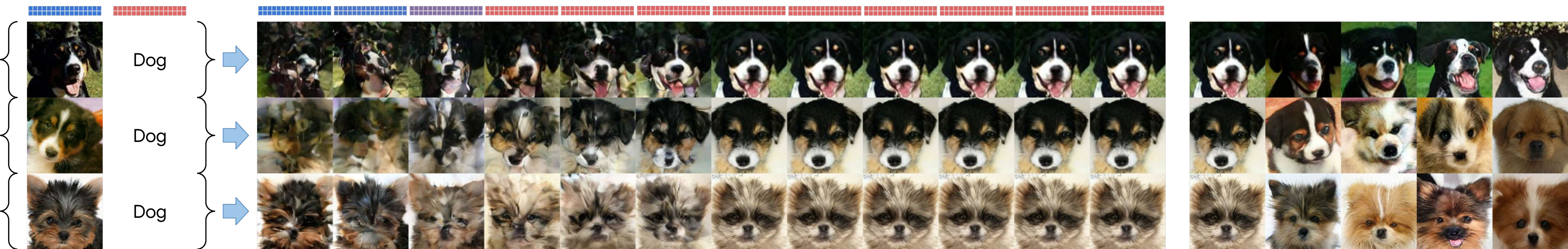}
        \caption{Image synthesis using a marquee header prompt between instance (blue) and class (red) conditioned prompts.}
        \label{fig:autoreg_gen_interp_instance_to_class}
    \end{subfigure}
    \begin{subfigure}[b]{\textwidth}
        \centering
        \includegraphics[width=0.95\textwidth]{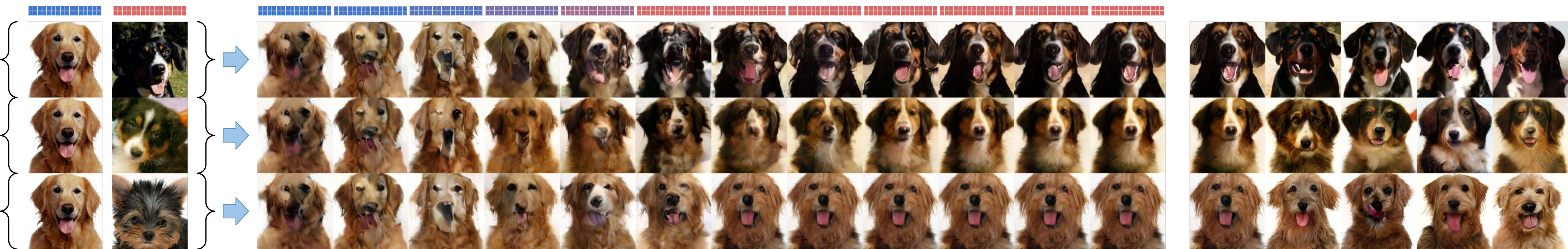}
        \caption{Image synthesis using a marquee header prompt between instance-conditioned prompts (blue and red).}
        \label{fig:autoreg_gen_interp_instance_to_instance}
    \end{subfigure}
    \caption{Iterative decoding of NAR transformers. (\ref{fig:autoreg_gen_prompt}) instance prompts generate images of high-fidelity but with low diversity. Marquee header prompts enhance generation diversity by interpolating (\ref{fig:autoreg_gen_interp_instance_to_class}) from instance to class prompts or (\ref{fig:autoreg_gen_interp_instance_to_instance}) between instance prompts.}
    \label{fig:autoreg_gen}
\end{figure*}

\subsection{Engineering Learned Prompts}
\label{sec:method_prompt_composition}

An interesting aspect of generative transformers in contrast to GANs is their iterative decoding. For example, as illustrated in \cref{fig:method}, AR transformers~\cite{esser2021taming} decode tokens sequentially given previously decoded tokens, and NAR transformers~\cite{chang2022maskgit} use a scheduled parallel decoding.

Given the wealth of learned prompts conditioned on the class and instance proposed in \cref{sec:method_prompt_design_and_learning}, we propose a novel prompt engineering strategy, a ``Marquee Header'' prompt, of the iterative transformer decoding, for enhancing the generation diversity. 
The idea is simple -- similarly to the latent variable interpolation of GANs, we interpolate the learned prompt representations (\eg, outputs of $\textsc{MLP}_{\textsc{C}}$). Yet, due to the iterative decoding, the interpolation between prompts is carried out over multiple decoding steps. This is illustrated in \cref{fig:autoreg_gen_interp_instance_to_class}, where we start the decoding process using instance-conditioned prompts (blue header) but gradually transition to a class-conditioned prompt (red header) over decoding steps. Compared to the generation in \cref{fig:autoreg_gen_prompt} where we use instance-conditioned prompts all along, the proposed prompt engineering strategy enhances the generation diversity while being controlled in that synthesized images follow certain characteristics (\eg, pose, color pattern, hairiness) of reference instances. 
In addition, it is also plausible to construct a marquee header prompt between instance-conditioned prompts, as in \cref{fig:autoreg_gen_interp_instance_to_instance}.

We provide a marquee header prompt formulation:
\begin{eqnarray}
\textsc{Pmt}(t) &=& (1-w_{t})\textsc{Pmt}_{1} + w_t \textsc{Pmt}_{2}\\
w_{t} &=& \min\Big\{\big(\frac{t-1}{T_{\mathrm{cutoff}}-1}\big)^2, 1\Big\} \label{eq:dynamic_prompt_schedule}
\end{eqnarray}
where $t\,{=}\,1, ..., T$ is a decoding step, $T_{\mathrm{cutoff}}\,{\leq}\,T$ is a cutoff step, and $\textsc{Pmt}_{i}$ is a prompt representation (\eg, an output of $\textsc{MLP}_{\textsc{C}}$).
The schedule in Eq.~\eqref{eq:dynamic_prompt_schedule} makes a smooth transition of prompts from $\textsc{Pmt}_{1}$ to $\textsc{Pmt}_{2}$.
Note that there could be various marquee header prompt formulations, which we leave their investigations as a future work.

\section{Experiments}
\label{sec:exp}
We evaluate the efficacy of the generative transfer learning on diverse visual domains and varying amounts of training data and compare with existing methods. In \cref{sec:exp_vtab}, we test on visual task adaptation benchmark (VTAB)~\cite{zhai2019large} and demonstrate state-of-the-art image generation performance with knowledge transfer. In \cref{sec:exp_fewshot}, we verify our method on diverse few-shot transfer learning tasks.

\begin{table*}[ht]
    \centering
    \resizebox{0.95\linewidth}{!}{%
    \begin{tabular}{llc||>{\columncolor[gray]{0.9}}c||c|c|c|c|c|c|c|c}
    \toprule
		\multicolumn{2}{l}{Model}			&	(\# tr params)	&	Mean	&	C101	&	Flowers	&	Pet 	&	DTD	&	Kitti	&	SUN	&	EuroSAT	&	Resisc	\\
\midrule																							
\multicolumn{2}{l}{MineGAN~\cite{wang2020minegan}}			&	(88M)	&	151.5	&	102.4	&	132.1	&	130.1	&	87.4	&	117.9	&	77.5	&	111.5	&	81.0	\\
\multicolumn{2}{l}{cGANTransfer~\cite{shahbazi2021efficient}}			&	(105M)	&	85.1	&	89.6	&	61.6	&	48.6	&	70.3	&	48.9	&	31.1	&	45.6	&	50.3	\\
\midrule
\multirow{4}{*}{Non-Autoregressive}	&	Prompt ($S\,{=}\,1$)	&	(0.67M)	&	53.7	&	13.5	&	13.8	&	11.9	&	\textbf{25.8}	&	32.3	&	\textbf{7.3}	&	45.9	&	28.5	\\
	&	Prompt ($S\,{=}\,16$)	&	(0.68M)	&	\underline{39.9}	&	\underline{12.7}	&	\textbf{13.2}	&	\underline{11.1}	&	26.0	&	\underline{30.0}	&	\underline{7.4}	&	\textbf{35.8}	&	\underline{24.9}	\\
	&	Prompt ($S\,{=}\,128$)	&	(0.76M)	&	\textbf{36.4}	&	\textbf{12.9}	&	\underline{13.4}	&	\textbf{10.9}	&	\textbf{25.9}	&	\textbf{29.9}	&	7.7	&	\underline{38.4}	&	\textbf{24.8}	\\
	&	Scratch	&	(172M)	&	42.7	&	72.7	&	57.2	&	70.3	&	66.1	&	33.8	&	9.2	&	39.5	&	32.0	\\
\midrule
\multirow{5}{*}{Autoregressive}	&	Prompt ($S\,{=}\,1$)	&	(0.86M)	&	58.4	&	45.5	&	28.9	&	42.4	&	37.1	&	66.9	&	18.9	&	37.3	&	35.1	\\
	&	Prompt ($S\,{=}\,16$)	&	(0.88M)	&	45.8	&	41.4	&	19.6	&	36.6	&	33.4	&	41.4	&	16.4	&	32.6	&	28.8	\\
	&	Prompt ($S\,{=}\,256$)	&	(1.06M)	&	\underline{39.0}	&	\underline{39.6}	&	\underline{17.3}	&	\underline{34.9}	&	\underline{32.5}	&	{37.1}	&	15.0	&	29.6	&	\underline{26.7}	\\
	&	Prompt ($S\,{=}\,256, F\,{=}\,16$)	&	(5.16M)	&	\textbf{36.9}	&	\textbf{27.2}	&	\textbf{14.1}	&	\textbf{27.2}	&	\textbf{30.0}	&	\underline{34.6}	&	\textbf{12.8}	&	\underline{26.4}	&	\textbf{22.2}	\\
	&	Scratch	&	(306M)	&	39.6	&	76.0	&	56.1	&	52.5	&	92.7	&	\textbf{31.6}	&	\underline{13.5}	&	\textbf{19.4}	&	29.5	\\
    \bottomrule
    \end{tabular}
    }
    \caption{FIDs (lower the better) of image generation models on VTAB tasks. The number of trainable parameters (second column) are computed assuming 100 classes. The mean FID over 19 VTAB tasks (third column) and those for dataset with a small to mid-scale training data are reported. Complete results are in \cref{sec:supp_exp_vtab_fids}. The \textbf{best} and the \underline{second best} results are highlighted in each column.}
    \label{tab:vtab_main}
\end{table*}
\begin{figure*}[t]
    \centering
    \begin{subfigure}[b]{0.31\linewidth}
        \centering
        \includegraphics[width=\linewidth]{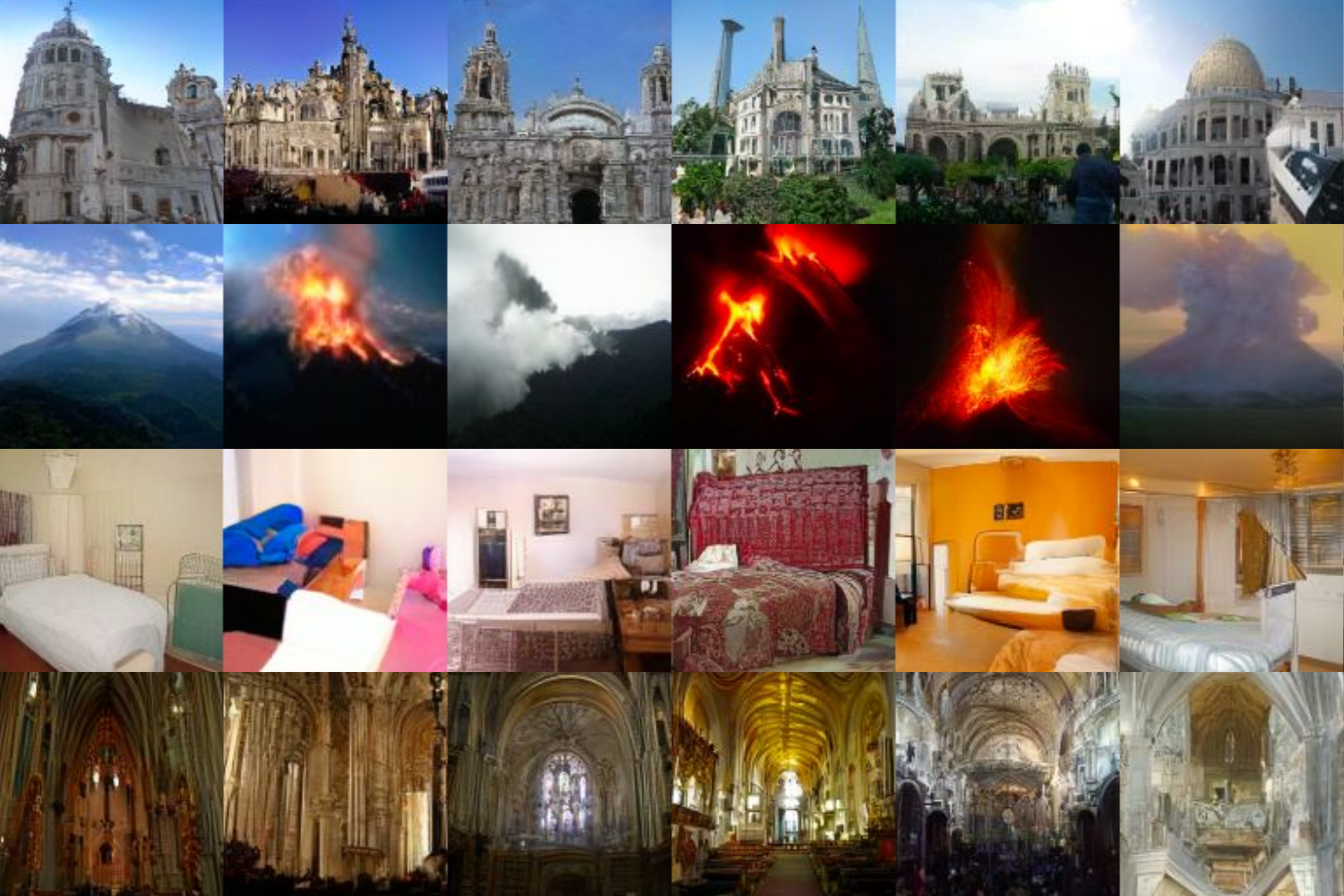}
        \caption{SUN397 (FID=$7.7$; NAR + Prompt)}
        \label{fig:vtab_maskgit_sun}
    \end{subfigure}
    \begin{subfigure}[b]{0.31\linewidth}
        \centering
        \includegraphics[width=\linewidth]{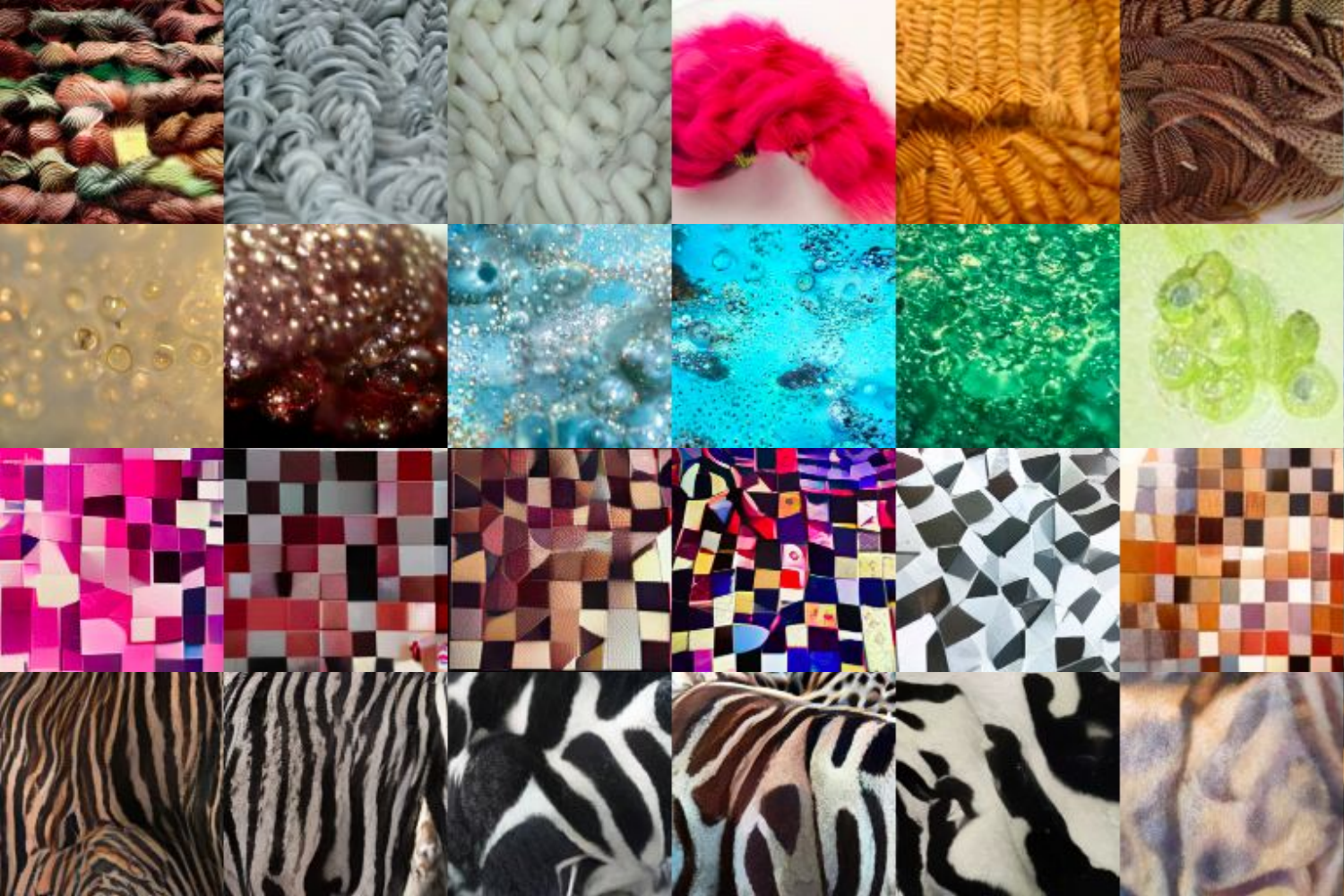}
        \caption{DTD (FID=$25.9$; NAR + Prompt)}
        \label{fig:vtab_maskgit_dtd}
    \end{subfigure}
    \begin{subfigure}[b]{0.31\linewidth}
        \centering
        \includegraphics[width=\linewidth]{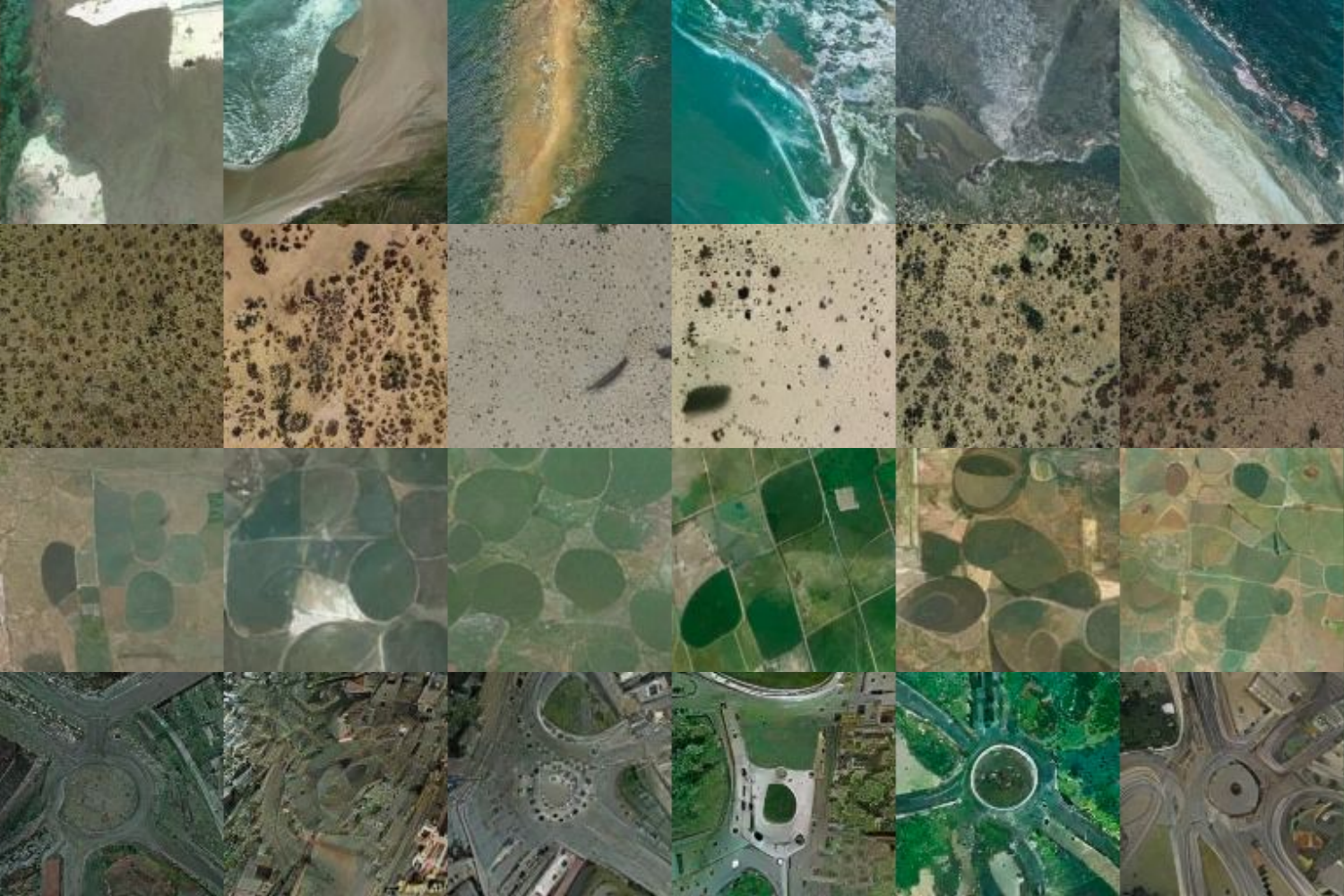}
        \caption{Resisc (FID=$24.8$; NAR + Prompt)}
        \label{fig:vtab_maskgit_resisc}
    \end{subfigure}
    \begin{subfigure}[b]{0.31\linewidth}
        \centering
        \includegraphics[width=\linewidth]{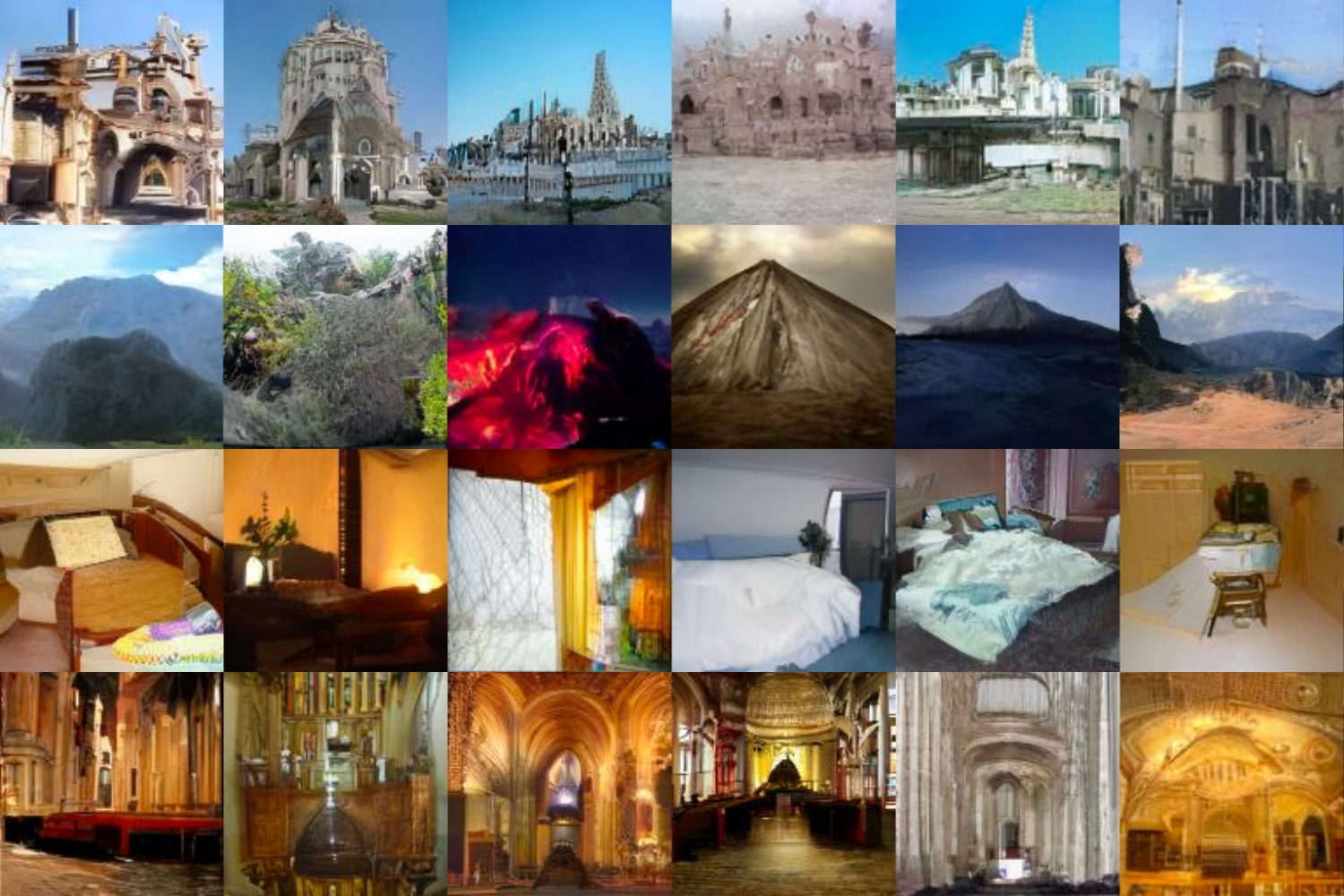}
        \caption{SUN397 (FID=$12.8$; AR + Prompt)}
        \label{fig:vtab_taming_sun}
    \end{subfigure}
    \begin{subfigure}[b]{0.31\linewidth}
        \centering
        \includegraphics[width=\linewidth]{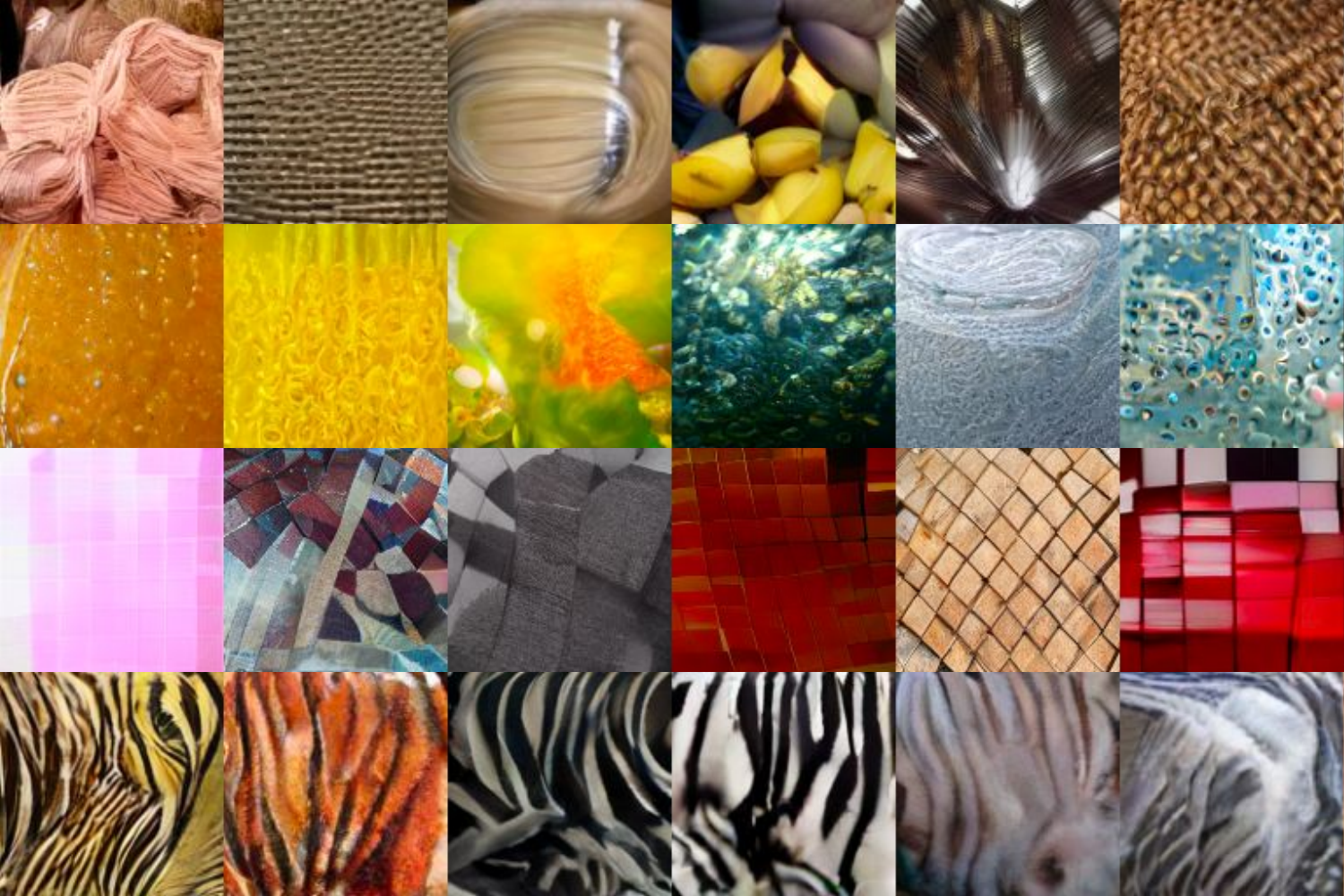}
        \caption{DTD (FID=$30.0$; AR + Prompt)}
        \label{fig:vtab_taming_dtd}
    \end{subfigure}
    \begin{subfigure}[b]{0.31\linewidth}
        \centering
        \includegraphics[width=\linewidth]{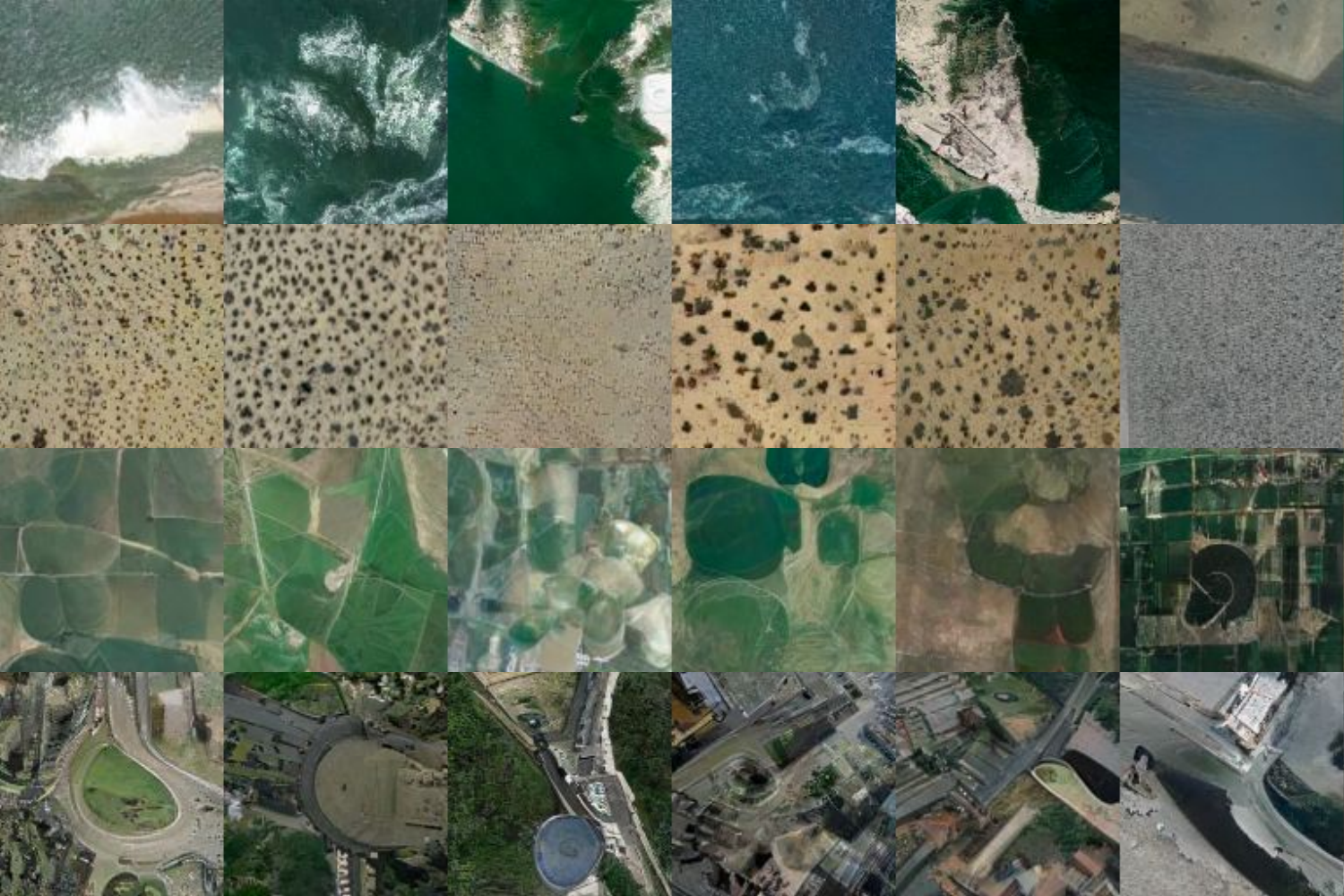}
        \caption{Resisc: (FID=$22.2$; AR + Prompt)}
        \label{fig:vtab_taming_resisc}
    \end{subfigure}
    \caption{Class conditional generation using NAR (top; $S{=}128$) and AR (bottom; $S{=}256, F{=}16$) transformers with prompt tuning.
    }
    \label{fig:vtab_vis}
\end{figure*}

\subsection{Generative Transfer on VTAB}
\label{sec:exp_vtab}

\vspace{0.05in}
\noindent\textbf{Dataset.} Towards developing a generative transfer method generalizable across domains and distributions, we employ the visual task adaptation benchmark (VTAB)~\cite{zhai2019large} -- a suite of 19 visual recognition tasks based on 16 datasets.
It covers diverse image domains (\eg, natural, structured, and specialized such as medical or satellite imagery) and tasks (\eg, object and scene recognition, distance classification, counting), making it a valuable asset not only for discriminative, but also for generative transfer learning.
The dataset information is provided in \cref{sec:supp_exp_vtab_info}.

\vspace{0.05in}
\noindent\textbf{Setting.} We study class-conditional image generation models on the VTAB (full) tasks. Class-conditional prompts are trained on the ``train'' split, using the same hyperparameters across tasks as provided in \cref{sec:supp_exp_vtab_hyper}.

We investigate generative transfer of AR and NAR transformers using class-conditional Taming Transformer~\cite{esser2021taming} and MaskGIT~\cite{chang2022maskgit}, respectively, trained on $256{\times}256$ images of ImageNet dataset as source models. Both models contain 24 transformer layers, comprised of $306$M and $172$M model parameters, respectively.

\vspace{0.05in}
\noindent\textbf{Baselines.} We compare our method with GAN-based generative transfer learning methods, including MineGAN~\cite{wang2020minegan} and cGANTransfer~\cite{shahbazi2021efficient}. Note that both of these algorithms use a BigGAN~\cite{brock2018large} trained on ImageNet as a source. It is worth noting that the BigGAN model is trained on $128{\times}128$ images and its validation FID on ImageNet is $7.4$. This is better than that of our pretrained AR transformer ($18.7$) and almost on par with that of NAR transformer ($6.2$).

We further compare with generative transformers trained from scratch on VTAB. To highlight the compute efficiency, models are trained with a comparable compute budget (\eg, same number of train epochs) to transfer learning models. Hyperparameters are provided in \cref{sec:supp_exp_vtab_hyper}. We provide more in-depth analysis without compute budget restrictions in \cref{sec:abl_beyond_prompt_tunig}.

\vspace{0.05in}
\noindent\textbf{Evaluation.} We use Frechet Inception Distance (FID)~\cite{heusel2017gans} as a quantitative metric. We generate $20k$ images from each model and compare with images from a respective dataset. We sample $20k$ images if the dataset is larger than $20k$.

\vspace{0.05in}
\noindent\textbf{Results.} We report FIDs of models trained and evaluated on VTAB tasks in \cref{tab:vtab_main} averaged over 3 runs. Due to limited space, we report results on tasks with small to mid-scale train set in addition to the mean FID over 19 datasets. Complete results are given in \cref{sec:supp_exp_vtab_fids}. In addition, we provide images generated by various transfer learning methods for thorough visual inspection. See \cref{sec:supp_exp_vtab} for evaluation details and more extensive comparison.
We see that prompt tuning is effective for both AR and NAR generative transformers, especially when the number of training images is small (\eg., $\leq{10k}$).
Between AR and NAR transformers, we find that NAR model transfers better than the AR counterpart. 
Nevertheless, both generative transformers with class-conditional prompt tuning show significant gain in performance when compared to GAN-based baselines.

We see that the prompt tuning of generative transformers benefits greatly from a long prompt, reducing mean FID from $53.7$ to $36.4$ by increasing the length from $1$ to $128$. This is achieved by only adding less than $0.1$M parameters, thanks to our parameter-efficient design of the prompt token generator. Nevertheless, this comes at an increased cost at generation time due to increased sequence length. Empirically, we find that using $128$ tokens for the prompt increases the overall generation time by 25\%, as shown in \cref{tab:method_comparison}.

AR transformers also benefit from the longer prompt. On the other hand, AR transformers generally requires prompts with more learnable parameters, which is achieved by increasing the number of factors. The performance is still on par with that achievable with the baseline prompt, while using significantly less number of parameters ($5.6$M instead of $20.5$M), as shown in \cref{sec:abl_nonfac_prompt}.

In \cref{fig:vtab_vis}, we show generated images using 128 prompt tokens for NAR transformers and 256 prompt tokens (with $F\,{=}\,16$) for AR transformers on a few VTAB tasks. More generated images are in \cref{sec:supp_exp_vtab_synth}.
Despite learning less than 0.5\% of the transformer parameters, the learned prompts are able to change the generation process of pretrained generative transformers to follow the target distribution.

\subsection{Few-shot Generative Transfer}
\label{sec:exp_fewshot}

After validation on VTAB, we delve deeper into a few-shot generative transfer, where the number of training images is further reduced. We limit our study to transfer of an NAR transformer, \ie, MaskGIT~\cite{chang2022maskgit}, but with more comparisons to existing few-shot image generation models, either with~\cite{wang2020minegan,shahbazi2021efficient} or without~\cite{zhao2020differentiable,tseng2021regularizing} knowledge transfer.

\vspace{0.05in}
\noindent\textbf{Dataset.} We study few-shot generative transfer learning on Places~\cite{zhou2014learning}, ImageNet~\cite{deng2009imagenet}, and Animal Face~\cite{si2011learning}. Following~\cite{wang2020minegan,shahbazi2021efficient}, for Places and ImageNet, we select 5 classes\footnote{Cock, Tape player, Broccoli, Fire engine, Harvester for ImageNet, and Alley, Arch, Art gallery, Auditorium, Ballroom for Places.} and use 500 images per class for training. For Animal Face, we consider two scenarios -- following \cite{shahbazi2021efficient}, we use 100 images per class for training from 20 classes (denoted as ``Animal Face'' in \cref{tab:fewshot_main}); alternatively, following \cite{zhao2020differentiable,tseng2021regularizing}, we use all images of dog (389) and cat (160) classes (denoted as ``dog face'' and ``cat face'' in \cref{tab:fewshot_main}) for training.

Moreover, we test our methods to more challenging off-manifold target tasks on DomainNet~\cite{peng2019moment} Infograph and Clipart (345 classes), and ImageNet sketch (1000 classes)~\cite{wang2019learning} with as low as 2 training images per class.

\vspace{0.05in}
\noindent\textbf{Setting.} We study a class-and-instance conditional generative transfer as in \cref{sec:method_prompt_token_generator}. Class-and-instance conditional prompts are particularly suitable for few-shot scenarios as there are only a limited number of training images.

\vspace{0.05in}
\noindent\textbf{Baselines.} GAN-based generative transfer learning methods, \eg, MineGAN~\cite{wang2020minegan} and cGANTransfer~\cite{shahbazi2021efficient}, are used as baselines. Moreover, we compare to few-shot image generation models, \eg, DiffAug~\cite{zhao2020differentiable} and LeCam GAN~\cite{tseng2021regularizing}.

\vspace{0.05in}
\noindent\textbf{Evaluation.} We report FIDs using $10k$ generated images, except for experiments on dog and cat faces, where we generate $5k$ images following~\cite{zhao2020differentiable}. For Places, ImageNet, and Animal Face, we use an entire training data (\ie, 2500 for Places and ImageNet, 2000 for Animal Face, 389 and 160 for dog and cat faces, respectively) for the reference distribution. We sample $10k$ images for the reference distribution to compute FID for DomainNet and ImageNet sketch.

\begin{table}[t]
    \centering
    \resizebox{0.99\linewidth}{!}{%
    \begin{tabular}{l|c|c|c|c|c}
        \toprule
        Dataset & ImageNet & Places & Animal Face & Dog Face & Cat Face \\
        (shot) & (500) & (500) & (100) & (389) & (160) \\
        \midrule
        MineGAN~\cite{wang2020minegan} & 61.8$^{\dagger}$ & 82.3$^{\dagger}$ & -- & 93.0$^{\ast}$ & 54.5$^{\ast}$ \\
        cGANTransfer~\cite{shahbazi2021efficient} & -- & 71.1$^{\ddagger}$ & 85.9$^{\ddagger}$ & -- & -- \\
        DiffAug~\cite{zhao2020differentiable} & -- & -- & -- & 58.5$^{\ast}$ & 42.4$^{\ast}$ \\
        LeCam GAN~\cite{tseng2021regularizing} & -- & -- & -- & 54.9$^{\ast}$ & 34.2$^{\ast}$ \\
        \midrule
        Ours (class) & \textbf{16.9} & {24.2} & {16.3} & 65.4 & 40.2 \\
        Ours (instance) & 19.6 & \textbf{19.5} & \textbf{13.3} & \textbf{26.0} & \textbf{31.2} \\
        \bottomrule
    \end{tabular}
    }
    \caption{FIDs of image generation models on few-shot benchmark. Numbers with $\dagger$, $\ddagger$, $\ast$ are from \cite{wang2020minegan}, \cite{shahbazi2021efficient}, \cite{tseng2021regularizing}, respectively.}
    \label{tab:fewshot_main}
\end{table}
\begin{figure*}[t]
    \centering
    \begin{subfigure}[b]{0.32\linewidth}
        \centering
        \includegraphics[height=1.8in]{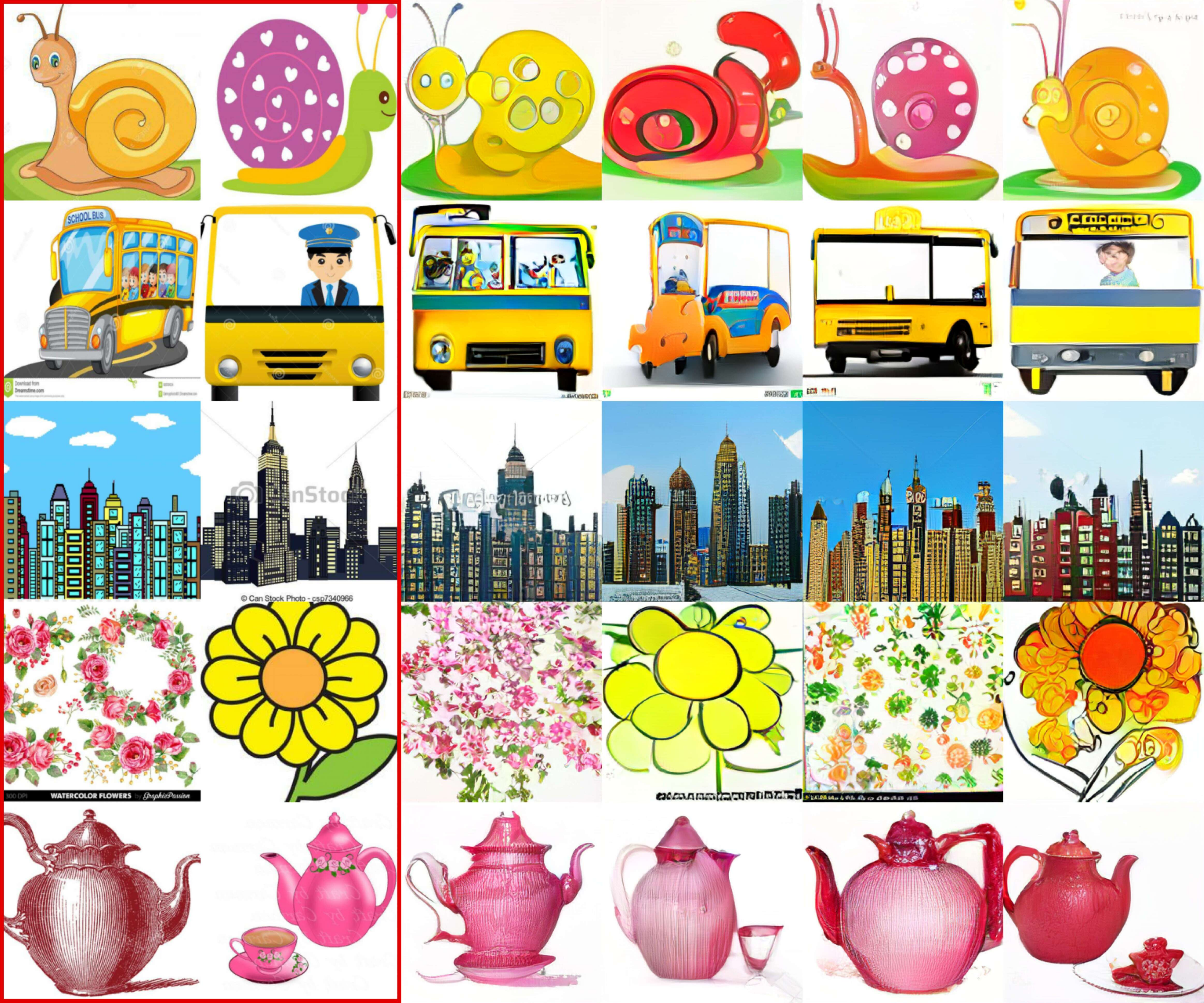}
        \caption{DomainNet Clipart (2 shot; FID=$22.4$)}
        \label{fig:fewshot_clipart}
    \end{subfigure}
    \hspace{0.05in}
    \begin{subfigure}[b]{0.32\linewidth}
        \centering
        \includegraphics[height=1.8in]{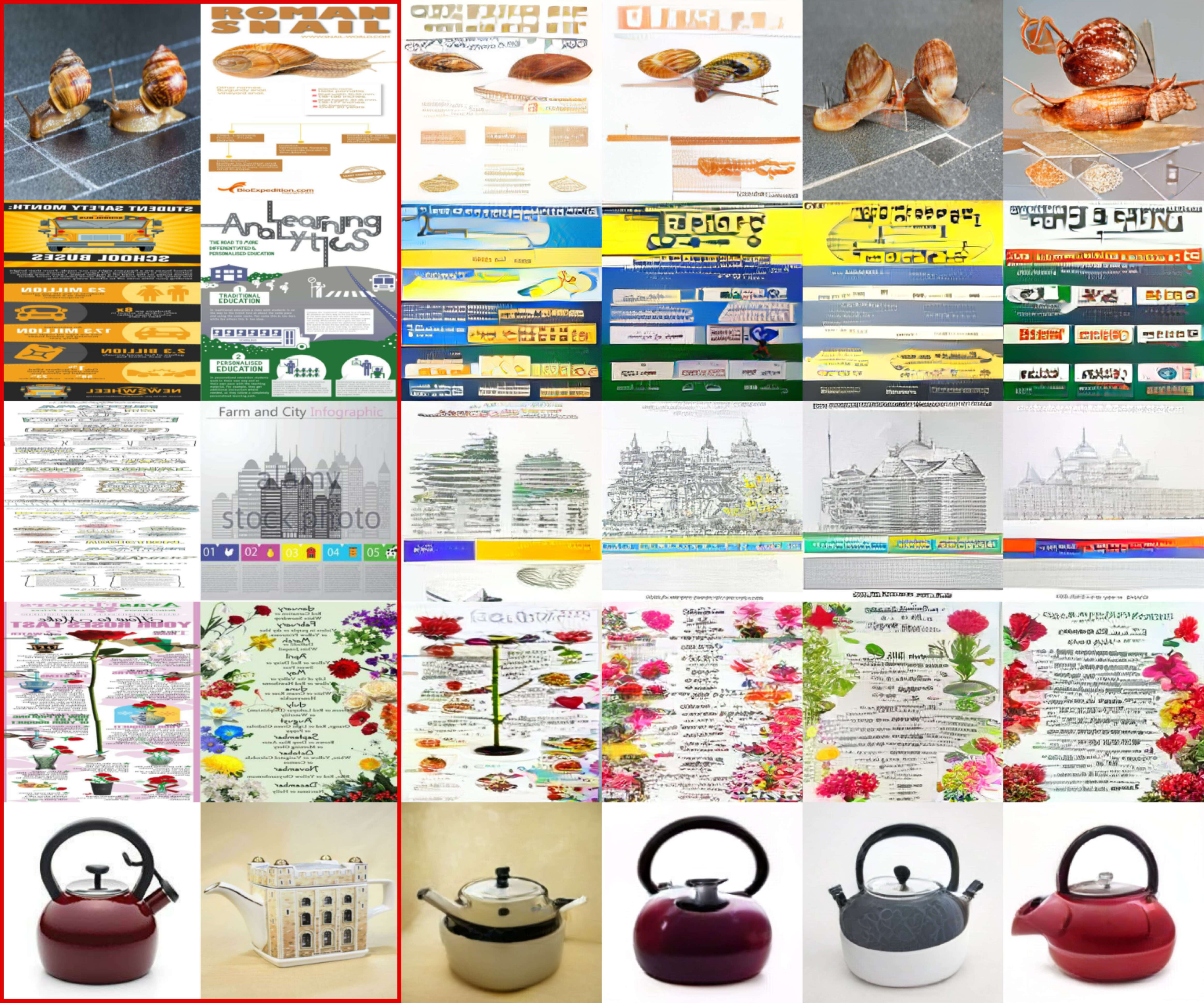}
        \caption{DomainNet Infograph (2 shot; FID=$20.6$)}
        \label{fig:fewshot_infograph}
    \end{subfigure}
    \hspace{0.05in}
    \begin{subfigure}[b]{0.32\linewidth}
        \centering
        \includegraphics[height=1.8in]{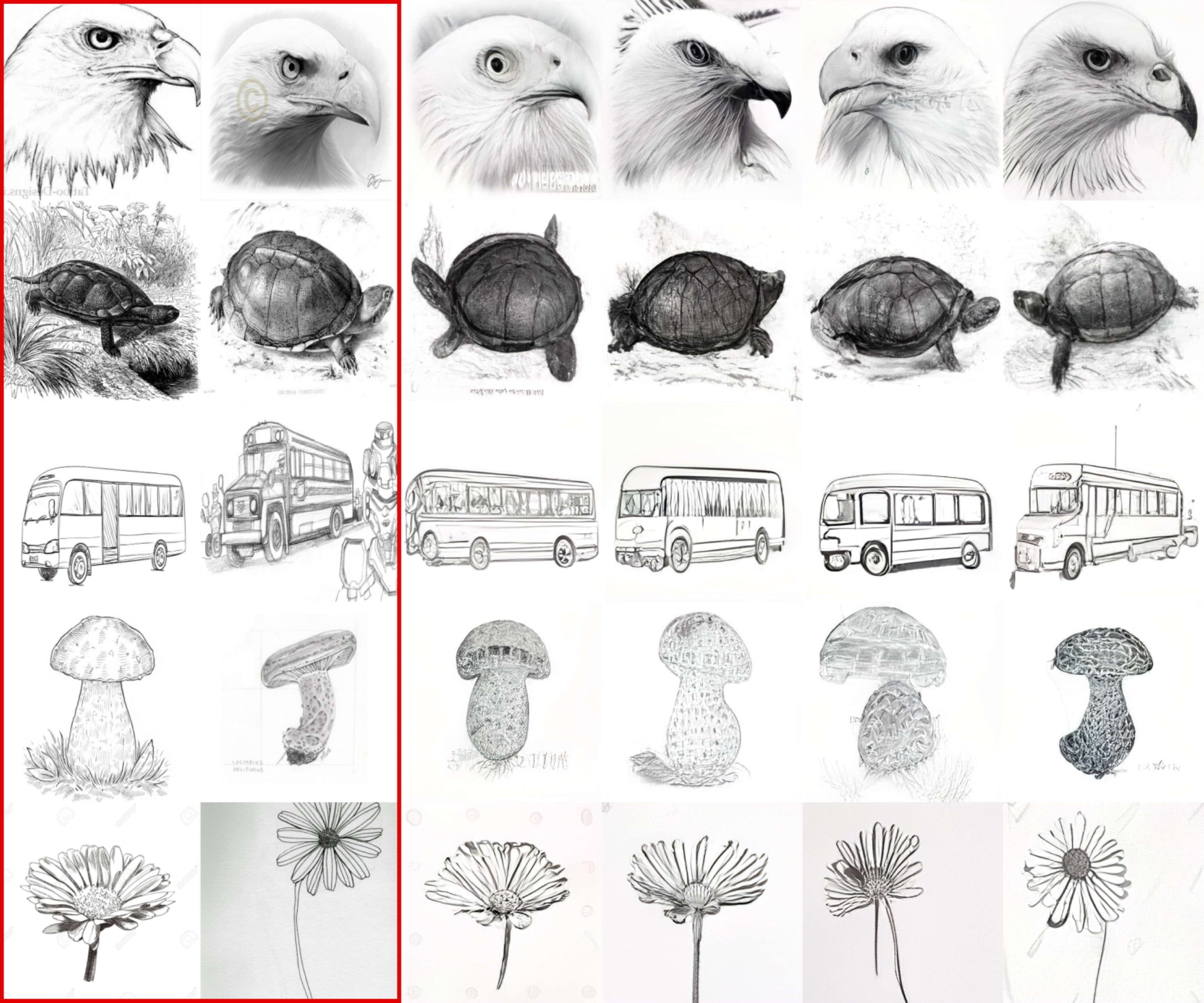}
        \caption{ImageNet Sketch (2 shot; FID=$14.4$)}
        \label{fig:fewshot_sketch}
    \end{subfigure}
    \caption{Class conditional generation of few-shot transfer models. Images in red boxes are two training images of each class.}
    \label{fig:fewshot_vis}
\end{figure*}

\vspace{0.05in}
\noindent\textbf{Results.} In \cref{tab:fewshot_main}, we report FIDs of our proposed method using prompts of $S\,{=}\,128$. When conditioned on the class, our method improves FIDs upon existing generative transfer learning methods. When comparing with few-shot generation methods on dog and cat face datasets, our method with a class condition slightly under-performs, likely due to that dataset having one class.
When conditioned on instances, our models outperform all GAN-based few-shot generation models. We provide visualizations in \cref{sec:supp_exp_fewshot_synth}.

We visualize generated images conditioned on the class by our models in \cref{fig:fewshot_vis}. We show 2 (and only) training data for each class in red boxes. We observe that, though images in these datasets are highly artificial and their distributions are different from the source dataset, our method is able to synthesize images from respective target distributions well. 
Moreover, as clearly seen from \cref{fig:fewshot_vis}, our models do more than simply memorizing the training data.

\vspace{0.05in}
\noindent\textbf{Data Efficiency.}
We conduct experiments with less training images to investigate the data efficiency. We train models on 5, 10, 50 and 100 images per class for ImageNet, Places and Animal Face datasets. We use a class-condition for image generation. The same number of images is used for the reference set to make FIDs comparable across settings.

\begin{figure}[t]
    \centering
    \begin{subfigure}[b]{0.35\linewidth}
        \centering
        \includegraphics[height=1.33in]{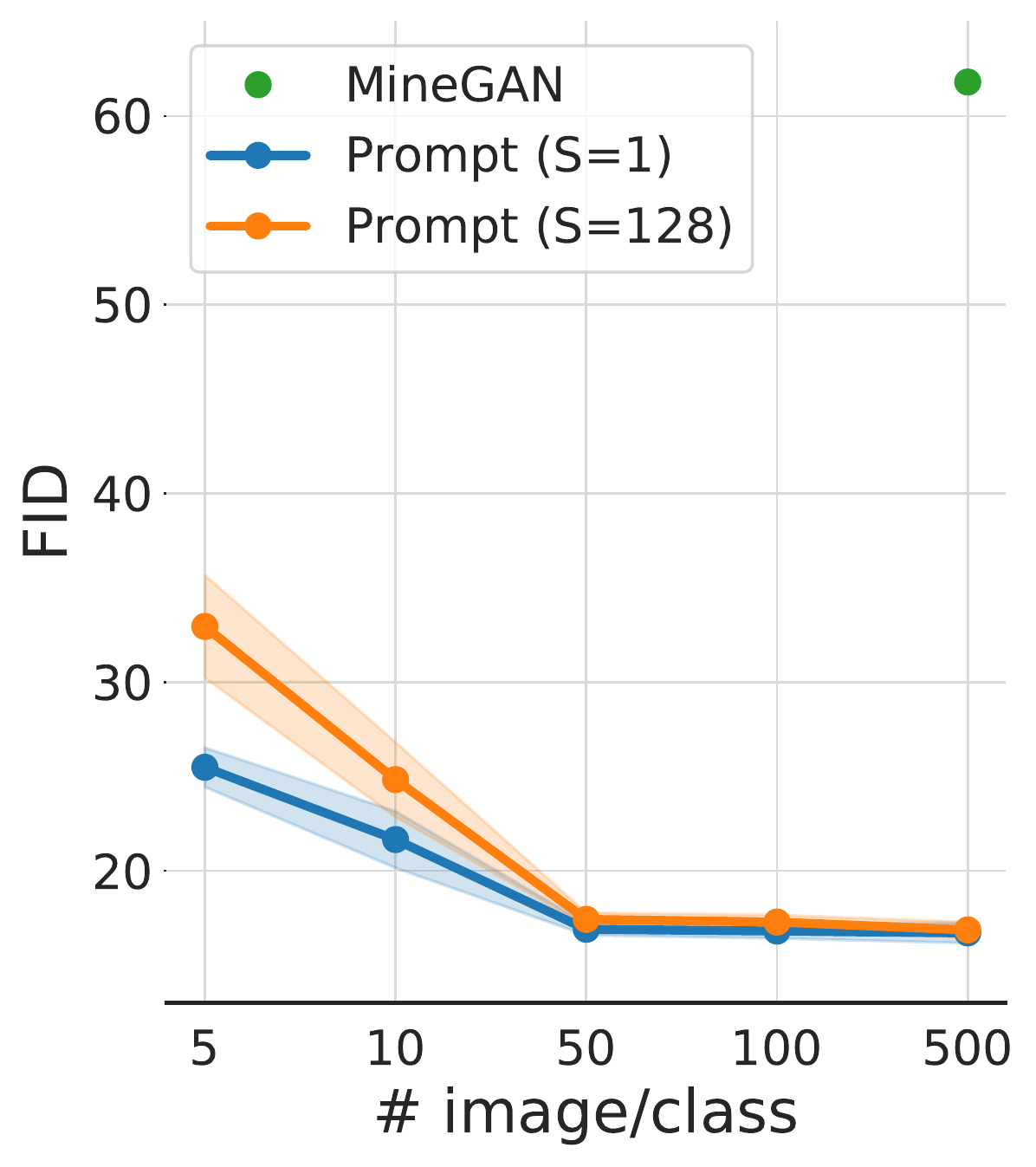}
        \caption{ImageNet}
        \label{fig:fewshot_imagenet}
    \end{subfigure}
    \begin{subfigure}[b]{0.35\linewidth}
        \centering
        \includegraphics[height=1.33in]{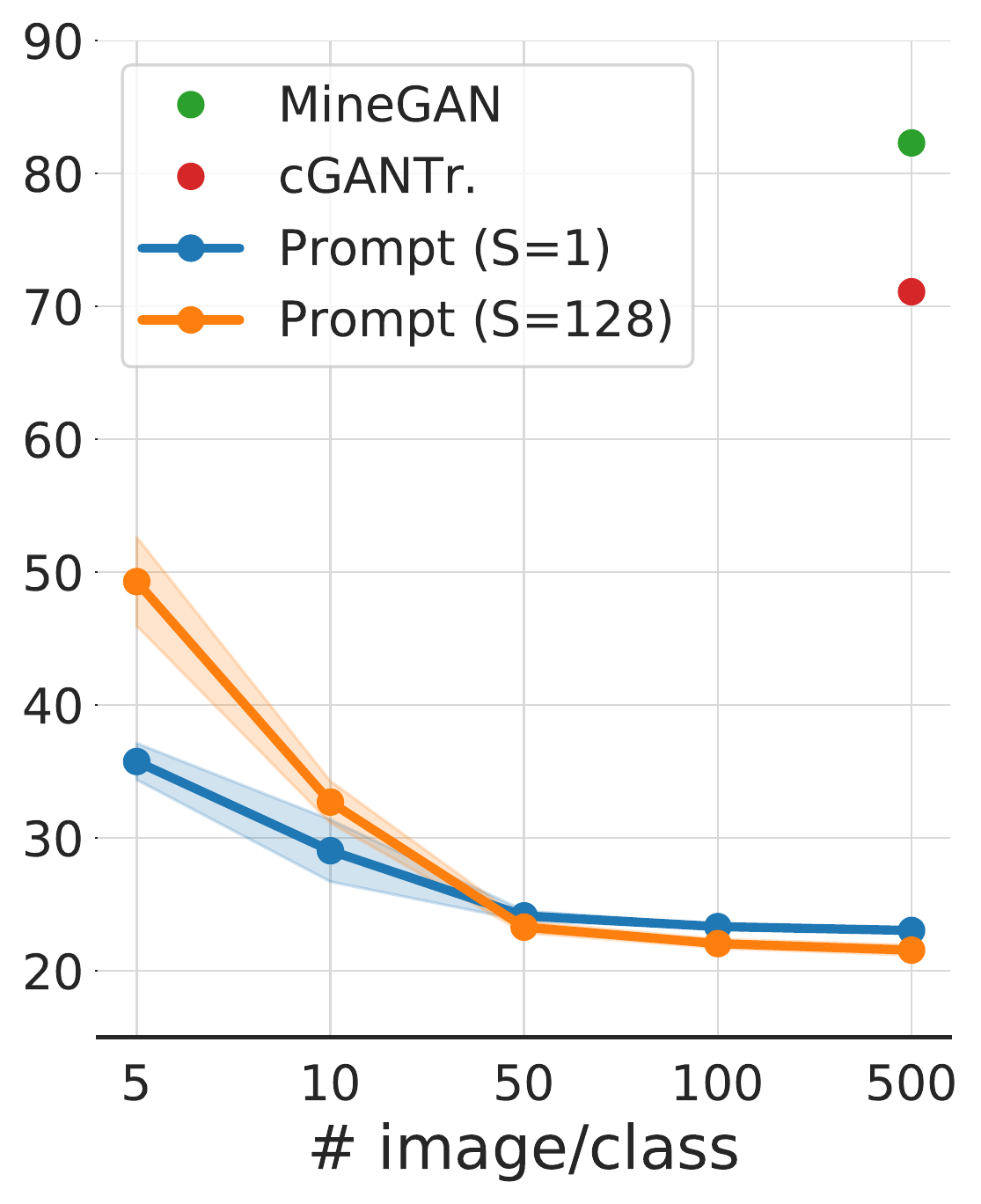}
        \caption{Places}
        \label{fig:fewshot_places}
    \end{subfigure}
    \begin{subfigure}[b]{.28\linewidth}
        \centering
        \includegraphics[height=1.33in]{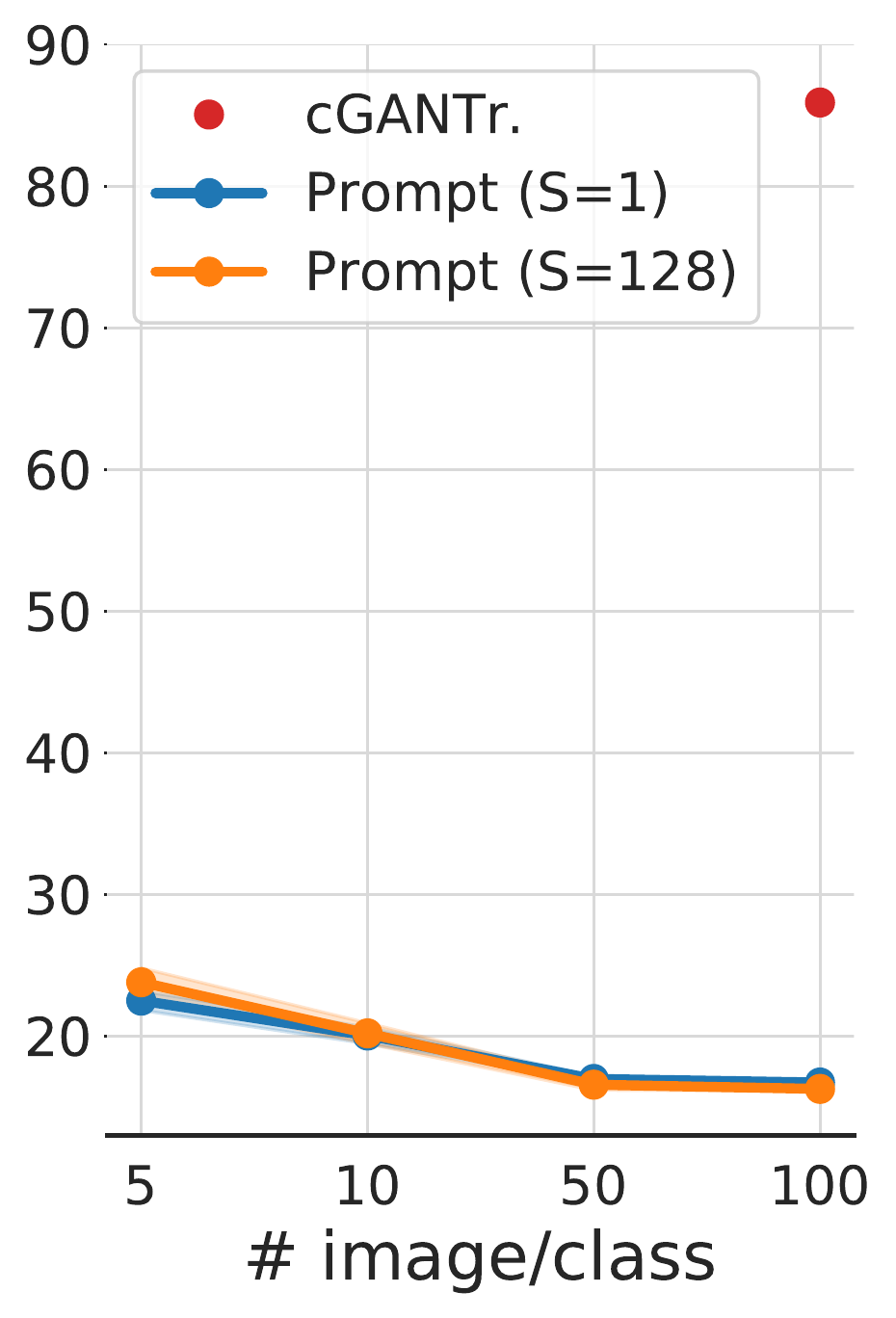}
        \caption{Animal Face}
        \label{fig:fewshot_animalface}
    \end{subfigure}
    \caption{FIDs for models trained with varying numbers of images per class for class-conditional few-shot generative transfer.}
    \label{fig:fewshot_fid_vs_numimg}
\end{figure}

Results are in \cref{fig:fewshot_fid_vs_numimg}. Our method shows far superior data efficiency, achieving substantially lower FIDs with only 5 training images per class, to GAN-based transfer learning methods trained with 20 or 100 times more images per class. 
We find that using long prompts is not favorable when the number of training images is too small (\eg, less than 10 images per class for ImageNet and Places, 50 in total), as models start to overfit to a few images in the train set. When the total number of images is larger than 250, we find that using a long prompt is still beneficial.

\vspace{0.05in}
\noindent\textbf{Enhancing Generation Diversity via Prompt Engineering.}
As in \cref{sec:method_prompt_composition,fig:autoreg_gen_interp_instance_to_class,fig:autoreg_gen_interp_instance_to_instance}, our model offers a way to enhance generation diversity by composing prompts. We report quantitative metrics to support our claim. 

We conduct experiments on the dog and cat faces dataset using marquee header prompts with different $T_{\mathrm{cutoff}}$ values. For the fidelity metric, we compute the FID. To measure the diversity, we follow \cite{ojha2021few} and report a intra-cluster pairwise LPIPS distance, where we generate $5k$ samples and map them into one of training images.\footnote{We use a pixel-wise L2 distance for computation efficiency instead of LPIPS distance in \cite{ojha2021few}.}

\begin{figure}[t]
    \centering
    \begin{subfigure}[b]{0.48\linewidth}
        \centering
        \includegraphics[width=\textwidth]{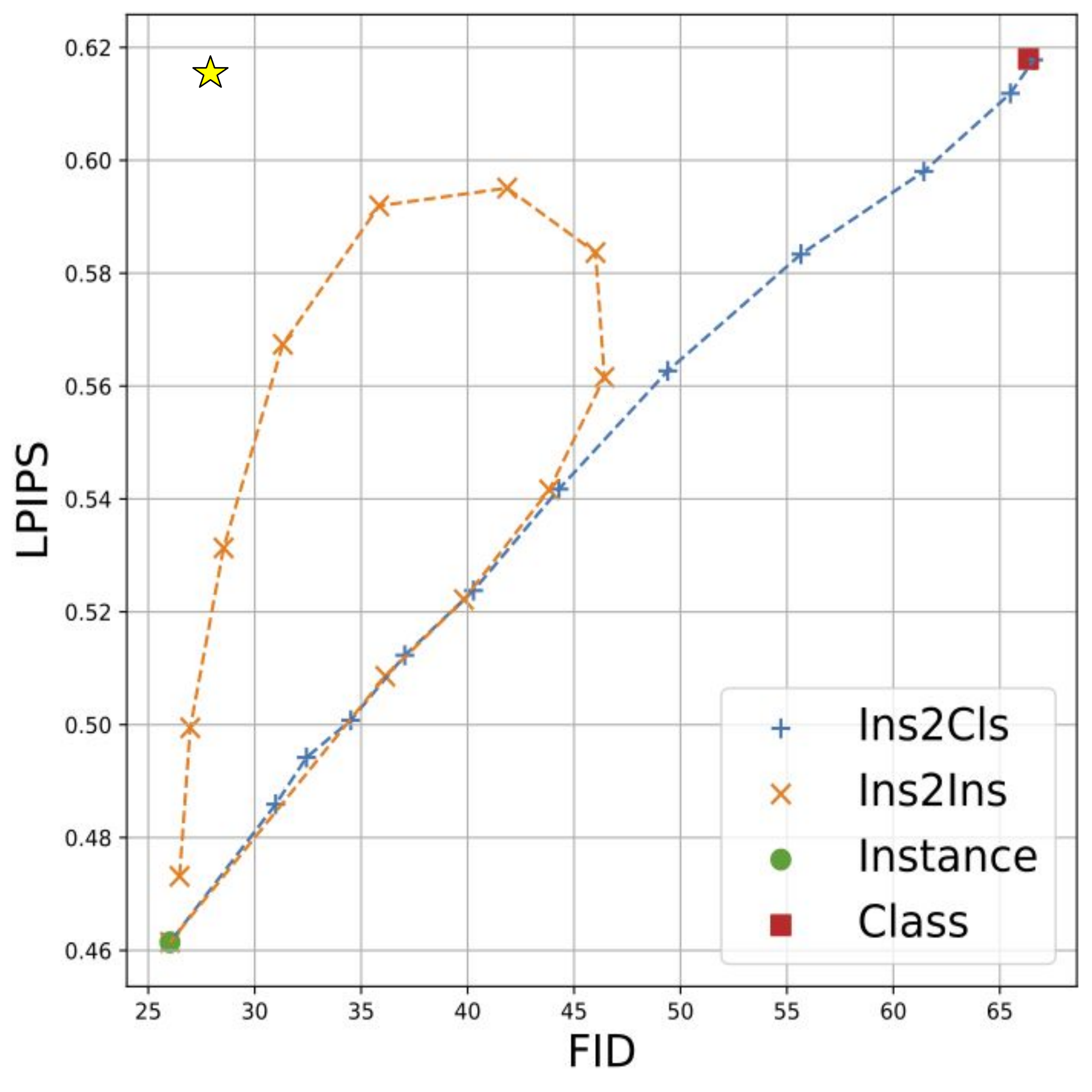}
        \caption{Dog Faces}
        \label{fig:fewshot_mht_dogface}
    \end{subfigure}
    \begin{subfigure}[b]{0.48\linewidth}
        \centering
        \includegraphics[width=\textwidth]{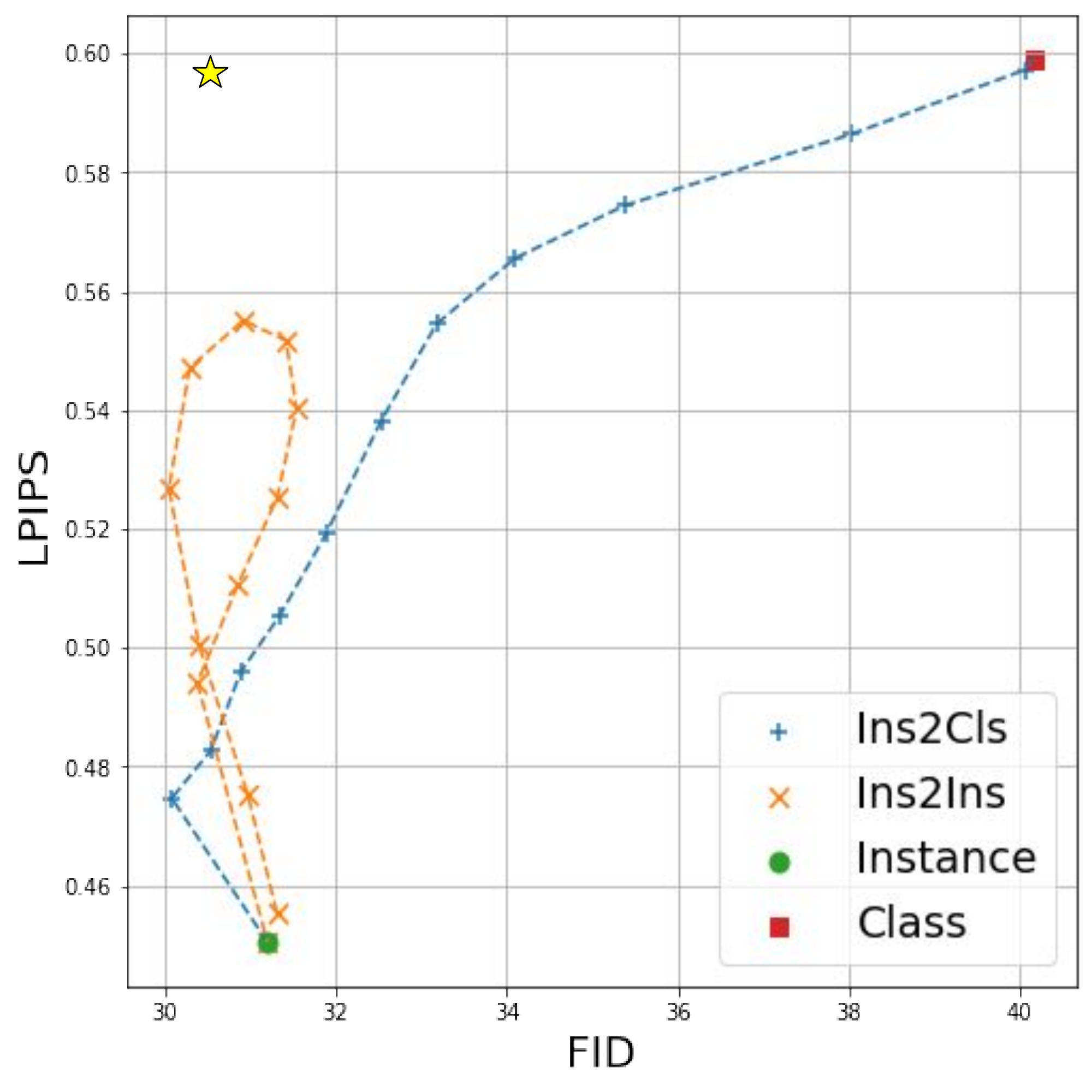}
        \caption{{Cat Faces}}
        \label{fig:fewshot_mht_catface}
    \end{subfigure}
    \caption{Marquee header prompt shows clear tradeoff between fidelity (FID) and diversity (LPIPS) when interpolating from instance to class (blue). It shows a better tradeoff when interpolating between instances (orange), achieving low FID and high LPIPS.}
    \label{fig:fewshot_mht}
\end{figure}

Results are shown in \cref{fig:fewshot_mht}. Ideally, we expect a model with low FID and high intra-cluster LPIPS scores (\eg, yellow star at top-left corner). When generating samples using a class-condition (red square), we generate diverse images, but with relatively poor fidelity. On the other hand, when conditioned on data instances (green dot), we improve the FID by a large margin, but at the cost of reduced diversity. Instance to class Marquee header prompts (blue) allow to control the generation diversity and fidelity. Moreover, instance to instance Marquee header prompts, which interpolates an instance prompt to another randomly selected instance prompt, shows much better tradeoff between fidelity and diversity.

\section{Analysis and Discussion}
\label{sec:abl}

With a successful demonstration of the power of prompt tuning for generative transfer learning, we further study to understand prompt representations (\cref{sec:abl_prompt_representation}, \cref{sec:abl_prompt_tradeoff}) and conduct an ablation study regarding design choices of prompt token generator (\cref{sec:abl_nonfac_prompt}) and transfer learning (\cref{sec:abl_beyond_prompt_tunig}).

\subsection{What does the Prompt Learn?}
\label{sec:abl_prompt_representation}

To understand what the prompt has learned, we study some properties of learned prompt representations. For this study, we train instance conditioned prompt models on flowers dataset of VTAB, with $S\,{=}\,1$ and $128$. Note that no class information is used for training in this experiment.

\begin{figure}
    \centering
    \begin{subfigure}[b]{0.45\linewidth}
        \centering
        \includegraphics[width=\linewidth]{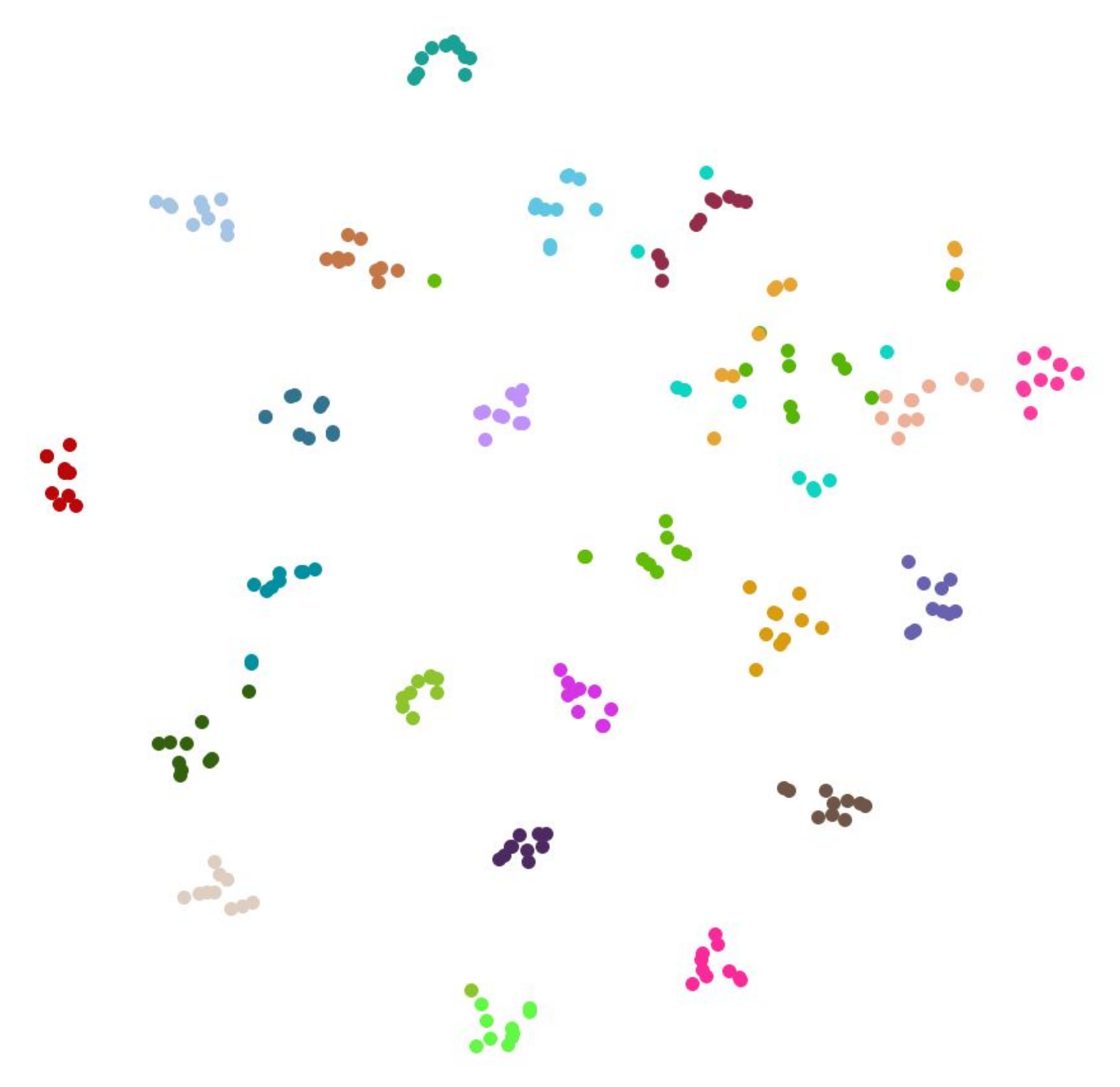}
        \caption{$S\,{=}\,1$ (NMI=$0.848$)}
        \label{fig:abl_instance_prompt_tsne_flower_s1}
    \end{subfigure}
    \hfill
    \begin{subfigure}[b]{0.45\linewidth}
        \centering
        \includegraphics[width=\linewidth]{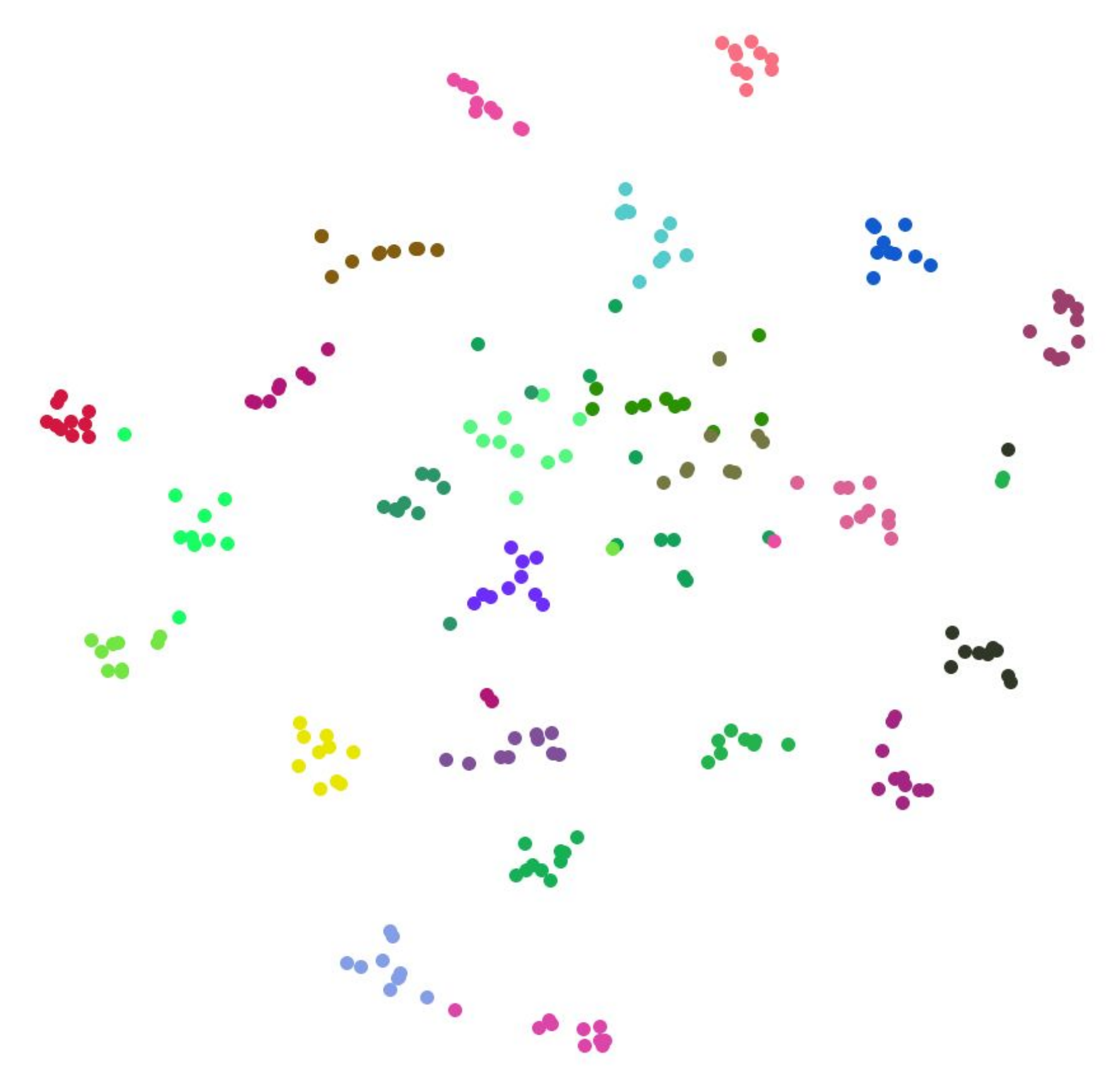}
        \caption{$S\,{=}\,128$ (NMI=$0.800$)}
        \label{fig:abl_instance_prompt_tsne_flower_s128}
    \end{subfigure}
    \caption{t-SNE plots of instance-conditioned prompt representations on flowers dataset. Points of the same color are from the same class. We also report normalized mutual information (NMI) score by clustering prompt representations using KMeans.}
    \label{fig:abl_instance_prompt_tsne}
\end{figure}

We draw t-SNE plots~\cite{van2008visualizing} of prompts in \cref{fig:abl_instance_prompt_tsne}. Here, we opt to use an output of an $\textsc{MLP}_{C}$ as a prompt representation instead of a token sequence (\eg, output of an $\textsc{MLP}_{T}$) due to its low dimensionality.
We see in \cref{fig:abl_instance_prompt_tsne_flower_s1} that points of the same color (\ie, same class) are grouped together, implying that the prompt representations learn discriminative class information. While we see a similar trend in \cref{fig:abl_instance_prompt_tsne_flower_s128}, there are clusters crowded with points of various colors. 
We quantify our observation using a normalized mutual information (NMI) computed by clustering prompts. Clustering is more consistent with the ground-truth class labels with higher NMIs. The model with $S\,{=}\,1$ achieves $0.848$ and the one with $S\,{=}\,128$ gets $0.800$. Note that these are even better results than the number obtained using an embedding from ImageNet pretrained ResNet-50~\cite{he2016deep} (NMI=$0.734$).

\subsection{Adaptation-Diversity Trade-Off}
\label{sec:abl_prompt_tradeoff}

\begin{figure}
    \centering
    \includegraphics[width=0.9\linewidth]{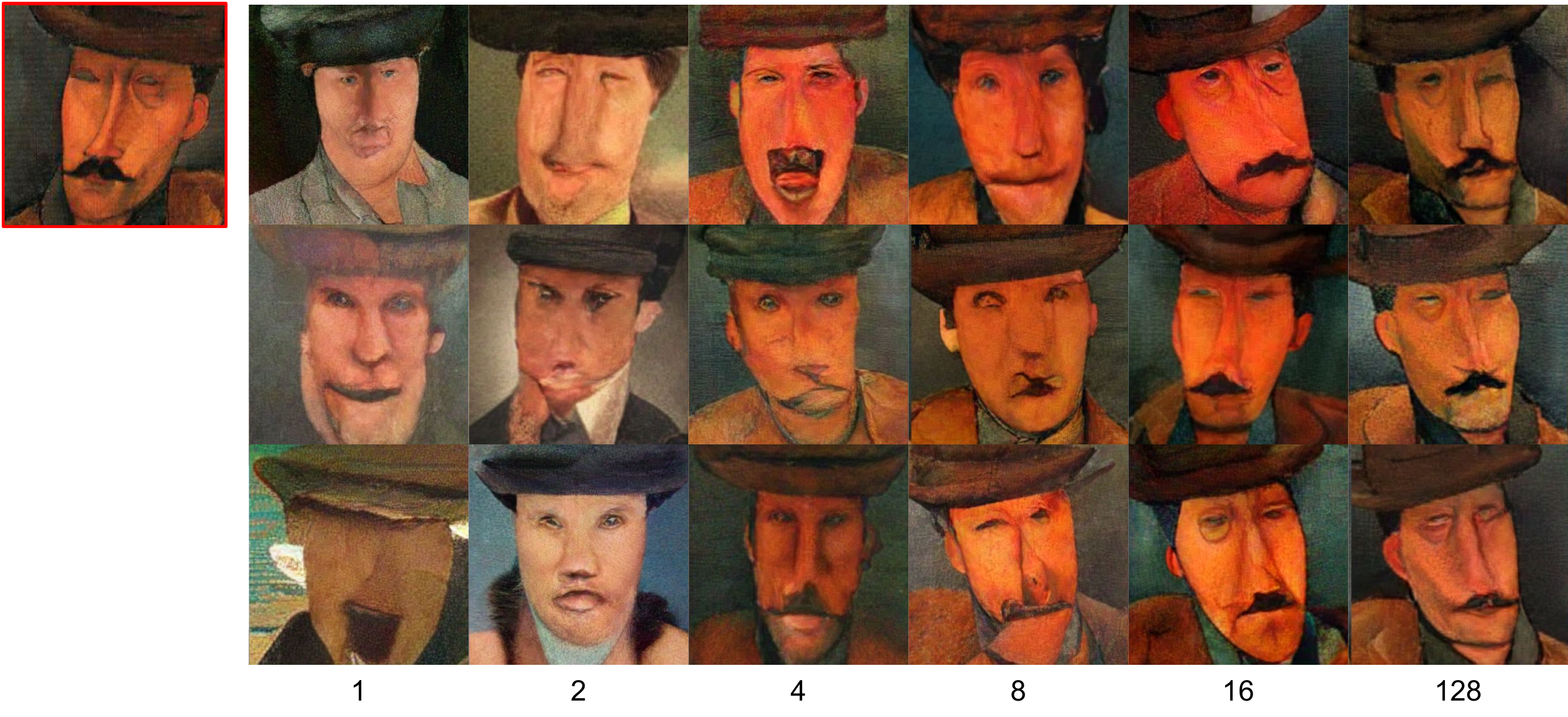}
    \caption{A single training image in red box and those generated by models using prompts of various lengths from $1$ to $128$.}
    \label{fig:abl_artface_256}
\end{figure}

We study prompts with various lengths, but on a \emph{single} image. We show generated images of models with different lengths in \cref{fig:abl_artface_256}. With short prompts, the model produces diverse but less detailed images. On the other hand, a long prompt model generates images of a higher quality, more faithful to the training image, but less diverse.
This implies that the short prompt learns concepts, while the long prompt learns fine details of training data. This is in line with our results in \cref{sec:abl_prompt_representation} where short prompts learn more discriminative information than long prompts.

\begin{figure*}
    \centering
    \begin{subfigure}[b]{0.48\linewidth}
        \centering
        \includegraphics[width=0.9\linewidth]{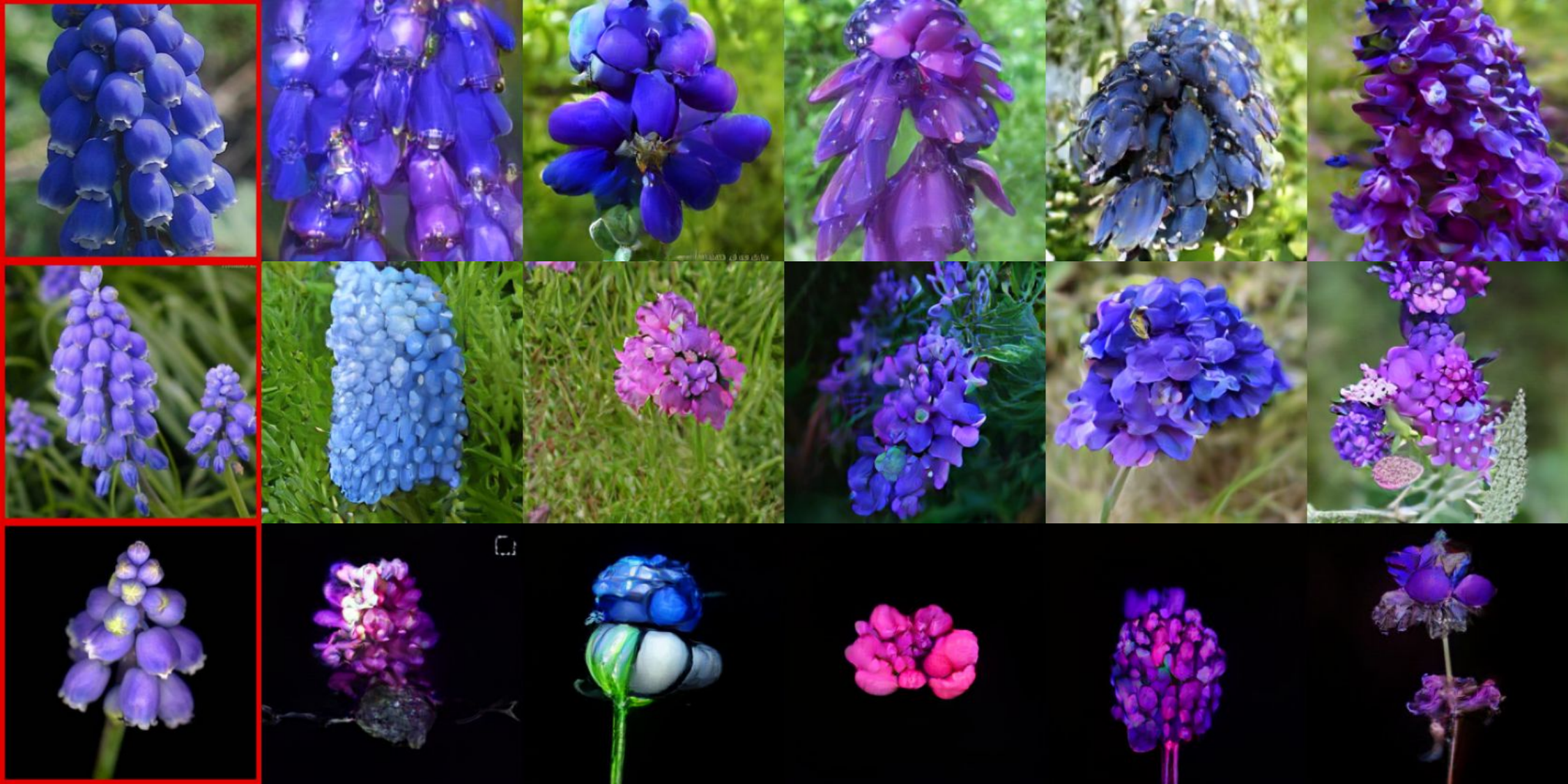}
        \caption{Oxford Flowers, ``Grape hyacinth'' ($S\,{=}\,1$)}
        \label{fig:abl_instance_gen_flower}
    \end{subfigure}
    \begin{subfigure}[b]{0.48\linewidth}
        \centering
        \includegraphics[width=0.9\linewidth]{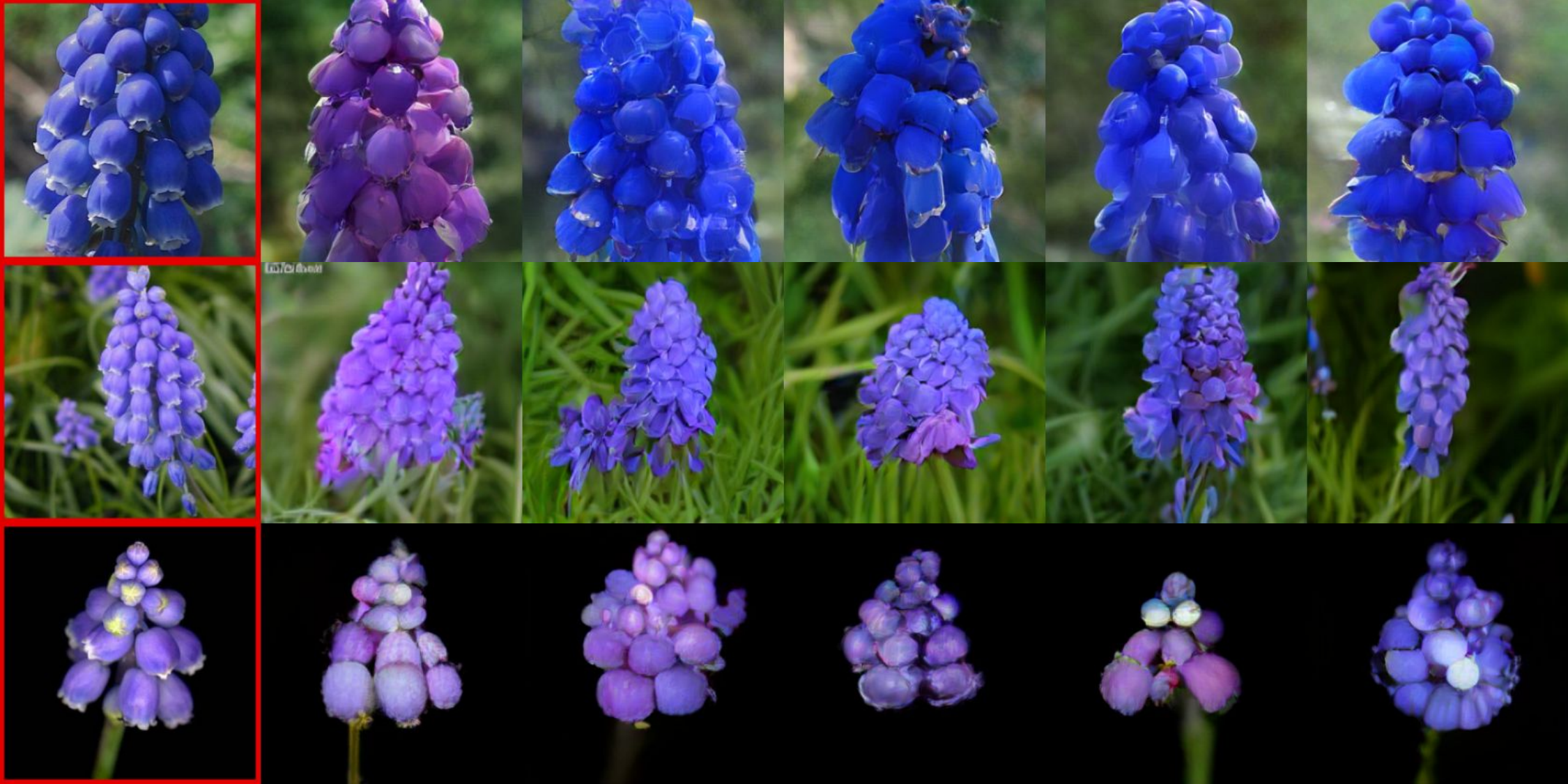}
        \caption{Oxford Flowers, ``Grape hyacinth'' ($S\,{=}\,128$)}
        \label{fig:abl_instance_gen_flower_seq}
    \end{subfigure}
    \caption{Instance-conditioned generation. For each row, leftmost image in red box is a training image and next five images are generated. When instance conditioned, generated images follow finer-grained details of the reference training image, such as color, shape, or background, beyond class information. Adaptation and diversity could be further controlled by the prompt length.}
    \label{fig:abl_instance_gen}
\end{figure*}

In \cref{fig:abl_instance_gen}, we visualize images generated by models of \cref{sec:abl_prompt_representation}. Compared to images in \cref{fig:abl_instance_gen_flower} whose model is trained with $S\,{=}\,1$, we clearly see in \cref{fig:abl_instance_gen_flower_seq} that the model trained with a long prompt generates images that are more consistent with training instances. 


\begin{table}[t]
    \centering
    \resizebox{0.99\linewidth}{!}{%
    \begin{tabular}{l|l|c||c|c|c||c|c|c}
        \toprule
        \multicolumn{2}{c|}{NAR} & \# params & Small & Medium & Large & Natural & Struct. & Spec. \\
        \midrule
        \multirow{4}{*}{$S{=}16$} & baseline & $1.81$M & 18.6 & 34.6 & 89.1 & 23.8 & 50.9 & 41.7 \\
         & $F{=}1$ & $0.68$M & 18.6 & 36.1 & 89.5 & 25.2 & 51.9 & 41.5 \\
         & $F{=}4$ & $0.95$M & 18.6 & 35.5 & 88.4 & 24.4 & 51.5 & 41.4 \\
         & $F{=}16$ & $2.02$M & 18.5 & 35.0 & 86.8 & 24.3 & 50.8 & 40.4 \\ 
        \midrule
        \multirow{4}{*}{$S{=}128$} & baseline & $10.4$M & 18.2 & 30.8 & 86.4 & 22.0 & 46.9 & 39.9 \\
         & $F{=}1$ & $0.76$M & 18.5 & 30.6 & 88.9 & 22.5 & 47.1 & 40.5 \\
         & $F{=}4$ & $1.30$M & 18.1 & 31.5 & 88.0 & 23.3 & 48.2 & 38.0 \\
         & $F{=}16$ & $3.39$M & 17.9 & 30.8 & 86.5 & 22.6 & 47.4 & 37.7 \\
        \bottomrule
    \end{tabular}
    }
    \vspace{0.02in}
    \resizebox{0.99\linewidth}{!}{%
    \begin{tabular}{l|l|c||c|c|c||c|c|c}
        \toprule
        \multicolumn{2}{c|}{AR} & \# params & Small & Medium & Large & Natural & Struct. & Spec. \\
        \midrule
        \multirow{4}{*}{$S{=}16$} & baseline & $2.02$M & 30.5 & 41.9 & 82.7 & 28.5 & 61.9 & 41.7 \\
         & $F{=}1$ & $0.88$M & 34.5 & 43.3 & 83.9 & 32.3 & 62.9 & 42.9 \\
         & $F{=}4$ & $1.14$M & 31.9 & 42.3 & 82.7 & 29.9 & 62.0 & 42.0 \\
         & $F{=}16$ & $2.21$M & 31.2 & 41.9 & 82.6 & 28.9 & 61.9 & 41.6 \\ \midrule
        \multirow{4}{*}{$S{=}256$} & baseline & $20.4$M & 25.7 & 32.7 & 71.6 & 23.7 & 52.1 & 35.9 \\
         & $F{=}1$ & $1.06$M & 32.3 & 33.5 & 70.5 & 29.0 & 49.1 & 36.4 \\
         & $F{=}4$ & $1.88$M & 31.2 & 41.9 & 82.6 & 28.9 & 61.9 & 41.6 \\
         & $F{=}16$ & $5.16$M & 26.6 & 32.6 & 69.9 & 24.5 & 48.9 & 34.6 \\
        \bottomrule
    \end{tabular}
    }
    \caption{Ablation on prompt token generators for (top) NAR and (bottom) AR transformers on VTAB. We report FIDs averaged by different categorizations of tasks.}
    \label{tab:ablation_token_generator}
\end{table}

\begin{figure*}[t]
    \centering
    \begin{subfigure}[b]{0.28\linewidth}
        \centering
        \includegraphics[width=\linewidth]{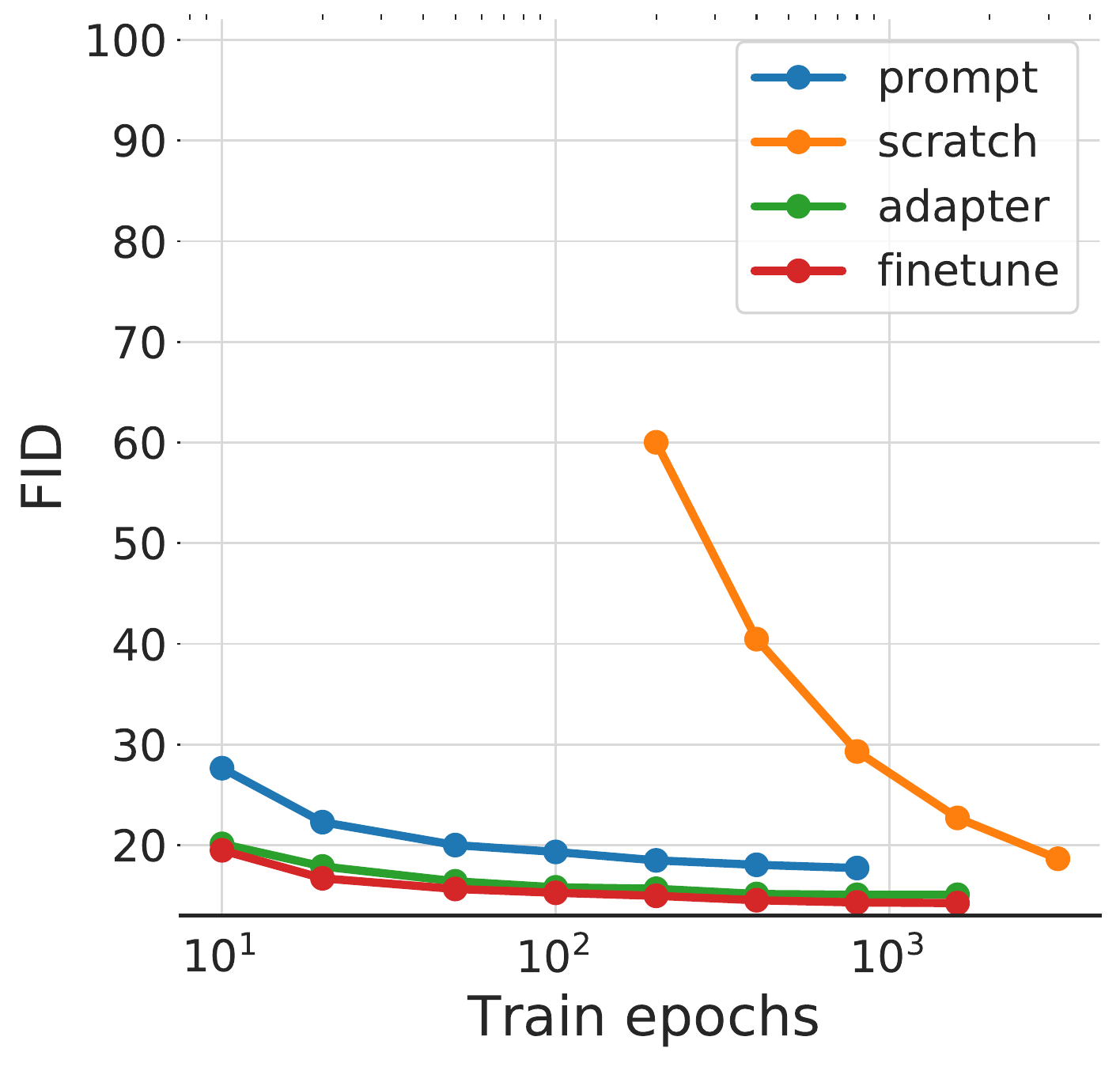}
        \caption{VTAB small (${<}10k$)}
        \label{fig:vtab_method_comparison_small}
    \end{subfigure}
    \hspace{0.24in}
    \begin{subfigure}[b]{0.28\linewidth}
        \centering
        \includegraphics[width=\linewidth]{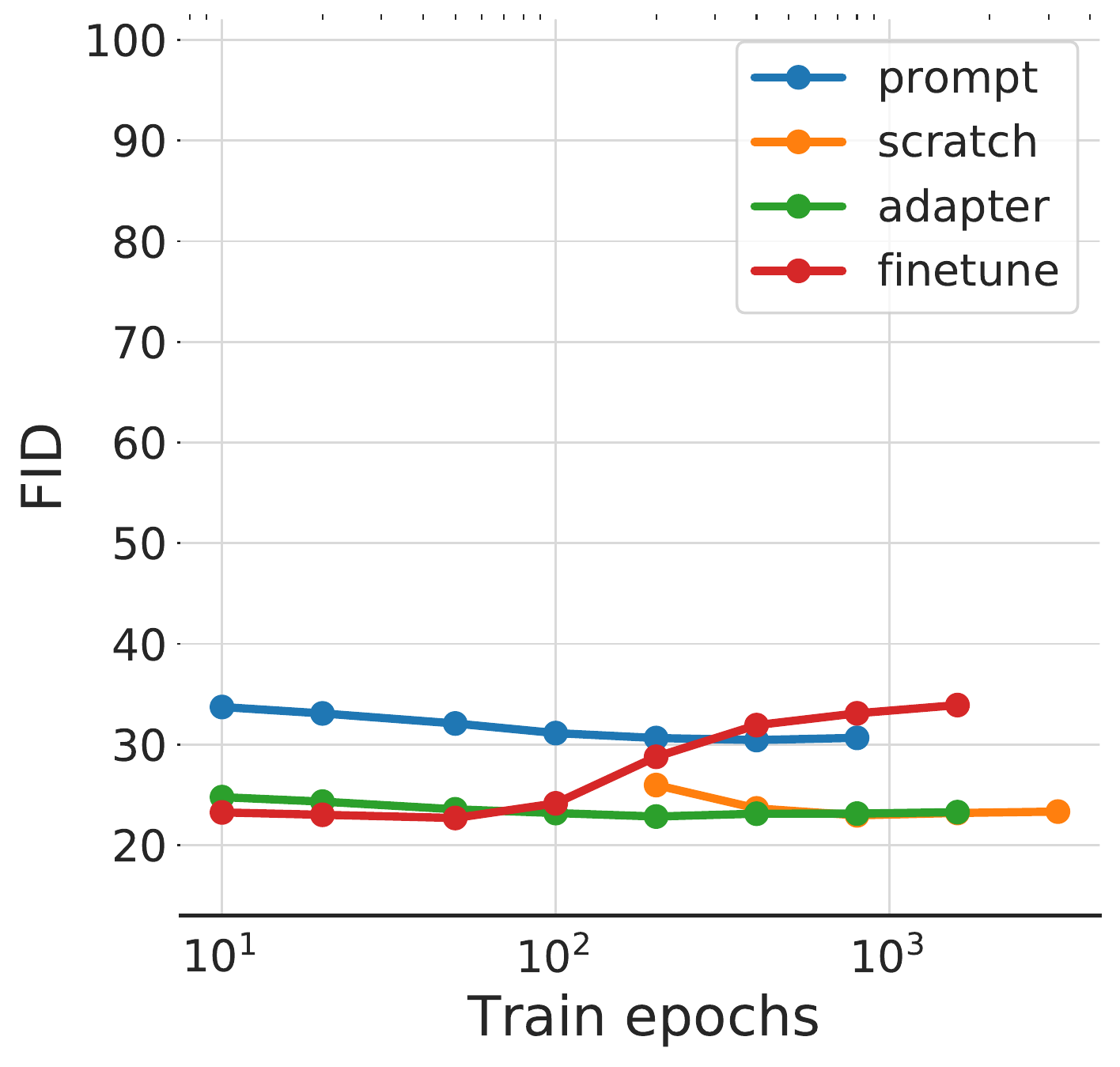}
        \caption{VTAB medium (${<}100k$)}
        \label{fig:vtab_method_comparison_medium}
    \end{subfigure}
    \hspace{0.24in}
    \begin{subfigure}[b]{0.28\linewidth}
        \centering
        \includegraphics[width=\linewidth]{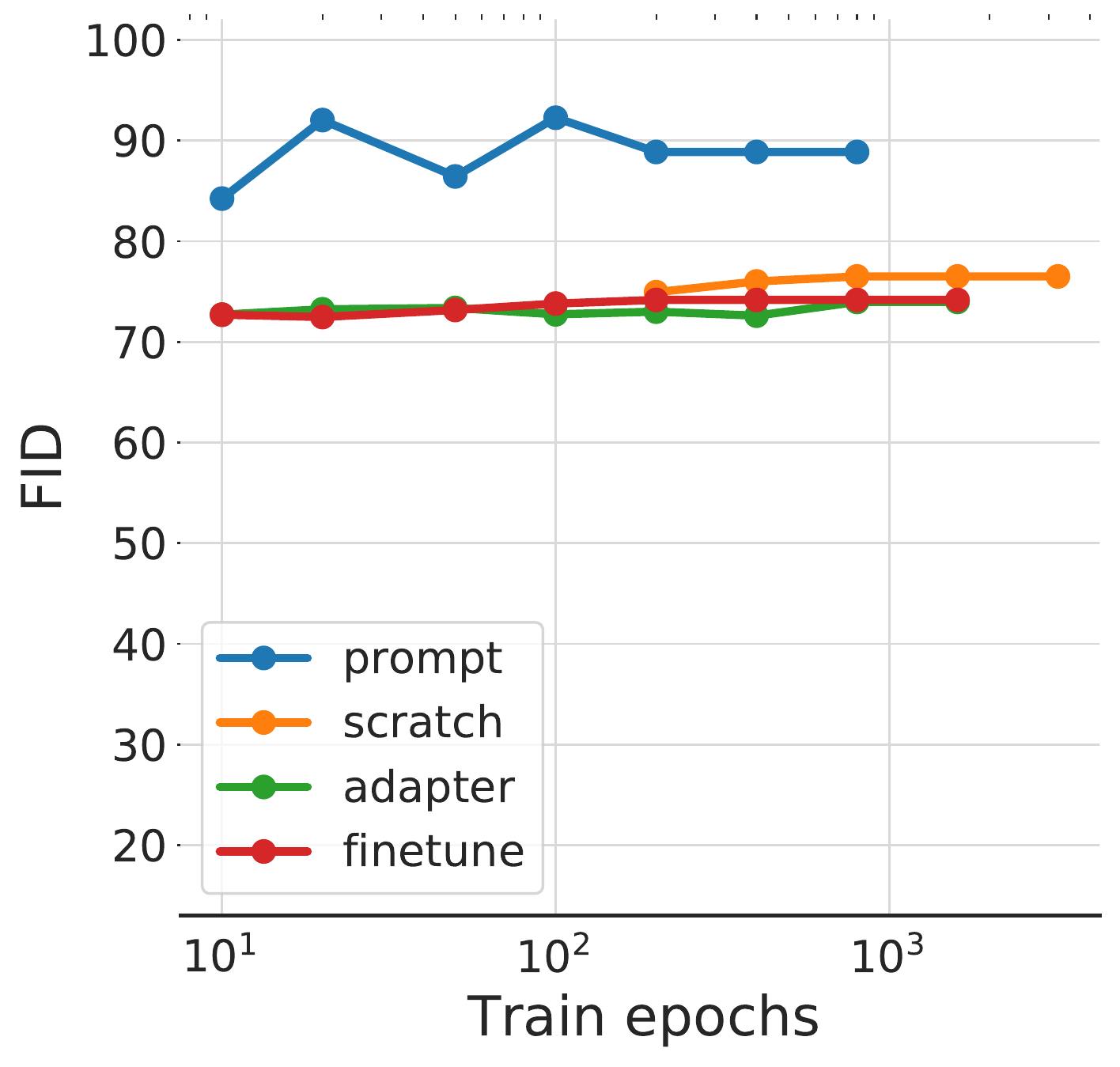}
        \caption{VTAB large (${>}100k$)}
        \label{fig:vtab_method_comparison_large}
    \end{subfigure}
    \caption{FID vs the number of train epochs for various learning methods for transformer-based sequence models. Knowledge transfer is essential for faster convergence when training data is small.}
    \label{fig:vtab_method_comparison}
    \vspace{-0.15in}
\end{figure*}

\subsection{Ablation on Prompt Token Generators}
\label{sec:abl_nonfac_prompt}

One of our technical novelties is the parameter-efficient design of the prompt token generator as in \cref{fig:prompt_design_proposed}. 
We provide in-depth study on different prompt token generators.

\cref{tab:ablation_token_generator} summarizes results. The key takeaway is that the performance, measured in FIDs, for models using prompts with the proposed factorization closely matches those using the baseline, non-factorized prompts. This is particularly true for NAR transformers. On the other hand, AR transformers still prefers prompt generators with more parameters. Nevertheless, we achieve on par results with the baseline using less than 30\% of parameters.

\subsection{Beyond Prompt Tuning for Generative Transfer}
\label{sec:abl_beyond_prompt_tunig}

We have studied applying a prompt tuning to learn generative vision transformers via knowledge transfer. We have seen promising results, \eg, excelling state-of-the-art GAN-based transfer learning methods at generative modeling. We also demonstrate the importance of knowledge transfer for fast and efficient learning of generative models from small training data. Despite the success, prompt tuning is not the only method for learning transformer-based sequence models. For the completeness, we conduct an extended study for various learning methods of generative vision transformers. 

To that end, we evaluate adapter tuning and fine-tuning in addition to the prompt tuning and learning from scratch. Adapter tuning~\cite{houlsby2019parameter} introduces learnable adapter modules to each transformer block. Fine-tuning unfreezes pretrained weights and updates them. All models are trained using the same loss (\eg, masked visual token model loss~\cite{chang2022maskgit} for NAR transformer). As we are interested in class-conditional generative models, we also introduce class-conditional prompts of length $1$ that are randomly initialized for adapter tuning and fine-tuning.

For experiments, we vary the number of training epochs from 10 to 3200,\footnote{We limit the maximum number of training steps to $500K$ to finish model training within a reasonable time window.} as training efficiency is one of the key differentiating factors across various learning strategies. 
For prompt tuning, we use $128$ prompt tokens with a single factor. For adapter tuning, we use $64$ hidden units for adapter modules.
We report the number of trainable parameters assuming $100$ classes, train time per step and generation time comparisons in \cref{tab:method_comparison}. Prompt tuning shows the best parameter and train time efficiency, where the number of trainable parameters is less than 0.5\% of those of fine-tuning and learning from scratch. On the other hand, due to the longer sequence, it takes more time for generation than those models with a single class token. 
Adapter tuning, together with a tunable class-conditional prompt, turns out to be a method with a good balance, with relatively few trainable parameters and efficiency at both train and test time.

\cref{fig:vtab_method_comparison} compares the generation performance in FID on VTAB. We see that models with a knowledge transfer converge faster than the ones without a transfer. For example, it requires almost 800 epochs for models learned from scratch to reach FIDs of the prompt tuning models trained for 10 epochs for tasks with a small data. Fine-tuning also adapts to new data distributions quickly, though it takes more time per step for model training. As in \cref{fig:vtab_method_comparison_small}, for tasks with a small training data, fine-tuning shows the best FIDs. On the other hand, we find that fine-tuning behaves unstable on some datasets (\eg, smallnorb), and the performance diverges as training goes, as in \cref{fig:vtab_method_comparison_medium}. Complete FID results are in \cref{tab:vtab_supp_eval_on_all} of Appendix. To our surprise, learning from scratch performs well even for tasks with a small training data when given sufficient compute resources and time.

Finally, we'd like to note that there is no single method that wins against the rest as each method has its own advantage. For example, for applications where the small number of parameter is critical, prompt tuning should be preferred despite slightly worse generation quality. Also, prompt and adapter tuning are preferred when there are many datasets and tasks as transformer parameters are shared across tasks.

\begin{table}[t]
    \centering
    \resizebox{0.99\linewidth}{!}{%
    \begin{tabular}{l|c|c|c}
        \toprule
         & \# params & train / step & generation \\
        \midrule
        Prompt tuning ($S\,{=}\,128$) & $0.76$M & $1{\times}$ & $1{\times}$ \\
        Adapter tuning & $5.43$M & $1.04{\times}$ & $0.84{\times}$\\
        Fine-tuning, Scratch & $172$M & $1.67{\times}$ & $0.80{\times}$\\
        \bottomrule
    \end{tabular}
    }
    \caption{Qualitative comparison (\eg, number of trainable parameters, train and generation time) among various learning strategies based on NAR transformers.}
    \label{tab:method_comparison}
\end{table}

\section{Related Work}
\label{sec:related_wor}

Transfer learning~\cite{pan2009survey,weiss2016survey,tan2018survey,zhuang2020comprehensive} is a method for improving the performance of downstream tasks using knowledge from the source domain and task. It is shown to be particularly effective when the amount of training data is limited for downstream tasks. Knowledge transfer of deep neural networks has been realized in various forms, such as linear probing~\cite{chen2018closer,henaff2020data}, fine-tuning~\cite{raffel2019exploring,kolesnikov2020big}, or adapter~\cite{rebuffi2017learning,rebuffi2018efficient,houlsby2019parameter}.
Recently, prompt tuning~\cite{li2021prefix,liu2021p,lester2021power} has emerged as a powerful tool for transfer learning of transformer-based large language models in NLP. Since the introduction of Vision Transformer~\cite{dosovitskiy2020image}, such approach has been studied for vision tasks as well~\cite{jia2022visual,bahng2022visual}. While previous works have shown effectiveness of prompt tuning for discriminative tasks (e.g., classification), we apply the technique for image synthesis.

Generative models have been extensively studied for image synthesis, including variational autoencoder~\cite{kingma2013auto,van2017neural,vahdat2020nvae}, diffusion~\cite{dhariwal2021diffusion,rombach2022high} and autoregressive~\cite{van2016pixel,van2016conditional,parmar2018image} models. A large volume of progress has been made around the generative adversarial network (GAN)~\cite{goodfellow2014generative} thanks to its ability at synthesizing high-fidelity images~\cite{brock2018large,karras2019style,karras2020analyzing,sauer2022stylegan}. As such, generative knowledge transfer has been studied to transfer knowledge of pretrained GAN models.
TransferGAN~\cite{wang2018transferring}, following a usual practice by fine-tune on the target dataset, has demonstrated transferring knowledge from pretraining improves the performance when training with limited data. Freezing a few layers of discriminator~\cite{mo2020freeze} further improves while stabilizing the training process.
MineGAN~\cite{wang2020minegan} introduces a miner, which projects random noise into the embedding space of the pretrained generator, and trains it with discriminator while fixing generator parameters. cGANTransfer~\cite{shahbazi2021efficient} makes explicit transfer of knowledge on classes of the source dataset to new classes.
Albeit showing improvement, these methods still require careful training (e.g., early stopping) and have evaluated on a few datasets. In our work, we extensively test methods on a wide variety of visual domains (e.g., VTAB) and show improvement by a large margin over existing GAN-based generative transfer methods.

\section{Conclusion}
\label{sec:conclusion}

We present a method for learning image generation models from diverse data distributions and varying amount of training data via knowledge transfer from the source model trained on a large dataset. A simple modification on prompt token designs allows to learn a parameter and compute efficient class and instance conditional image generation models of autoregressive and non-autoregressive vision transformers. Comprehensive experimental results of image synthesis are provided across diverse visual domains, tasks, and the number of training images. In addition, we show how to use learned prompts for novel image synthesis in the form of marquee header prompts, which is particularly useful when synthesizing images using generative models learned from a few images.

\section*{Acknowledgment}
\noindent We thank Brian Lester for helpful discussion on prompt tuning, Boqing Gong and David Salesin for their feedback on the manuscript.

\newpage
{\small
\bibliographystyle{ieee_fullname}
\bibliography{reference}

\begin{thebibliography}{10}\itemsep=-1pt

\bibitem{bahng2022visual}
Hyojin Bahng, Ali Jahanian, Swami Sankaranarayanan, and Phillip Isola.
\newblock Visual prompting: Modifying pixel space to adapt pre-trained models.
\newblock {\em arXiv preprint arXiv:2203.17274}, 2022.

\bibitem{brock2018large}
Andrew Brock, Jeff Donahue, and Karen Simonyan.
\newblock Large scale gan training for high fidelity natural image synthesis.
\newblock In {\em International Conference on Learning Representations}, 2018.

\bibitem{gpt3}
Tom Brown, Benjamin Mann, Nick Ryder, Melanie Subbiah, Jared~D Kaplan, Prafulla
  Dhariwal, Arvind Neelakantan, Pranav Shyam, Girish Sastry, Amanda Askell,
  Sandhini Agarwal, Ariel Herbert-Voss, Gretchen Krueger, Tom Henighan, Rewon
  Child, Aditya Ramesh, Daniel Ziegler, Jeffrey Wu, Clemens Winter, Chris
  Hesse, Mark Chen, Eric Sigler, Mateusz Litwin, Scott Gray, Benjamin Chess,
  Jack Clark, Christopher Berner, Sam McCandlish, Alec Radford, Ilya Sutskever,
  and Dario Amodei.
\newblock Language models are few-shot learners.
\newblock In {\em NeurIPS}, 2020.

\bibitem{brown2020language}
Tom Brown, Benjamin Mann, Nick Ryder, Melanie Subbiah, Jared~D Kaplan, Prafulla
  Dhariwal, Arvind Neelakantan, Pranav Shyam, Girish Sastry, Amanda Askell,
  et~al.
\newblock Language models are few-shot learners.
\newblock {\em Advances in neural information processing systems},
  33:1877--1901, 2020.

\bibitem{casanova2021instance}
Arantxa Casanova, Marlene Careil, Jakob Verbeek, Michal Drozdzal, and Adriana
  Romero~Soriano.
\newblock Instance-conditioned gan.
\newblock {\em Advances in Neural Information Processing Systems},
  34:27517--27529, 2021.

\bibitem{chang2022maskgit}
Huiwen Chang, Han Zhang, Lu Jiang, Ce Liu, and William~T Freeman.
\newblock Maskgit: Masked generative image transformer.
\newblock {\em arXiv preprint arXiv:2202.04200}, 2022.

\bibitem{chen2020imagegpt}
Mark Chen, Alec Radford, Rewon Child, Jeffrey Wu, Heewoo Jun, David Luan, and
  Ilya Sutskever.
\newblock Generative pretraining from pixels.
\newblock In {\em International Conference on Machine Learning}, pages
  1691--1703. PMLR, 2020.

\bibitem{chen2018closer}
Wei-Yu Chen, Yen-Cheng Liu, Zsolt Kira, Yu-Chiang~Frank Wang, and Jia-Bin
  Huang.
\newblock A closer look at few-shot classification.
\newblock In {\em International Conference on Learning Representations}, 2018.

\bibitem{deng2009imagenet}
Jia Deng, Wei Dong, Richard Socher, Li-Jia Li, Kai Li, and Li Fei-Fei.
\newblock Imagenet: A large-scale hierarchical image database.
\newblock In {\em 2009 IEEE conference on computer vision and pattern
  recognition}, pages 248--255. Ieee, 2009.

\bibitem{devlin2018bert}
Jacob Devlin, Ming-Wei Chang, Kenton Lee, and Kristina Toutanova.
\newblock Bert: Pre-training of deep bidirectional transformers for language
  understanding.
\newblock {\em arXiv preprint arXiv:1810.04805}, 2018.

\bibitem{dhariwal2021diffusion}
Prafulla Dhariwal and Alexander Nichol.
\newblock Diffusion models beat gans on image synthesis.
\newblock {\em Advances in Neural Information Processing Systems}, 34, 2021.

\bibitem{ding2021cogview}
Ming Ding, Zhuoyi Yang, Wenyi Hong, Wendi Zheng, Chang Zhou, Da Yin, Junyang
  Lin, Xu Zou, Zhou Shao, Hongxia Yang, et~al.
\newblock Cogview: Mastering text-to-image generation via transformers.
\newblock {\em Advances in Neural Information Processing Systems},
  34:19822--19835, 2021.

\bibitem{dosovitskiy2020image}
Alexey Dosovitskiy, Lucas Beyer, Alexander Kolesnikov, Dirk Weissenborn,
  Xiaohua Zhai, Thomas Unterthiner, Mostafa Dehghani, Matthias Minderer, Georg
  Heigold, Sylvain Gelly, et~al.
\newblock An image is worth 16x16 words: Transformers for image recognition at
  scale.
\newblock {\em arXiv preprint arXiv:2010.11929}, 2020.

\bibitem{esser2021taming}
Patrick Esser, Robin Rombach, and Bjorn Ommer.
\newblock Taming transformers for high-resolution image synthesis.
\newblock In {\em Proceedings of the IEEE/CVF Conference on Computer Vision and
  Pattern Recognition}, pages 12873--12883, 2021.

\bibitem{ghazvininejad2019mask}
Marjan Ghazvininejad, Omer Levy, Yinhan Liu, and Luke Zettlemoyer.
\newblock Mask-predict: Parallel decoding of conditional masked language
  models.
\newblock In {\em EMNLP-IJCNLP}, 2019.

\bibitem{girshick2015fast}
Ross Girshick.
\newblock Fast r-cnn.
\newblock In {\em Proceedings of the IEEE international conference on computer
  vision}, pages 1440--1448, 2015.

\bibitem{girshick2014rich}
Ross Girshick, Jeff Donahue, Trevor Darrell, and Jitendra Malik.
\newblock Rich feature hierarchies for accurate object detection and semantic
  segmentation.
\newblock In {\em Proceedings of the IEEE conference on computer vision and
  pattern recognition}, pages 580--587, 2014.

\bibitem{goodfellow2014generative}
Ian Goodfellow, Jean Pouget-Abadie, Mehdi Mirza, Bing Xu, David Warde-Farley,
  Sherjil Ozair, Aaron Courville, and Yoshua Bengio.
\newblock Generative adversarial nets.
\newblock {\em Advances in neural information processing systems}, 27, 2014.

\bibitem{gu-kong-2021-fully}
Jiatao Gu and Xiang Kong.
\newblock Fully non-autoregressive neural machine translation: Tricks of the
  trade.
\newblock In {\em Findings of ACL-IJCNLP}, 2021.

\bibitem{he2020momentum}
Kaiming He, Haoqi Fan, Yuxin Wu, Saining Xie, and Ross Girshick.
\newblock Momentum contrast for unsupervised visual representation learning.
\newblock In {\em Proceedings of the IEEE/CVF conference on computer vision and
  pattern recognition}, pages 9729--9738, 2020.

\bibitem{he2016deep}
Kaiming He, Xiangyu Zhang, Shaoqing Ren, and Jian Sun.
\newblock Deep residual learning for image recognition.
\newblock In {\em Proceedings of the IEEE conference on computer vision and
  pattern recognition}, pages 770--778, 2016.

\bibitem{flax2020github}
Jonathan Heek, Anselm Levskaya, Avital Oliver, Marvin Ritter, Bertrand
  Rondepierre, Andreas Steiner, and Marc van {Z}ee.
\newblock {F}lax: A neural network library and ecosystem for {JAX}, 2020.

\bibitem{henaff2020data}
Olivier Henaff.
\newblock Data-efficient image recognition with contrastive predictive coding.
\newblock In {\em International conference on machine learning}, pages
  4182--4192. PMLR, 2020.

\bibitem{heusel2017gans}
Martin Heusel, Hubert Ramsauer, Thomas Unterthiner, Bernhard Nessler, and Sepp
  Hochreiter.
\newblock Gans trained by a two time-scale update rule converge to a local nash
  equilibrium.
\newblock {\em Advances in neural information processing systems}, 30, 2017.

\bibitem{houlsby2019parameter}
Neil Houlsby, Andrei Giurgiu, Stanislaw Jastrzebski, Bruna Morrone, Quentin
  De~Laroussilhe, Andrea Gesmundo, Mona Attariyan, and Sylvain Gelly.
\newblock Parameter-efficient transfer learning for nlp.
\newblock In {\em International Conference on Machine Learning}, pages
  2790--2799. PMLR, 2019.

\bibitem{jia2022visual}
Menglin Jia, Luming Tang, Bor-Chun Chen, Claire Cardie, Serge Belongie, Bharath
  Hariharan, and Ser-Nam Lim.
\newblock Visual prompt tuning.
\newblock {\em arXiv preprint arXiv:2203.12119}, 2022.

\bibitem{karras2019style}
Tero Karras, Samuli Laine, and Timo Aila.
\newblock A style-based generator architecture for generative adversarial
  networks.
\newblock In {\em Proceedings of the IEEE/CVF conference on computer vision and
  pattern recognition}, pages 4401--4410, 2019.

\bibitem{karras2020analyzing}
Tero Karras, Samuli Laine, Miika Aittala, Janne Hellsten, Jaakko Lehtinen, and
  Timo Aila.
\newblock Analyzing and improving the image quality of stylegan.
\newblock In {\em Proceedings of the IEEE/CVF conference on computer vision and
  pattern recognition}, pages 8110--8119, 2020.

\bibitem{kingma2014adam}
Diederik~P Kingma and Jimmy Ba.
\newblock Adam: A method for stochastic optimization.
\newblock {\em arXiv preprint arXiv:1412.6980}, 2014.

\bibitem{kingma2013auto}
Diederik~P Kingma and Max Welling.
\newblock Auto-encoding variational bayes.
\newblock {\em arXiv preprint arXiv:1312.6114}, 2013.

\bibitem{kolesnikov2020big}
Alexander Kolesnikov, Lucas Beyer, Xiaohua Zhai, Joan Puigcerver, Jessica Yung,
  Sylvain Gelly, and Neil Houlsby.
\newblock Big transfer (bit): General visual representation learning.
\newblock In {\em European conference on computer vision}, pages 491--507.
  Springer, 2020.

\bibitem{kong2021blt}
Xiang Kong, Lu Jiang, Huiwen Chang, Han Zhang, Yuan Hao, Haifeng Gong, and
  Irfan Essa.
\newblock Blt: Bidirectional layout transformer for controllable layout
  generation.
\newblock {\em arXiv preprint arXiv:2112.05112}, 2021.

\bibitem{kong2020incorporating}
Xiang Kong, Zhisong Zhang, and Eduard Hovy.
\newblock Incorporating a local translation mechanism into non-autoregressive
  translation.
\newblock {\em arXiv preprint arXiv:2011.06132}, 2020.

\bibitem{lester2021power}
Brian Lester, Rami Al-Rfou, and Noah Constant.
\newblock The power of scale for parameter-efficient prompt tuning.
\newblock In {\em Proceedings of the 2021 Conference on Empirical Methods in
  Natural Language Processing}, pages 3045--3059, 2021.

\bibitem{lezama2022improved}
Jose Lezama, Huiwen Chang, Lu Jiang, and Irfan Essa.
\newblock Improved masked image generation with token-critic.
\newblock In {\em ECCV}, 2022.

\bibitem{li2021prefix}
Xiang~Lisa Li and Percy Liang.
\newblock Prefix-tuning: Optimizing continuous prompts for generation.
\newblock {\em arXiv preprint arXiv:2101.00190}, 2021.

\bibitem{liu2021p}
Xiao Liu, Kaixuan Ji, Yicheng Fu, Zhengxiao Du, Zhilin Yang, and Jie Tang.
\newblock P-tuning v2: Prompt tuning can be comparable to fine-tuning
  universally across scales and tasks.
\newblock {\em arXiv preprint arXiv:2110.07602}, 2021.

\bibitem{loshchilov2016sgdr}
Ilya Loshchilov and Frank Hutter.
\newblock Sgdr: Stochastic gradient descent with warm restarts.
\newblock {\em arXiv preprint arXiv:1608.03983}, 2016.

\bibitem{mikolov2010recurrent}
Tomas Mikolov, Martin Karafi{\'a}t, Lukas Burget, Jan Cernock{\`y}, and Sanjeev
  Khudanpur.
\newblock Recurrent neural network based language model.
\newblock In {\em Interspeech}, volume~2, pages 1045--1048. Makuhari, 2010.

\bibitem{mo2020freeze}
Sangwoo Mo, Minsu Cho, and Jinwoo Shin.
\newblock Freeze the discriminator: a simple baseline for fine-tuning gans.
\newblock {\em arXiv preprint arXiv:2002.10964}, 2020.

\bibitem{nie2022protuning}
Xing Nie, Bolin Ni, Jianlong Chang, Gaomeng Meng, Chunlei Huo, Zhaoxiang Zhang,
  Shiming Xiang, Qi Tian, and Chunhong Pan.
\newblock Pro-tuning: Unified prompt tuning for vision tasks.
\newblock {\em arXiv preprint arXiv:2207.14381}, 2022.

\bibitem{ojha2021few}
Utkarsh Ojha, Yijun Li, Jingwan Lu, Alexei~A Efros, Yong~Jae Lee, Eli
  Shechtman, and Richard Zhang.
\newblock Few-shot image generation via cross-domain correspondence.
\newblock In {\em Proceedings of the IEEE/CVF Conference on Computer Vision and
  Pattern Recognition}, pages 10743--10752, 2021.

\bibitem{pan2009survey}
Sinno~Jialin Pan and Qiang Yang.
\newblock A survey on transfer learning.
\newblock {\em IEEE Transactions on knowledge and data engineering},
  22(10):1345--1359, 2009.

\bibitem{parmar2018image}
Niki Parmar, Ashish Vaswani, Jakob Uszkoreit, Lukasz Kaiser, Noam Shazeer,
  Alexander Ku, and Dustin Tran.
\newblock Image transformer.
\newblock In {\em International conference on machine learning}, pages
  4055--4064. PMLR, 2018.

\bibitem{peng2019moment}
Xingchao Peng, Qinxun Bai, Xide Xia, Zijun Huang, Kate Saenko, and Bo Wang.
\newblock Moment matching for multi-source domain adaptation.
\newblock In {\em Proceedings of the IEEE/CVF international conference on
  computer vision}, pages 1406--1415, 2019.

\bibitem{raffel2019exploring}
Colin Raffel, Noam Shazeer, Adam Roberts, Katherine Lee, Sharan Narang, Michael
  Matena, Yanqi Zhou, Wei Li, and Peter~J Liu.
\newblock Exploring the limits of transfer learning with a unified text-to-text
  transformer.
\newblock {\em arXiv preprint arXiv:1910.10683}, 2019.

\bibitem{Ramesh21dalle}
Aditya Ramesh, Mikhail Pavlov, Gabriel Goh, Scott Gray, Chelsea Voss, Alec
  Radford, Mark Chen, and Ilya Sutskever.
\newblock Zero-shot text-to-image generation.
\newblock In Marina Meila and Tong Zhang, editors, {\em {ICML}}, 2021.

\bibitem{razavi2019generating}
Ali Razavi, Aaron Van~den Oord, and Oriol Vinyals.
\newblock Generating diverse high-fidelity images with vq-vae-2.
\newblock {\em Advances in neural information processing systems}, 32, 2019.

\bibitem{rebuffi2017learning}
Sylvestre-Alvise Rebuffi, Hakan Bilen, and Andrea Vedaldi.
\newblock Learning multiple visual domains with residual adapters.
\newblock {\em Advances in neural information processing systems}, 30, 2017.

\bibitem{rebuffi2018efficient}
Sylvestre-Alvise Rebuffi, Hakan Bilen, and Andrea Vedaldi.
\newblock Efficient parametrization of multi-domain deep neural networks.
\newblock In {\em Proceedings of the IEEE Conference on Computer Vision and
  Pattern Recognition}, pages 8119--8127, 2018.

\bibitem{rombach2022high}
Robin Rombach, Andreas Blattmann, Dominik Lorenz, Patrick Esser, and Bj{\"o}rn
  Ommer.
\newblock High-resolution image synthesis with latent diffusion models.
\newblock In {\em Proceedings of the IEEE/CVF Conference on Computer Vision and
  Pattern Recognition}, pages 10684--10695, 2022.

\bibitem{sauer2022stylegan}
Axel Sauer, Katja Schwarz, and Andreas Geiger.
\newblock Stylegan-xl: Scaling stylegan to large diverse datasets.
\newblock {\em arXiv preprint arXiv:2202.00273}, 1, 2022.

\bibitem{shahbazi2021efficient}
Mohamad Shahbazi, Zhiwu Huang, Danda~Pani Paudel, Ajad Chhatkuli, and Luc
  Van~Gool.
\newblock Efficient conditional gan transfer with knowledge propagation across
  classes.
\newblock In {\em Proceedings of the IEEE/CVF Conference on Computer Vision and
  Pattern Recognition}, pages 12167--12176, 2021.

\bibitem{si2011learning}
Zhangzhang Si and Song-Chun Zhu.
\newblock Learning hybrid image templates (hit) by information projection.
\newblock {\em IEEE Transactions on pattern analysis and machine intelligence},
  34(7):1354--1367, 2011.

\bibitem{tan2018survey}
Chuanqi Tan, Fuchun Sun, Tao Kong, Wenchang Zhang, Chao Yang, and Chunfang Liu.
\newblock A survey on deep transfer learning.
\newblock In {\em International conference on artificial neural networks},
  pages 270--279. Springer, 2018.

\bibitem{tseng2021regularizing}
Hung-Yu Tseng, Lu Jiang, Ce Liu, Ming-Hsuan Yang, and Weilong Yang.
\newblock Regularizing generative adversarial networks under limited data.
\newblock In {\em Proceedings of the IEEE/CVF Conference on Computer Vision and
  Pattern Recognition}, pages 7921--7931, 2021.

\bibitem{vahdat2020nvae}
Arash Vahdat and Jan Kautz.
\newblock Nvae: A deep hierarchical variational autoencoder.
\newblock {\em Advances in Neural Information Processing Systems},
  33:19667--19679, 2020.

\bibitem{van2016conditional}
Aaron Van~den Oord, Nal Kalchbrenner, Lasse Espeholt, Oriol Vinyals, Alex
  Graves, et~al.
\newblock Conditional image generation with pixelcnn decoders.
\newblock {\em Advances in neural information processing systems}, 29, 2016.

\bibitem{van2017neural}
Aaron Van Den~Oord, Oriol Vinyals, et~al.
\newblock Neural discrete representation learning.
\newblock {\em Advances in neural information processing systems}, 30, 2017.

\bibitem{Oord17vqvae}
A{\"{a}}ron van~den Oord, Oriol Vinyals, and Koray Kavukcuoglu.
\newblock Neural discrete representation learning.
\newblock In Isabelle Guyon, Ulrike von Luxburg, Samy Bengio, Hanna~M. Wallach,
  Rob Fergus, S.~V.~N. Vishwanathan, and Roman Garnett, editors, {\em
  {NeurIPS}}, 2017.

\bibitem{van2008visualizing}
Laurens Van~der Maaten and Geoffrey Hinton.
\newblock Visualizing data using t-sne.
\newblock {\em Journal of machine learning research}, 9(11), 2008.

\bibitem{van2016pixel}
Aaron Van~Oord, Nal Kalchbrenner, and Koray Kavukcuoglu.
\newblock Pixel recurrent neural networks.
\newblock In {\em International conference on machine learning}, pages
  1747--1756. PMLR, 2016.

\bibitem{wang2019learning}
Haohan Wang, Songwei Ge, Zachary Lipton, and Eric~P Xing.
\newblock Learning robust global representations by penalizing local predictive
  power.
\newblock {\em Advances in Neural Information Processing Systems}, 32, 2019.

\bibitem{wang2020minegan}
Yaxing Wang, Abel Gonzalez-Garcia, David Berga, Luis Herranz, Fahad~Shahbaz
  Khan, and Joost van~de Weijer.
\newblock Minegan: effective knowledge transfer from gans to target domains
  with few images.
\newblock In {\em Proceedings of the IEEE/CVF Conference on Computer Vision and
  Pattern Recognition}, pages 9332--9341, 2020.

\bibitem{wang2018transferring}
Yaxing Wang, Chenshen Wu, Luis Herranz, Joost van~de Weijer, Abel
  Gonzalez-Garcia, and Bogdan Raducanu.
\newblock Transferring gans: generating images from limited data.
\newblock In {\em Proceedings of the European Conference on Computer Vision
  (ECCV)}, pages 218--234, 2018.

\bibitem{weiss2016survey}
Karl Weiss, Taghi~M Khoshgoftaar, and DingDing Wang.
\newblock A survey of transfer learning.
\newblock {\em Journal of Big data}, 3(1):1--40, 2016.

\bibitem{wu2021n}
Chenfei Wu, Jian Liang, Lei Ji, Fan Yang, Yuejian Fang, Daxin Jiang, and Nan
  Duan.
\newblock N$\backslash$" uwa: Visual synthesis pre-training for neural visual
  world creation.
\newblock {\em arXiv preprint arXiv:2111.12417}, 2021.

\bibitem{yang2021one}
Ceyuan Yang, Yujun Shen, Zhiyi Zhang, Yinghao Xu, Jiapeng Zhu, Zhirong Wu, and
  Bolei Zhou.
\newblock One-shot generative domain adaptation.
\newblock {\em arXiv preprint arXiv:2111.09876}, 2021.

\bibitem{vim2021}
Jiahui Yu, Xin Li, Jing~Yu Koh, Han Zhang, Ruoming Pang, James Qin, Alexander
  Ku, Yuanzhong Xu, Jason Baldridge, and Yonghui Wu.
\newblock Vector-quantized image modeling with improved {VQGAN}.
\newblock {\em arXiv preprint arXiv:2110.04627}, 2021.

\bibitem{yu2022scaling}
Jiahui Yu, Yuanzhong Xu, Jing~Yu Koh, Thang Luong, Gunjan Baid, Zirui Wang,
  Vijay Vasudevan, Alexander Ku, Yinfei Yang, Burcu~Karagol Ayan, et~al.
\newblock Scaling autoregressive models for content-rich text-to-image
  generation.
\newblock {\em arXiv preprint arXiv:2206.10789}, 2022.

\bibitem{zhai2019large}
Xiaohua Zhai, Joan Puigcerver, Alexander Kolesnikov, Pierre Ruyssen, Carlos
  Riquelme, Mario Lucic, Josip Djolonga, Andre~Susano Pinto, Maxim Neumann,
  Alexey Dosovitskiy, et~al.
\newblock A large-scale study of representation learning with the visual task
  adaptation benchmark.
\newblock {\em arXiv preprint arXiv:1910.04867}, 2019.

\bibitem{zhang2021m6}
Zhu Zhang, Jianxin Ma, Chang Zhou, Rui Men, Zhikang Li, Ming Ding, Jie Tang,
  Jingren Zhou, and Hongxia Yang.
\newblock M6-ufc: Unifying multi-modal controls for conditional image
  synthesis.
\newblock {\em arXiv preprint arXiv:2105.14211}, 2021.

\bibitem{zhao2020differentiable}
Shengyu Zhao, Zhijian Liu, Ji Lin, Jun-Yan Zhu, and Song Han.
\newblock Differentiable augmentation for data-efficient gan training.
\newblock {\em Advances in Neural Information Processing Systems},
  33:7559--7570, 2020.

\bibitem{zhou2014learning}
Bolei Zhou, Agata Lapedriza, Jianxiong Xiao, Antonio Torralba, and Aude Oliva.
\newblock Learning deep features for scene recognition using places database.
\newblock {\em Advances in neural information processing systems}, 27, 2014.

\bibitem{zhuang2020comprehensive}
Fuzhen Zhuang, Zhiyuan Qi, Keyu Duan, Dongbo Xi, Yongchun Zhu, Hengshu Zhu, Hui
  Xiong, and Qing He.
\newblock A comprehensive survey on transfer learning.
\newblock {\em Proceedings of the IEEE}, 109(1):43--76, 2020.

\end{thebibliography}
}

\newpage
\onecolumn

\appendix

\section{Pseudo-code for Token Generator}

In \cref{fig:prompt_generator} we provide an example code that implements the prompt token generator in Flax~\cite{flax2020github} format. 

\section{Comprehensive Experiment Description}
\label{sec:supp_exp_setting}

\subsection{Visual Task Adaptation Benchmark (VTAB)}
\label{sec:supp_exp_vtab}

\subsubsection{Dataset Meta Information of Visual Task Adaptation Benchmark}
\label{sec:supp_exp_vtab_info}

In \cref{tab:vtab_supp_info} we provide a dataset meta information, including the number of class and the number of images in each data split, of VTAB.

\subsubsection{Hyperparameters}
\label{sec:supp_exp_vtab_hyper}

We provide hyperparameters used in our experiments in \cref{tab:vtab_hparams}. Note that most hyperparameters are shared across datasets, except the number of training epochs. We use Adam optimizer~\cite{kingma2014adam} with a cosine learning rate decay~\cite{loshchilov2016sgdr}. When learning models from scratch, we find that learning rate warm-up is essential. To this end, we use a warm-up for the first two epochs for AR models, and 80 train epochs for NAR transformers.

\subsubsection{Experimental Results}
\label{sec:supp_exp_vtab_fids}

We provide complete results in \cref{tab:vtab_supp_eval_on_all} for autoregressive transformers, non-autoregressive transformers as well as GAN-based generative model transfer learning methods including MineGAN~\cite{wang2020minegan} and cGANTransfer~\cite{shahbazi2021efficient}. For AR and NAR transformers, we report FIDs for prompt tuning, learning from scratch, as well as different transfer learning techniques including adapter~\cite{houlsby2019parameter} and fine-tuning~\cite{kolesnikov2020big}.
%

\subsubsection{Visualization of Generated Images}
\label{sec:supp_exp_vtab_synth}

\begin{figure}
  \centering
  \begin{subfigure}[b]{0.48\linewidth}
    \centering
    \includegraphics[width=\linewidth]{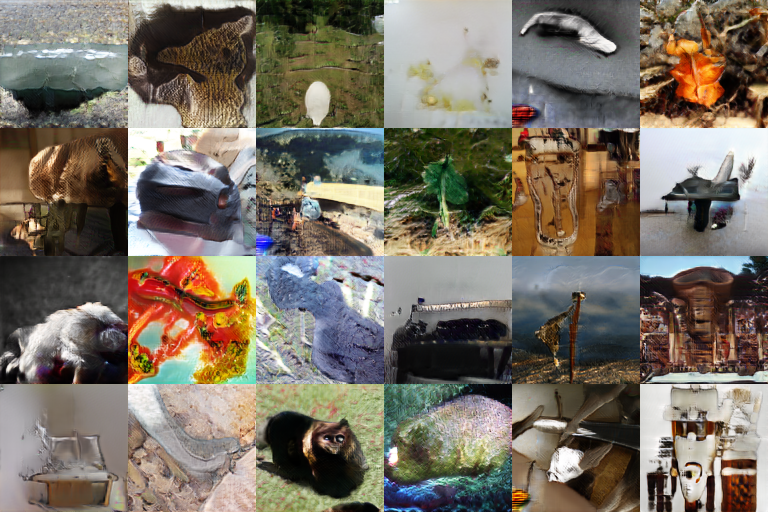}
    \caption{MineGAN}
    \label{fig:vtab_minegan_caltech101}
  \end{subfigure}
  \hspace{0.02in}
  \begin{subfigure}[b]{0.48\linewidth}
    \centering
    \includegraphics[width=\linewidth]{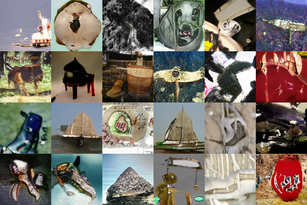}
    \caption{cGANTransfer}
    \label{fig:vtab_cgantrasnfer_caltech101}
  \end{subfigure}
  \begin{subfigure}[b]{0.48\linewidth}
    \centering
    \includegraphics[width=\linewidth]{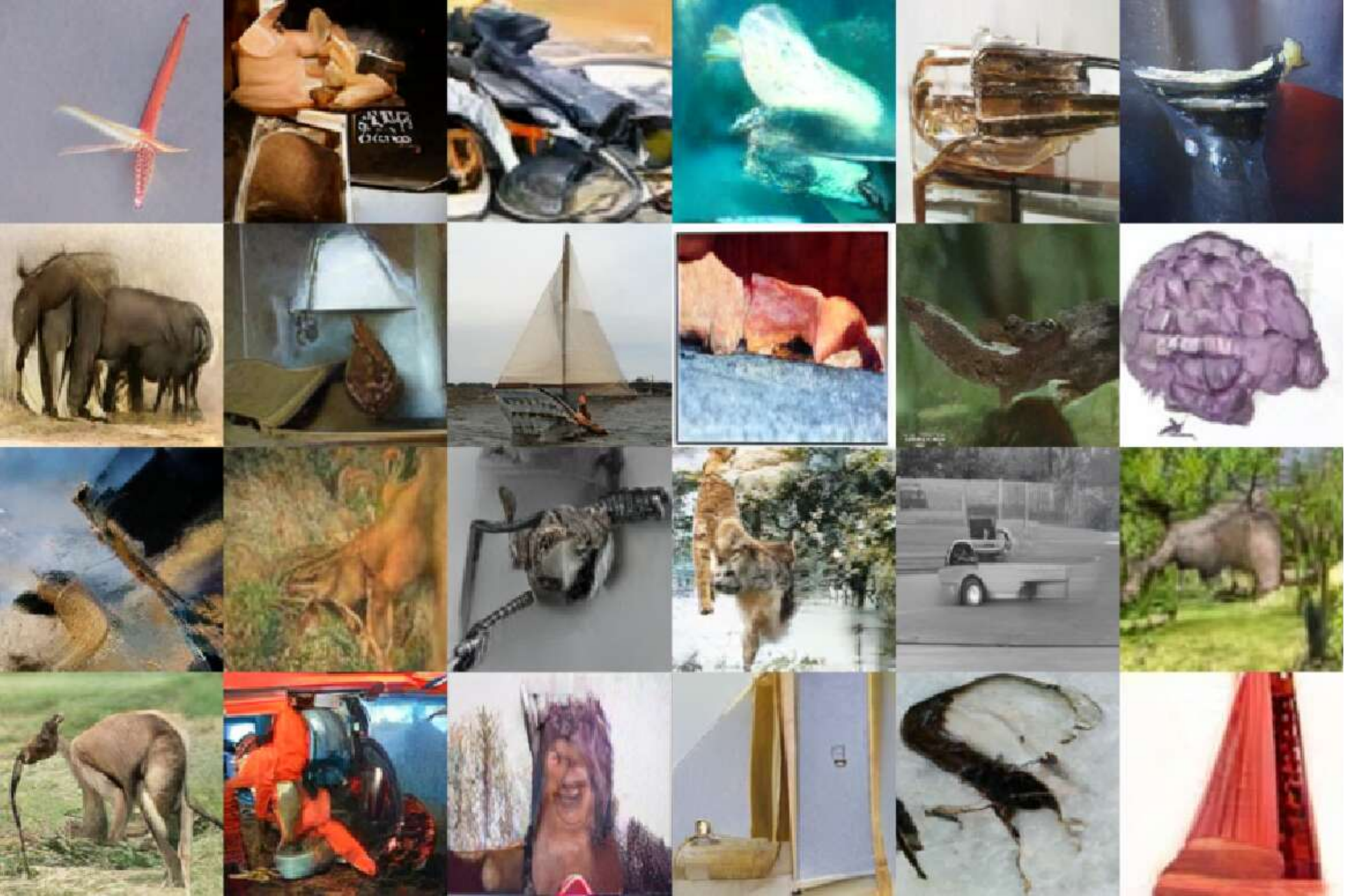}
    \caption{AR transformer with prompt tuning ($S\,{=}\,1$)}
    \label{fig:vtab_taming_prompt_s1_caltech101}
  \end{subfigure}
  \hspace{0.02in}
  \begin{subfigure}[b]{0.48\linewidth}
    \centering
    \includegraphics[width=\linewidth]{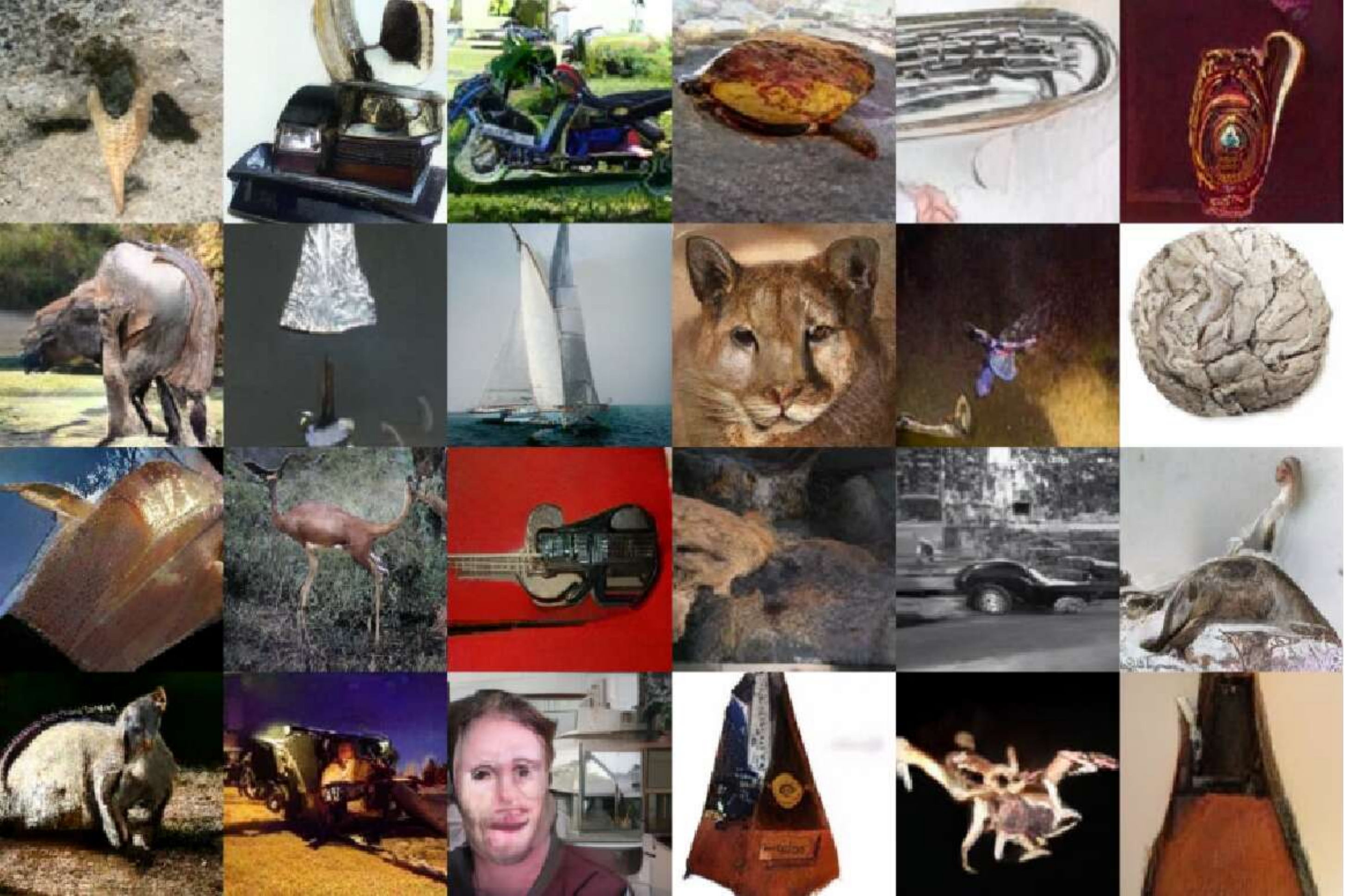}
    \caption{AR transformer with prompt tuning ($S\,{=}\,256$, $F\,{=}\,16$)}
    \label{fig:vtab_taming_prompt_s128_caltech101}
  \end{subfigure}
  \begin{subfigure}[b]{0.48\linewidth}
    \centering
    \includegraphics[width=\linewidth]{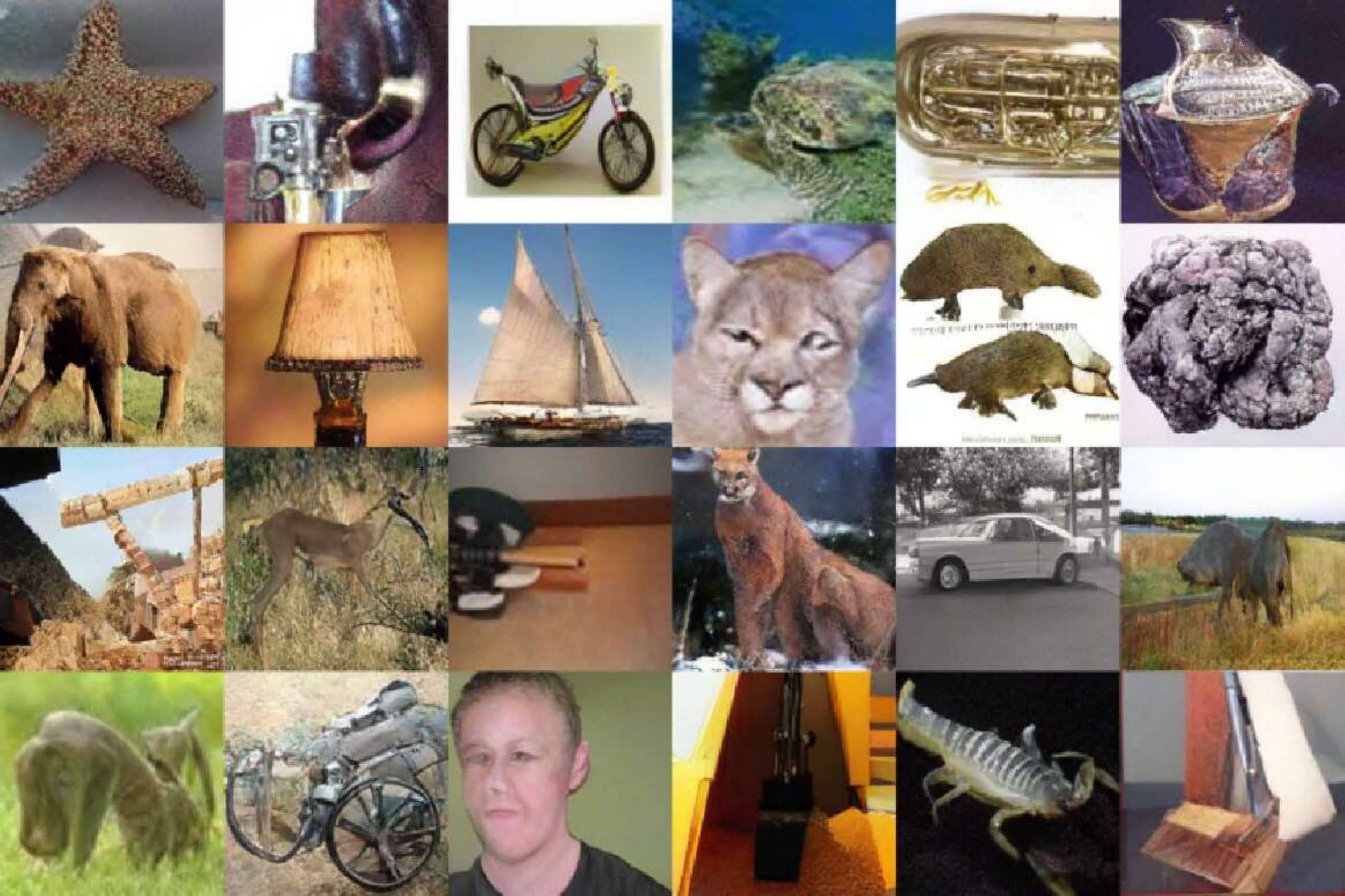}
    \caption{NAR transformer with prompt tuning ($S\,{=}\,1$)}
    \label{fig:vtab_maskgit_prompt_s1_caltech101}
  \end{subfigure}
  \hspace{0.02in}
  \begin{subfigure}[b]{0.48\linewidth}
    \centering
    \includegraphics[width=\linewidth]{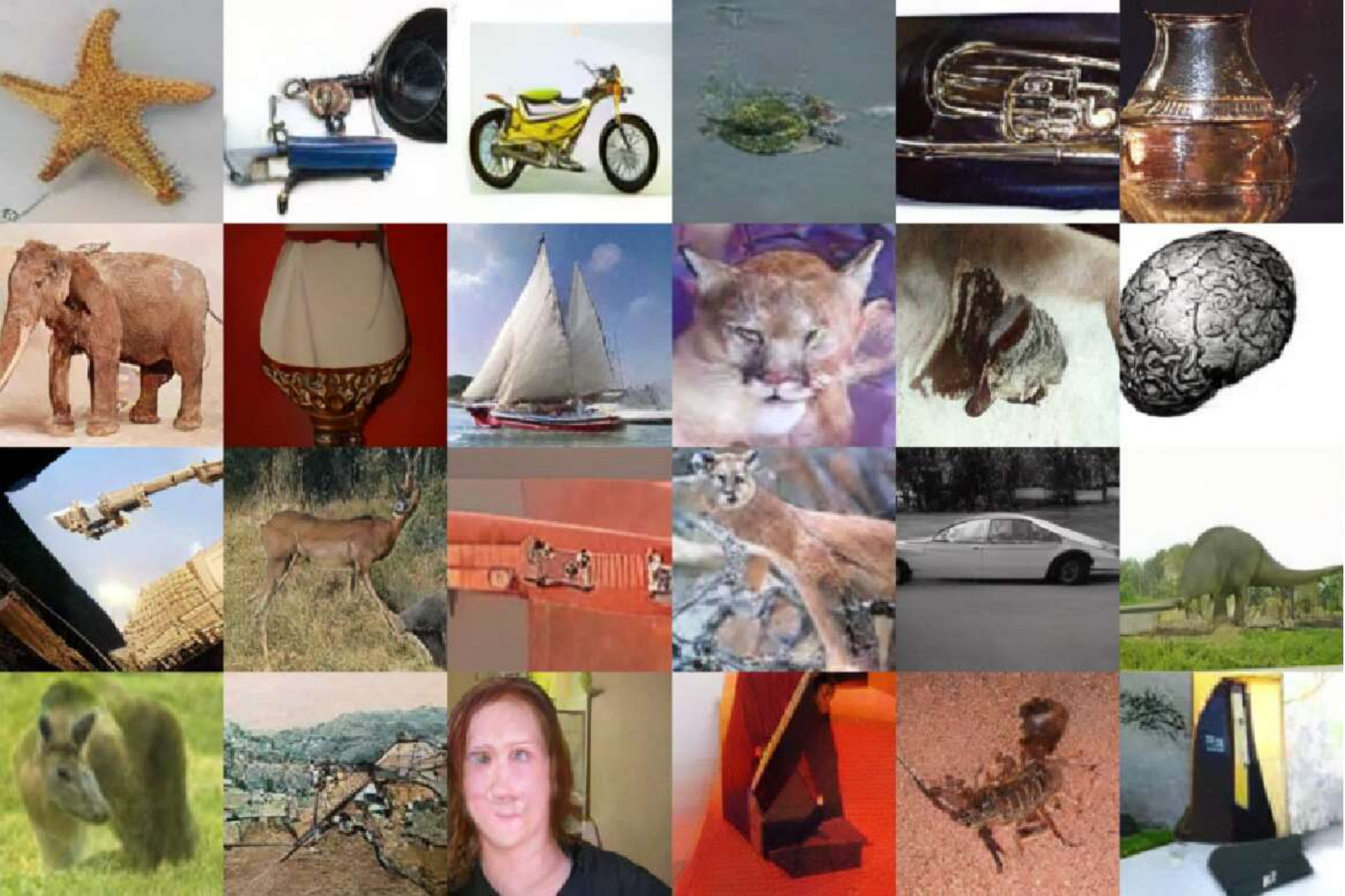}
    \caption{NAR transformer with prompt tuning ($S\,{=}\,128$)}
    \label{fig:vtab_maskgit_prompt_s128_caltech101}
  \end{subfigure}
  \caption{Visualization of generated images with different models on Caltech101 of VTAB.}
  \label{fig:vtab_supp_caltech101}
\end{figure}

\begin{figure}
  \centering
  \begin{subfigure}[b]{0.48\linewidth}
    \centering
    \includegraphics[width=\linewidth]{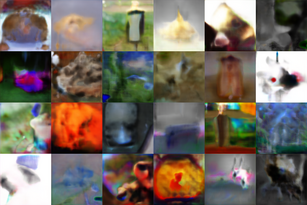}
    \caption{MineGAN}
    \label{fig:vtab_minegan_cifar}
  \end{subfigure}
  \hspace{0.02in}
  \begin{subfigure}[b]{0.48\linewidth}
    \centering
    \includegraphics[width=\linewidth]{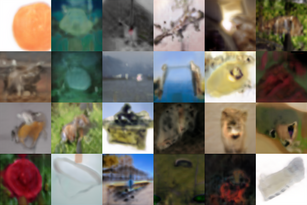}
    \caption{cGANTransfer}
    \label{fig:vtab_cgantrasnfer_cifar}
  \end{subfigure}
  \begin{subfigure}[b]{0.48\linewidth}
    \centering
    \includegraphics[width=\linewidth]{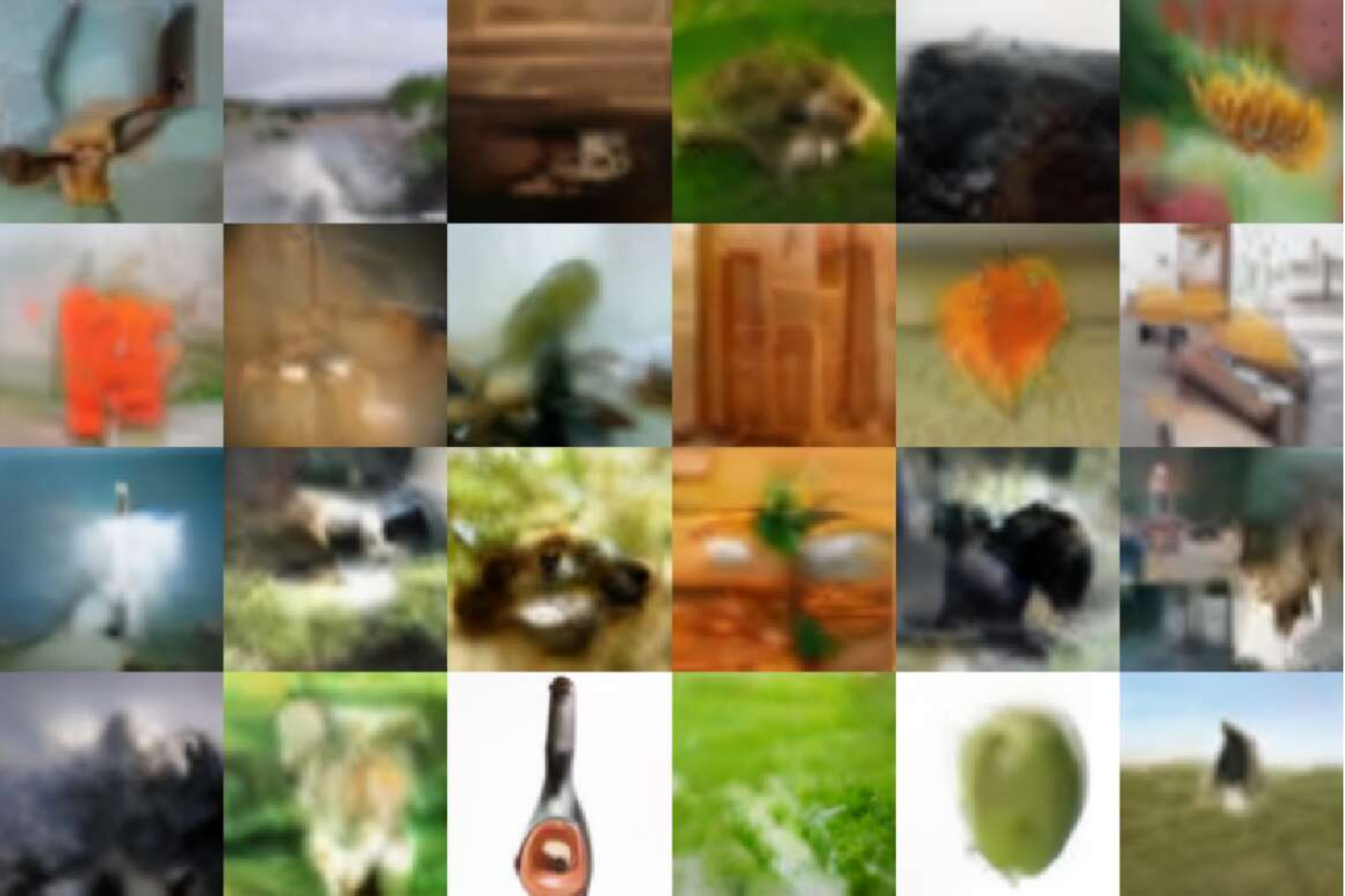}
    \caption{AR transformer with prompt tuning ($S\,{=}\,1$)}
    \label{fig:vtab_taming_prompt_s1_cifar}
  \end{subfigure}
  \hspace{0.02in}
  \begin{subfigure}[b]{0.48\linewidth}
    \centering
    \includegraphics[width=\linewidth]{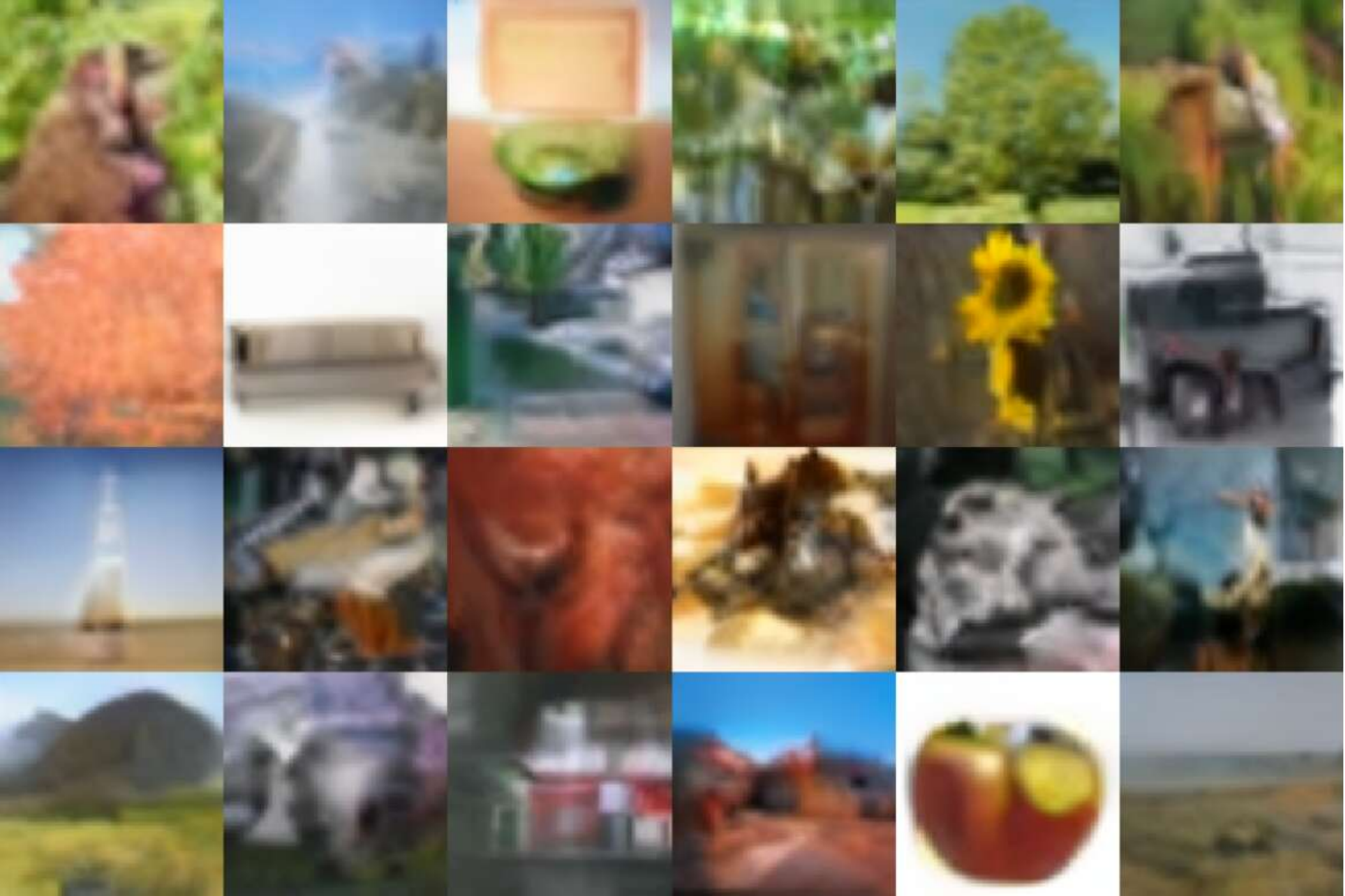}
    \caption{AR transformer with prompt tuning ($S\,{=}\,256$, $F\,{=}\,16$)}
    \label{fig:vtab_taming_prompt_s128_cifar}
  \end{subfigure}
  \begin{subfigure}[b]{0.48\linewidth}
    \centering
    \includegraphics[width=\linewidth]{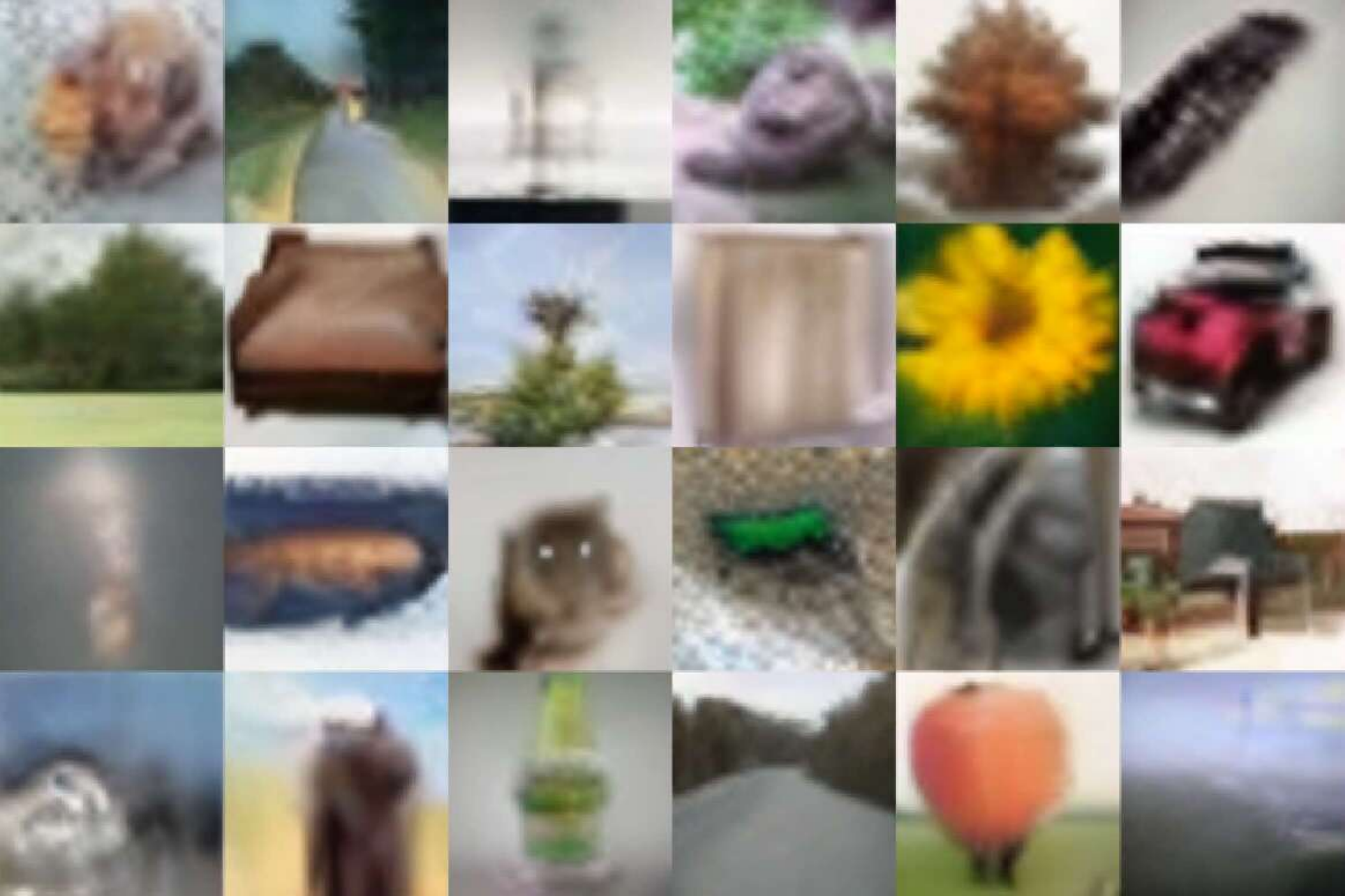}
    \caption{NAR transformer with prompt tuning ($S\,{=}\,1$)}
    \label{fig:vtab_maskgit_prompt_s1_cifar}
  \end{subfigure}
  \hspace{0.02in}
  \begin{subfigure}[b]{0.48\linewidth}
    \centering
    \includegraphics[width=\linewidth]{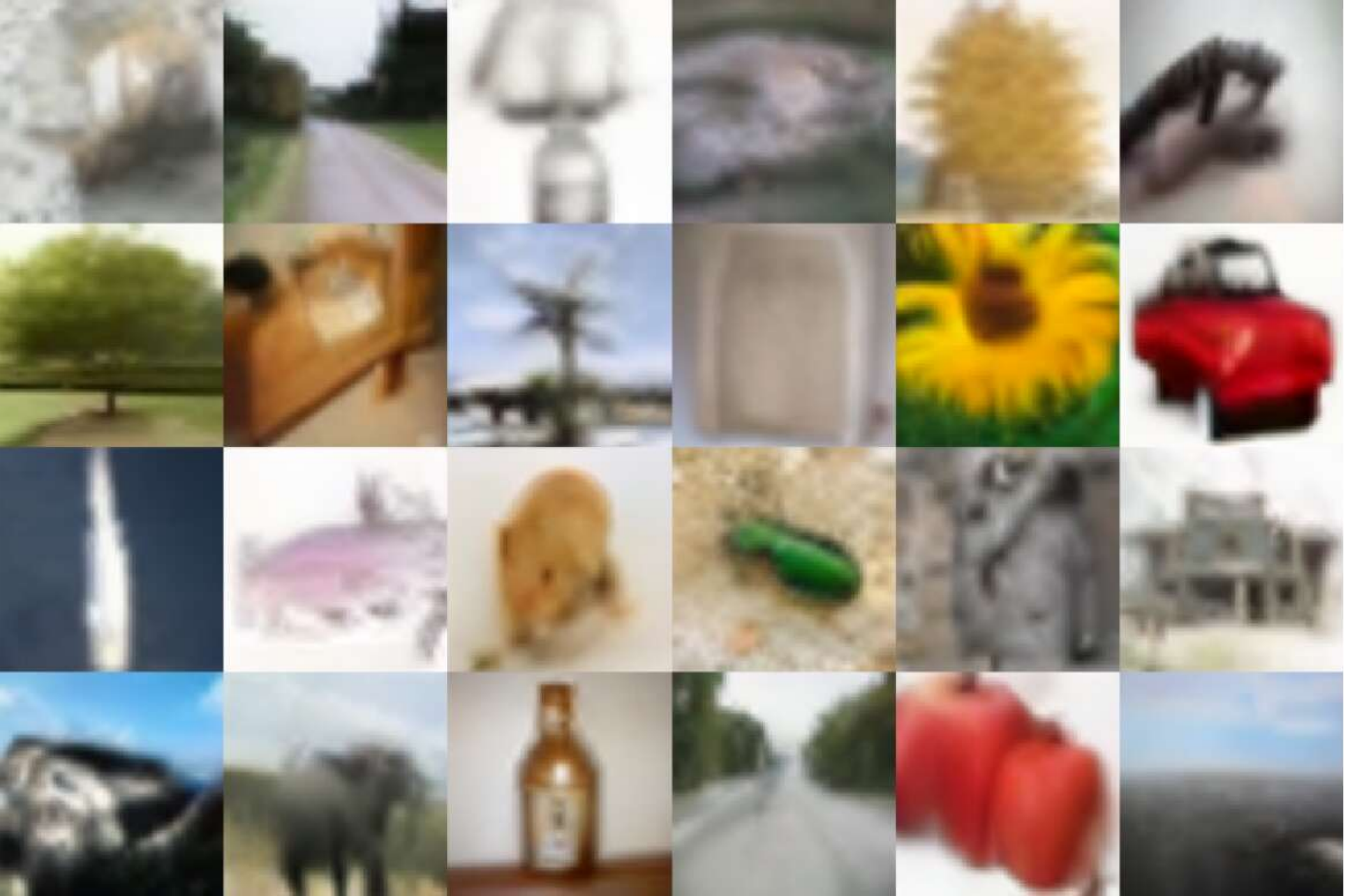}
    \caption{NAR transformer with prompt tuning ($S\,{=}\,128$)}
    \label{fig:vtab_maskgit_prompt_s128_cifar}
  \end{subfigure}
  \caption{Visualization of generated images with different models on CIFAR100 of VTAB.}
  \label{fig:vtab_supp_cifar}
\end{figure}

\begin{figure}
  \centering
  \begin{subfigure}[b]{0.48\linewidth}
    \centering
    \includegraphics[width=\linewidth]{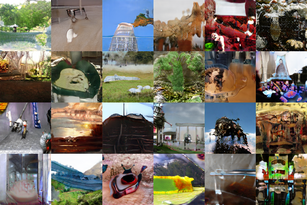}
    \caption{MineGAN}
    \label{fig:vtab_minegan_sun397}
  \end{subfigure}
  \hspace{0.02in}
  \begin{subfigure}[b]{0.48\linewidth}
    \centering
    \includegraphics[width=\linewidth]{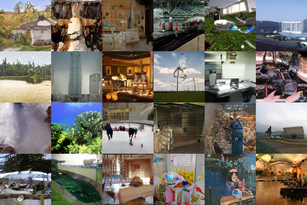}
    \caption{cGANTransfer}
    \label{fig:vtab_cgantrasnfer_sun397}
  \end{subfigure}
  \begin{subfigure}[b]{0.48\linewidth}
    \centering
    \includegraphics[width=\linewidth]{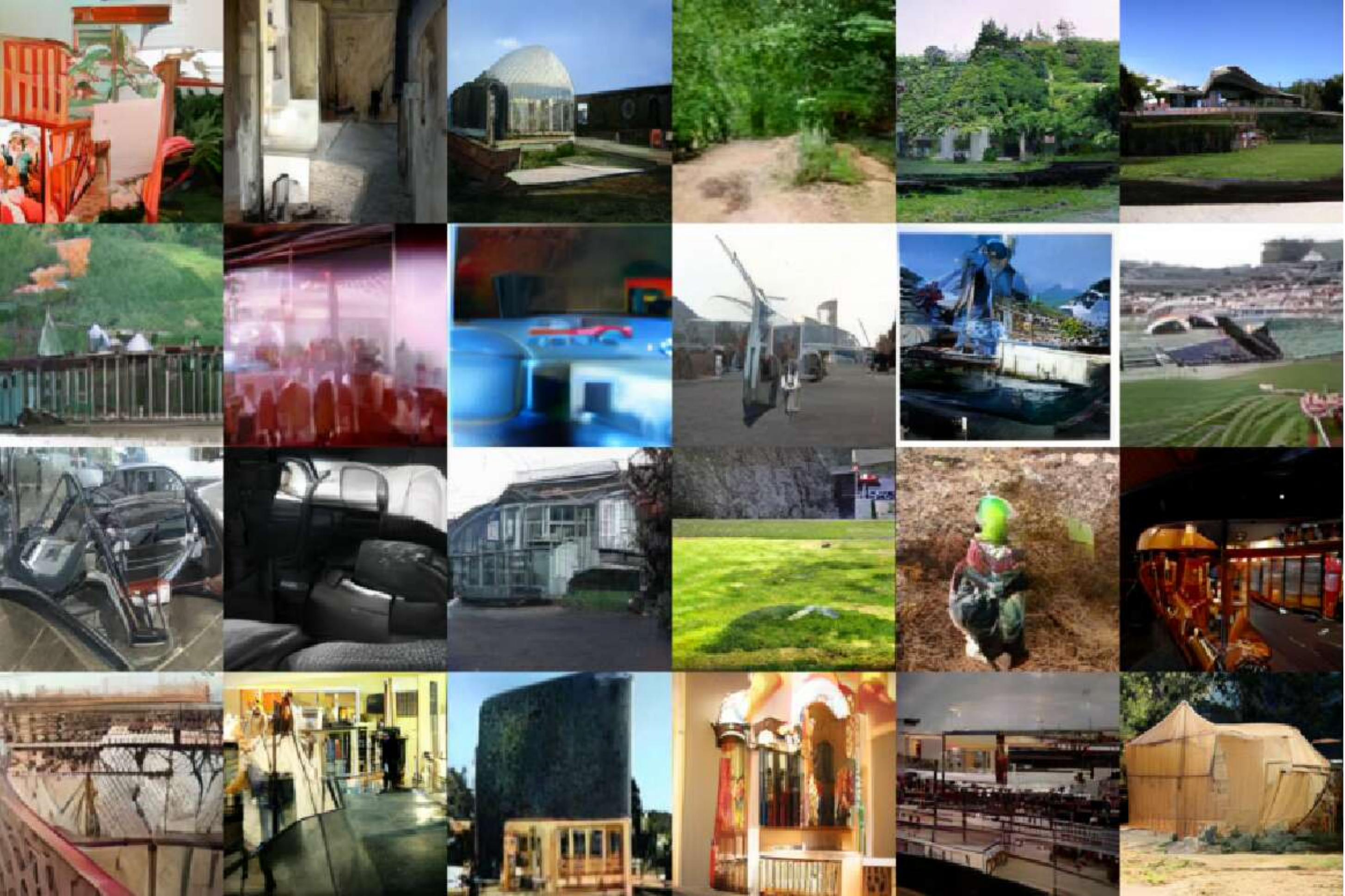}
    \caption{AR transformer with prompt tuning ($S\,{=}\,1$)}
    \label{fig:vtab_taming_prompt_s1_sun397}
  \end{subfigure}
  \hspace{0.02in}
  \begin{subfigure}[b]{0.48\linewidth}
    \centering
    \includegraphics[width=\linewidth]{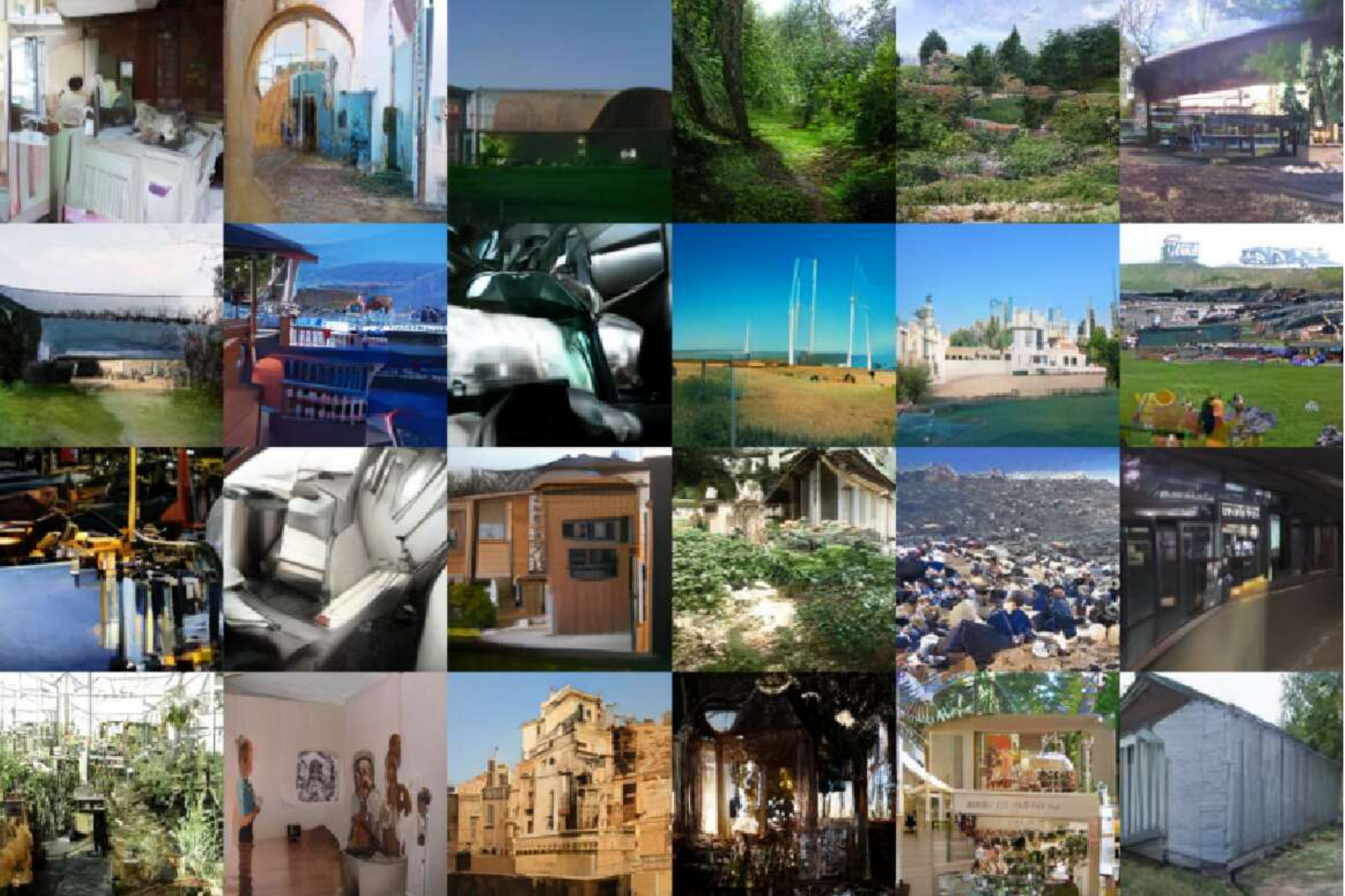}
    \caption{AR transformer with prompt tuning ($S\,{=}\,256$, $F\,{=}\,16$)}
    \label{fig:vtab_taming_prompt_s128_sun397}
  \end{subfigure}
  \begin{subfigure}[b]{0.48\linewidth}
    \centering
    \includegraphics[width=\linewidth]{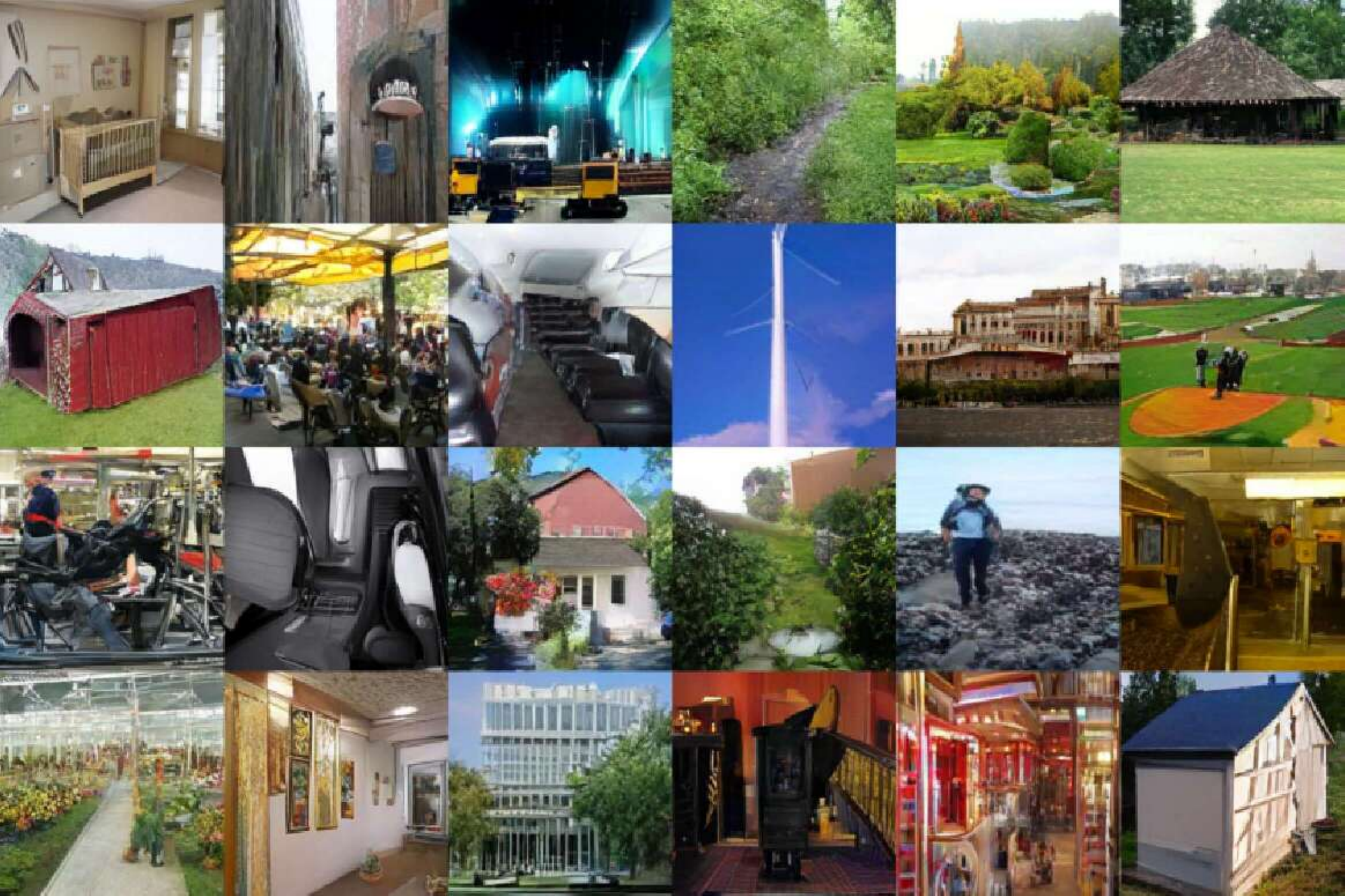}
    \caption{NAR transformer with prompt tuning ($S\,{=}\,1$)}
    \label{fig:vtab_maskgit_prompt_s1_sun397}
  \end{subfigure}
  \hspace{0.02in}
  \begin{subfigure}[b]{0.48\linewidth}
    \centering
    \includegraphics[width=\linewidth]{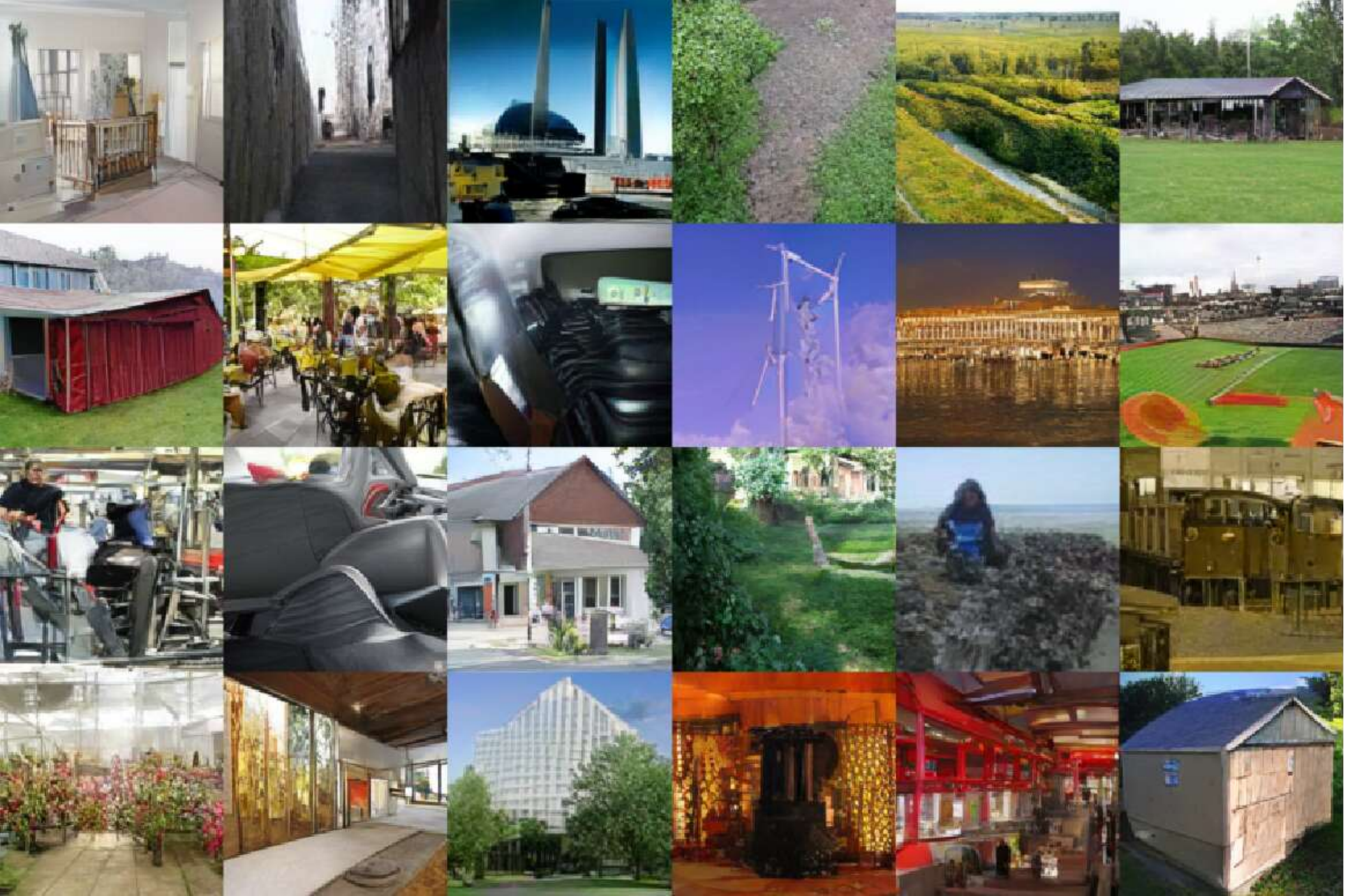}
    \caption{NAR transformer with prompt tuning ($S\,{=}\,128$)}
    \label{fig:vtab_maskgit_prompt_s128_sun397}
  \end{subfigure}
  \caption{Visualization of generated images with different models on SUN397 of VTAB.}
  \label{fig:vtab_supp_sun397}
\end{figure}

\begin{figure}
  \centering
  \begin{subfigure}[b]{0.48\linewidth}
    \centering
    \includegraphics[width=\linewidth]{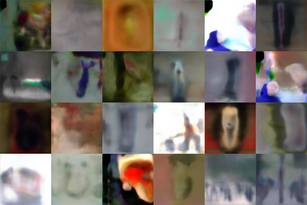}
    \caption{MineGAN}
    \label{fig:vtab_minegan_svhn}
  \end{subfigure}
  \hspace{0.02in}
  \begin{subfigure}[b]{0.48\linewidth}
    \centering
    \includegraphics[width=\linewidth]{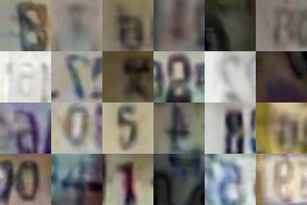}
    \caption{cGANTransfer}
    \label{fig:vtab_cgantrasnfer_svhn}
  \end{subfigure}
  \begin{subfigure}[b]{0.48\linewidth}
    \centering
    \includegraphics[width=\linewidth]{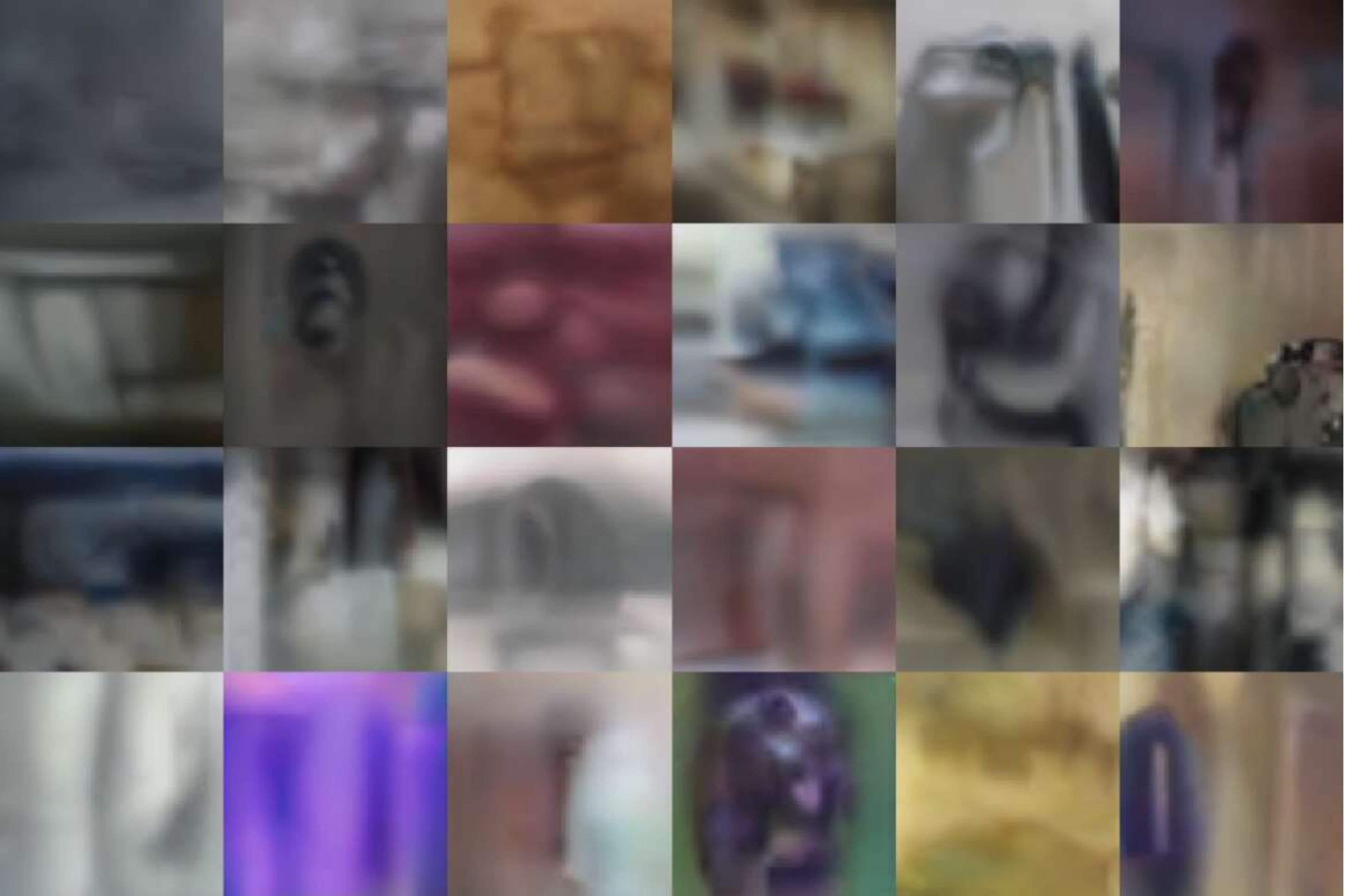}
    \caption{AR transformer with prompt tuning ($S\,{=}\,1$)}
    \label{fig:vtab_taming_prompt_s1_svhn}
  \end{subfigure}
  \hspace{0.02in}
  \begin{subfigure}[b]{0.48\linewidth}
    \centering
    \includegraphics[width=\linewidth]{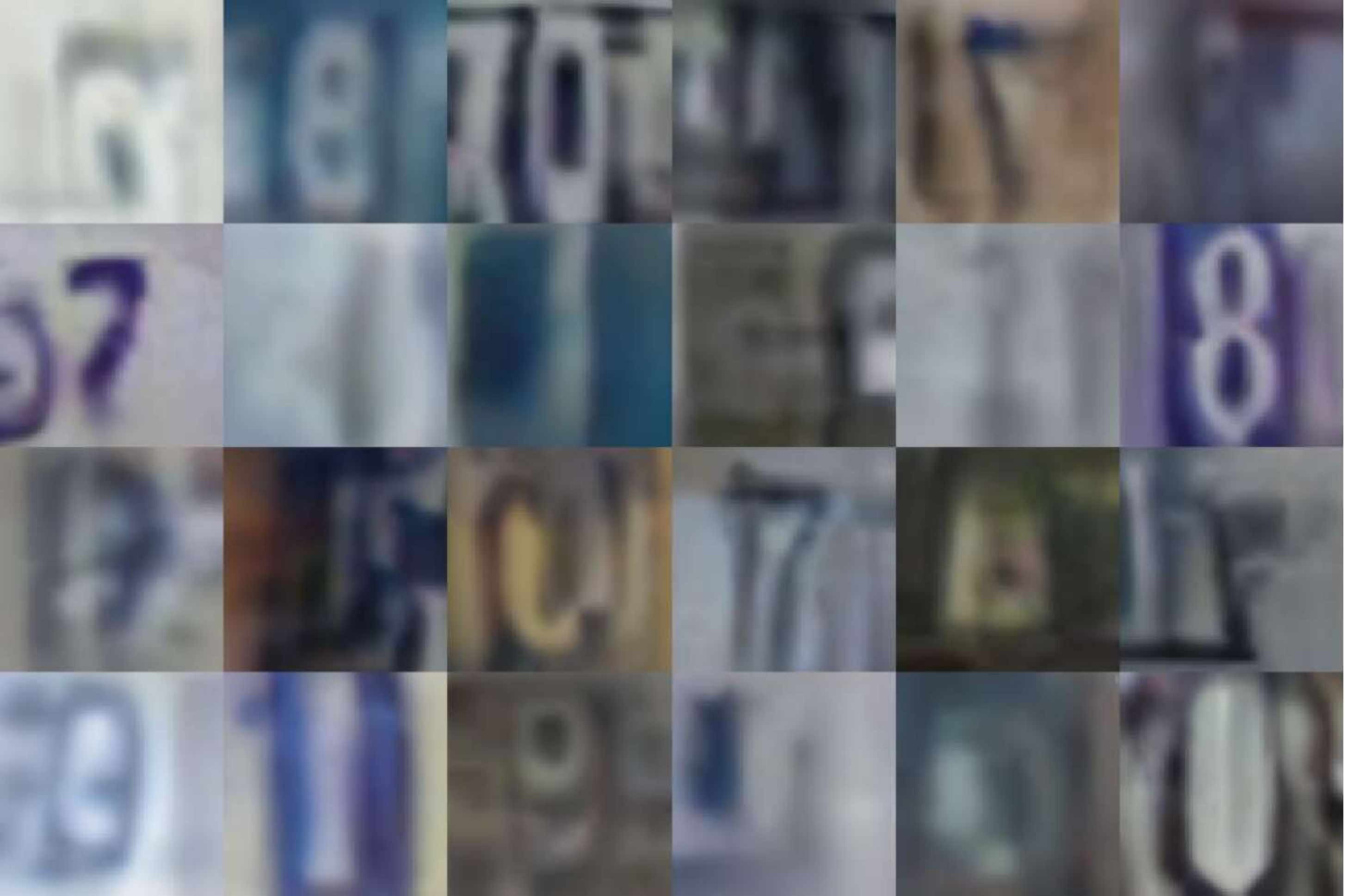}
    \caption{AR transformer with prompt tuning ($S\,{=}\,256$, $F\,{=}\,16$)}
    \label{fig:vtab_taming_prompt_s128_svhn}
  \end{subfigure}
  \begin{subfigure}[b]{0.48\linewidth}
    \centering
    \includegraphics[width=\linewidth]{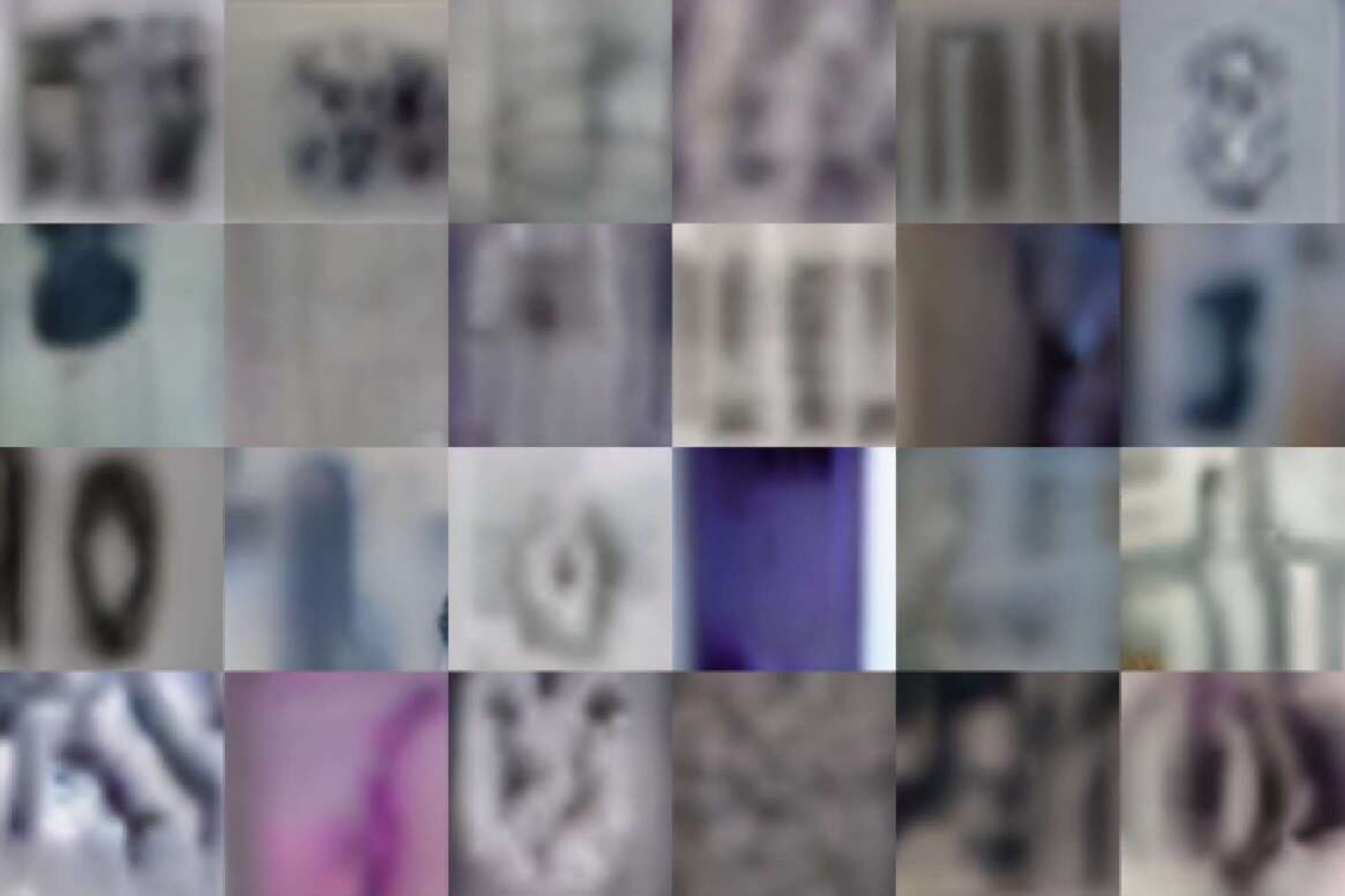}
    \caption{NAR transformer with prompt tuning ($S\,{=}\,1$)}
    \label{fig:vtab_maskgit_prompt_s1_svhn}
  \end{subfigure}
  \hspace{0.02in}
  \begin{subfigure}[b]{0.48\linewidth}
    \centering
    \includegraphics[width=\linewidth]{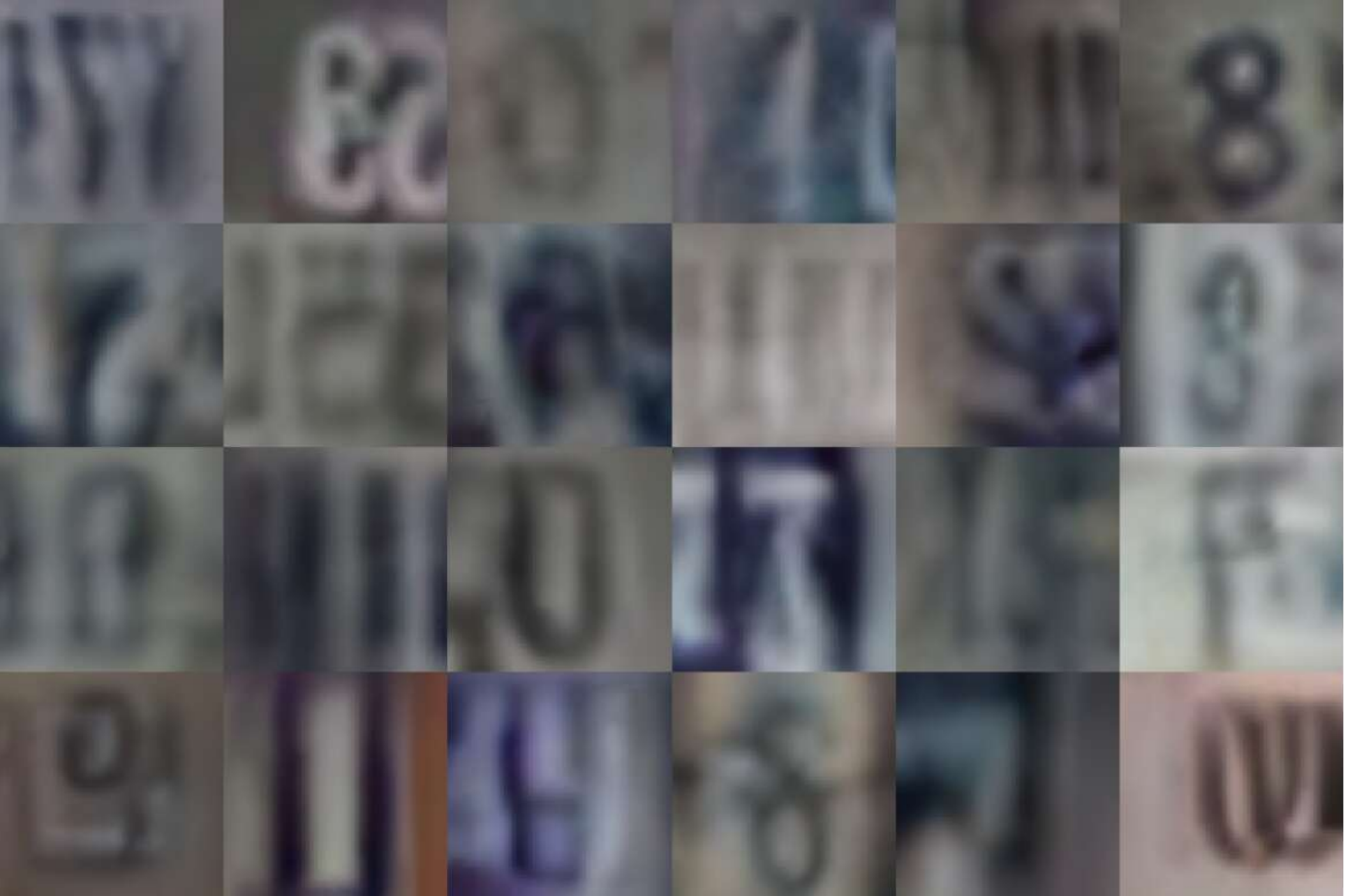}
    \caption{NAR transformer with prompt tuning ($S\,{=}\,128$)}
    \label{fig:vtab_maskgit_prompt_s128_svhn}
  \end{subfigure}
  \caption{Visualization of generated images with different models on SVHN of VTAB.}
  \label{fig:vtab_supp_svhn}
\end{figure}

\begin{figure}
  \centering
  \begin{subfigure}[b]{0.48\linewidth}
    \centering
    \includegraphics[width=\linewidth]{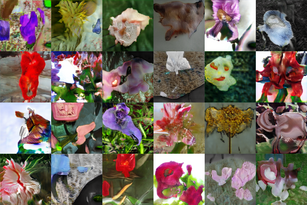}
    \caption{MineGAN}
    \label{fig:vtab_minegan_oxford_flowers102}
  \end{subfigure}
  \hspace{0.02in}
  \begin{subfigure}[b]{0.48\linewidth}
    \centering
    \includegraphics[width=\linewidth]{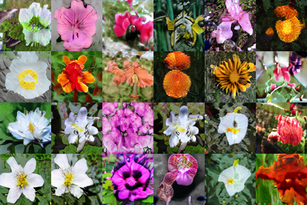}
    \caption{cGANTransfer}
    \label{fig:vtab_cgantrasnfer_oxford_flowers102}
  \end{subfigure}
  \begin{subfigure}[b]{0.48\linewidth}
    \centering
    \includegraphics[width=\linewidth]{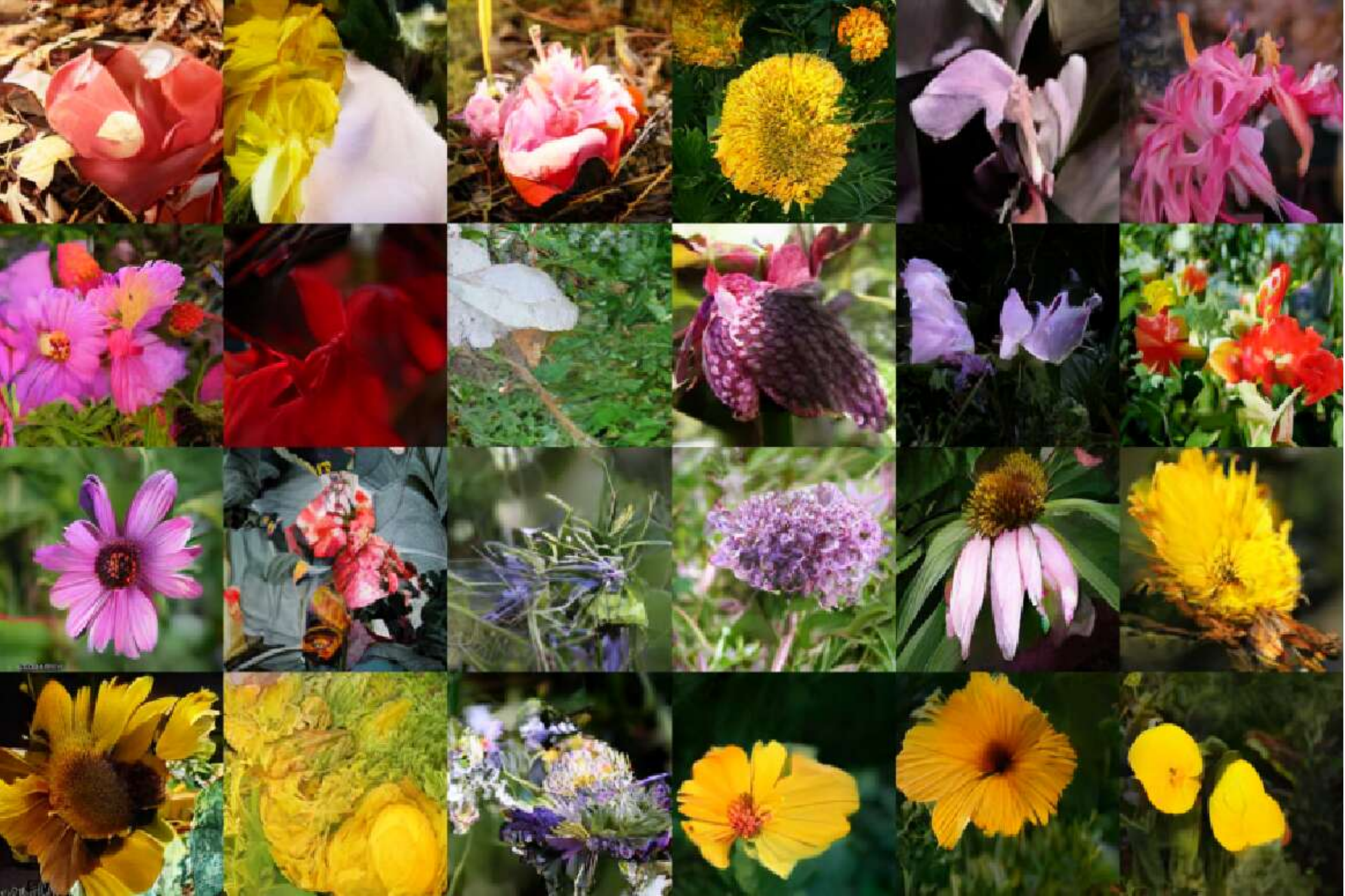}
    \caption{AR transformer with prompt tuning ($S\,{=}\,1$)}
    \label{fig:vtab_taming_prompt_s1_oxford_flowers102}
  \end{subfigure}
  \hspace{0.02in}
  \begin{subfigure}[b]{0.48\linewidth}
    \centering
    \includegraphics[width=\linewidth]{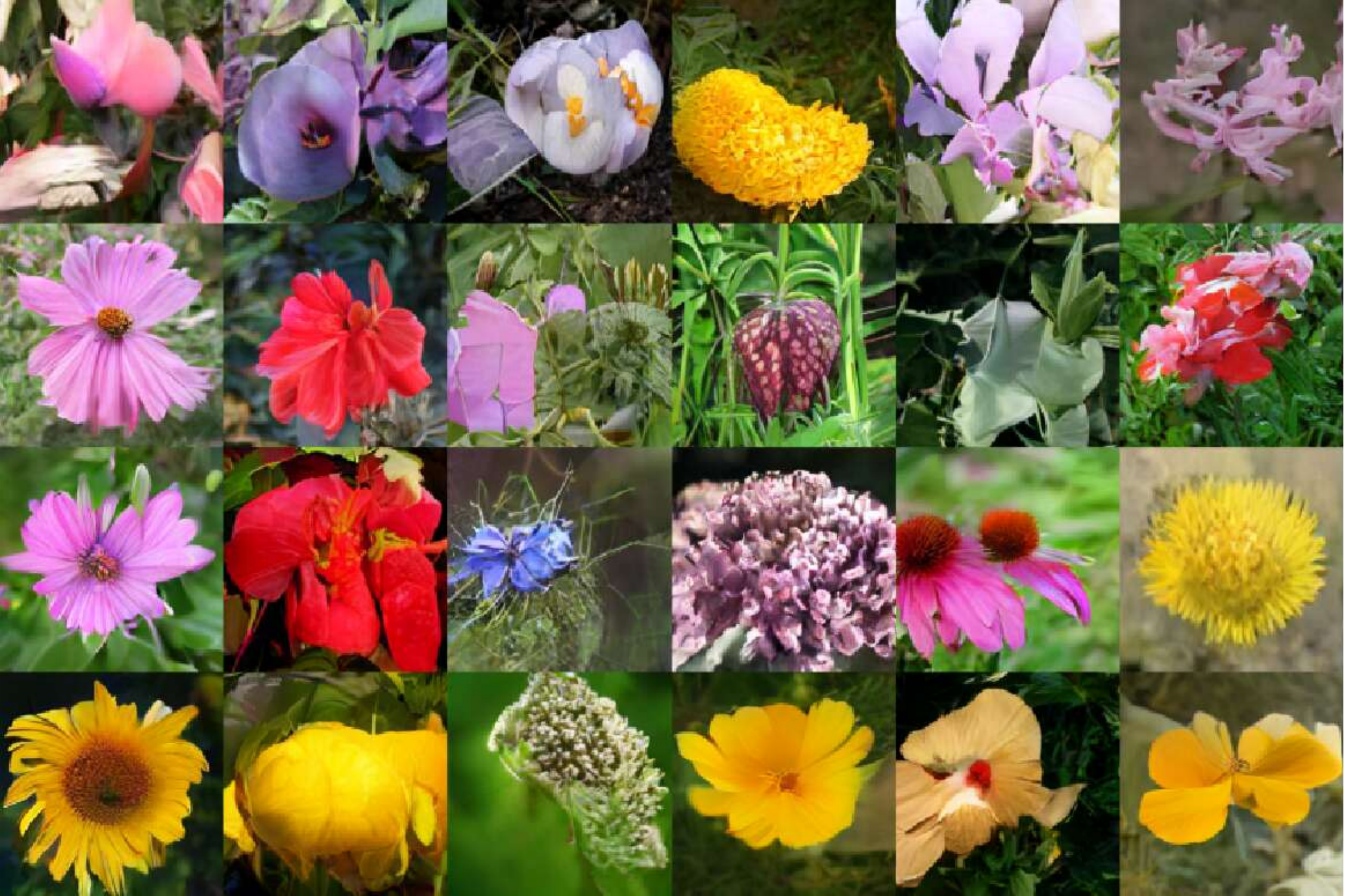}
    \caption{AR transformer with prompt tuning ($S\,{=}\,256$, $F\,{=}\,16$)}
    \label{fig:vtab_taming_prompt_s128_oxford_flowers102}
  \end{subfigure}
  \begin{subfigure}[b]{0.48\linewidth}
    \centering
    \includegraphics[width=\linewidth]{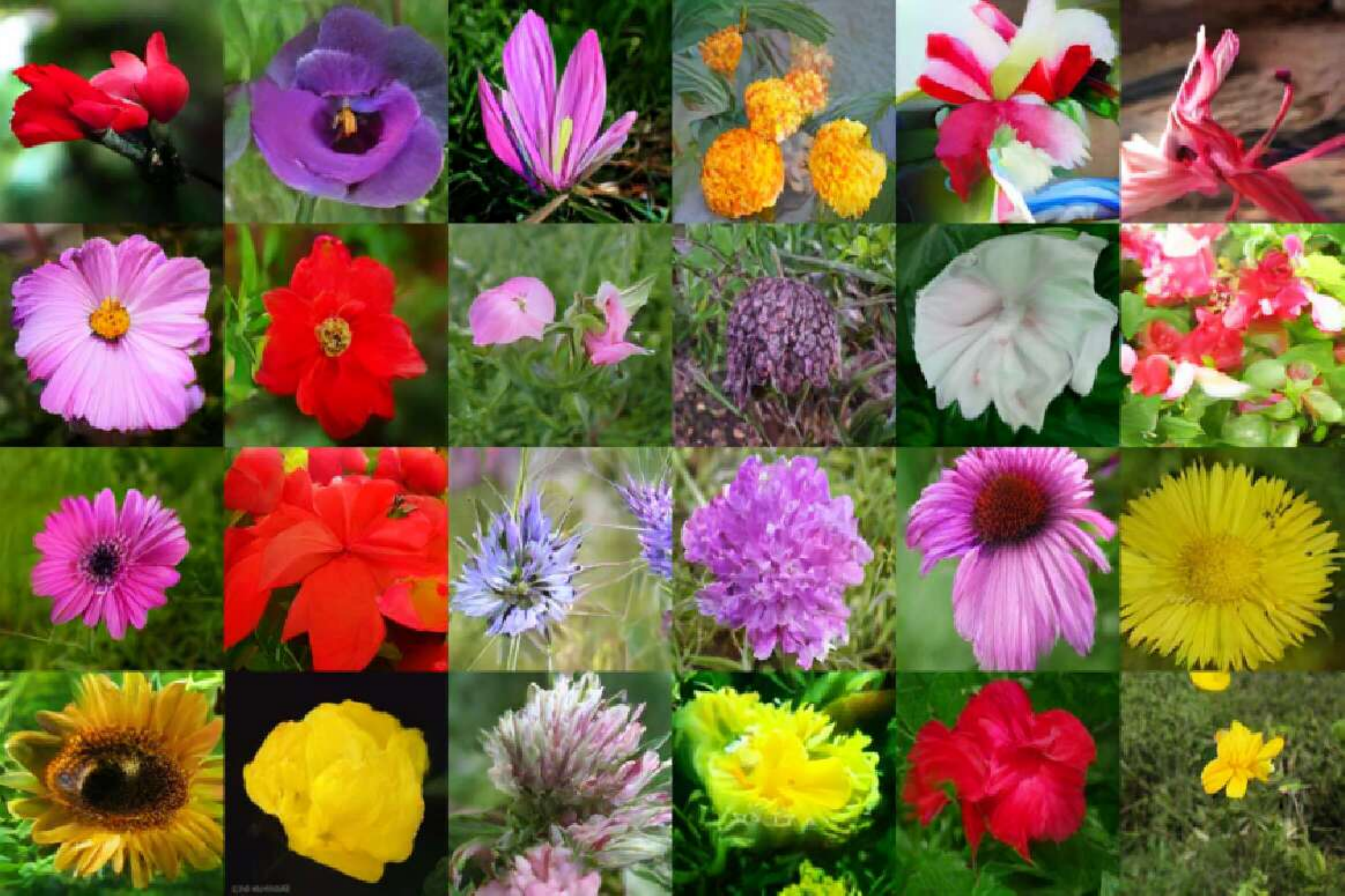}
    \caption{NAR transformer with prompt tuning ($S\,{=}\,1$)}
    \label{fig:vtab_maskgit_prompt_s1_oxford_flowers102}
  \end{subfigure}
  \hspace{0.02in}
  \begin{subfigure}[b]{0.48\linewidth}
    \centering
    \includegraphics[width=\linewidth]{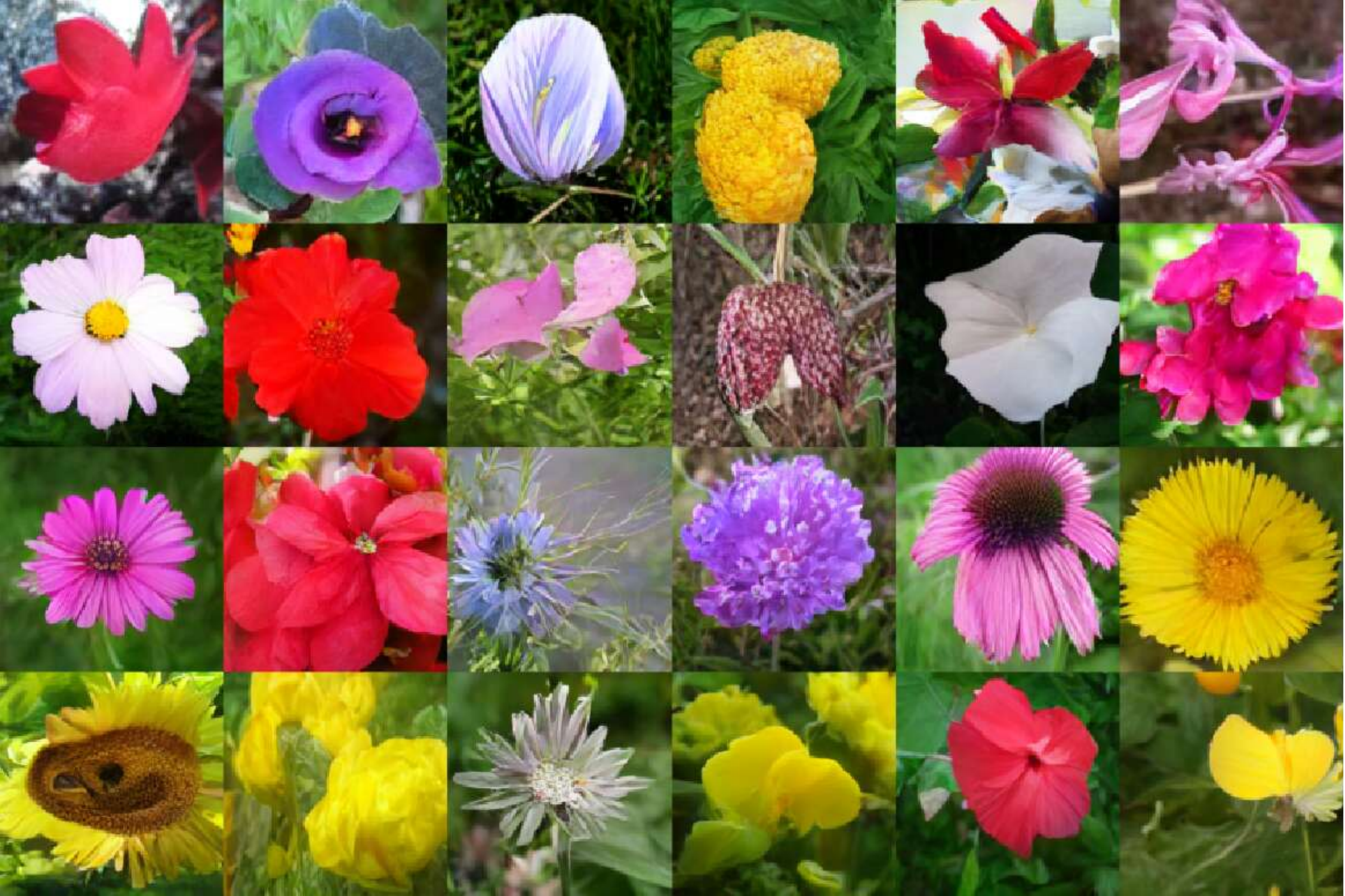}
    \caption{NAR transformer with prompt tuning ($S\,{=}\,128$)}
    \label{fig:vtab_maskgit_prompt_s128_oxford_flowers102}
  \end{subfigure}
  \caption{Visualization of generated images with different models on Oxford Flowers102 of VTAB.}
  \label{fig:vtab_supp_oxford_flowers102}
\end{figure}

\begin{figure}
  \centering
  \begin{subfigure}[b]{0.48\linewidth}
    \centering
    \includegraphics[width=\linewidth]{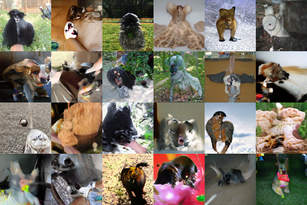}
    \caption{MineGAN}
    \label{fig:vtab_minegan_oxford_iiit_pet}
  \end{subfigure}
  \hspace{0.02in}
  \begin{subfigure}[b]{0.48\linewidth}
    \centering
    \includegraphics[width=\linewidth]{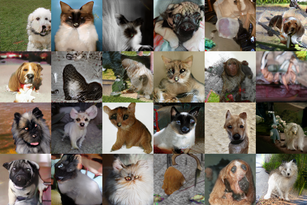}
    \caption{cGANTransfer}
    \label{fig:vtab_cgantrasnfer_oxford_iiit_pet}
  \end{subfigure}
  \begin{subfigure}[b]{0.48\linewidth}
    \centering
    \includegraphics[width=\linewidth]{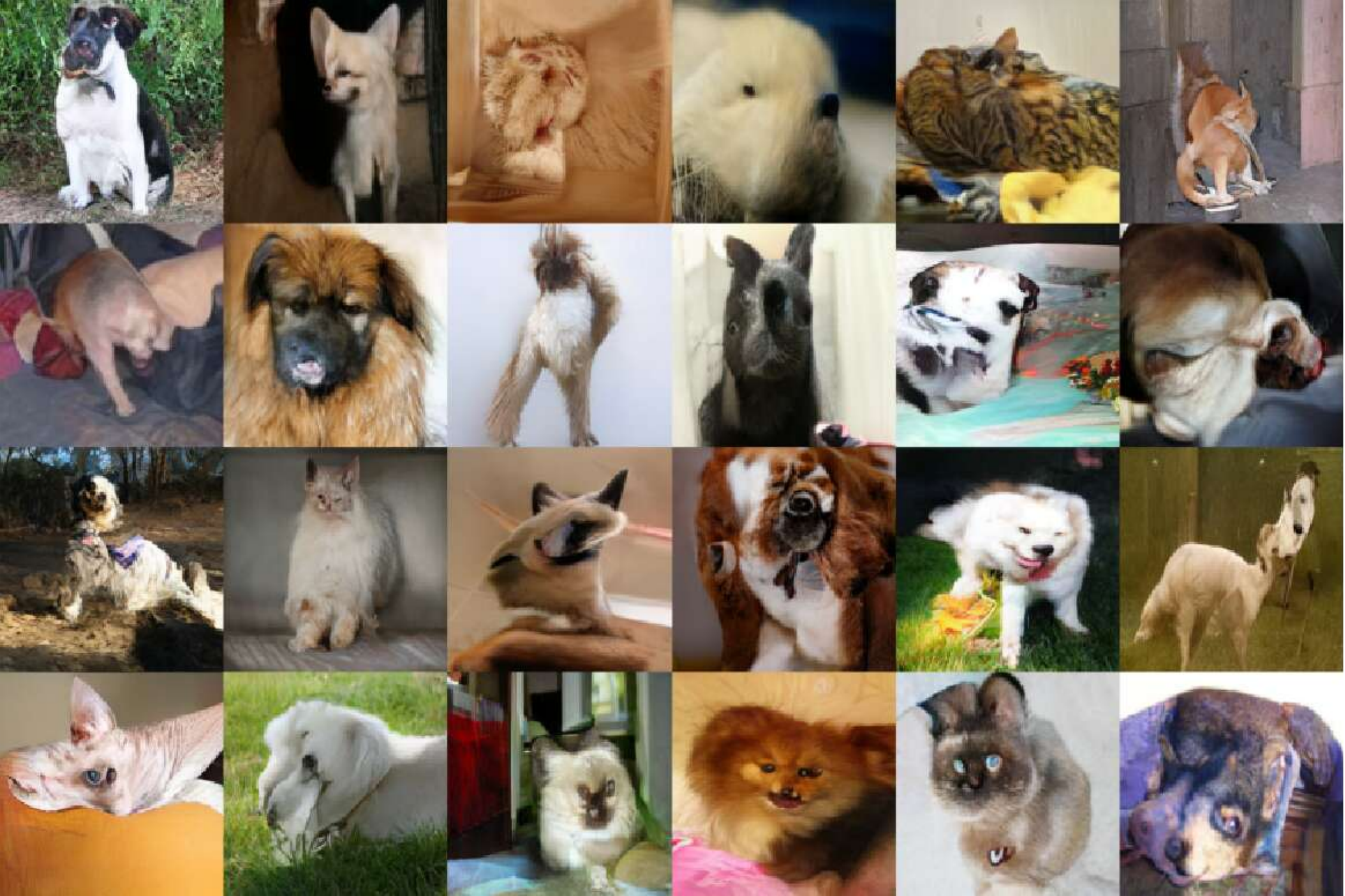}
    \caption{AR transformer with prompt tuning ($S\,{=}\,1$)}
    \label{fig:vtab_taming_prompt_s1_oxford_iiit_pet}
  \end{subfigure}
  \hspace{0.02in}
  \begin{subfigure}[b]{0.48\linewidth}
    \centering
    \includegraphics[width=\linewidth]{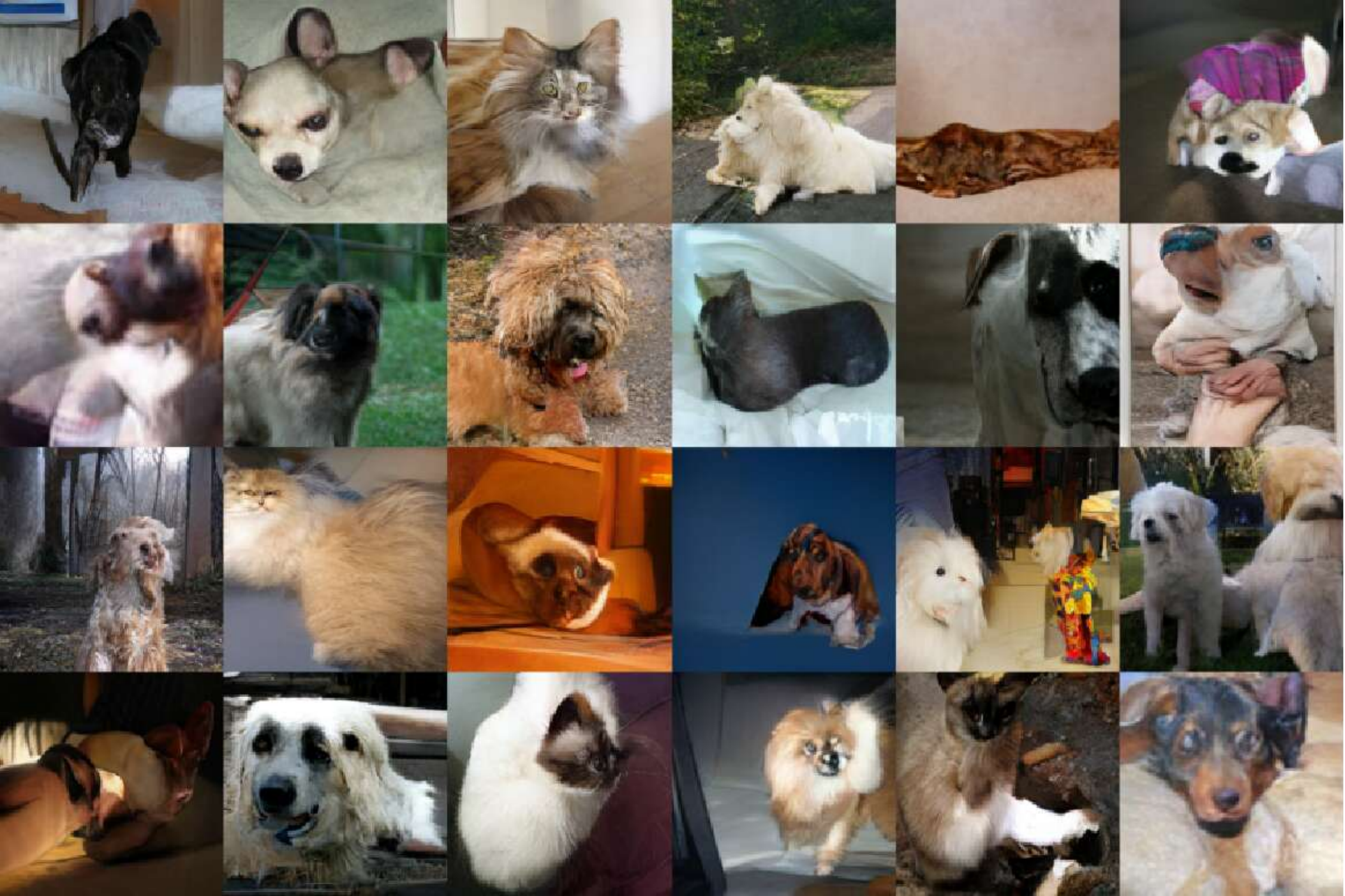}
    \caption{AR transformer with prompt tuning ($S\,{=}\,256$, $F\,{=}\,16$)}
    \label{fig:vtab_taming_prompt_s128_oxford_iiit_pet}
  \end{subfigure}
  \begin{subfigure}[b]{0.48\linewidth}
    \centering
    \includegraphics[width=\linewidth]{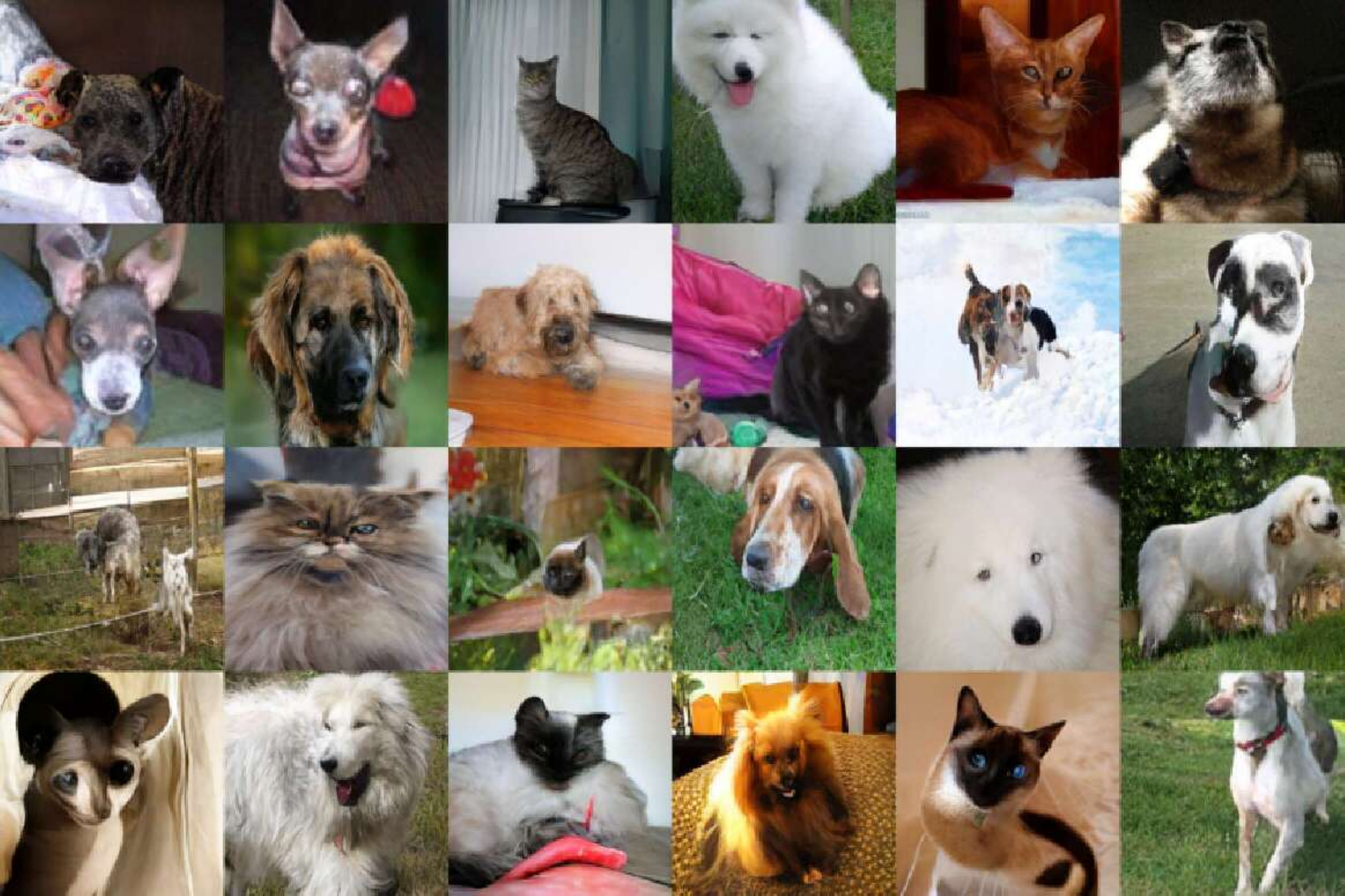}
    \caption{NAR transformer with prompt tuning ($S\,{=}\,1$)}
    \label{fig:vtab_maskgit_prompt_s1_oxford_iiit_pet}
  \end{subfigure}
  \hspace{0.02in}
  \begin{subfigure}[b]{0.48\linewidth}
    \centering
    \includegraphics[width=\linewidth]{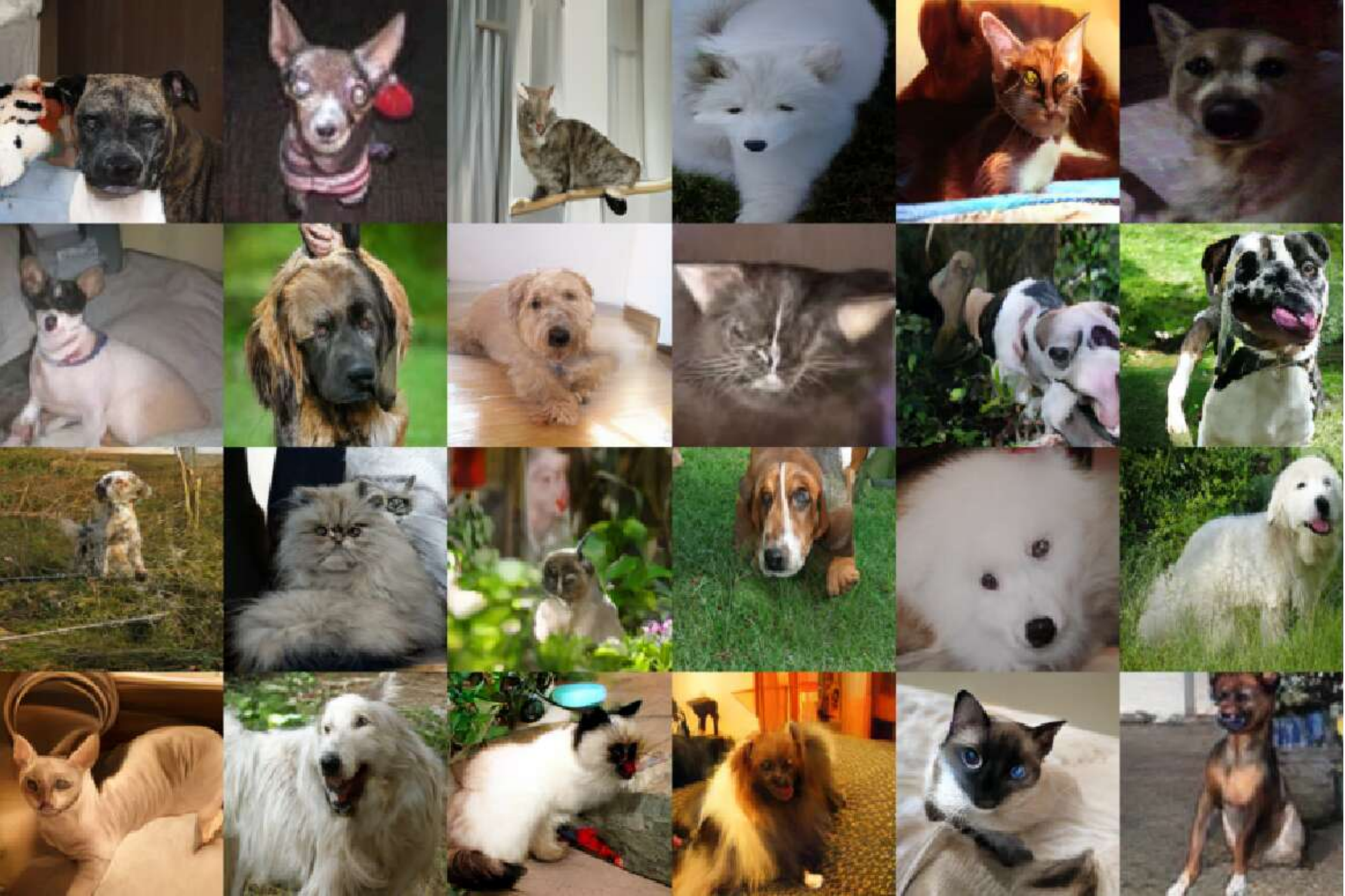}
    \caption{NAR transformer with prompt tuning ($S\,{=}\,128$)}
    \label{fig:vtab_maskgit_prompt_s128_oxford_iiit_pet}
  \end{subfigure}
  \caption{Visualization of generated images with different models on Oxford iiit Pet of VTAB.}
  \label{fig:vtab_supp_oxford_iiit_pet}
\end{figure}

\begin{figure}
  \centering
  \begin{subfigure}[b]{0.48\linewidth}
    \centering
    \includegraphics[width=\linewidth]{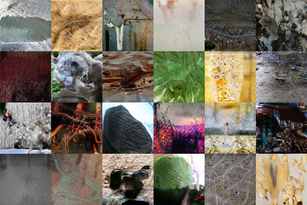}
    \caption{MineGAN}
    \label{fig:vtab_minegan_dtd}
  \end{subfigure}
  \hspace{0.02in}
  \begin{subfigure}[b]{0.48\linewidth}
    \centering
    \includegraphics[width=\linewidth]{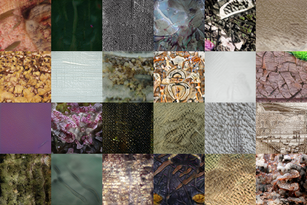}
    \caption{cGANTransfer}
    \label{fig:vtab_cgantrasnfer_dtd}
  \end{subfigure}
  \begin{subfigure}[b]{0.48\linewidth}
    \centering
    \includegraphics[width=\linewidth]{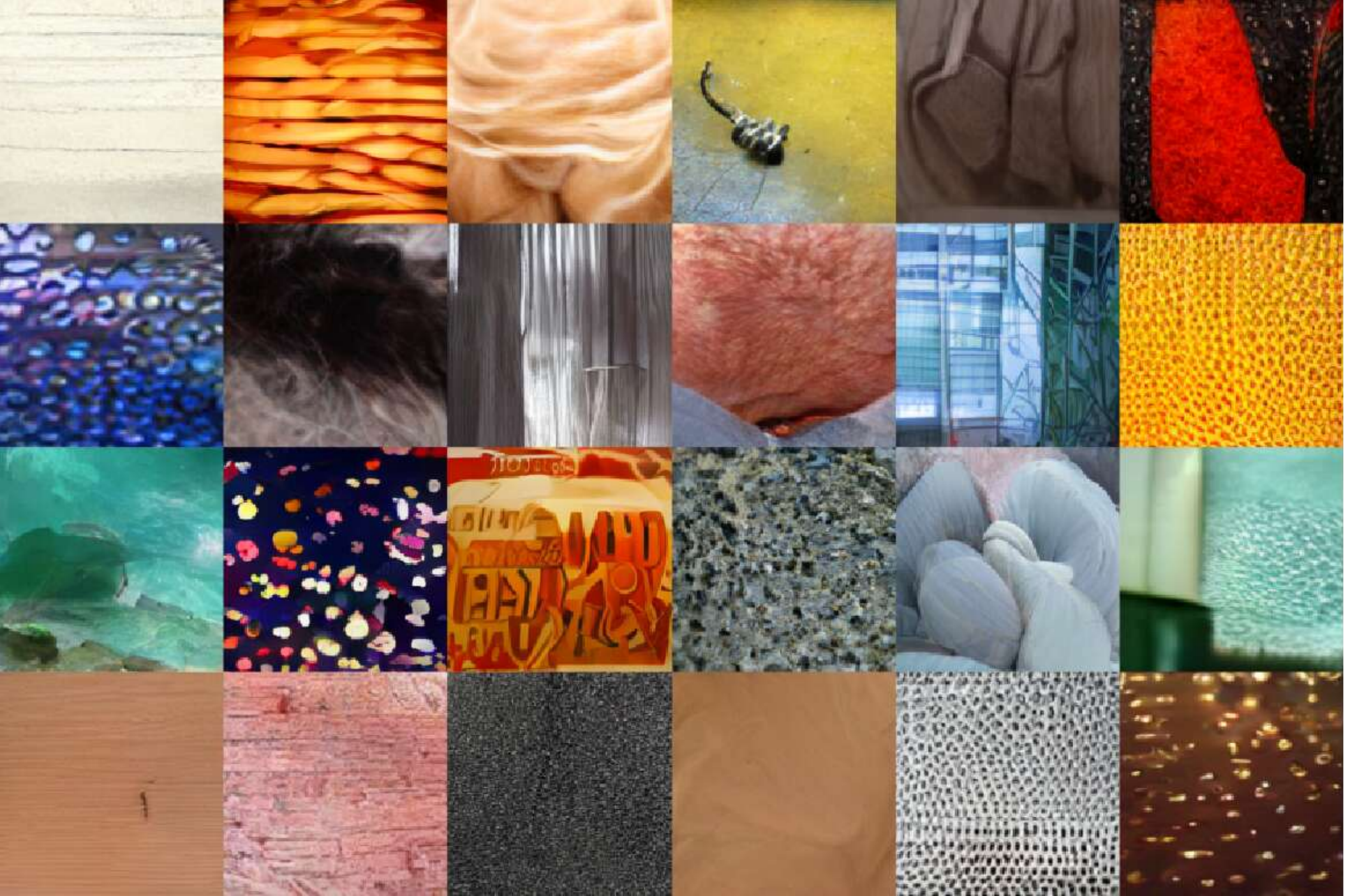}
    \caption{AR transformer with prompt tuning ($S\,{=}\,1$)}
    \label{fig:vtab_taming_prompt_s1_dtd}
  \end{subfigure}
  \hspace{0.02in}
  \begin{subfigure}[b]{0.48\linewidth}
    \centering
    \includegraphics[width=\linewidth]{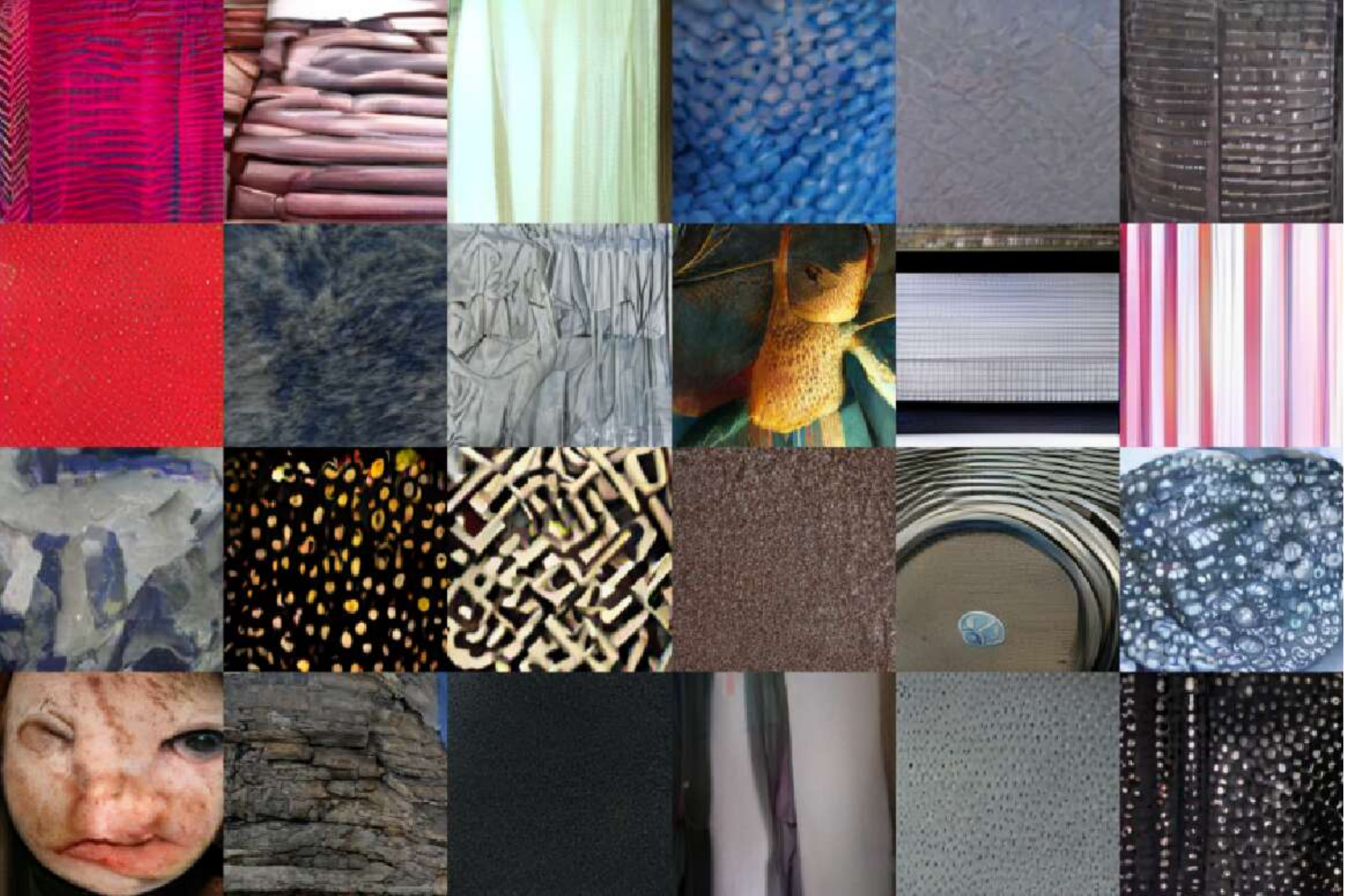}
    \caption{AR transformer with prompt tuning ($S\,{=}\,256$, $F\,{=}\,16$)}
    \label{fig:vtab_taming_prompt_s128_dtd}
  \end{subfigure}
  \begin{subfigure}[b]{0.48\linewidth}
    \centering
    \includegraphics[width=\linewidth]{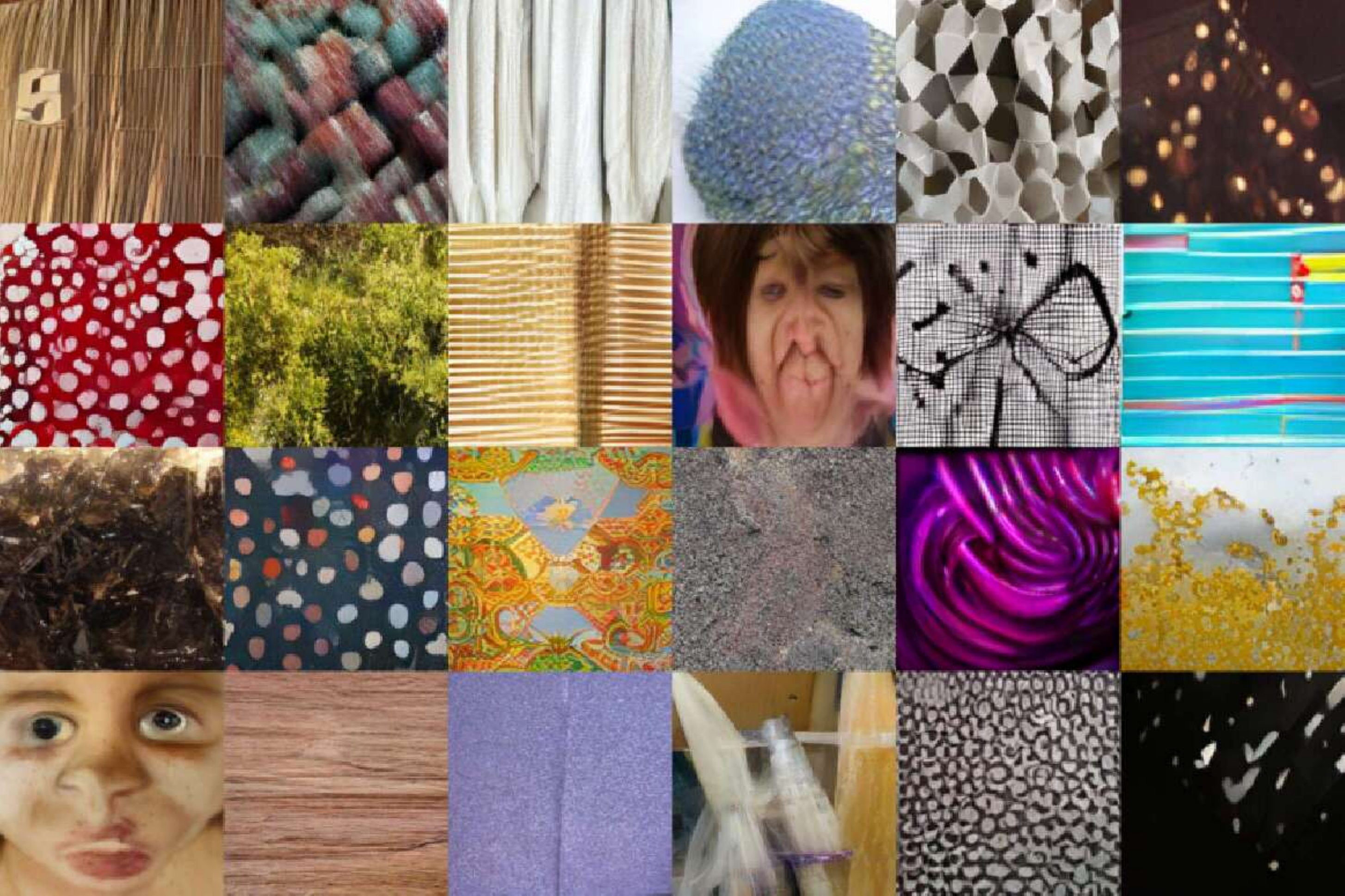}
    \caption{NAR transformer with prompt tuning ($S\,{=}\,1$)}
    \label{fig:vtab_maskgit_prompt_s1_dtd}
  \end{subfigure}
  \hspace{0.02in}
  \begin{subfigure}[b]{0.48\linewidth}
    \centering
    \includegraphics[width=\linewidth]{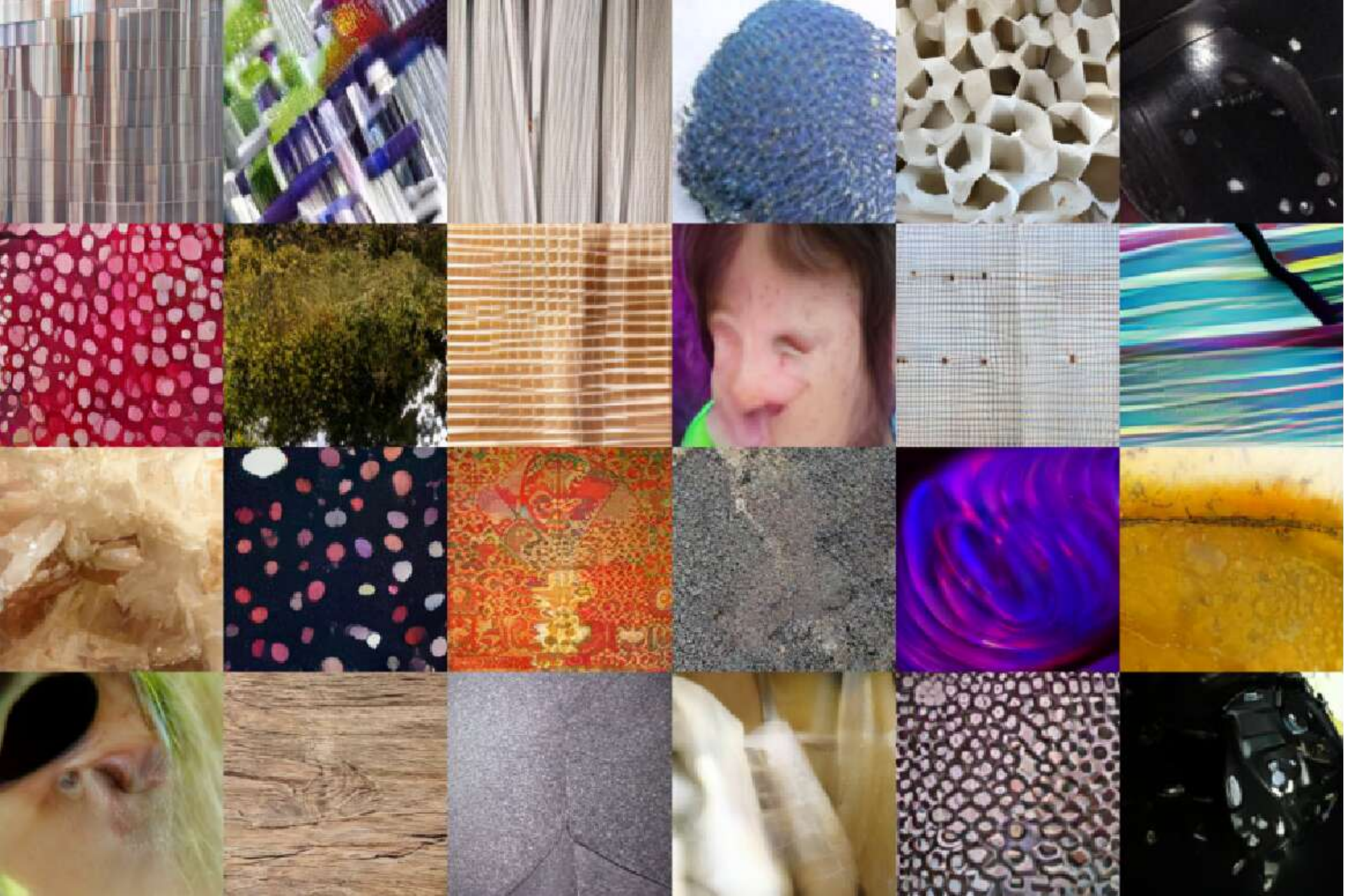}
    \caption{NAR transformer with prompt tuning ($S\,{=}\,128$)}
    \label{fig:vtab_maskgit_prompt_s128_dtd}
  \end{subfigure}
  \caption{Visualization of generated images with different models on DTD of VTAB.}
  \label{fig:vtab_supp_dtd}
\end{figure}

\begin{figure}
  \centering
  \begin{subfigure}[b]{0.48\linewidth}
    \centering
    \includegraphics[width=\linewidth]{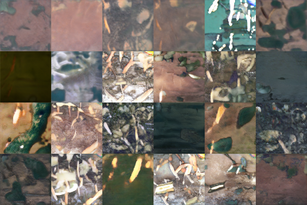}
    \caption{MineGAN}
    \label{fig:vtab_minegan_eurosat}
  \end{subfigure}
  \hspace{0.02in}
  \begin{subfigure}[b]{0.48\linewidth}
    \centering
    \includegraphics[width=\linewidth]{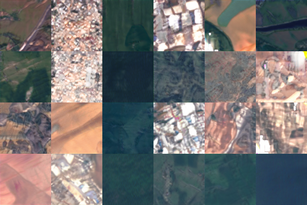}
    \caption{cGANTransfer}
    \label{fig:vtab_cgantrasnfer_eurosat}
  \end{subfigure}
  \begin{subfigure}[b]{0.48\linewidth}
    \centering
    \includegraphics[width=\linewidth]{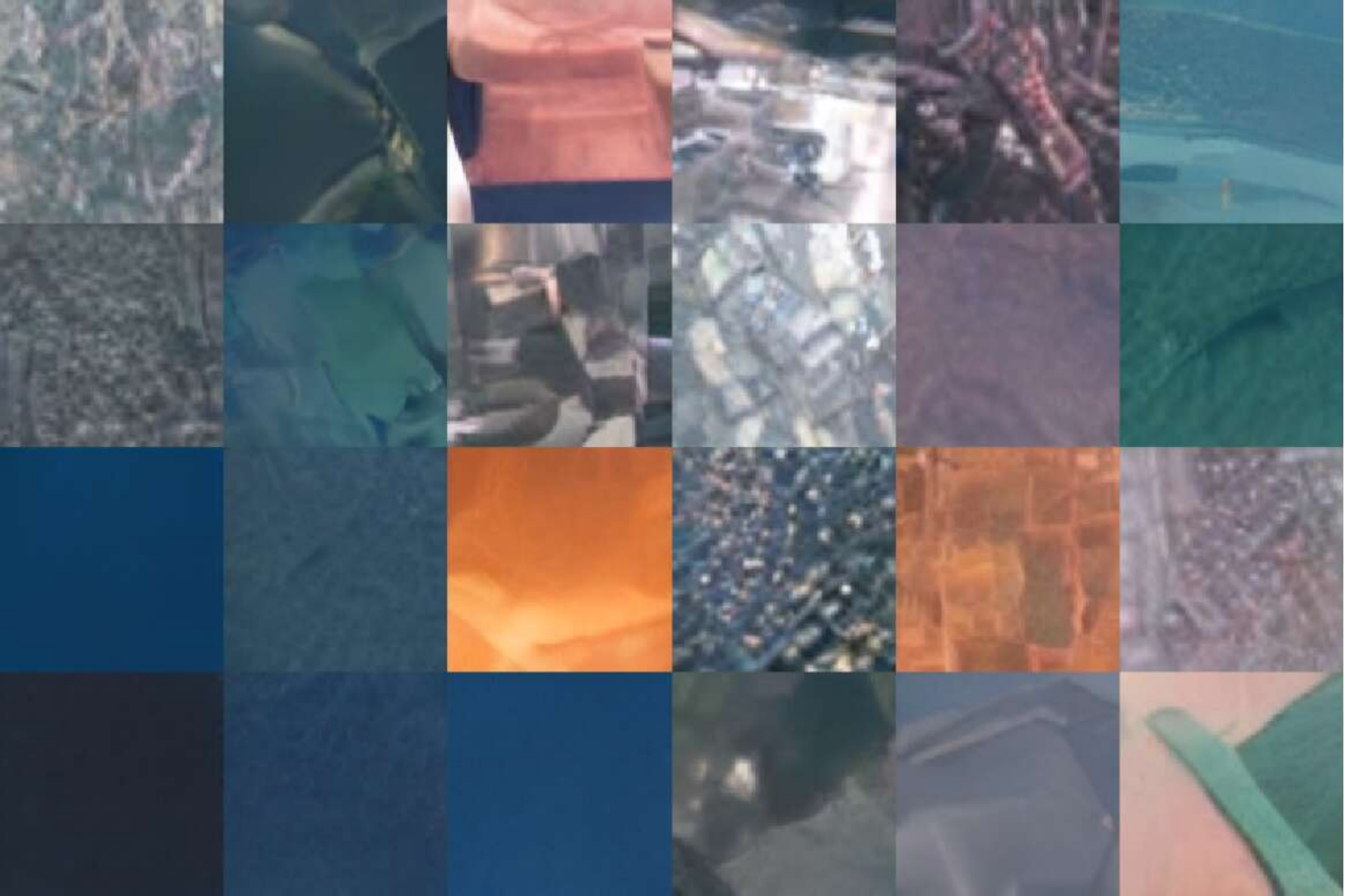}
    \caption{AR transformer with prompt tuning ($S\,{=}\,1$)}
    \label{fig:vtab_taming_prompt_s1_eurosat}
  \end{subfigure}
  \hspace{0.02in}
  \begin{subfigure}[b]{0.48\linewidth}
    \centering
    \includegraphics[width=\linewidth]{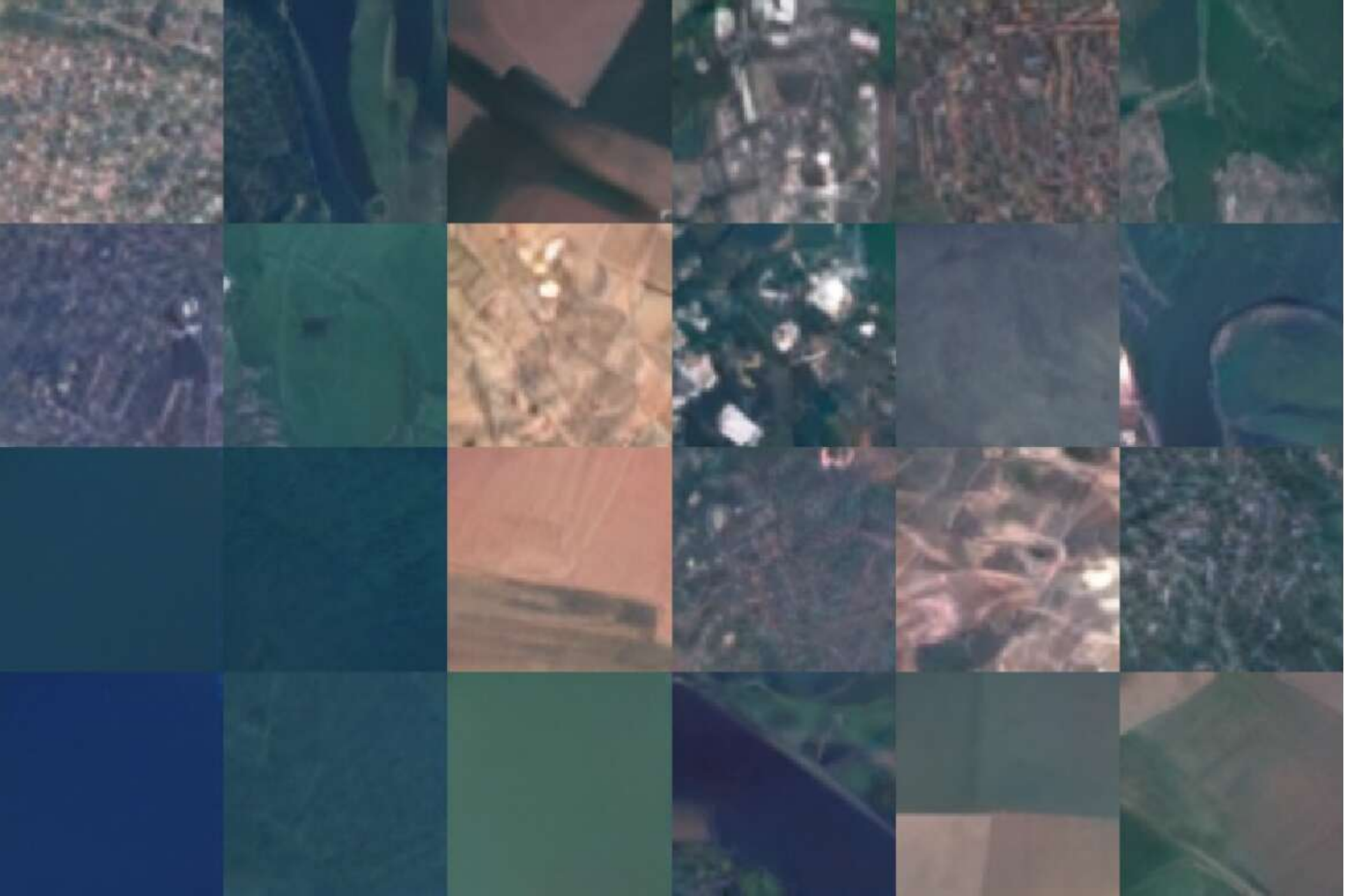}
    \caption{AR transformer with prompt tuning ($S\,{=}\,256$, $F\,{=}\,16$)}
    \label{fig:vtab_taming_prompt_s128_eurosat}
  \end{subfigure}
  \begin{subfigure}[b]{0.48\linewidth}
    \centering
    \includegraphics[width=\linewidth]{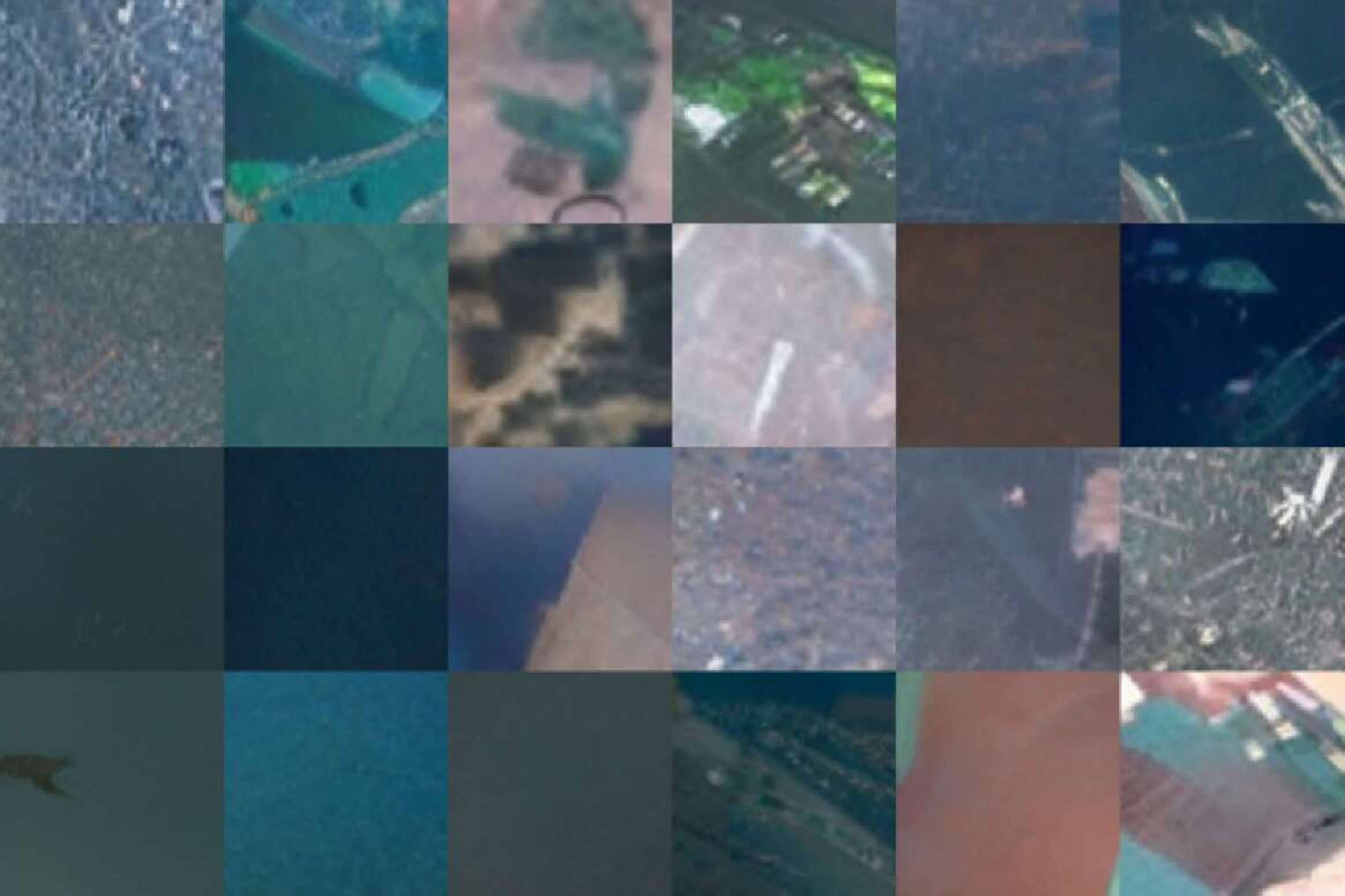}
    \caption{NAR transformer with prompt tuning ($S\,{=}\,1$)}
    \label{fig:vtab_maskgit_prompt_s1_eurosat}
  \end{subfigure}
  \hspace{0.02in}
  \begin{subfigure}[b]{0.48\linewidth}
    \centering
    \includegraphics[width=\linewidth]{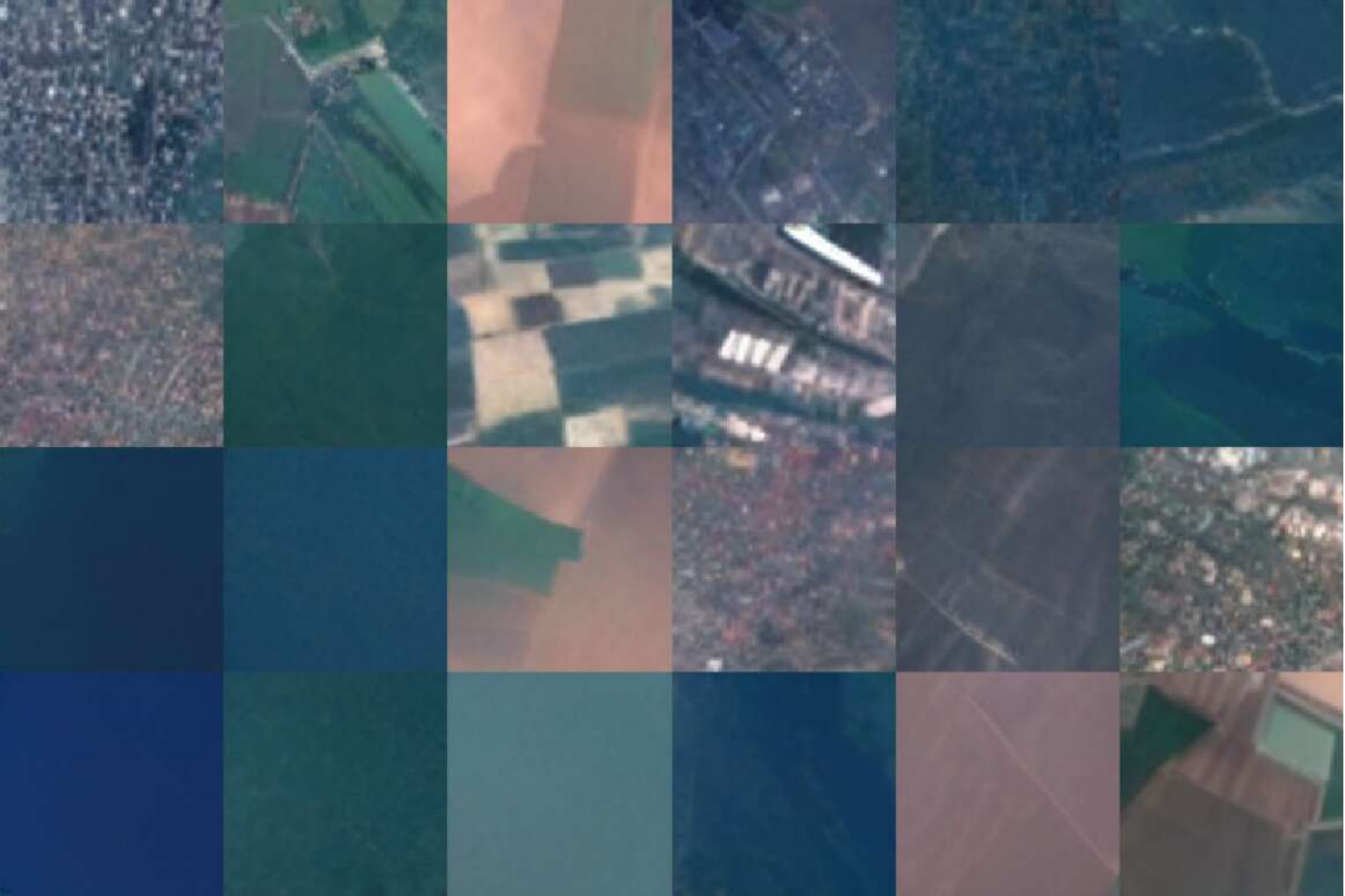}
    \caption{NAR transformer with prompt tuning ($S\,{=}\,128$)}
    \label{fig:vtab_maskgit_prompt_s128_eurosat}
  \end{subfigure}
  \caption{Visualization of generated images with different models on EuroSAT of VTAB.}
  \label{fig:vtab_supp_eurosat}
\end{figure}

\begin{figure}
  \centering
  \begin{subfigure}[b]{0.48\linewidth}
    \centering
    \includegraphics[width=\linewidth]{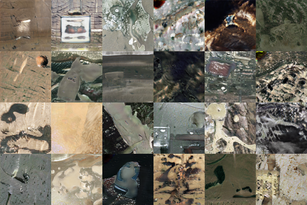}
    \caption{MineGAN}
    \label{fig:vtab_minegan_resisc45}
  \end{subfigure}
  \hspace{0.02in}
  \begin{subfigure}[b]{0.48\linewidth}
    \centering
    \includegraphics[width=\linewidth]{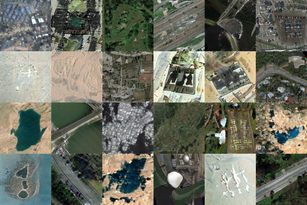}
    \caption{cGANTransfer}
    \label{fig:vtab_cgantrasnfer_resisc45}
  \end{subfigure}
  \begin{subfigure}[b]{0.48\linewidth}
    \centering
    \includegraphics[width=\linewidth]{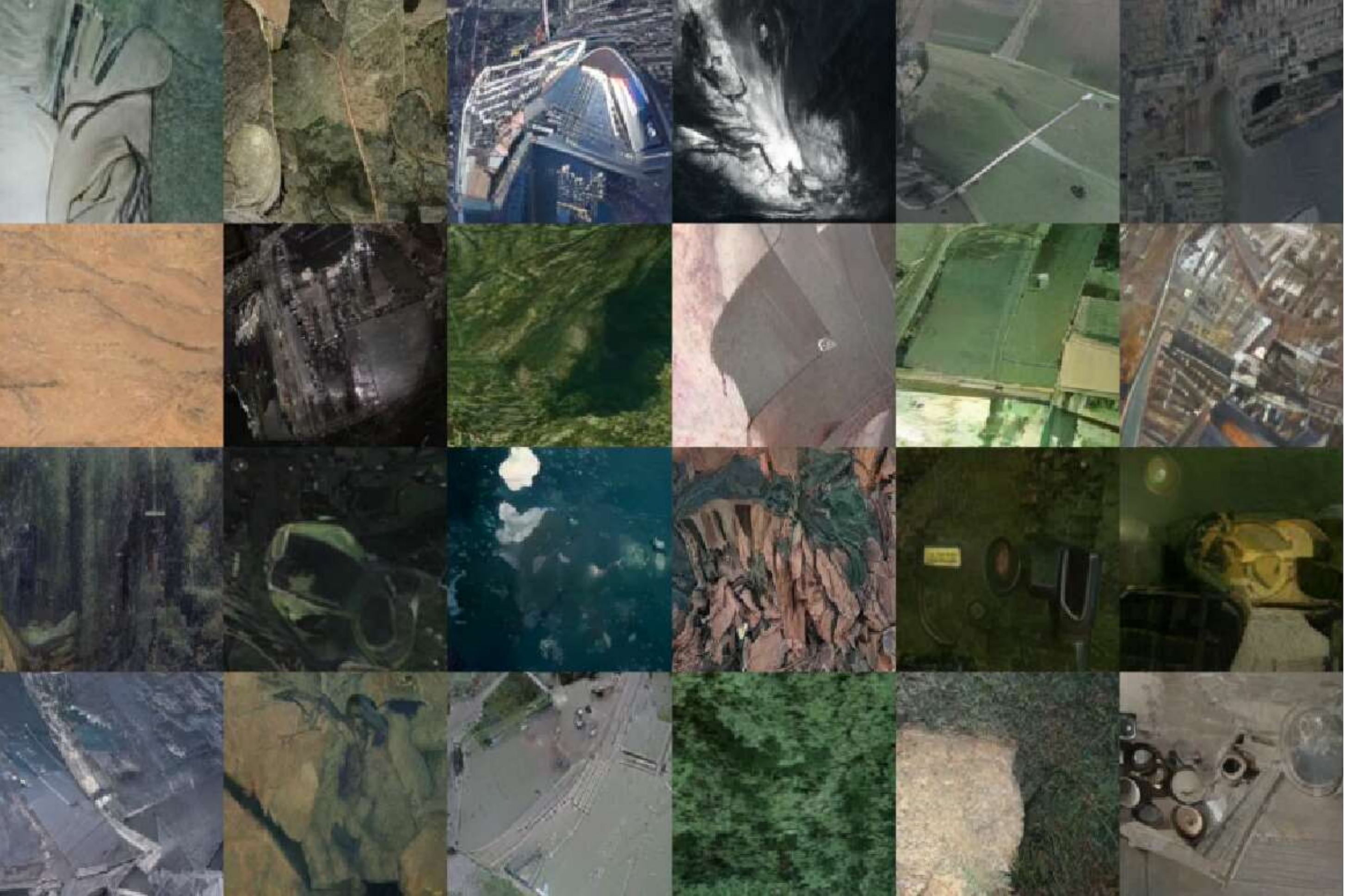}
    \caption{AR transformer with prompt tuning ($S\,{=}\,1$)}
    \label{fig:vtab_taming_prompt_s1_resisc45}
  \end{subfigure}
  \hspace{0.02in}
  \begin{subfigure}[b]{0.48\linewidth}
    \centering
    \includegraphics[width=\linewidth]{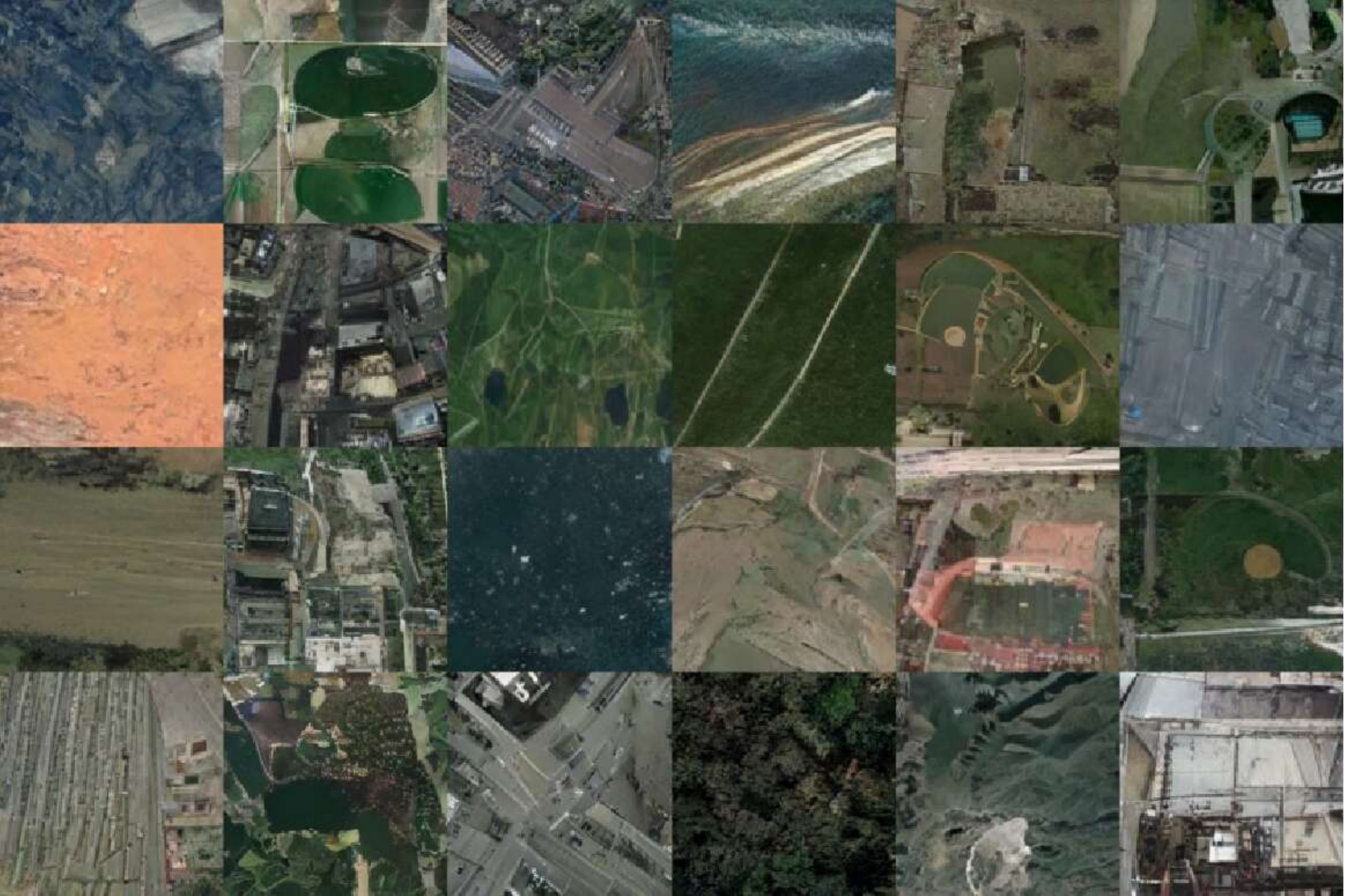}
    \caption{AR transformer with prompt tuning ($S\,{=}\,256$, $F\,{=}\,16$)}
    \label{fig:vtab_taming_prompt_s128_resisc45}
  \end{subfigure}
  \begin{subfigure}[b]{0.48\linewidth}
    \centering
    \includegraphics[width=\linewidth]{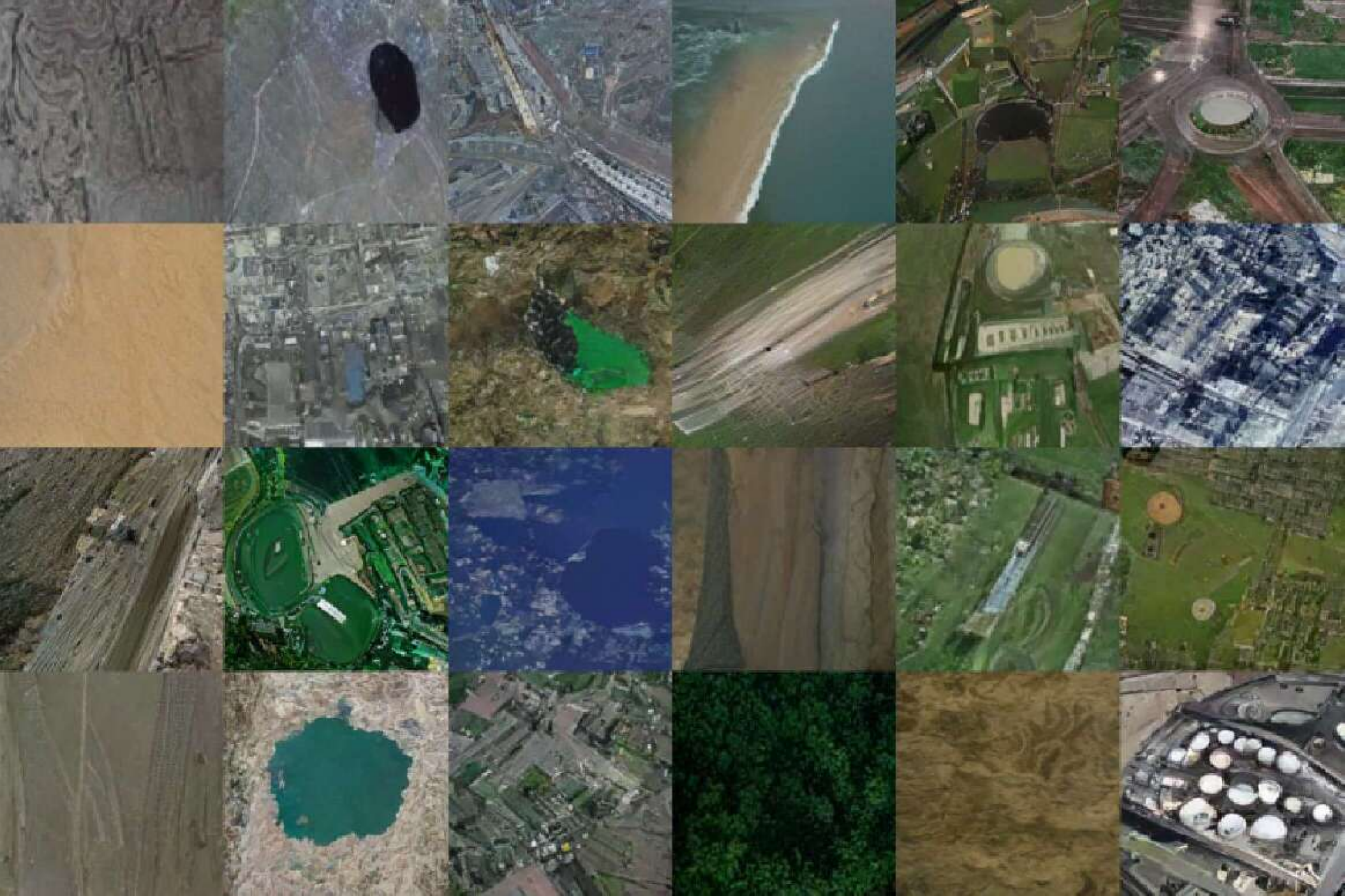}
    \caption{NAR transformer with prompt tuning ($S\,{=}\,1$)}
    \label{fig:vtab_maskgit_prompt_s1_resisc45}
  \end{subfigure}
  \hspace{0.02in}
  \begin{subfigure}[b]{0.48\linewidth}
    \centering
    \includegraphics[width=\linewidth]{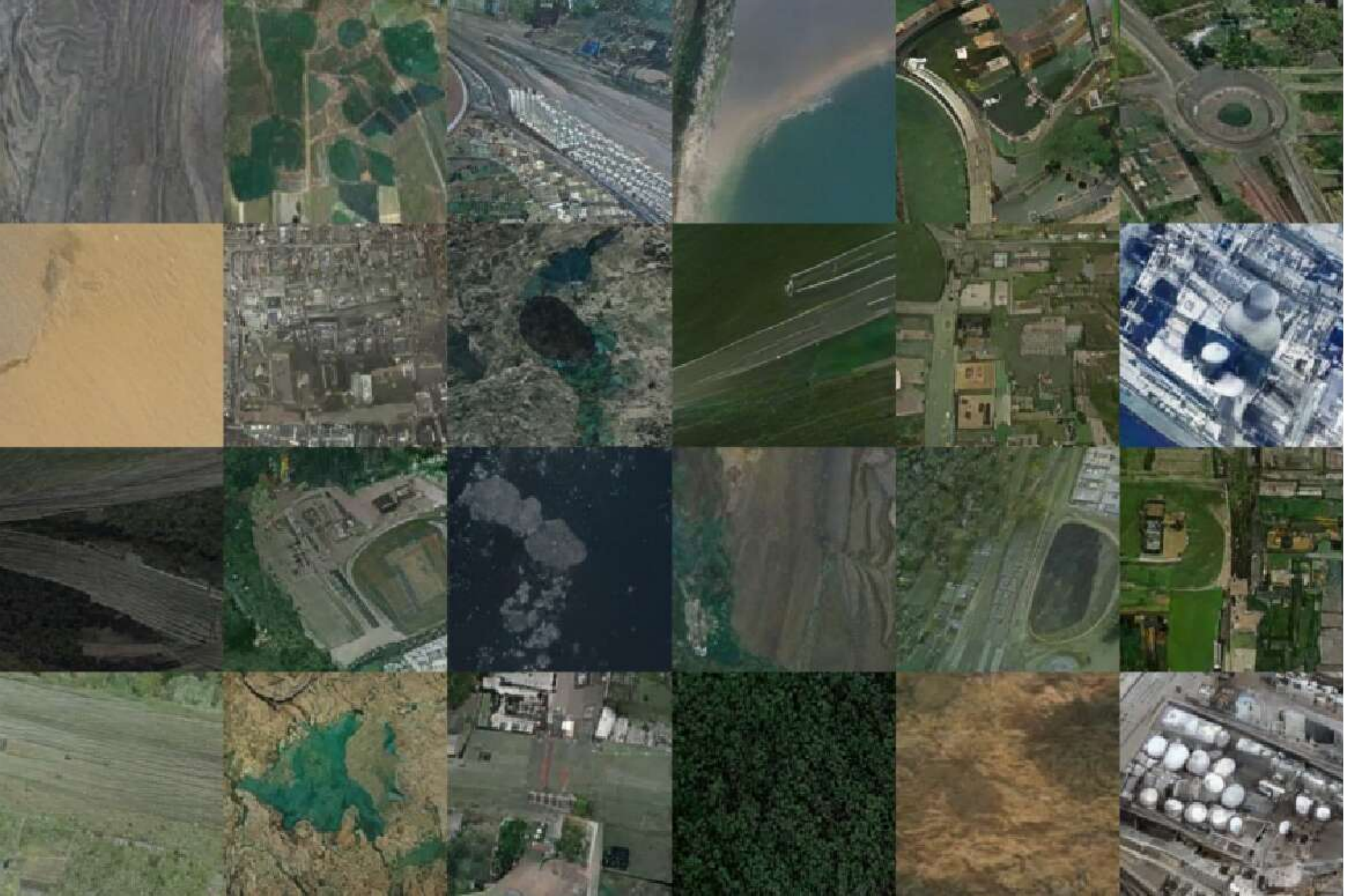}
    \caption{NAR transformer with prompt tuning ($S\,{=}\,128$)}
    \label{fig:vtab_maskgit_prompt_s128_resisc45}
  \end{subfigure}
  \caption{Visualization of generated images with different models on Resisc45 of VTAB.}
  \label{fig:vtab_supp_resisc45}
\end{figure}

\begin{figure}
  \centering
  \begin{subfigure}[b]{0.48\linewidth}
    \centering
    \includegraphics[width=\linewidth]{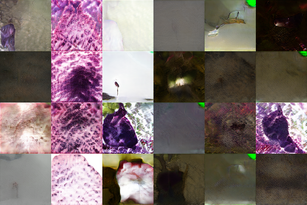}
    \caption{MineGAN}
    \label{fig:vtab_minegan_patch_camelyon}
  \end{subfigure}
  \hspace{0.02in}
  \begin{subfigure}[b]{0.48\linewidth}
    \centering
    \includegraphics[width=\linewidth]{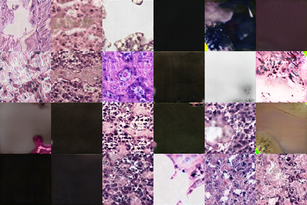}
    \caption{cGANTransfer}
    \label{fig:vtab_cgantrasnfer_patch_camelyon}
  \end{subfigure}
  \begin{subfigure}[b]{0.48\linewidth}
    \centering
    \includegraphics[width=\linewidth]{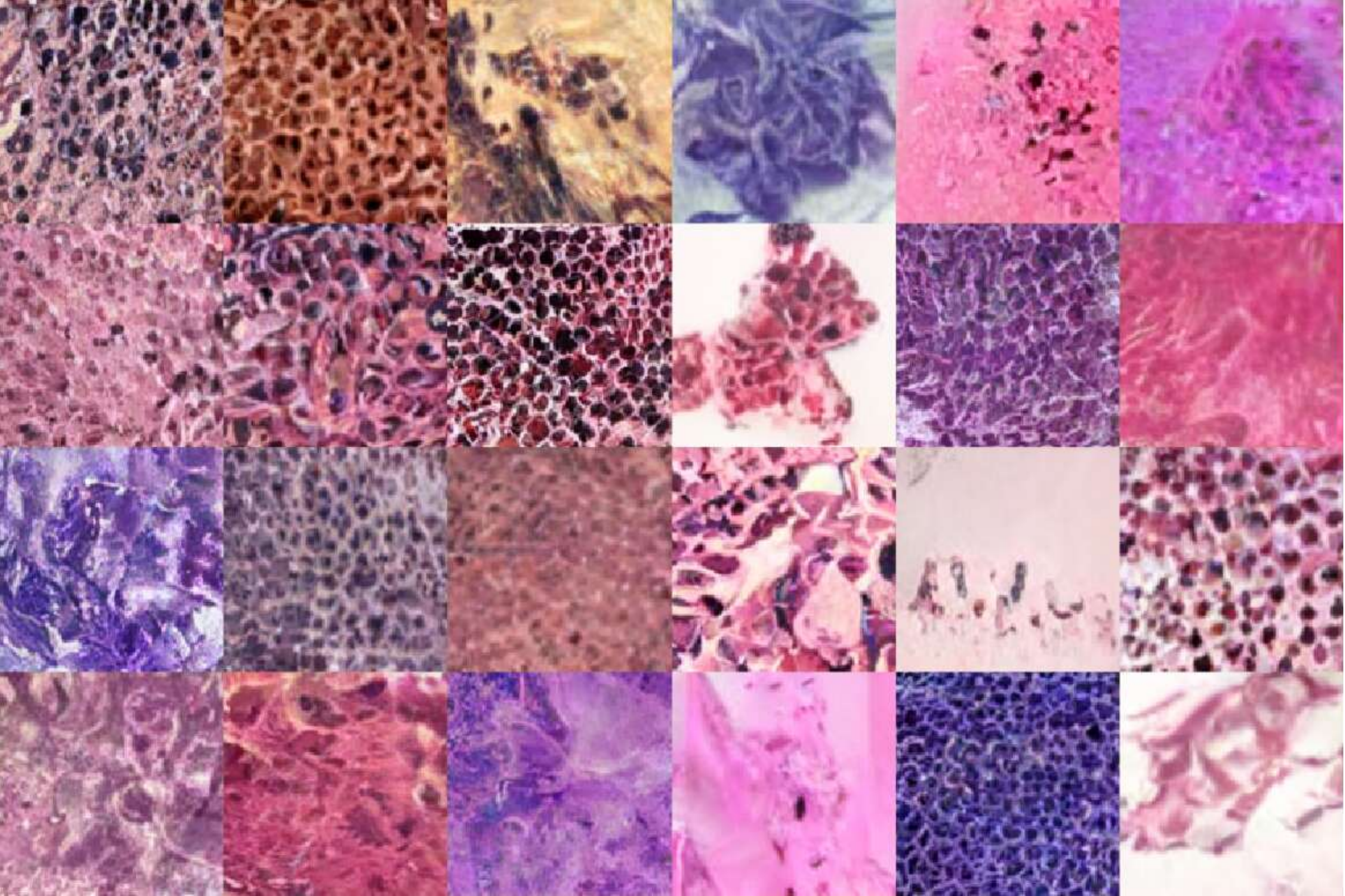}
    \caption{AR transformer with prompt tuning ($S\,{=}\,1$)}
    \label{fig:vtab_taming_prompt_s1_patch_camelyon}
  \end{subfigure}
  \hspace{0.02in}
  \begin{subfigure}[b]{0.48\linewidth}
    \centering
    \includegraphics[width=\linewidth]{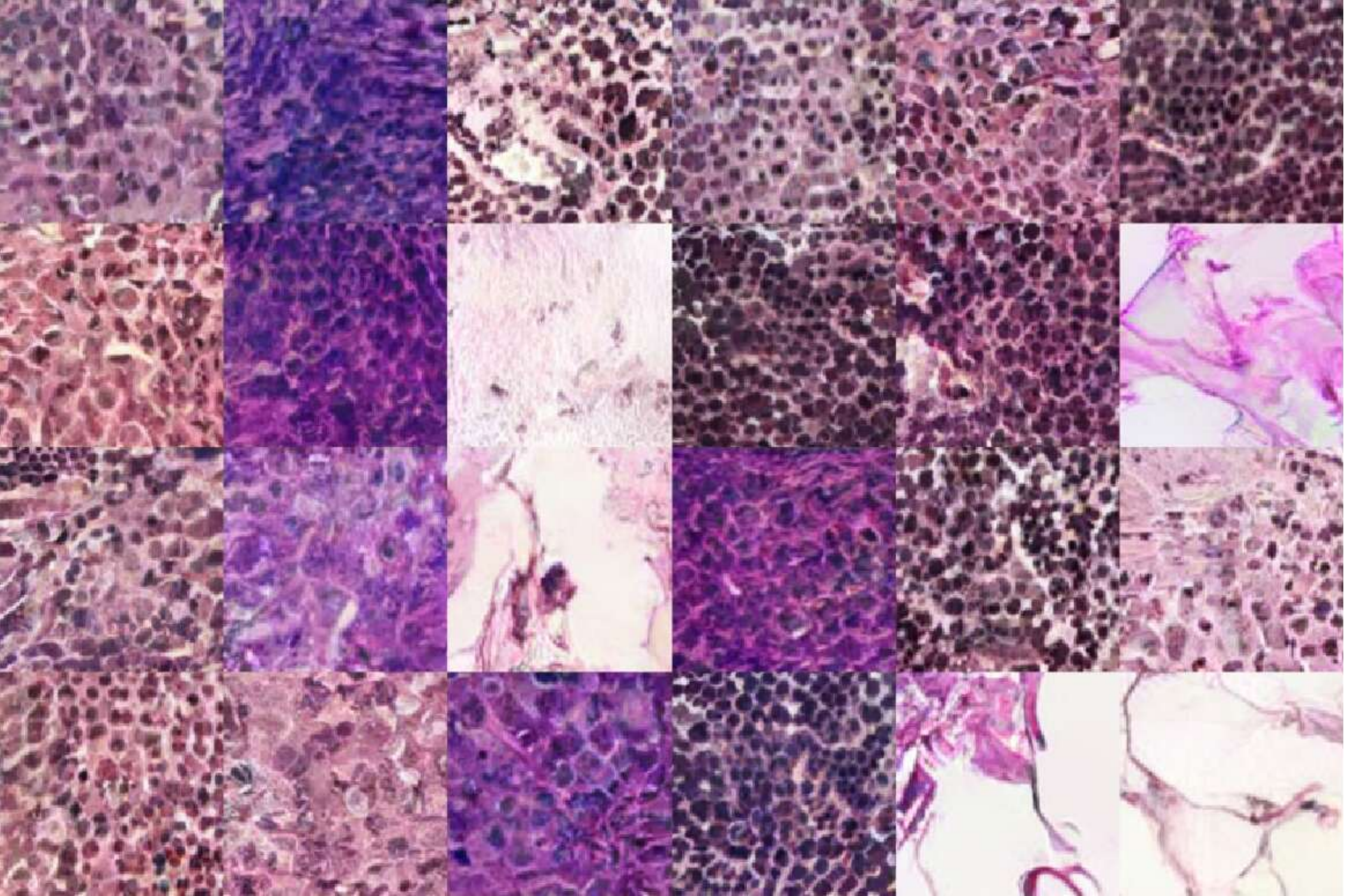}
    \caption{AR transformer with prompt tuning ($S\,{=}\,256$, $F\,{=}\,16$)}
    \label{fig:vtab_taming_prompt_s128_patch_camelyon}
  \end{subfigure}
  \begin{subfigure}[b]{0.48\linewidth}
    \centering
    \includegraphics[width=\linewidth]{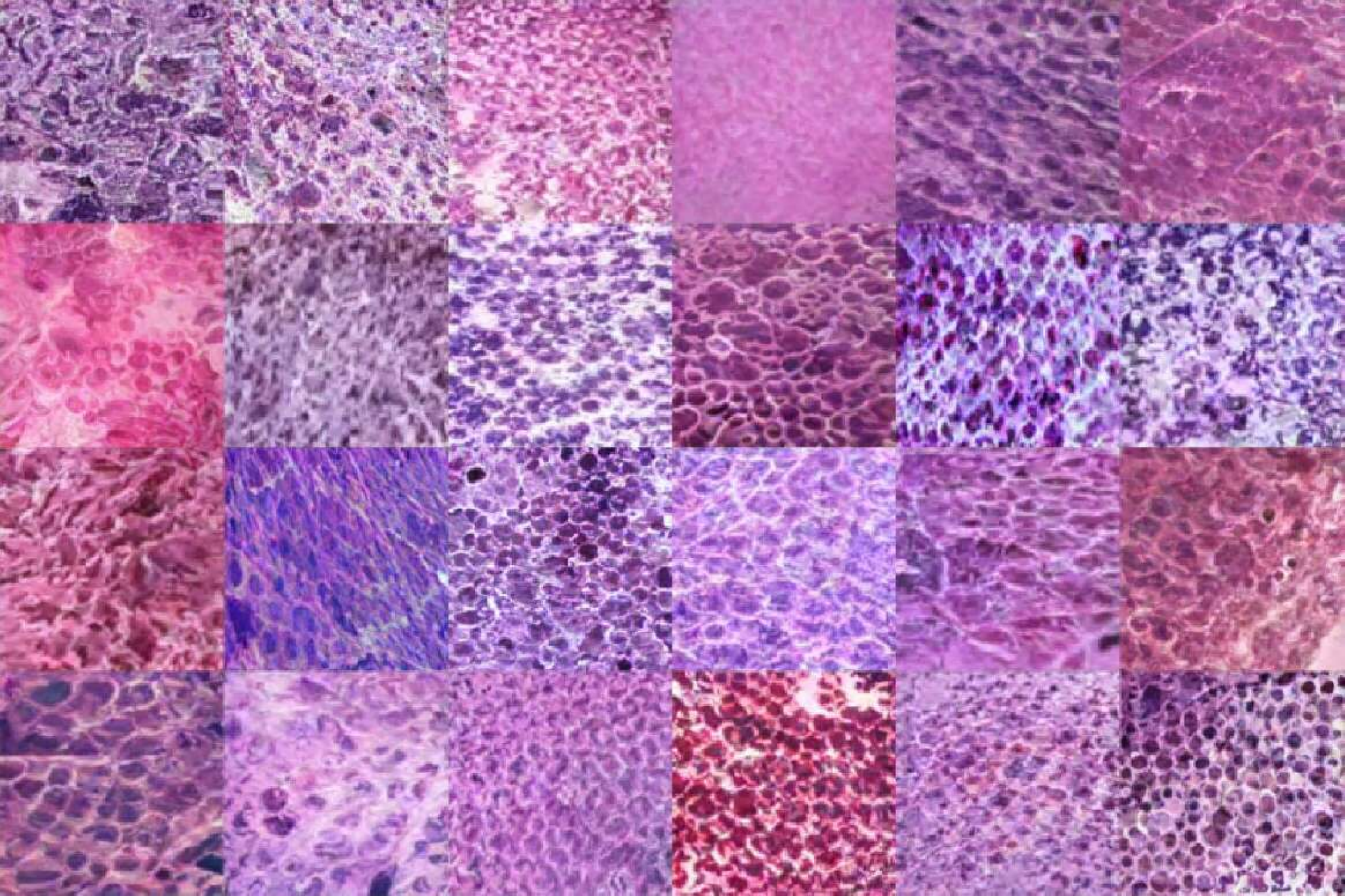}
    \caption{NAR transformer with prompt tuning ($S\,{=}\,1$)}
    \label{fig:vtab_maskgit_prompt_s1_patch_camelyon}
  \end{subfigure}
  \hspace{0.02in}
  \begin{subfigure}[b]{0.48\linewidth}
    \centering
    \includegraphics[width=\linewidth]{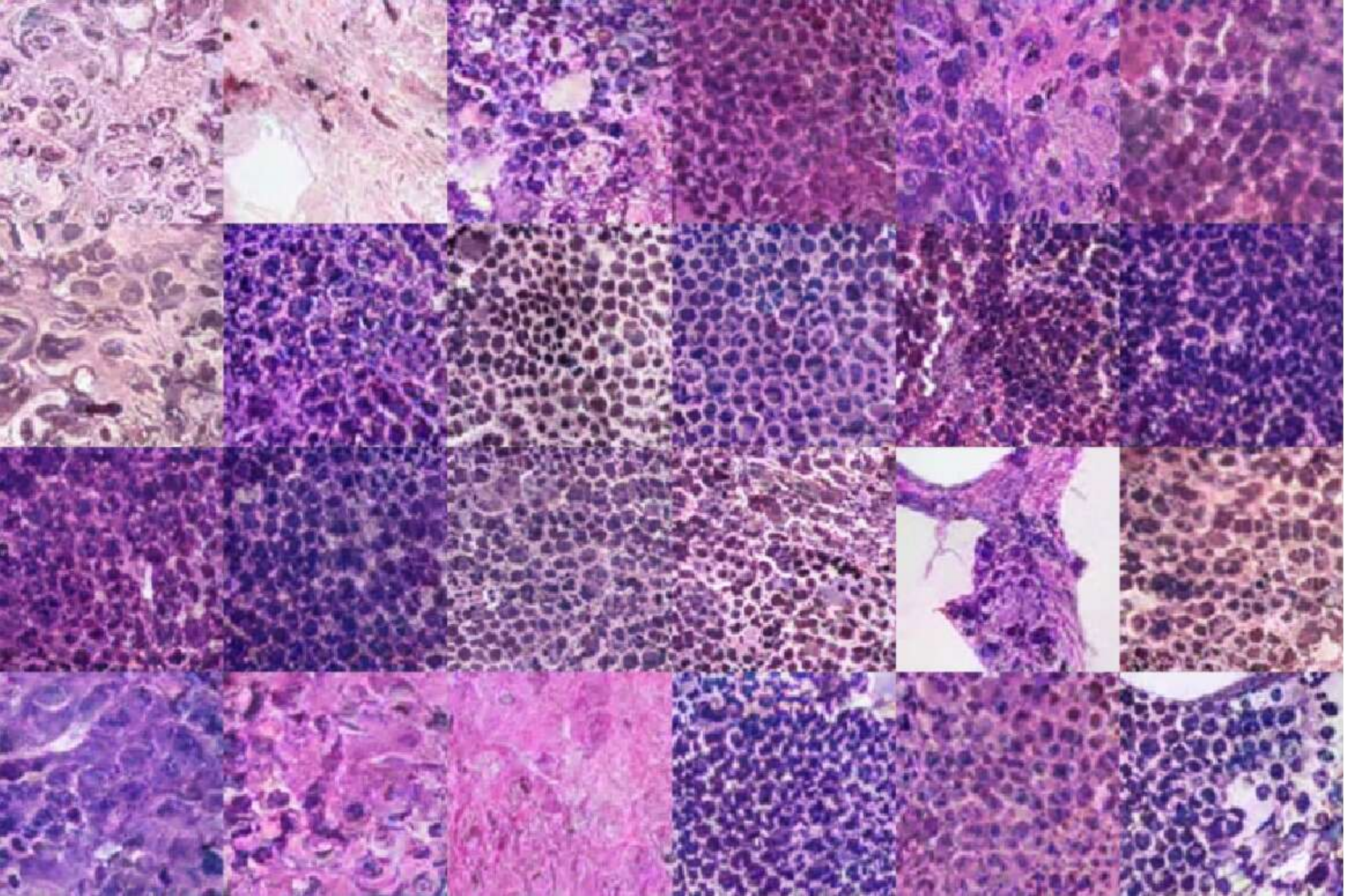}
    \caption{NAR transformer with prompt tuning ($S\,{=}\,128$)}
    \label{fig:vtab_maskgit_prompt_s128_patch_camelyon}
  \end{subfigure}
  \caption{Visualization of generated images with different models on Patch Camelyon of VTAB.}
  \label{fig:vtab_supp_patch_camelyon}
\end{figure}

\begin{figure}
  \centering
  \begin{subfigure}[b]{0.48\linewidth}
    \centering
    \includegraphics[width=\linewidth]{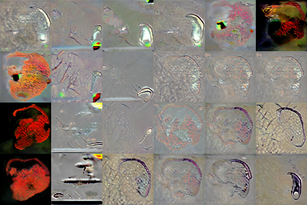}
    \caption{MineGAN}
    \label{fig:vtab_minegan_diabetic_retinopathy}
  \end{subfigure}
  \hspace{0.02in}
  \begin{subfigure}[b]{0.48\linewidth}
    \centering
    \includegraphics[width=\linewidth]{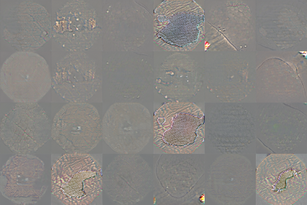}
    \caption{cGANTransfer}
    \label{fig:vtab_cgantrasnfer_diabetic_retinopathy}
  \end{subfigure}
  \begin{subfigure}[b]{0.48\linewidth}
    \centering
    \includegraphics[width=\linewidth]{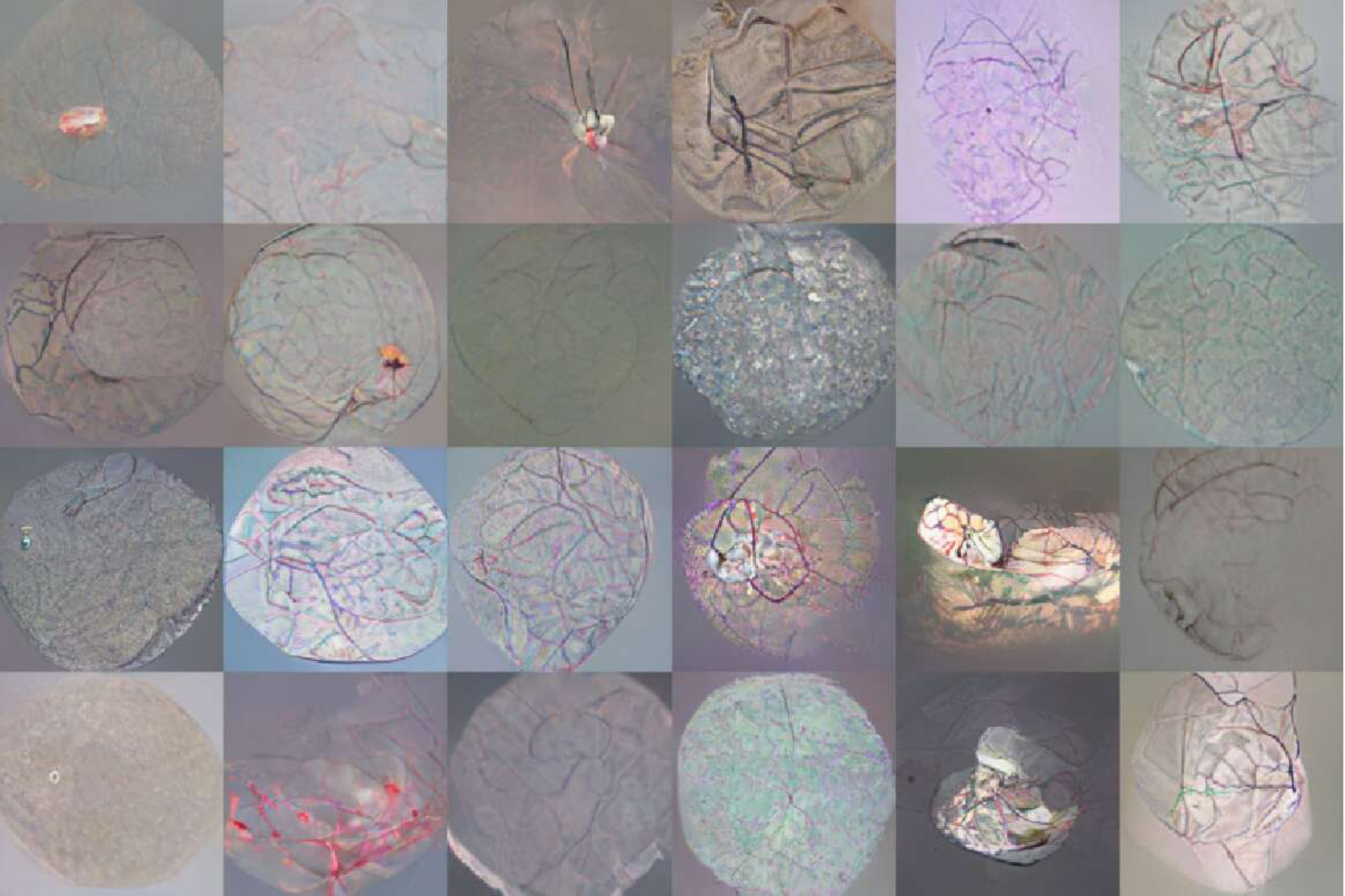}
    \caption{AR transformer with prompt tuning ($S\,{=}\,1$)}
    \label{fig:vtab_taming_prompt_s1_diabetic_retinopathy}
  \end{subfigure}
  \hspace{0.02in}
  \begin{subfigure}[b]{0.48\linewidth}
    \centering
    \includegraphics[width=\linewidth]{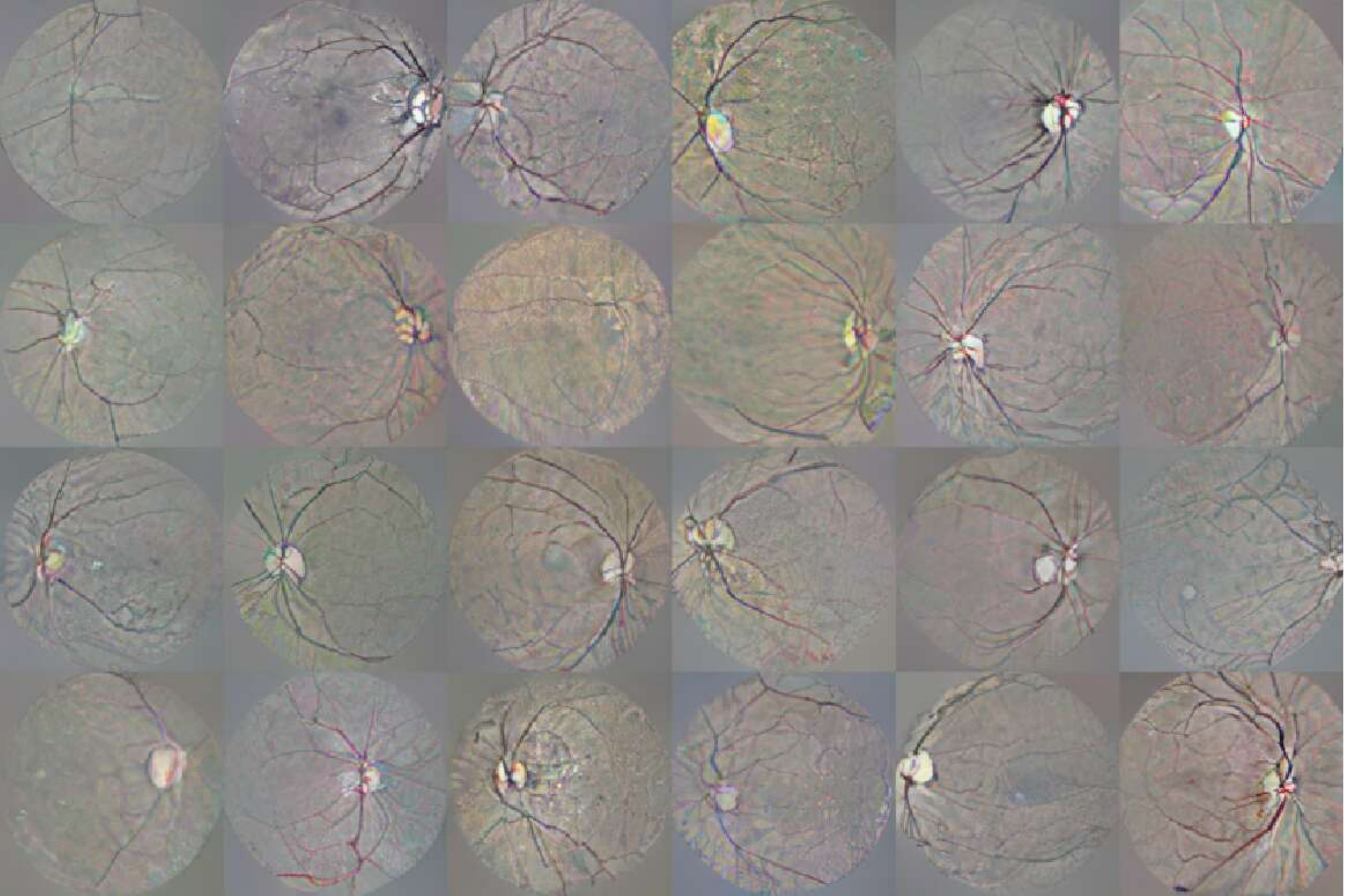}
    \caption{AR transformer with prompt tuning ($S\,{=}\,256$, $F\,{=}\,16$)}
    \label{fig:vtab_taming_prompt_s128_diabetic_retinopathy}
  \end{subfigure}
  \begin{subfigure}[b]{0.48\linewidth}
    \centering
    \includegraphics[width=\linewidth]{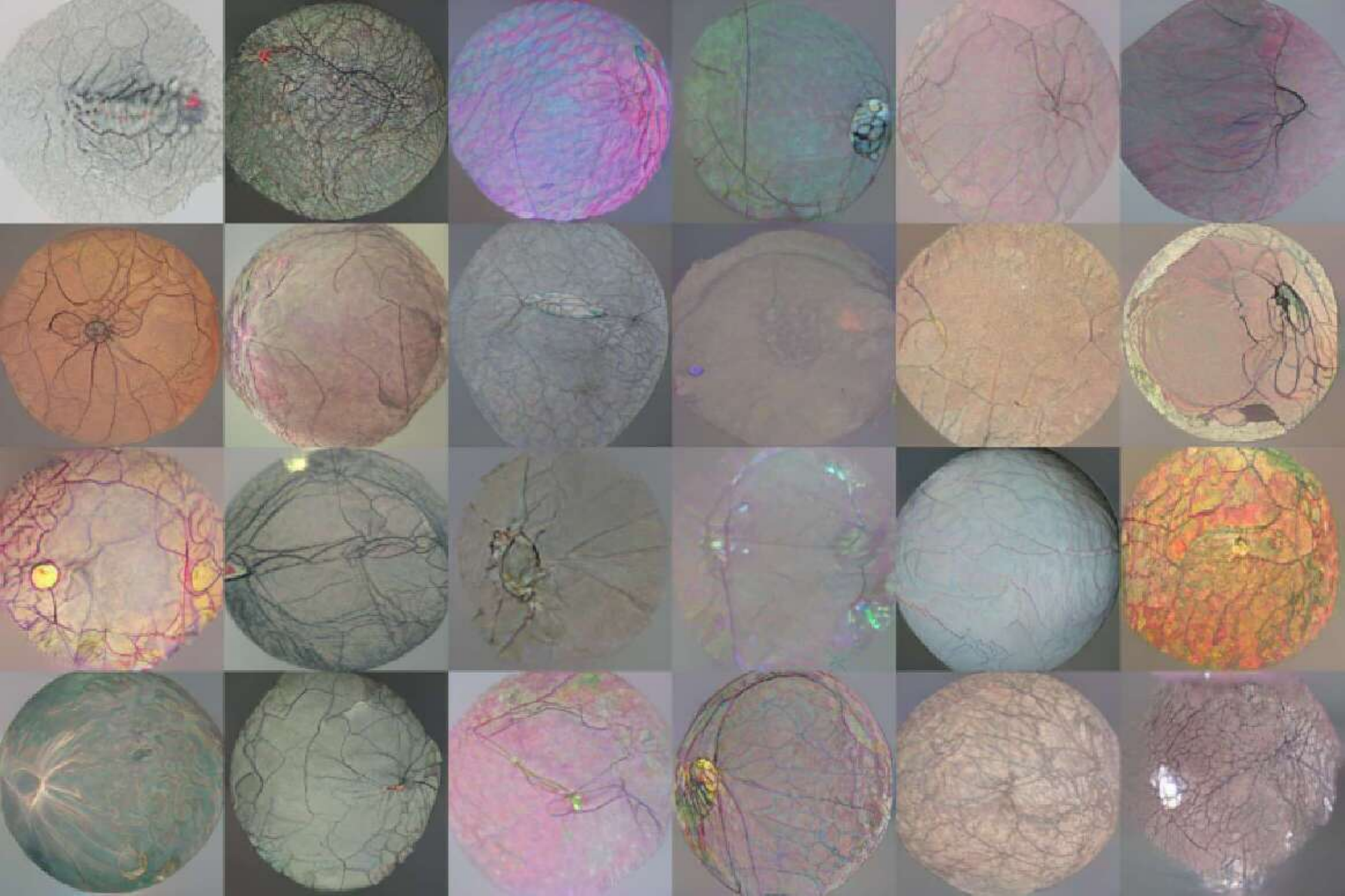}
    \caption{NAR transformer with prompt tuning ($S\,{=}\,1$)}
    \label{fig:vtab_maskgit_prompt_s1_diabetic_retinopathy}
  \end{subfigure}
  \hspace{0.02in}
  \begin{subfigure}[b]{0.48\linewidth}
    \centering
    \includegraphics[width=\linewidth]{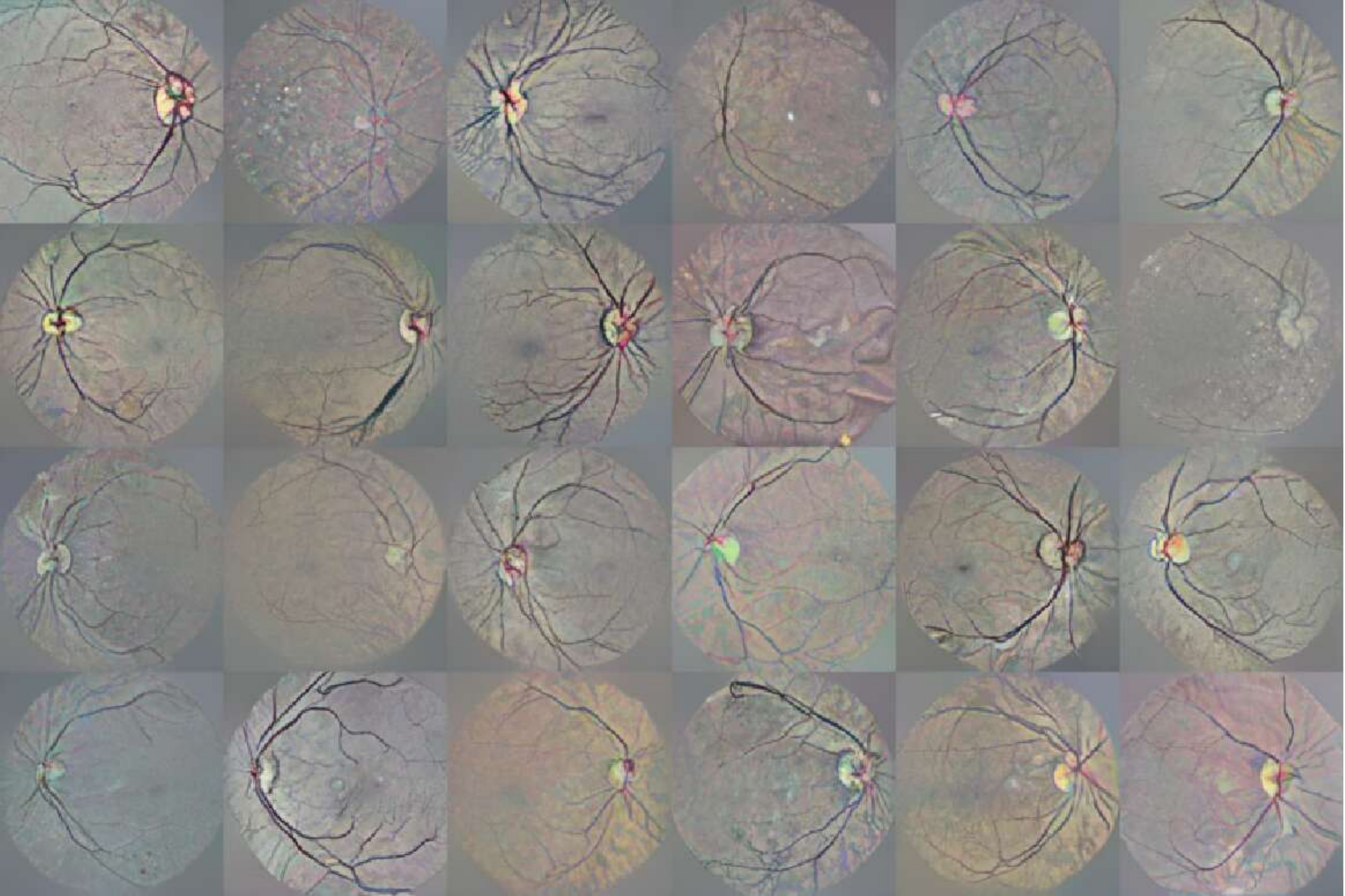}
    \caption{NAR transformer with prompt tuning ($S\,{=}\,128$)}
    \label{fig:vtab_maskgit_prompt_s128_diabetic_retinopathy}
  \end{subfigure}
  \caption{Visualization of generated images with different models on Diabetic Retinopathy of VTAB.}
  \label{fig:vtab_supp_diabetic_retinopathy}
\end{figure}

\begin{figure}
  \centering
  \begin{subfigure}[b]{0.48\linewidth}
    \centering
    \includegraphics[width=\linewidth]{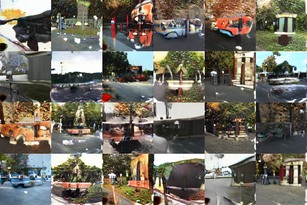}
    \caption{MineGAN}
    \label{fig:vtab_minegan_kitti}
  \end{subfigure}
  \hspace{0.02in}
  \begin{subfigure}[b]{0.48\linewidth}
    \centering
    \includegraphics[width=\linewidth]{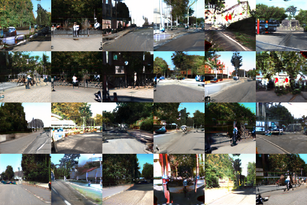}
    \caption{cGANTransfer}
    \label{fig:vtab_cgantrasnfer_kitti}
  \end{subfigure}
  \begin{subfigure}[b]{0.48\linewidth}
    \centering
    \includegraphics[width=\linewidth]{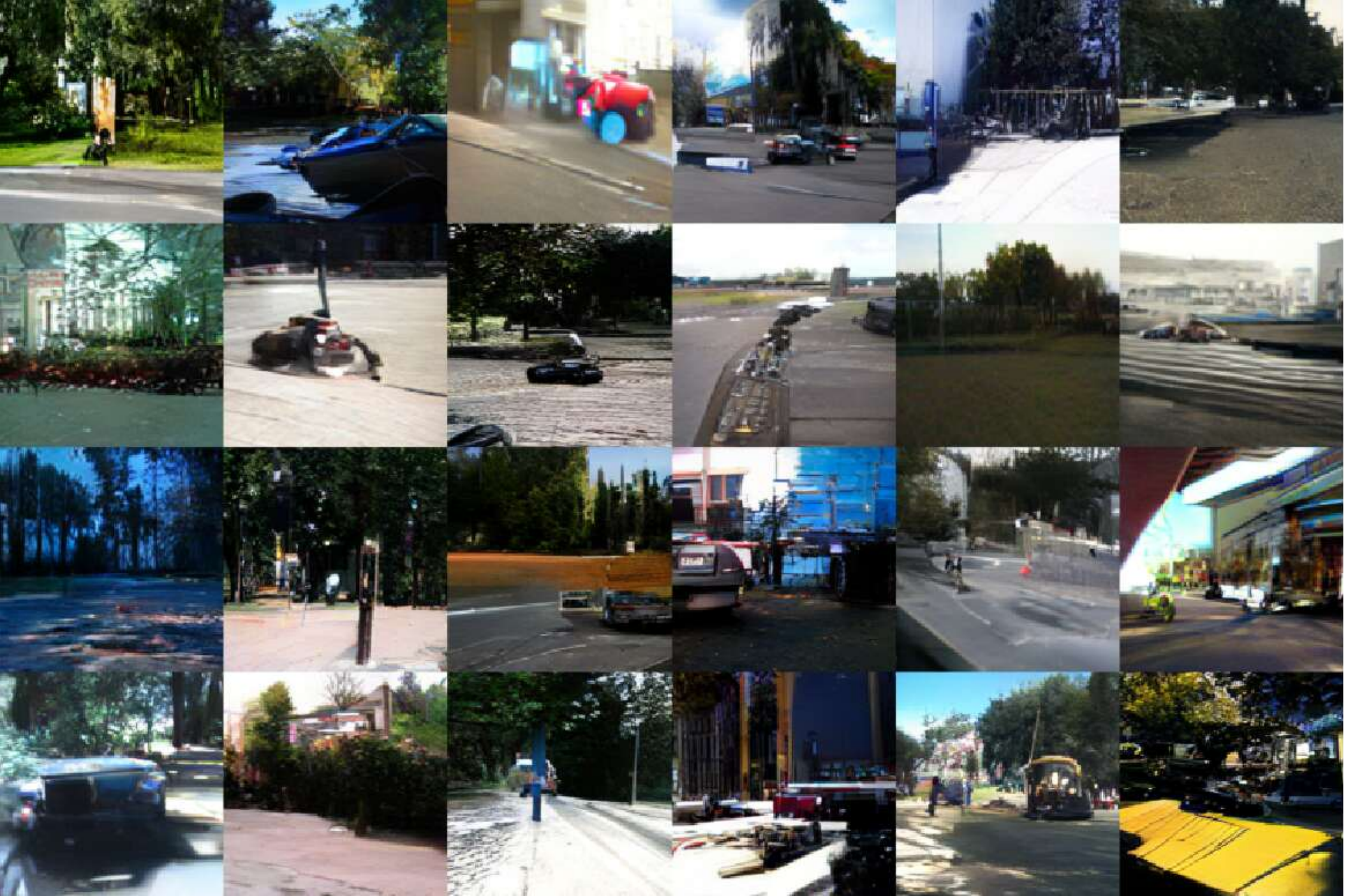}
    \caption{AR transformer with prompt tuning ($S\,{=}\,1$)}
    \label{fig:vtab_taming_prompt_s1_kitti}
  \end{subfigure}
  \hspace{0.02in}
  \begin{subfigure}[b]{0.48\linewidth}
    \centering
    \includegraphics[width=\linewidth]{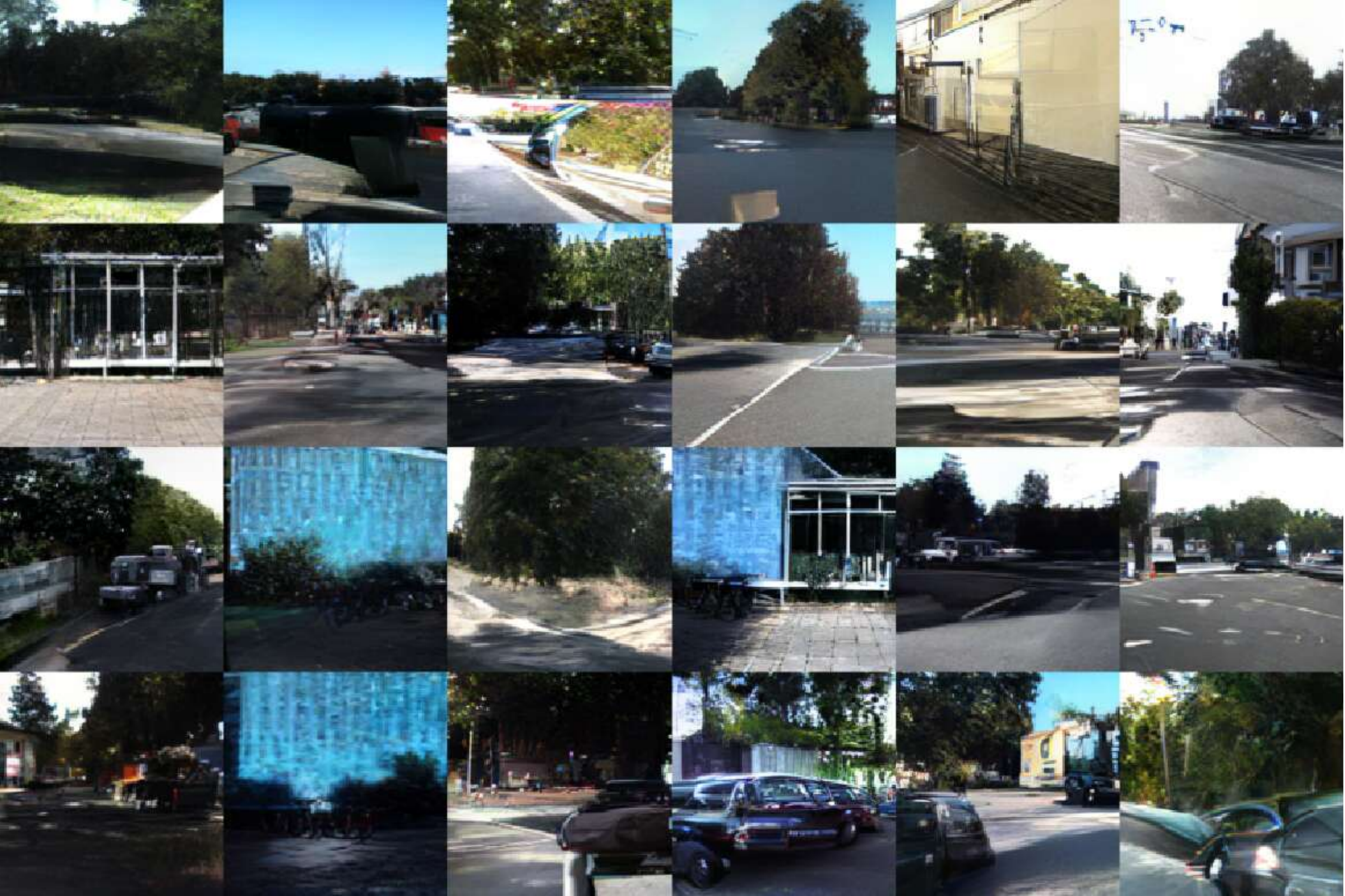}
    \caption{AR transformer with prompt tuning ($S\,{=}\,256$, $F\,{=}\,16$)}
    \label{fig:vtab_taming_prompt_s128_kitti}
  \end{subfigure}
  \begin{subfigure}[b]{0.48\linewidth}
    \centering
    \includegraphics[width=\linewidth]{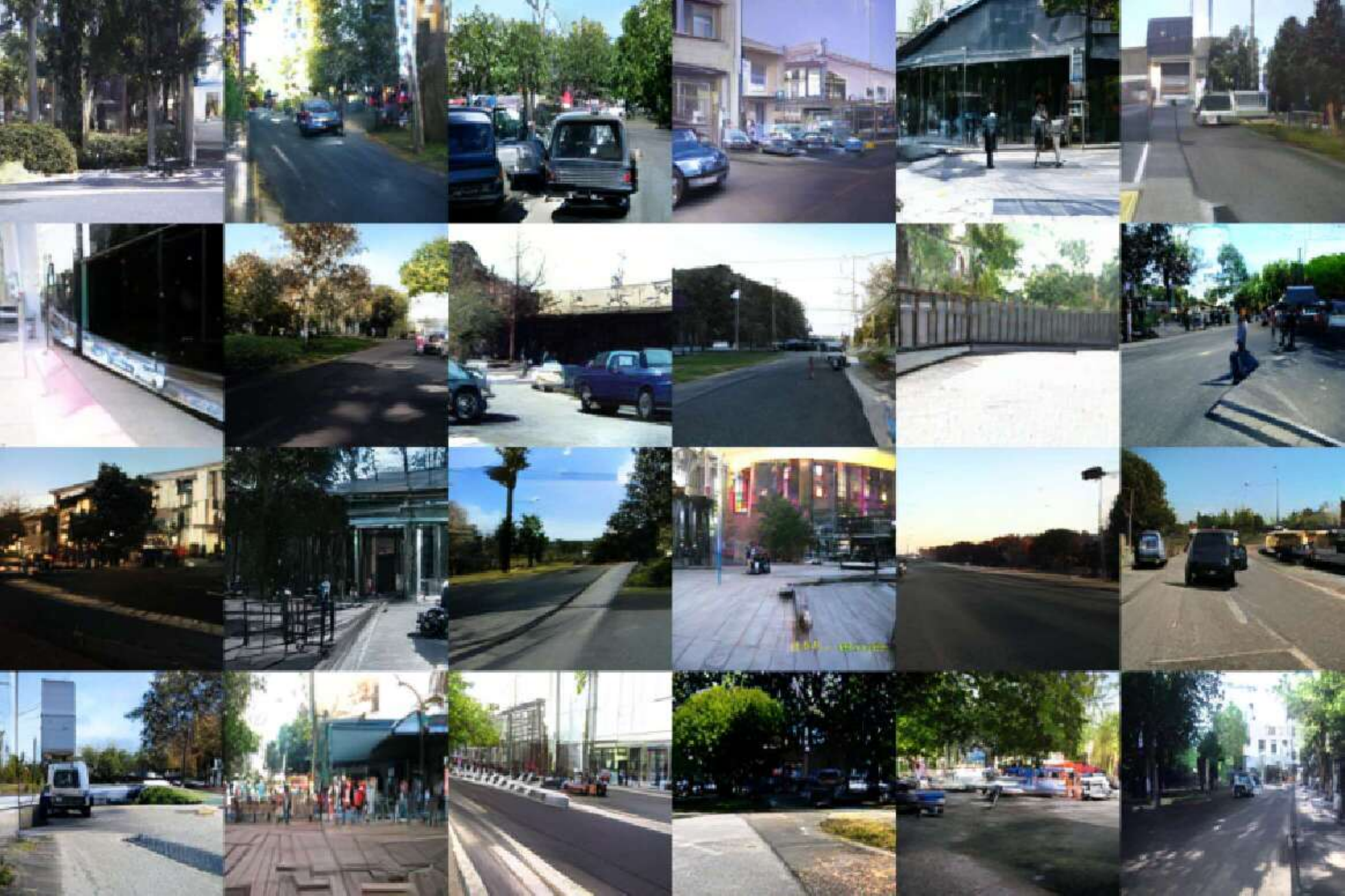}
    \caption{NAR transformer with prompt tuning ($S\,{=}\,1$)}
    \label{fig:vtab_maskgit_prompt_s1_kitti}
  \end{subfigure}
  \hspace{0.02in}
  \begin{subfigure}[b]{0.48\linewidth}
    \centering
    \includegraphics[width=\linewidth]{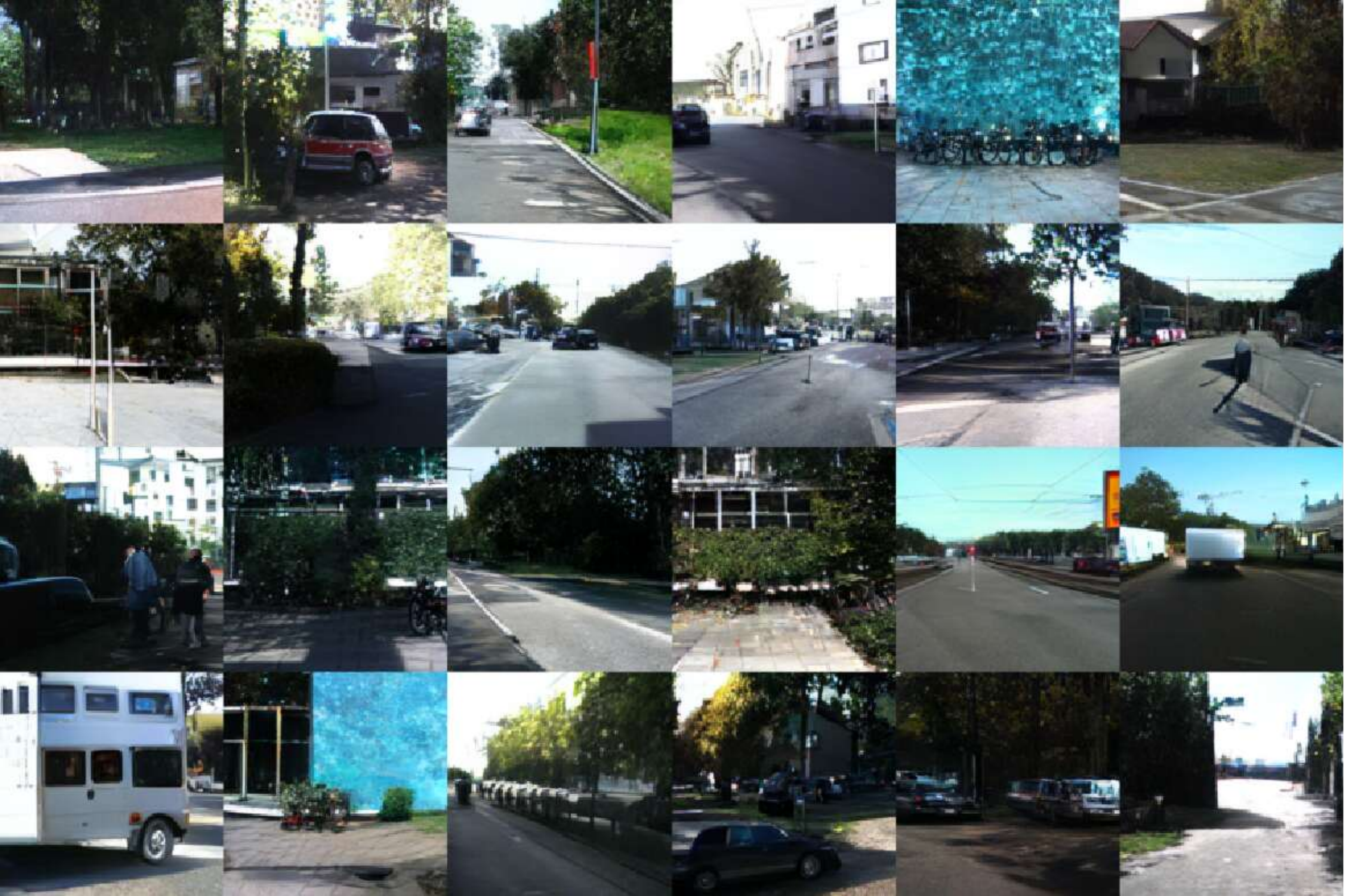}
    \caption{NAR transformer with prompt tuning ($S\,{=}\,128$)}
    \label{fig:vtab_maskgit_prompt_s128_kitti}
  \end{subfigure}
  \caption{Visualization of generated images with different models on Kitti of VTAB.}
  \label{fig:vtab_supp_kitti}
\end{figure}

\begin{figure}
  \centering
  \begin{subfigure}[b]{0.48\linewidth}
    \centering
    \includegraphics[width=\linewidth]{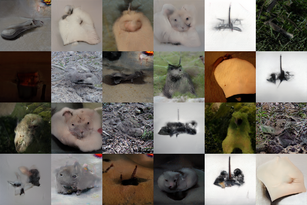}
    \caption{MineGAN}
    \label{fig:vtab_minegan_smallnorb}
  \end{subfigure}
  \hspace{0.02in}
  \begin{subfigure}[b]{0.48\linewidth}
    \centering
    \includegraphics[width=\linewidth]{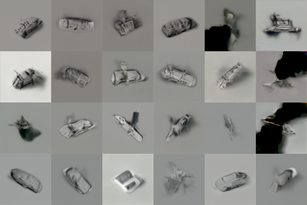}
    \caption{cGANTransfer}
    \label{fig:vtab_cgantrasnfer_smallnorb}
  \end{subfigure}
  \begin{subfigure}[b]{0.48\linewidth}
    \centering
    \includegraphics[width=\linewidth]{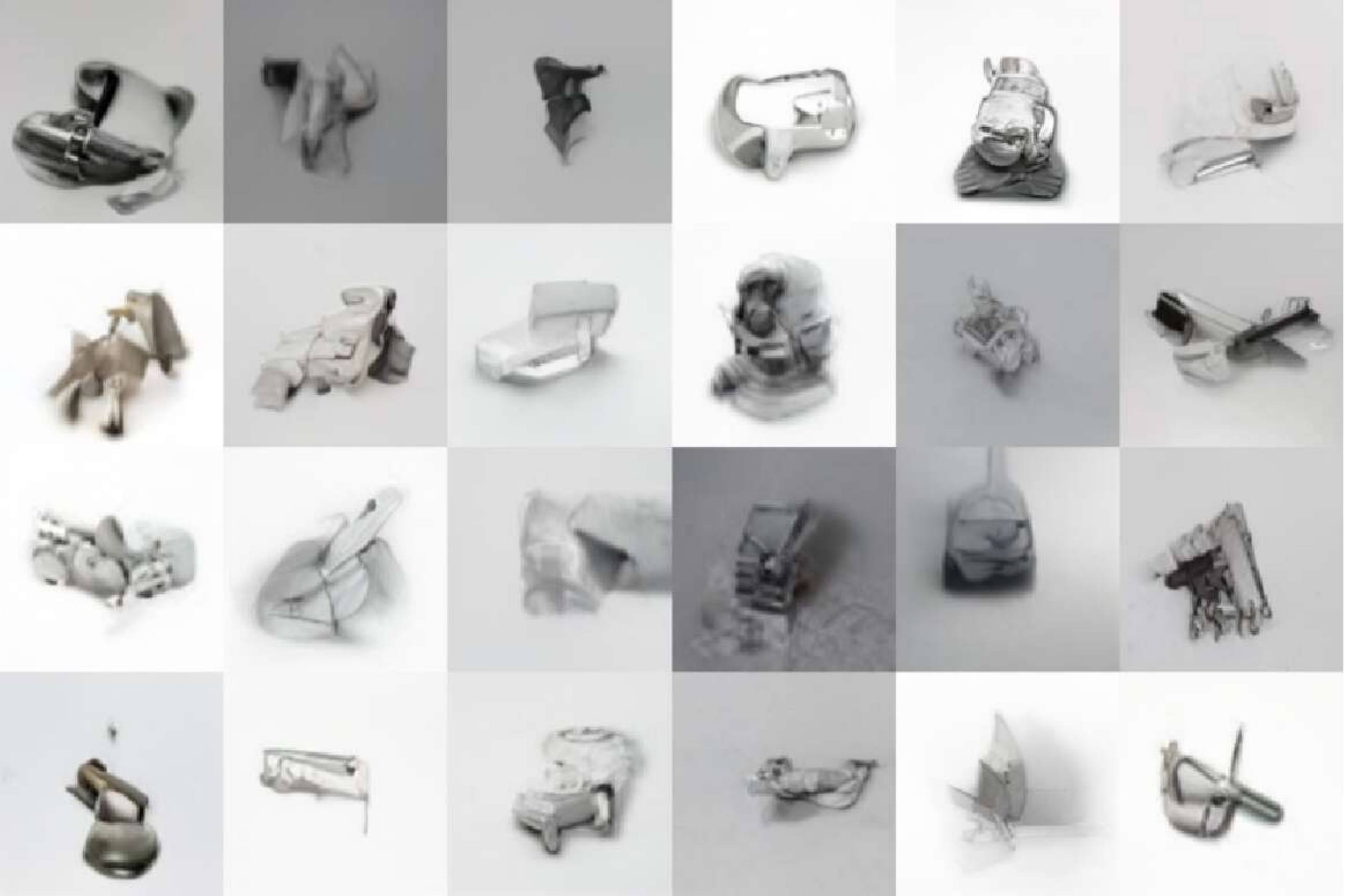}
    \caption{AR transformer with prompt tuning ($S\,{=}\,1$)}
    \label{fig:vtab_taming_prompt_s1_smallnorb}
  \end{subfigure}
  \hspace{0.02in}
  \begin{subfigure}[b]{0.48\linewidth}
    \centering
    \includegraphics[width=\linewidth]{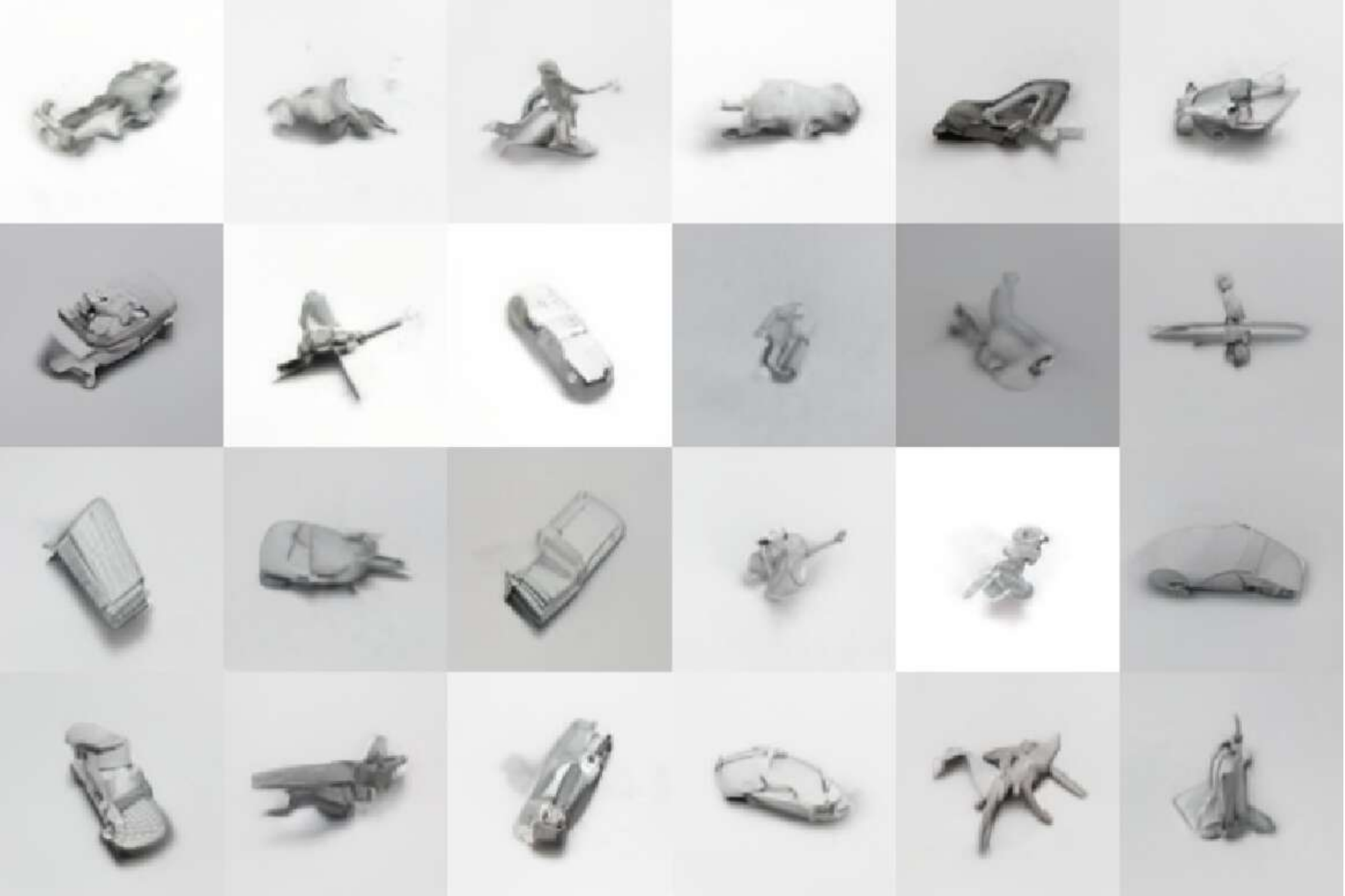}
    \caption{AR transformer with prompt tuning ($S\,{=}\,256$, $F\,{=}\,16$)}
    \label{fig:vtab_taming_prompt_s128_smallnorb}
  \end{subfigure}
  \begin{subfigure}[b]{0.48\linewidth}
    \centering
    \includegraphics[width=\linewidth]{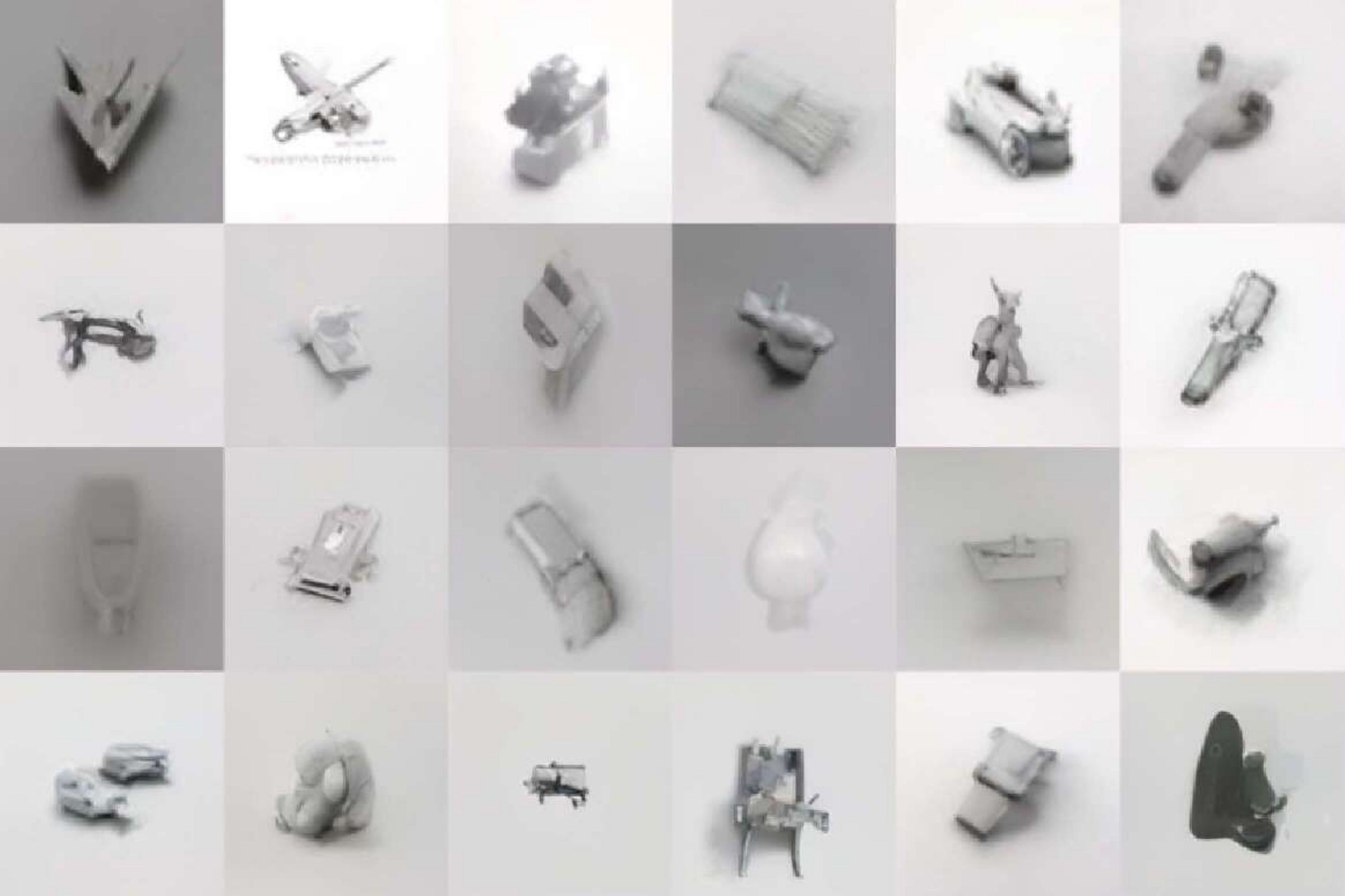}
    \caption{NAR transformer with prompt tuning ($S\,{=}\,1$)}
    \label{fig:vtab_maskgit_prompt_s1_smallnorb}
  \end{subfigure}
  \hspace{0.02in}
  \begin{subfigure}[b]{0.48\linewidth}
    \centering
    \includegraphics[width=\linewidth]{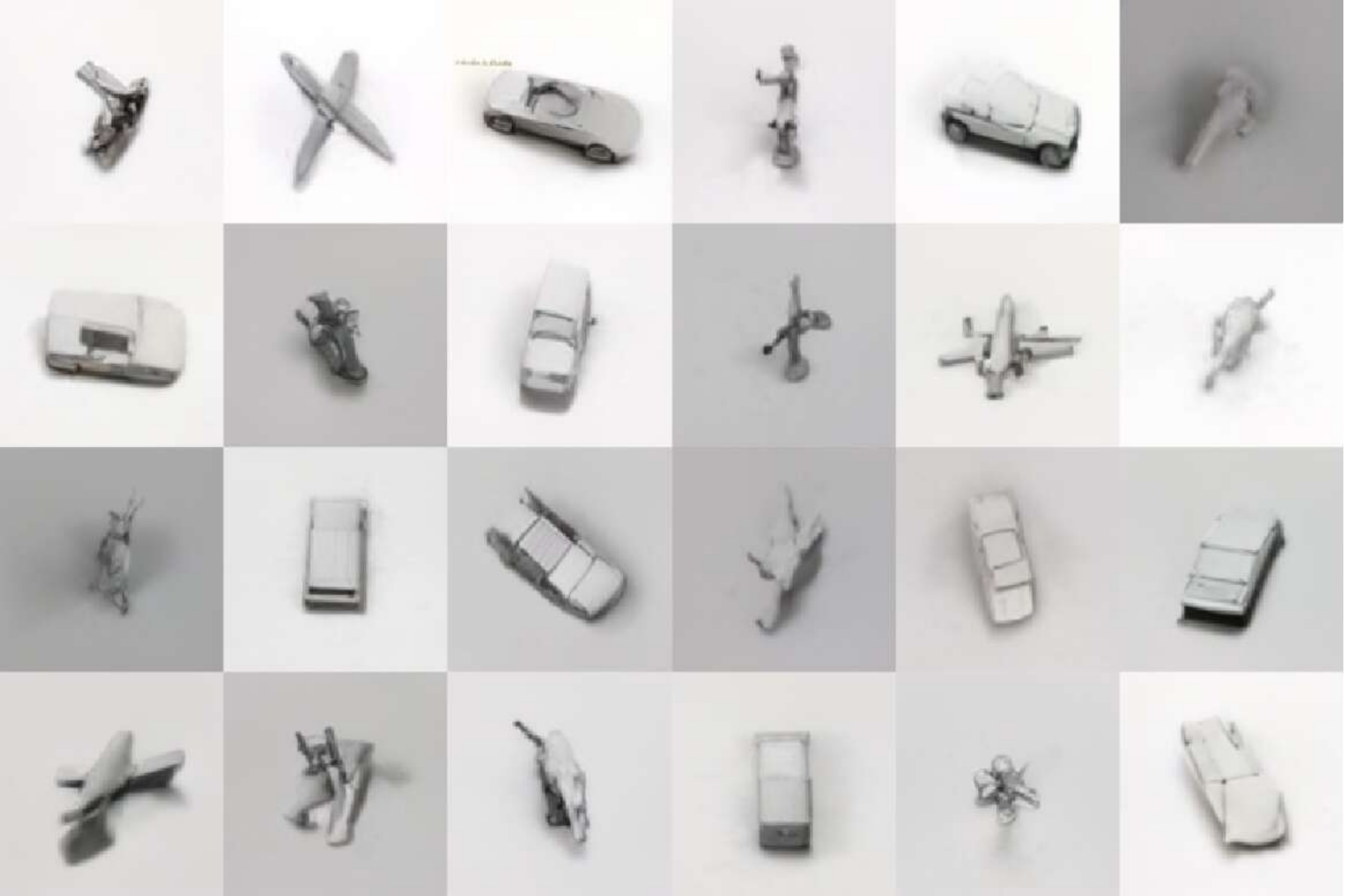}
    \caption{NAR transformer with prompt tuning ($S\,{=}\,128$)}
    \label{fig:vtab_maskgit_prompt_s128_smallnorb}
  \end{subfigure}
  \caption{Visualization of generated images with different models on Smallnorb of VTAB.}
  \label{fig:vtab_supp_smallnorb}
\end{figure}

\begin{figure}
  \centering
  \begin{subfigure}[b]{0.48\linewidth}
    \centering
    \includegraphics[width=\linewidth]{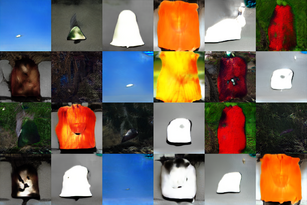}
    \caption{MineGAN}
    \label{fig:vtab_minegan_dsprites}
  \end{subfigure}
  \hspace{0.02in}
  \begin{subfigure}[b]{0.48\linewidth}
    \centering
    \includegraphics[width=\linewidth]{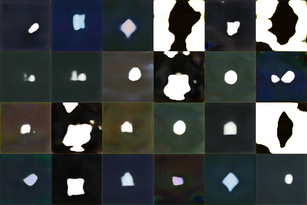}
    \caption{cGANTransfer}
    \label{fig:vtab_cgantrasnfer_dsprites}
  \end{subfigure}
  \begin{subfigure}[b]{0.48\linewidth}
    \centering
    \includegraphics[width=\linewidth]{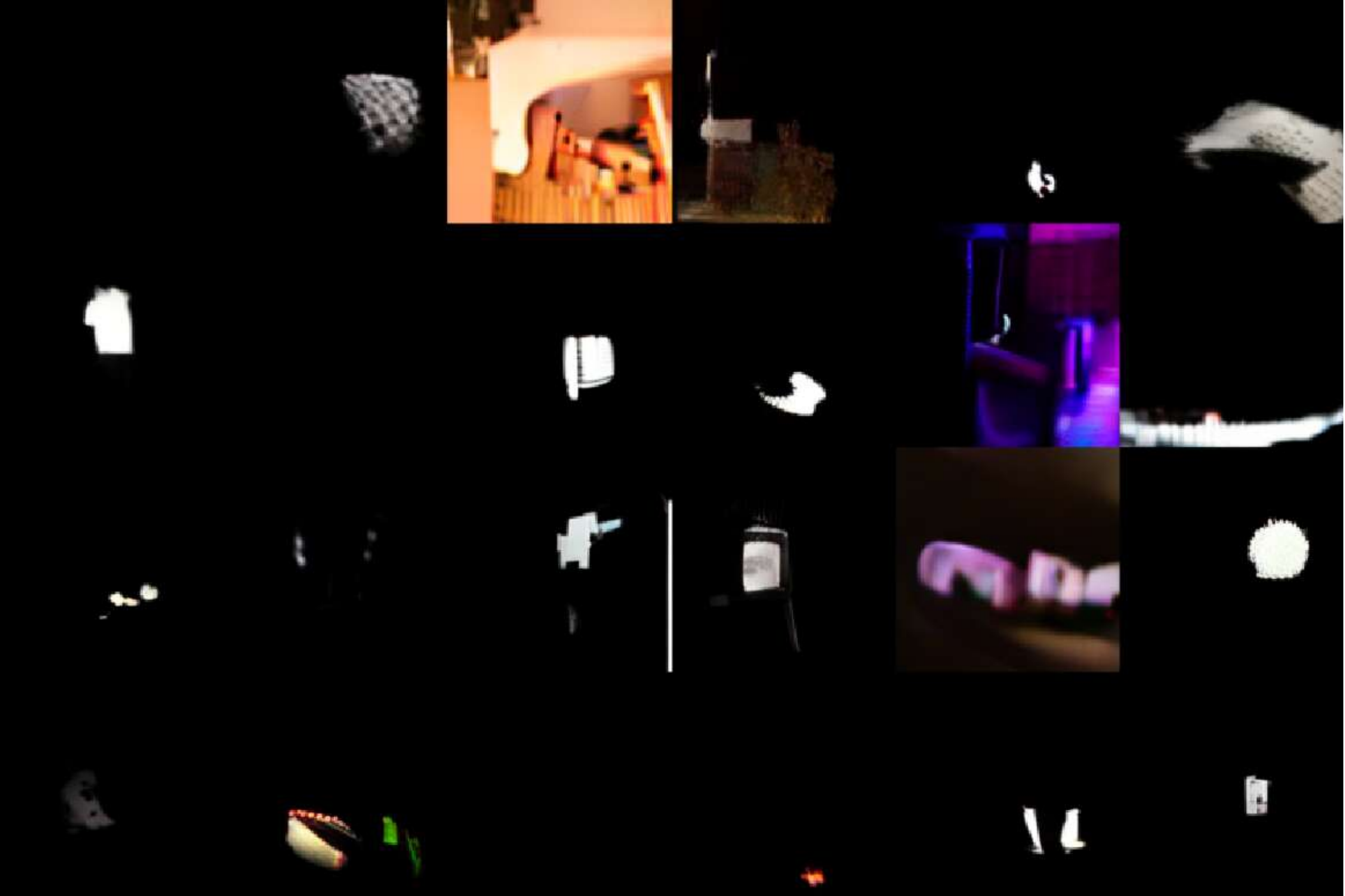}
    \caption{AR transformer with prompt tuning ($S\,{=}\,1$)}
    \label{fig:vtab_taming_prompt_s1_dsprites}
  \end{subfigure}
  \hspace{0.02in}
  \begin{subfigure}[b]{0.48\linewidth}
    \centering
    \includegraphics[width=\linewidth]{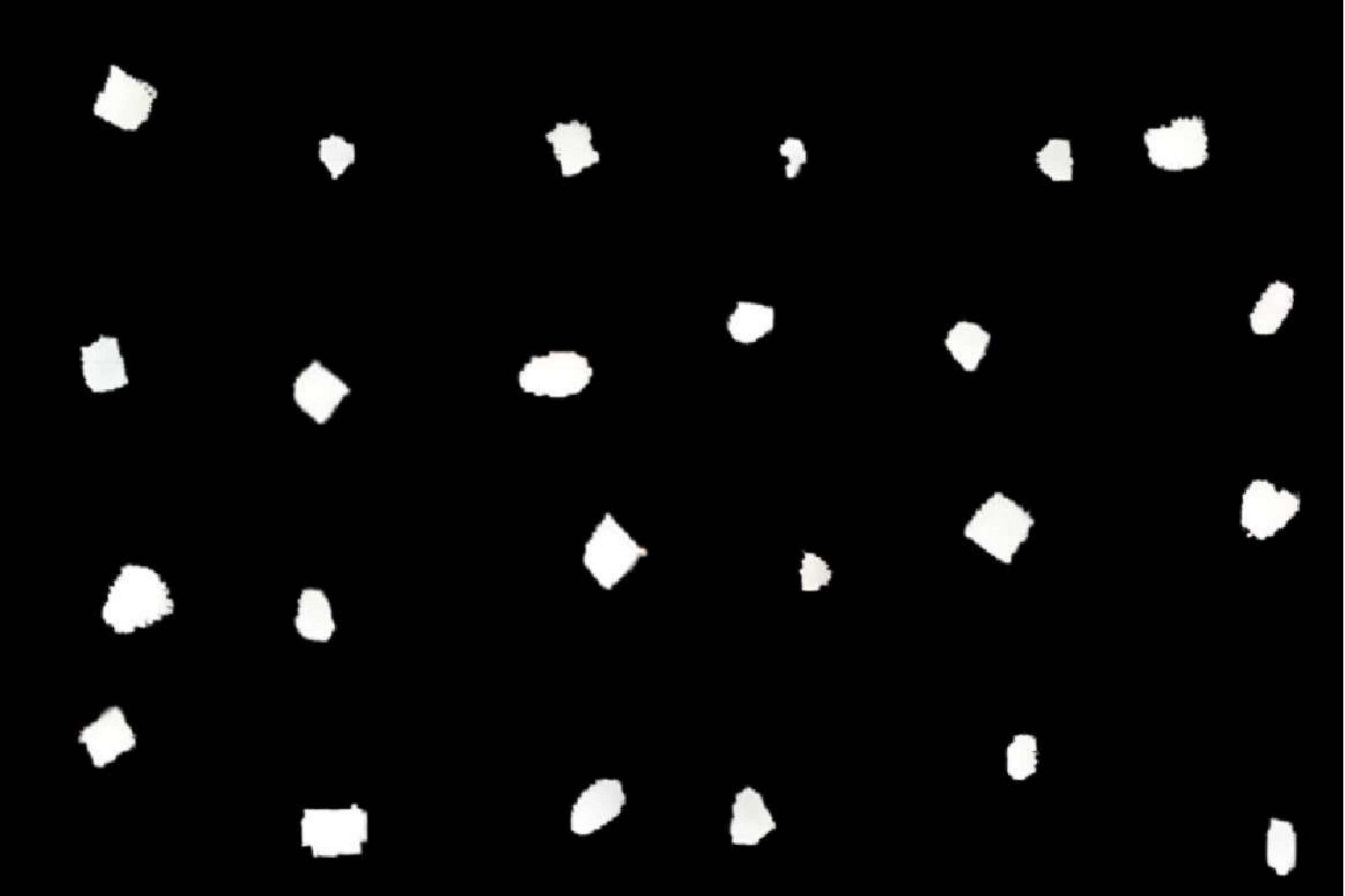}
    \caption{AR transformer with prompt tuning ($S\,{=}\,256$, $F\,{=}\,16$)}
    \label{fig:vtab_taming_prompt_s128_dsprites}
  \end{subfigure}
  \begin{subfigure}[b]{0.48\linewidth}
    \centering
    \includegraphics[width=\linewidth]{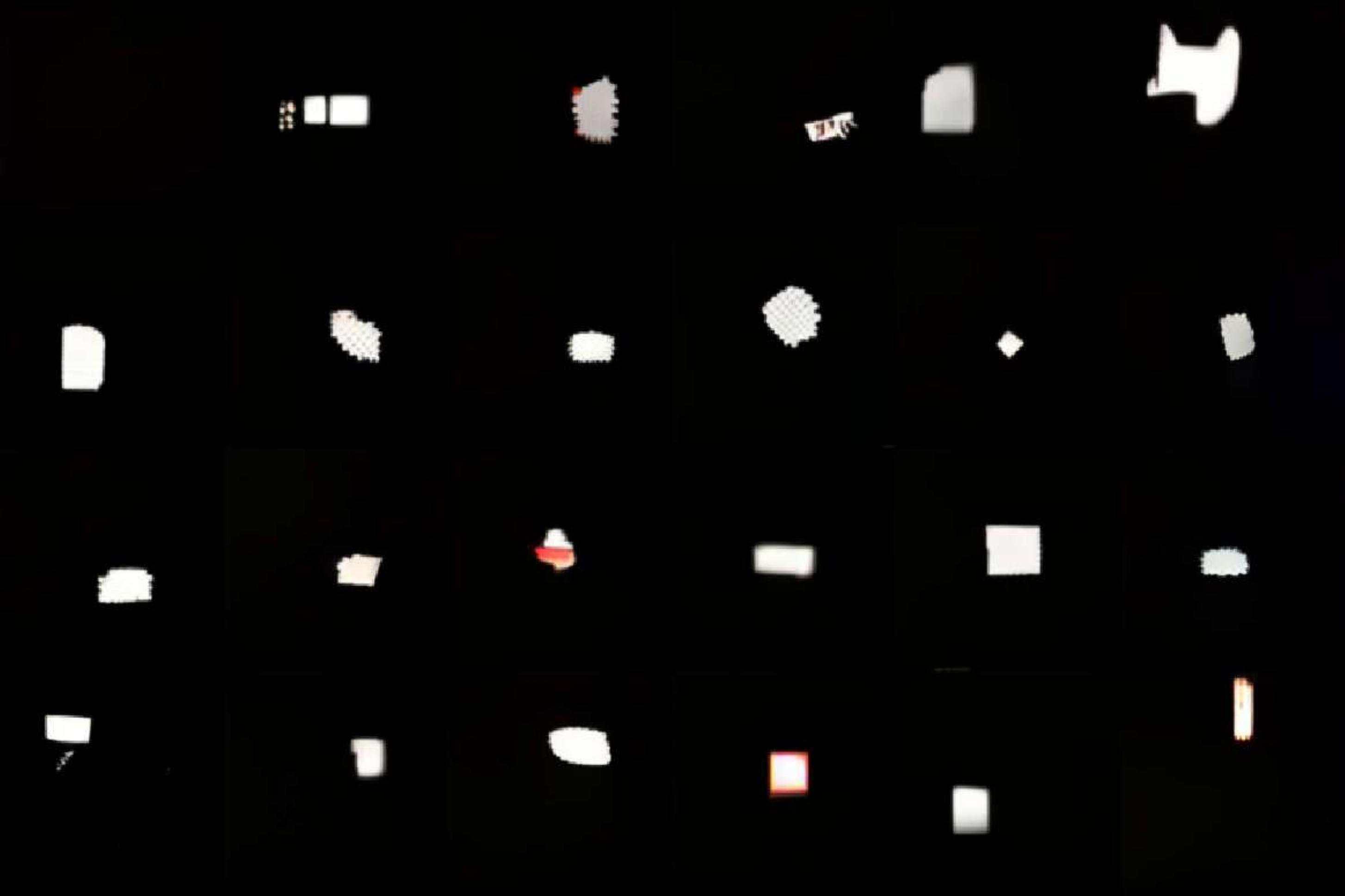}
    \caption{NAR transformer with prompt tuning ($S\,{=}\,1$)}
    \label{fig:vtab_maskgit_prompt_s1_dsprites}
  \end{subfigure}
  \hspace{0.02in}
  \begin{subfigure}[b]{0.48\linewidth}
    \centering
    \includegraphics[width=\linewidth]{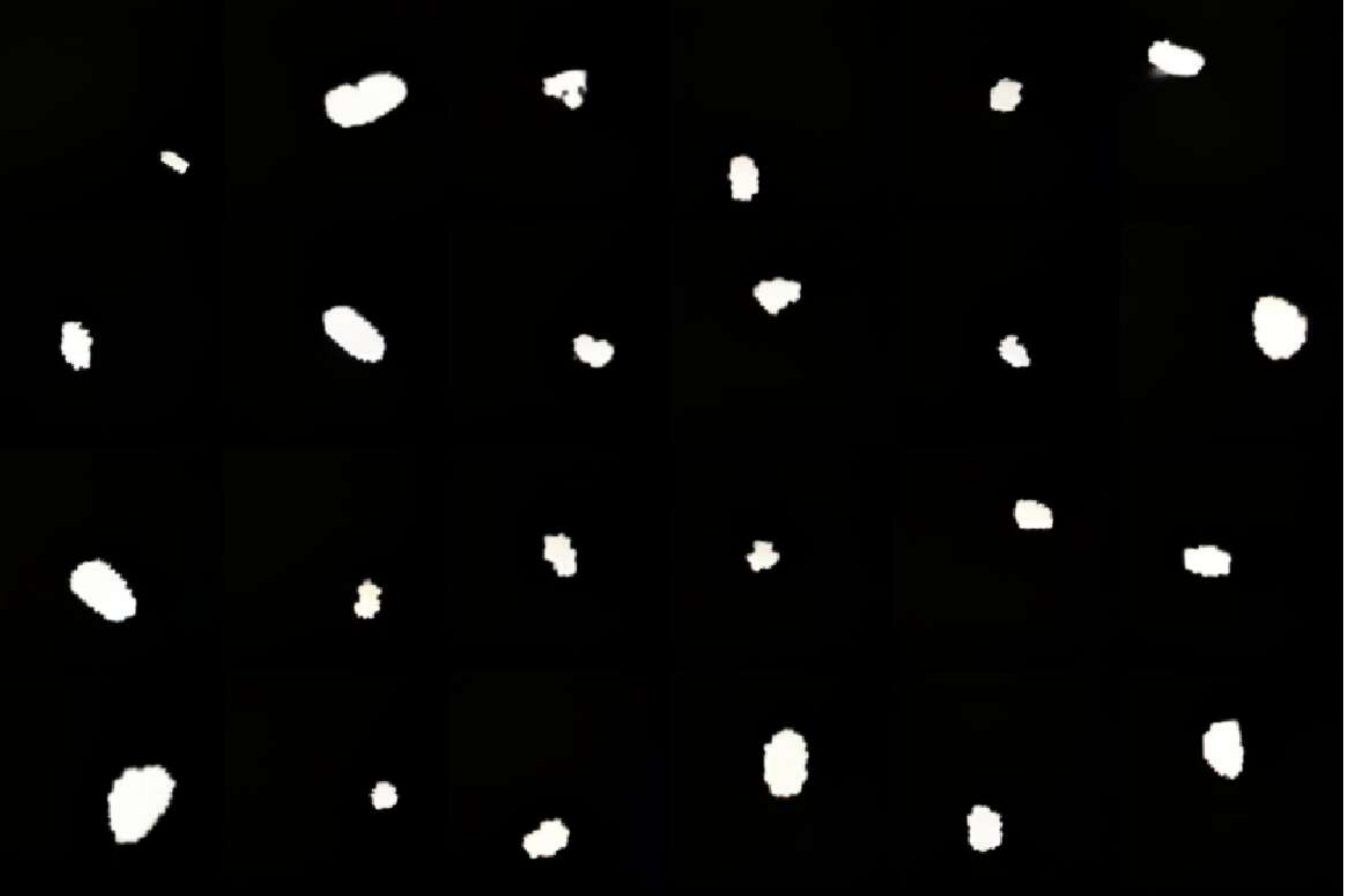}
    \caption{NAR transformer with prompt tuning ($S\,{=}\,128$)}
    \label{fig:vtab_maskgit_prompt_s128_dsprites}
  \end{subfigure}
  \caption{Visualization of generated images with different models on Dsprites of VTAB.}
  \label{fig:vtab_supp_dsprites}
\end{figure}

\begin{figure}
  \centering
  \begin{subfigure}[b]{0.48\linewidth}
    \centering
    \includegraphics[width=\linewidth]{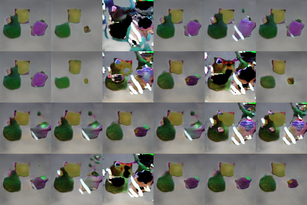}
    \caption{MineGAN}
    \label{fig:vtab_minegan_clevr}
  \end{subfigure}
  \hspace{0.02in}
  \begin{subfigure}[b]{0.48\linewidth}
    \centering
    \includegraphics[width=\linewidth]{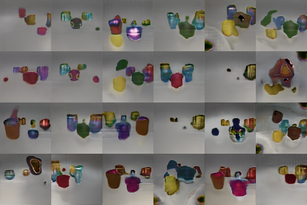}
    \caption{cGANTransfer}
    \label{fig:vtab_cgantrasnfer_clevr}
  \end{subfigure}
  \begin{subfigure}[b]{0.48\linewidth}
    \centering
    \includegraphics[width=\linewidth]{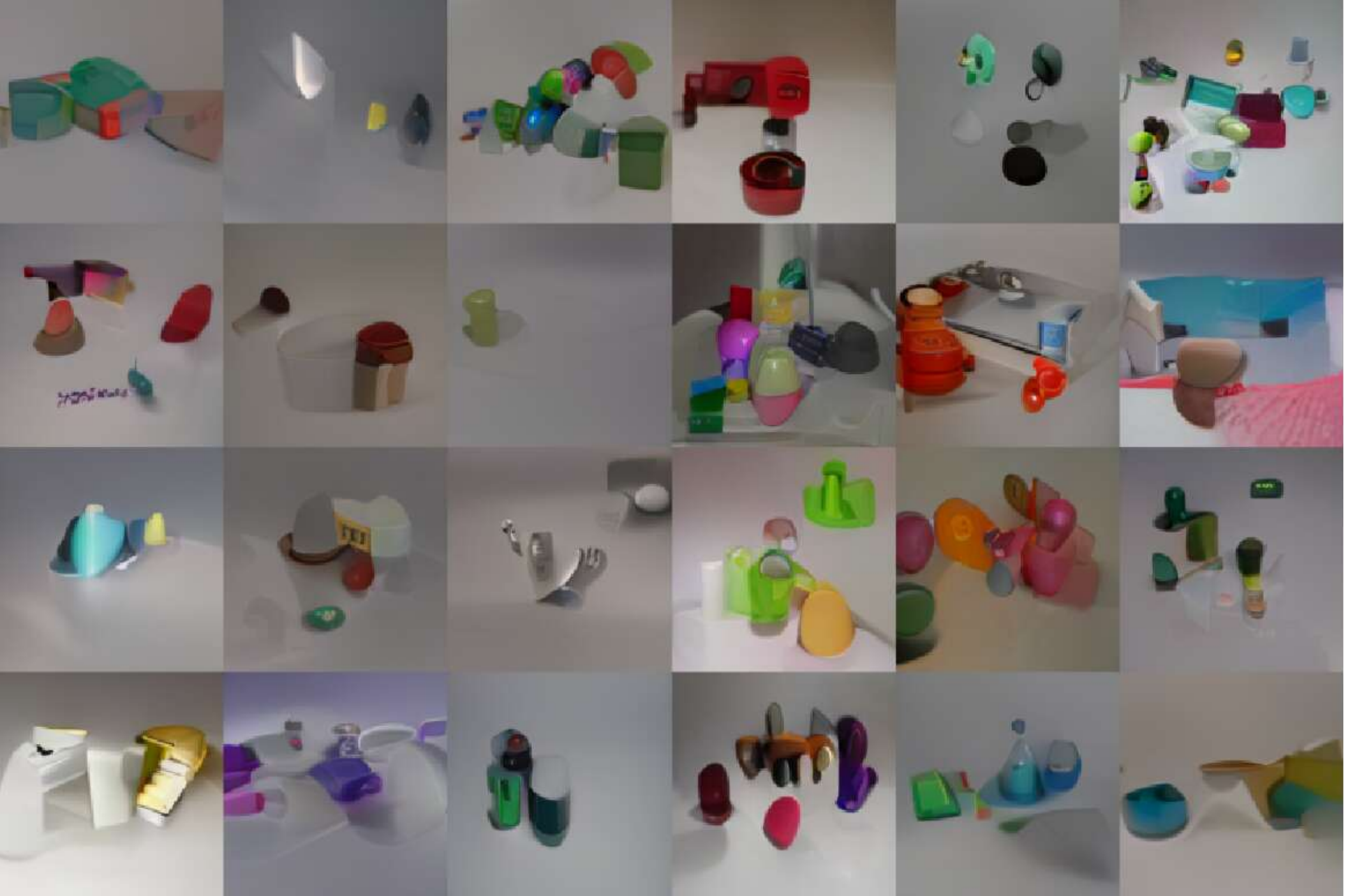}
    \caption{AR transformer with prompt tuning ($S\,{=}\,1$)}
    \label{fig:vtab_taming_prompt_s1_clevr}
  \end{subfigure}
  \hspace{0.02in}
  \begin{subfigure}[b]{0.48\linewidth}
    \centering
    \includegraphics[width=\linewidth]{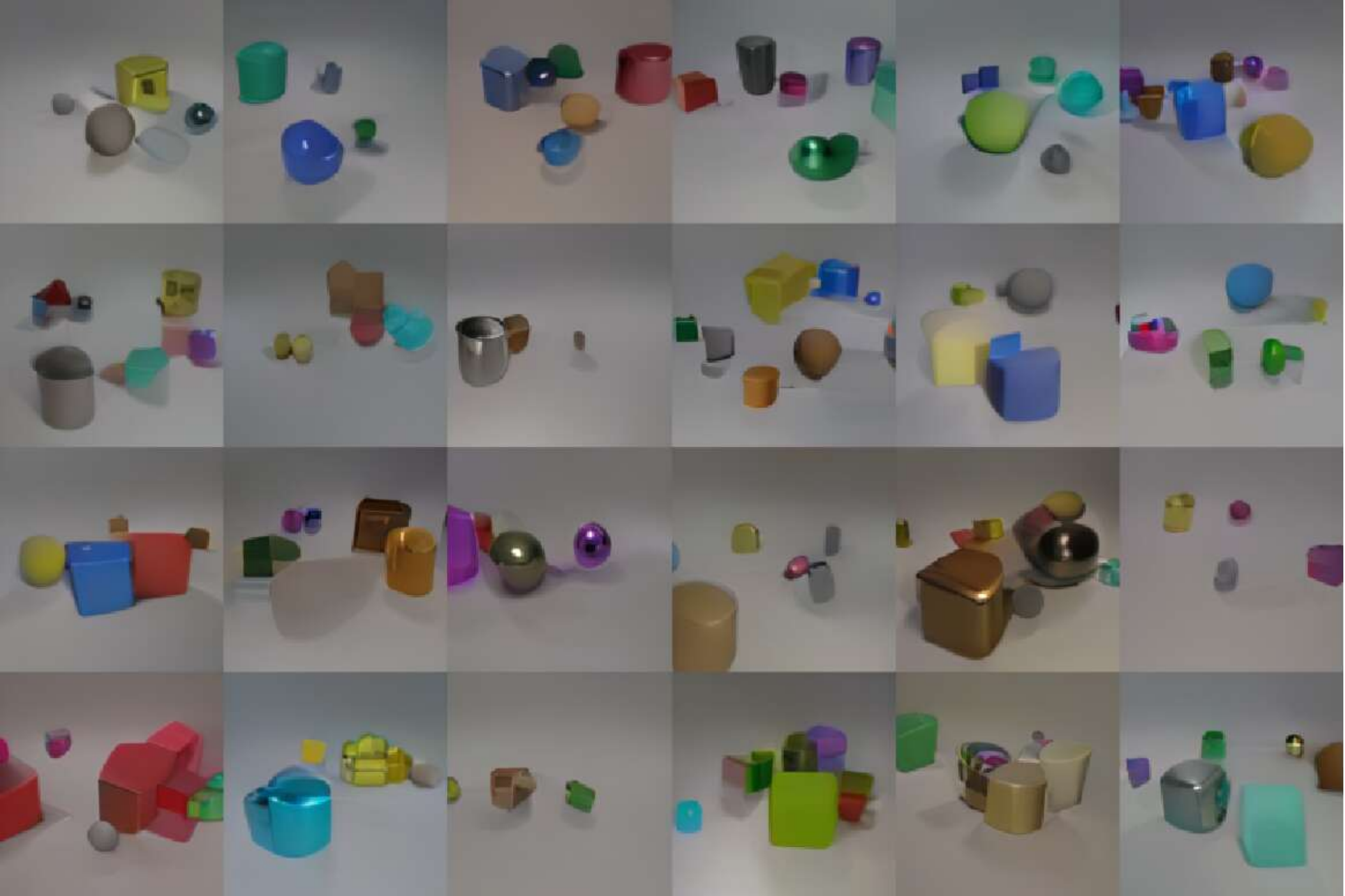}
    \caption{AR transformer with prompt tuning ($S\,{=}\,256$, $F\,{=}\,16$)}
    \label{fig:vtab_taming_prompt_s128_clevr}
  \end{subfigure}
  \begin{subfigure}[b]{0.48\linewidth}
    \centering
    \includegraphics[width=\linewidth]{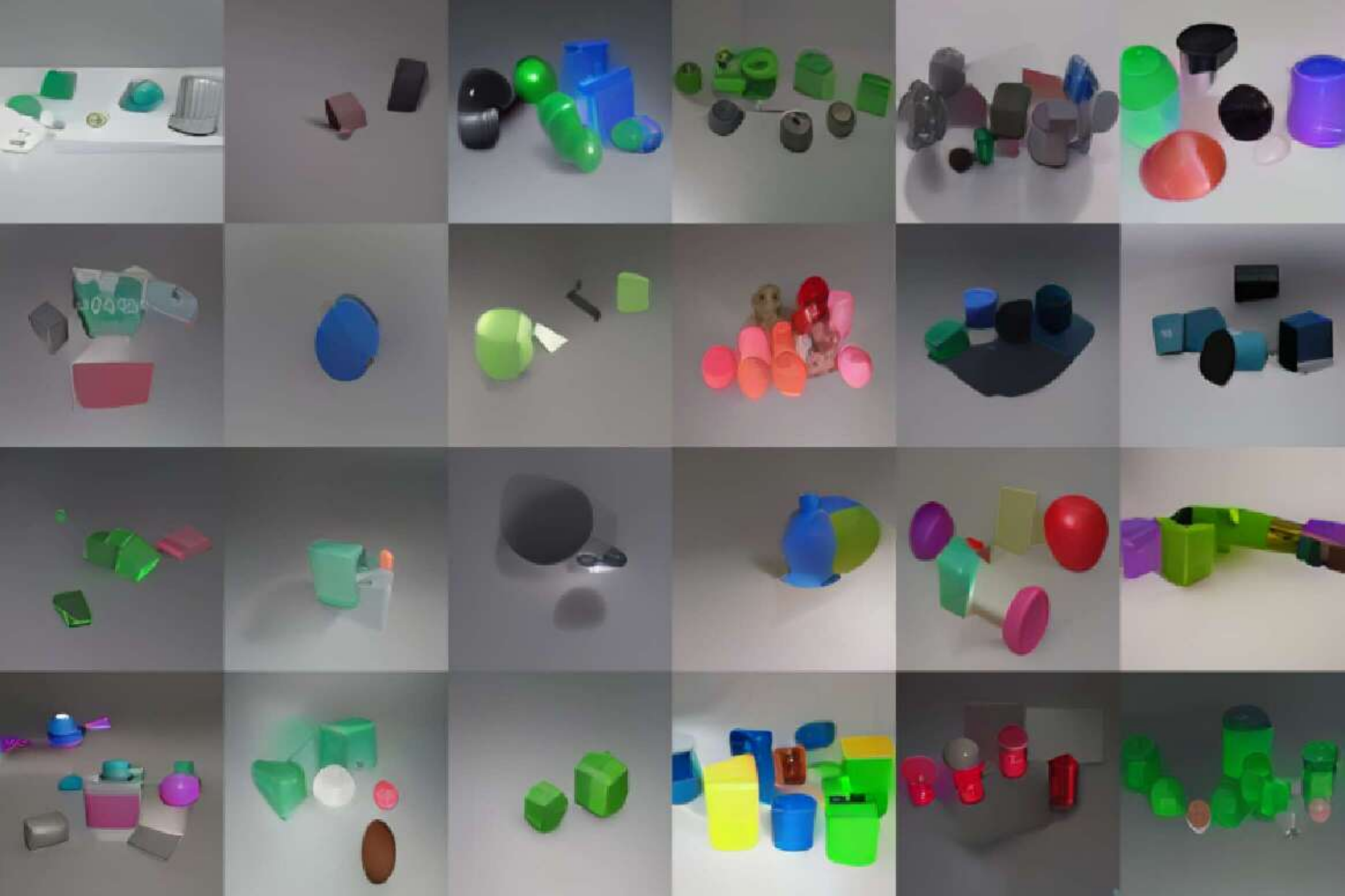}
    \caption{NAR transformer with prompt tuning ($S\,{=}\,1$)}
    \label{fig:vtab_maskgit_prompt_s1_clevr}
  \end{subfigure}
  \hspace{0.02in}
  \begin{subfigure}[b]{0.48\linewidth}
    \centering
    \includegraphics[width=\linewidth]{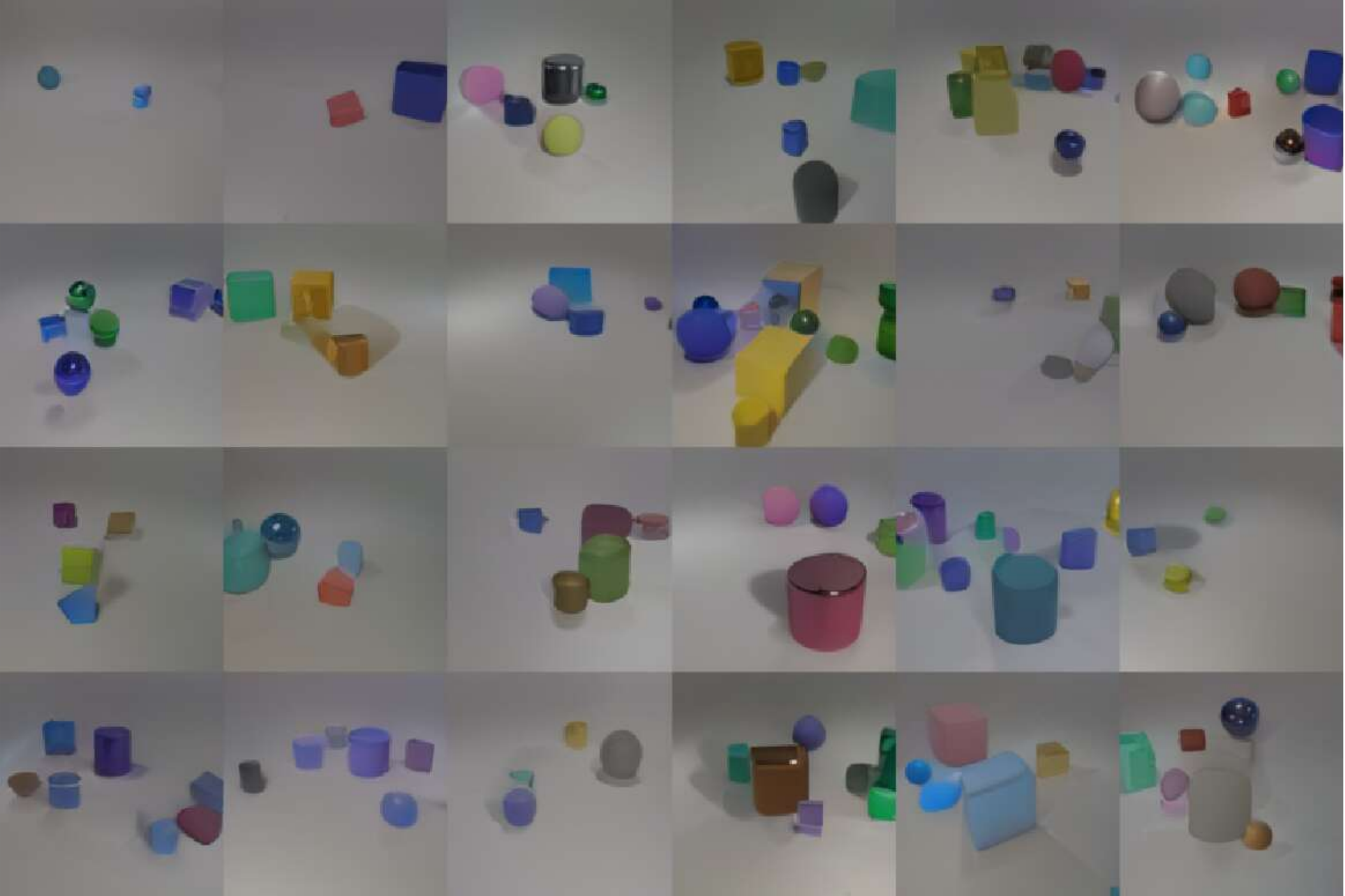}
    \caption{NAR transformer with prompt tuning ($S\,{=}\,128$)}
    \label{fig:vtab_maskgit_prompt_s128_clevr}
  \end{subfigure}
  \caption{Visualization of generated images with different models on Clevr of VTAB.}
  \label{fig:vtab_supp_clevr}
\end{figure}

\begin{figure}
  \centering
  \begin{subfigure}[b]{0.48\linewidth}
    \centering
    \includegraphics[width=\linewidth]{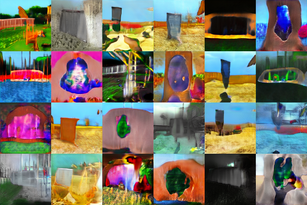}
    \caption{MineGAN}
    \label{fig:vtab_minegan_dmlab}
  \end{subfigure}
  \hspace{0.02in}
  \begin{subfigure}[b]{0.48\linewidth}
    \centering
    \includegraphics[width=\linewidth]{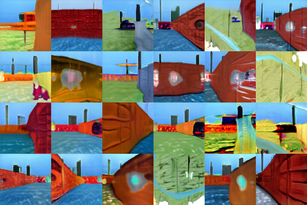}
    \caption{cGANTransfer}
    \label{fig:vtab_cgantrasnfer_dmlab}
  \end{subfigure}
  \begin{subfigure}[b]{0.48\linewidth}
    \centering
    \includegraphics[width=\linewidth]{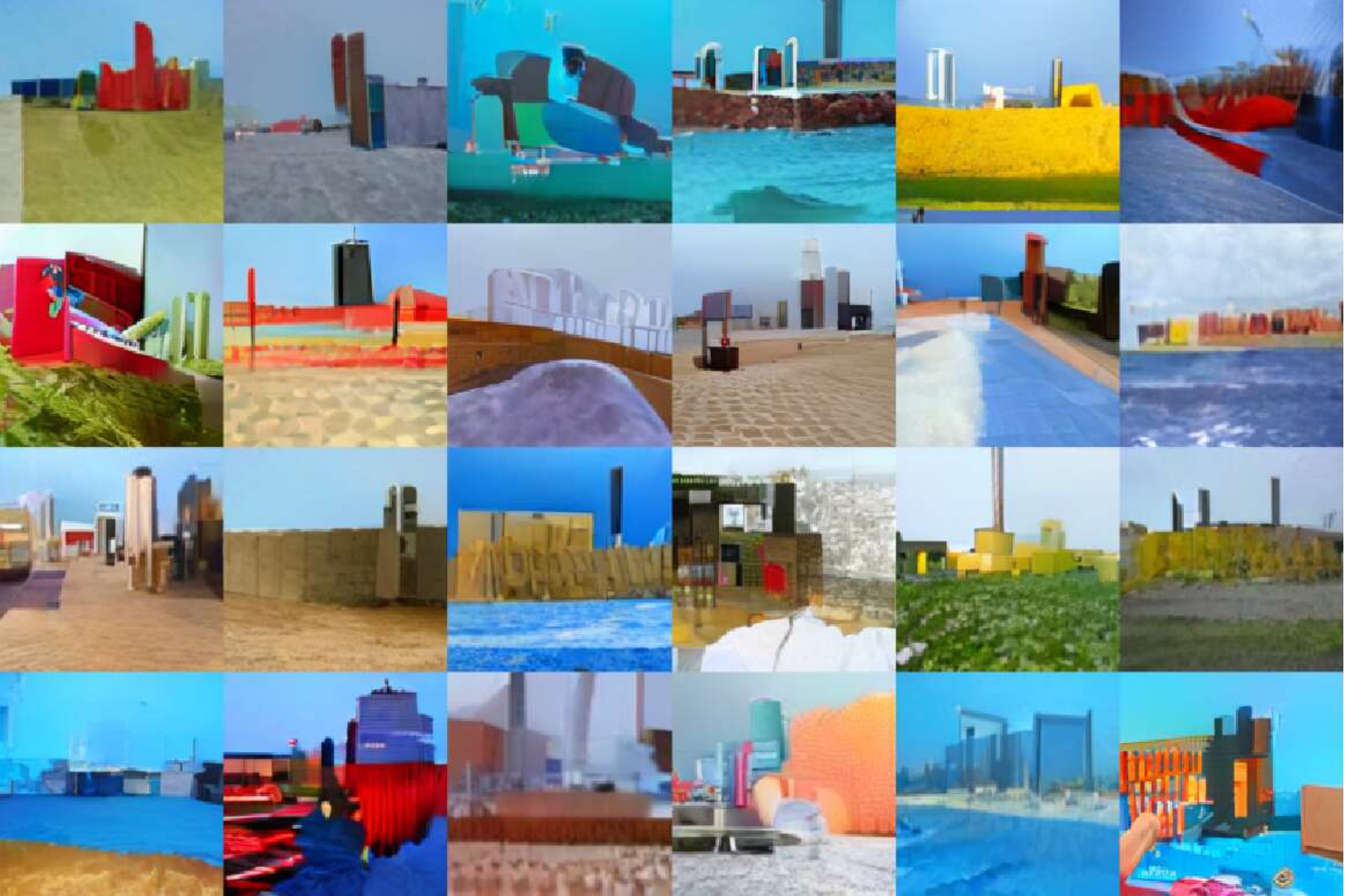}
    \caption{AR transformer with prompt tuning ($S\,{=}\,1$)}
    \label{fig:vtab_taming_prompt_s1_dmlab}
  \end{subfigure}
  \hspace{0.02in}
  \begin{subfigure}[b]{0.48\linewidth}
    \centering
    \includegraphics[width=\linewidth]{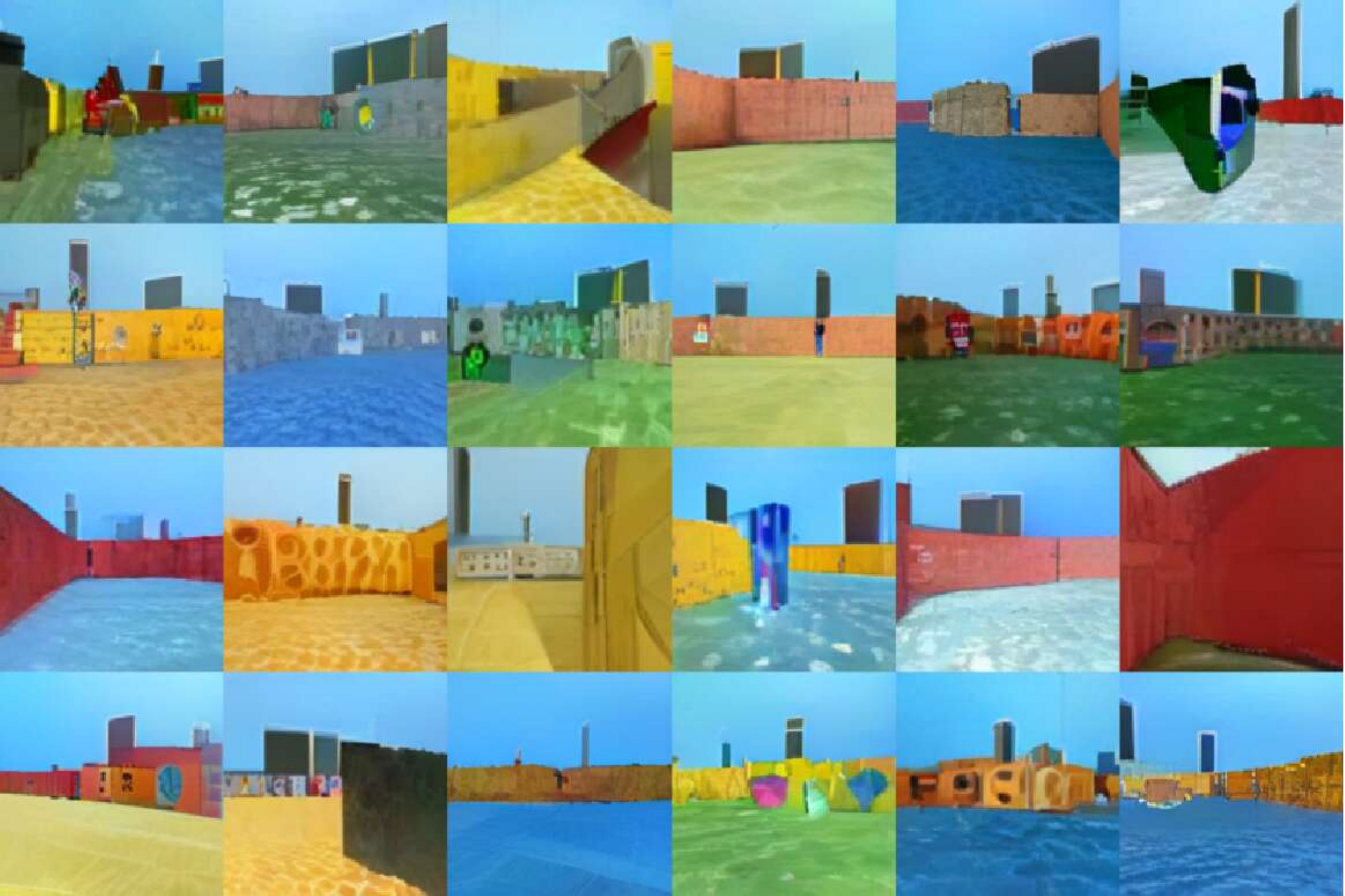}
    \caption{AR transformer with prompt tuning ($S\,{=}\,256$, $F\,{=}\,16$)}
    \label{fig:vtab_taming_prompt_s128_dmlab}
  \end{subfigure}
  \begin{subfigure}[b]{0.48\linewidth}
    \centering
    \includegraphics[width=\linewidth]{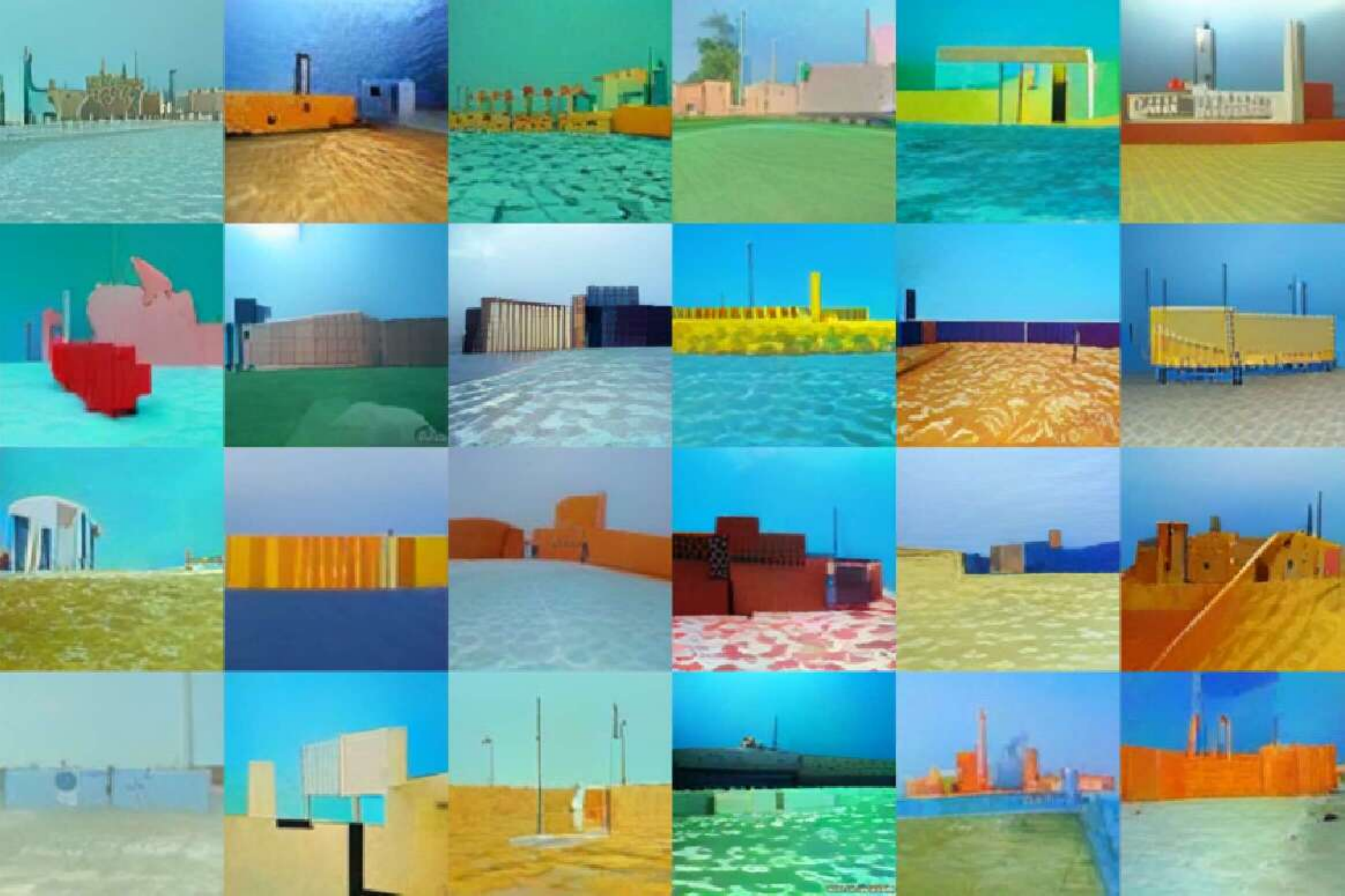}
    \caption{NAR transformer with prompt tuning ($S\,{=}\,1$)}
    \label{fig:vtab_maskgit_prompt_s1_dmlab}
  \end{subfigure}
  \hspace{0.02in}
  \begin{subfigure}[b]{0.48\linewidth}
    \centering
    \includegraphics[width=\linewidth]{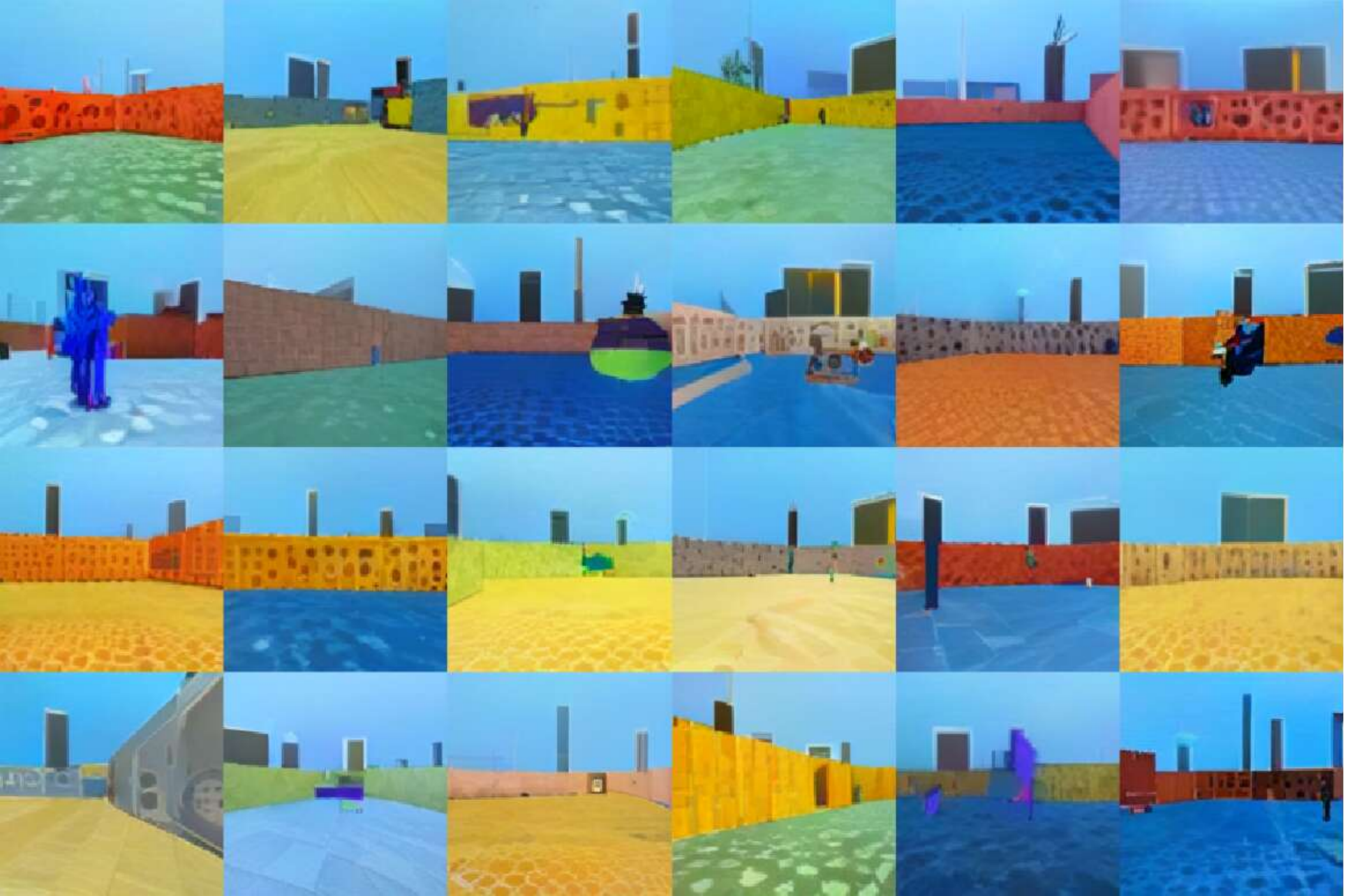}
    \caption{NAR transformer with prompt tuning ($S\,{=}\,128$)}
    \label{fig:vtab_maskgit_prompt_s128_dmlab}
  \end{subfigure}
  \caption{Visualization of generated images with different models on DMLab of VTAB.}
  \label{fig:vtab_supp_dmlab}
\end{figure}

We visualize images generated by the models trained on each of VTAB tasks from \cref{fig:vtab_supp_caltech101} to \cref{fig:vtab_supp_dmlab}.

\begin{figure}[t]
\centering
\begin{lstlisting}[language=Python]
import flax.linen as nn
import jax.numpy as jnp

class TokenGenerator(nn.Module):
  n_token: int  # Number of token (S)
  n_class: int  # Number of class (C)
  n_factor: int # Number of factors (F)
  d_embed: int  # Embed dimension (P)
  d_token: int  # Token dimension (D)

  @nn.compact
  def __call__(self, cls_ids: jnp.ndarray):
    MLP_p = nn.Embed(self.n_token, [self.d_embed, self.n_factor])
    MLP_c = nn.Embed(self.n_class, [self.d_embed, self.n_factor])
    MLP_t = nn.Dense(self.d_token)
    MLP_f = nn.Embed(1, self.n_factor)

    pos_ids = jnp.arange(self.n_token)
    factor_ids = jnp.arange(1)[None, None, ...]
    pos_embed = MLP_p(pos_ids[None, ...])  # 1 x S x P x F
    cls_embed = MLP_c(cls_ids[..., None])  # B x 1 x P x F
    fac_embed = MLP_f([None, None, ...])  # 1 x 1 x 1 x F
    embed = (fac_embed * (pos_embed + cls_embed)).sum(-1)
    return MLP_t(nn.LayerNorm(embed))
\end{lstlisting}
\caption{An example code for the token generator in Flax-ish~\cite{flax2020github} format.}
\label{fig:prompt_generator}
\end{figure}

\begin{table}[ht]
    \centering
    \resizebox{0.6\linewidth}{!}{%
    \begin{tabular}{l|c|c|c|c|c}
    \toprule
    Dataset & \# class & train & val & test & all \\
    \midrule
         Caltech-101	&	102	&	2754	&	306	&	6084	&	9144	\\
CIFAR-100	&	100	&	45000	&	5000	&	10000	&	60000	\\
SUN397	&	397	&	76128	&	10875	&	21750	&	108753	\\
SVHN	&	10	&	65931	&	7326	&	26032	&	99289	\\
Flowers102	&	102	&	1020	&	1020	&	6149	&	8189	\\
Pet	&	37	&	2944	&	736	&	3669	&	7349	\\
DTD	&	47	&	1880	&	1880	&	1880	&	5640	\\
EuroSAT	&	10	&	16200	&	5400	&	5400	&	27000	\\
Resisc45	&	45	&	18900	&	6300	&	6300	&	31500	\\
Patch Camelyon	&	2	&	262144	&	32768	&	32768	&	327680	\\
Diabetic Retinopathy	&	5	&	35126	&	10906	&	42670	&	88702	\\
Kitti	&	4	&	6347	&	423	&	711	&	7481	\\
Smallnorb (azimuth)	&	18	&	24300	&	12150	&	12150	&	48600	\\
Smallnorb (elevation)	&	9	&	24300	&	12150	&	12150	&	48600	\\
Dsprites (x position)	&	16	&	589824	&	73728	&	73728	&	737280	\\
Dsprites (orientation)	&	16	&	589824	&	73728	&	73728	&	737280	\\
Clevr (object distance)	&	6	&	63000	&	7000	&	15000	&	85000	\\
Clevr (count)	&	8	&	63000	&	7000	&	15000	&	85000	\\
DMLab	&	6	&	65550	&	22628	&	22735	&	110913	\\ \midrule
Mean	&	49.5	&	102851.2	&	15332.8	&	20416.0	&	138600.0	\\
\bottomrule
    \end{tabular}
    }
    \caption{Dataset meta information (\eg, number of images, number of class) for tasks in VTAB.}
    \label{tab:vtab_supp_info}
\end{table}

\begin{table}[ht]
    \centering
    \resizebox{0.95\linewidth}{!}{%
    \begin{tabular}{c||c|c|c|c||c|c|c|c}
        \toprule
        & AR & AR & AR & AR & NAR & NAR & NAR & NAR \\
        & scratch & + Prompt & + Adapter & + Fine-tune & scratch & + Prompt & + Adapter & + Fine-tune \\
        \midrule
        Learning rate & 0.0005 & 0.001 & 0.001 & 0.0005 & 0.0001 & 0.001 & 0.001 & 0.001 / 0.0001 \\
        Batch size & 128 & 256 & 256 & 128 & 128 & 256 & 256 & 128 \\
        Weight decay & 0.045 & 0 & 0 & 0.045 & 0.045 & 0 & 0 & 0.045 \\
        Warmup epochs & 2 & 0 & 0 & 0 & 80 & 0 & 0 & 0 \\
        \bottomrule
    \end{tabular}
    }
    \caption{Hyperparameter used for experiments. For NAR + Fine-tune, we use the learning rate of 0.001 for new model parameters (\eg, prompt) while using 0.0001 for pretrained ones (\eg, transformer). The same hyperparameter is used across all datasets and scenarios.}
    \label{tab:vtab_hparams}
\end{table}

\begin{table}[t]
    \centering
    \resizebox{0.85\linewidth}{!}{%
    \begin{tabular}{c|l||c|c|c|c|c|c|c|c|c|c|c|c}
    \toprule
          \multicolumn{2}{c||}{Models} & Caltech101 & CIFAR100 & SUN397 & SVHN & Flower & Pet & DTD & EuroSAT & Resisc45 & PC & DR & Kitti \\ \midrule
 \multicolumn{2}{c||}{MineGAN} & 102.4 & 82.6 & 77.5 & 144.7 & 132.1 & 130.1 & 87.4 & 111.5 & 81.0 & 170.3 & 192.2 & 117.9 \\
 \multicolumn{2}{c||}{cGANTransfer} & 89.6 & 31.4 & 31.1 & 64.7 & 61.6 & 48.6 & 70.3 & 45.6 & 50.3 & 119.9 & 149.8 & 48.9 \\ \midrule
\multirow{10}{*}{NAR} & Scratch & 72.7 & 24.2 & 9.2 & 44.4 & 57.2 & 70.3 & 66.1 & 39.5 & 32.0 & 48.3 & 25.6 & 33.8 \\
 & Scratch (3200 ep.) & 14.5 & 22.5 & 7.3 & 43.5 & 14.9 & 8.5 & 29.2 & 26.4 & 24.2 & 51.1 & 26.0 & 26.1 \\
 & P ($S{=}1$) & 13.4 & 26.9 & 7.2 & 83.0 & 13.8 & 11.8 & 25.7 & 45.9 & 28.7 & 107.9 & 84.2 & 32.2 \\
 & P ($S{=}16$) & 12.7 & 25.5 & 7.3 & 80.8 & 13.2 & 11.0 & 26.0 & 35.8 & 25.1 & 71.0 & 34.2 & 30.0 \\
 & P ($S{=}128$) & 12.9 & 25.0 & 7.7 & 62.3 & 13.4 & 10.9 & 25.9 & 38.4 & 24.8 & 67.4 & 30.8 & 29.9 \\
 & P ($S{=}128$, $F{=}16$) & 11.8 & 25.0 & 7.5 & 63.4 & 13.3 & 11.5 & 26.0 & 35.8 & 24.3 & 61.4 & 29.2 & 27.0 \\
 & P$^{\dagger}$ ($S{=}16$) & 12.4 & 25.3 & 7.3 & 72.5 & 12.7 & 11.2 & 25.4 & 36.9 & 23.7 & 71.7 & 34.3 & 31.2 \\
 & P$^{\dagger}$ ($S{=}128$) & 12.2 & 25.2 & 7.5 & 60.4 & 12.3 & 11.0 & 25.7 & 35.4 & 24.3 & 71.7 & 28.2 & 29.6 \\
 & Adapter & 11.3 & 20.3 & 6.7 & 43.7 & 11.0 & 6.9 & 25.1 & 28.2 & 19.9 & 46.4 & 24.9 & 24.0 \\
 & Fine-tune & 11.3 & 18.2 & 6.5 & 43.9 & 10.2 & 6.3 & 24.2 & 23.1 & 18.2 & 48.0 & 24.4 & 22.8 \\ \midrule
\multirow{10}{*}{AR} & Scratch & 76.1 & 27.1 & 13.5 & 31.2 & 56.1 & 52.5 & 92.7 & 19.4 & 29.5 & 32.9 & 37.0 & 31.6 \\
 & Scratch (3200 ep.) & 30.5 & 25.8 & 14.4 & 27.9 & 24.3 & 28.1 & 45.1 & 15.5 & 11.5 & 32.3 & 37.7 & 33.2 \\
 & P ($S{=}1$) & 45.4 & 25.7 & 18.8 & 80.4 & 28.9 & 42.2 & 37.1 & 37.3 & 35.1 & 74.9 & 93.1 & 66.8 \\
 & P ($S{=}16$) & 41.4 & 22.5 & 16.4 & 55.5 & 19.6 & 36.6 & 33.4 & 32.6 & 28.8 & 49.8 & 60.7 & 41.3 \\
 & P ($S{=}256$) & 39.6 & 19.8 & 15.0 & 44.0 & 17.3 & 34.9 & 32.5 & 29.6 & 26.7 & 44.0 & 45.4 & 37.1 \\
 & P ($S{=}256$, $F{=}16$) & 27.2 & 17.6 & 12.8 & 42.8 & 14.1 & 27.2 & 30.0 & 26.4 & 22.2 & 44.3 & 45.4 & 34.6 \\
 & P$^{\dagger}$ ($S{=}16$) & 30.9 & 19.4 & 13.7 & 53.7 & 15.4 & 30.8 & 30.8 & 30.2 & 25.7 & 49.0 & 60.4 & 39.7 \\
 & P$^{\dagger}$ ($S{=}256$) & 24.6 & 17.5 & 12.3 & 43.1 & 13.7 & 25.1 & 29.8 & 26.7 & 20.9 & 43.6 & 46.1 & 35.1 \\
 & Adapter & 27.0 & 16.7 & 12.6 & 29.9 & 11.8 & 19.1 & 30.8 & 22.4 & 22.0 & 39.4 & 37.3 & 29.0 \\
 & Fine-tune & 17.6 & 13.2 & 9.1 & 27.7 & 17.7 & 10.7 & 35.4 & 15.1 & 11.6 & 30.9 & 34.5 & 29.6 \\
 \bottomrule
    \end{tabular}
    }
    \resizebox{0.98\linewidth}{!}{%
    \begin{tabular}{c|l||c|c|c|c|c|c|c||c|c|c|c|c|c|c}
    \toprule
          \multicolumn{2}{c||}{Models} & SNorb$^{A}$ & SNorb$^{B}$ & Dspr.$^{A}$ & Dspr.$^{B}$ & Clevr$^{A}$ & Clevr$^{B}$ & DMLab & Mean & $\leq 10K$ & $\leq 100K$ & $\geq 100K$ & Natural & Special. & Struct. \\ \midrule
  \multicolumn{2}{c||}{MineGAN} & 160.4 & 161.1 & 252.7 & 285.1 & 212.1 & 225.6 & 152.4 & 151.5 & 114.0 & 145.6 & 236.0 & 108.1 & 138.7 & 195.9 \\
 \multicolumn{2}{c||}{cGANTransfer} & 93.3 & 90.5 & 133.7 & 165.4 & 109.4 & 115.0 & 98.8 & 85.1 & 63.8 & 80.0 & 139.7 & 56.8 & 91.4 & 106.9 \\ \midrule
\multirow{10}{*}{NAR} & Scratch & 31.4 & 32.9 & 87.5 & 89.0 & 12.5 & 13.3 & 20.6 & 42.7 & 60.0 & 26.0 & 75.0 & 49.2 & 36.4 & 40.1 \\
 & Scratch (3200 ep.) & 29.4 & 30.5 & 90.1 & 88.3 & 13.7 & 13.5 & 19.6 & 30.5 & 18.6 & 23.3 & 76.5 & 20.1 & 31.9 & 38.9 \\
 & P ($S{=}1$) & 58.6 & 58.7 & 119.5 & 121.3 & 58.5 & 57.9 & 64.4 & 53.7 & 19.4 & 52.2 & 116.2 & 26.0 & 66.7 & 71.4 \\
 & P ($S{=}16$) & 46.1 & 42.8 & 98.7 & 98.8 & 27.3 & 28.2 & 43.4 & 39.9 & 18.6 & 36.1 & 89.5 & 25.2 & 41.5 & 51.9 \\
 & P ($S{=}128$) & 33.6 & 35.2 & 100.9 & 92.8 & 21.9 & 23.6 & 33.5 & 36.4 & 18.6 & 30.6 & 87.0 & 22.6 & 40.3 & 46.4 \\
 & P ($S{=}128$, $F{=}16$) & 36.0 & 36.1 & 98.7 & 99.3 & 25.6 & 24.1 & 32.0 & 36.2 & 17.9 & 30.8 & 86.5 & 22.6 & 37.7 & 47.4 \\
 & P$^{\dagger}$ ($S{=}16$) & 44.1 & 44.7 & 96.5 & 99.0 & 26.0 & 27.1 & 38.9 & 39.0 & 18.6 & 34.6 & 89.1 & 23.8 & 41.7 & 50.9 \\
 & P$^{\dagger}$ ($S{=}128$) & 34.6 & 38.4 & 92.2 & 95.4 & 24.7 & 27.5 & 32.9 & 36.3 & 18.2 & 30.8 & 86.4 & 22.0 & 39.9 & 46.9 \\
 & Adapter & 29.2 & 28.7 & 85.7 & 86.9 & 14.6 & 15.0 & 20.0 & 28.9 & 15.7 & 22.9 & 73.0 & 17.9 & 29.9 & 38.0 \\
 & Fine-tune & 67.2 & 51.3 & 86.5 & 88.0 & 20.6 & 19.7 & 23.4 & 32.3 & 15.0 & 28.8 & 74.1 & 17.2 & 28.4 & 47.4 \\ \midrule
\multirow{10}{*}{AR} & Scratch & 23.1 & 23.4 & 76.5 & 76.6 & 12.3 & 12.2 & 27.8 & 39.6 & 61.8 & 23.3 & 62.0 & 49.9 & 29.7 & 35.4 \\
 & Scratch (3200 ep.) & 23.4 & 23.3 & 76.5 & 75.1 & 12.1 & 11.4 & 25.5 & 30.2 & 32.2 & 20.8 & 61.3 & 28.0 & 24.3 & 35.1 \\
 & P ($S{=}1$) & 62.2 & 62.0 & 215.9 & 214.0 & 90.6 & 91.6 & 69.0 & 73.2 & 44.1 & 60.5 & 168.3 & 39.8 & 60.1 & 109.0 \\
 & P ($S{=}16$) & 52.9 & 52.6 & 102.3 & 99.8 & 51.0 & 49.8 & 53.6 & 47.4 & 34.5 & 43.3 & 83.9 & 32.2 & 42.9 & 62.9 \\
 & P ($S{=}256$) & 42.4 & 42.3 & 83.7 & 83.7 & 29.5 & 28.9 & 45.2 & 39.0 & 32.3 & 33.5 & 70.5 & 29.0 & 36.4 & 49.1 \\
 & P ($S{=}256$, $F{=}16$) & 43.4 & 42.7 & 83.8 & 81.6 & 30.2 & 29.0 & 45.9 & 36.9 & 26.6 & 32.6 & 69.9 & 24.5 & 34.6 & 48.9 \\
 & P$^{\dagger}$ ($S{=}16$) & 51.3 & 52.2 & 100.3 & 97.3 & 49.6 & 49.0 & 54.0 & 44.9 & 29.5 & 41.7 & 82.2 & 27.8 & 41.3 & 61.7 \\
 & P$^{\dagger}$ ($S{=}256$) & 43.5 & 43.5 & 86.9 & 84.3 & 30.4 & 29.8 & 45.7 & 37.0 & 25.7 & 32.7 & 71.6 & 23.7 & 34.3 & 49.9 \\
 & Adapter & 36.0 & 36.3 & 77.8 & 77.9 & 15.5 & 14.9 & 29.6 & 30.8 & 23.5 & 24.8 & 65.1 & 21.1 & 30.3 & 39.6 \\
 & Fine-tune & 23.2 & 23.2 & 76.8 & 77.2 & 11.8 & 11.5 & 25.6 & 26.4 & 22.2 & 18.8 & 61.6 & 18.8 & 23.0 & 34.9 \\
 \bottomrule
    \end{tabular}
    }
    \caption{FIDs on VTAB tasks tested with various models. We use the ``all'' set as a reference set for computing FIDs. Unless otherwise stated, all NAR models are trained for 200 epochs and AR models are trained for 400 epochs with the same hyperparameter settings specified in \cref{tab:vtab_hparams}. ``P'' refers to the prompt tuning with the sequence length $S$ and the number of factors $F$. ``DTD'': Describable Textures Dataset, ``PC'': Patch Camelyon, ``DR'': Diabetic Retinopathy, ``SNorb$^{A}$'': SmallNorb (azimuth), ``SNorb$^{B}$'': SmallNorb (elevation), ``Dspr$^{A}$'': Dsprites (x position), ``Dspr$^{B}$'': Dsprites (orientation), ``Clevr$^{A}$'': Clevr (object distance), ``Clevr$^{B}$'': Clevr (count).}
    \label{tab:vtab_supp_eval_on_all}
\end{table}


\subsection{Few-shot Generative Transfer}
\label{sec:supp_exp_fewshot}

\subsubsection{Visualization of Generated Images}
\label{sec:supp_exp_fewshot_synth}

\begin{figure}
    \centering
    \begin{subfigure}[b]{\textwidth}
        \centering
        \includegraphics[width=0.95\linewidth]{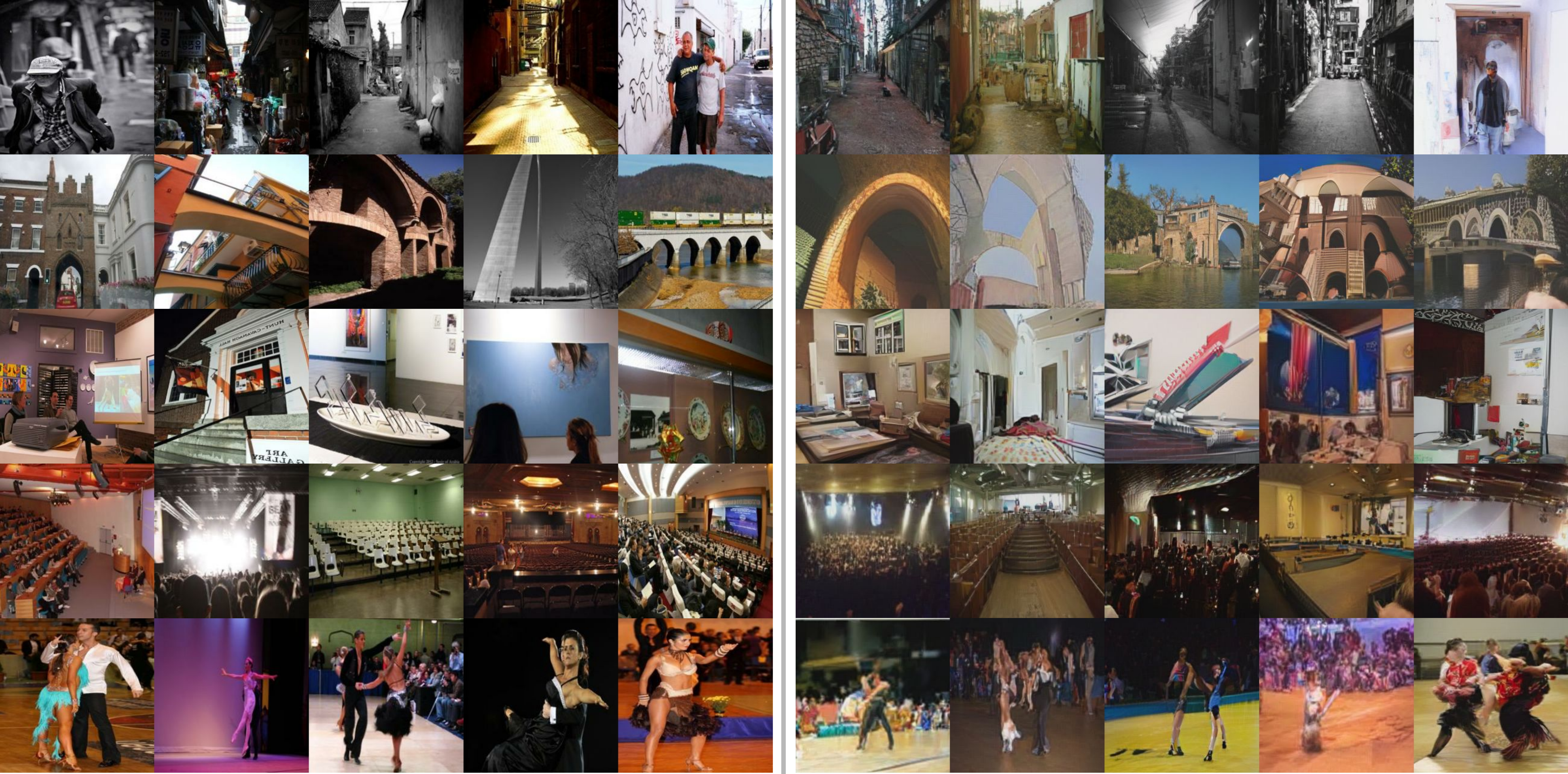}
        \caption{Places, 5-shot, Left: real, Right: generation.}
        \label{fig:fewshot_places_5shot_more}
    \end{subfigure}
    \begin{subfigure}[b]{\textwidth}
        \centering
        \includegraphics[width=0.95\linewidth]{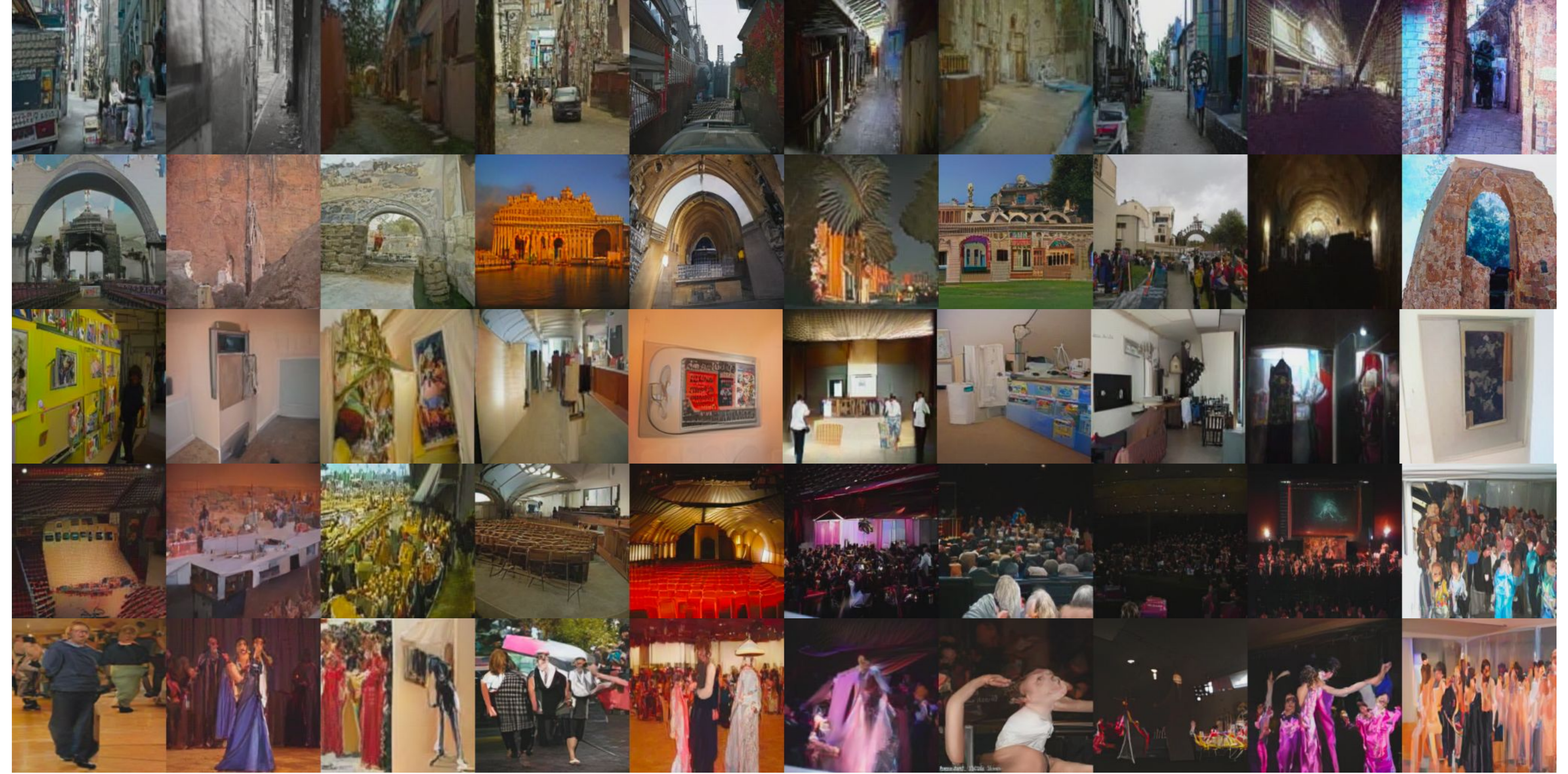}
        \caption{Places, 500-shot, All generation, without cherry-picking.}
        \label{fig:fewshot_places_500shot_more}
    \end{subfigure}
    \caption{Fewshot generation on places.}
    \label{fig:fewshot_places_more}
\end{figure}

\begin{figure}
    \centering
    \begin{subfigure}[b]{\textwidth}
        \centering
        \includegraphics[width=0.95\linewidth]{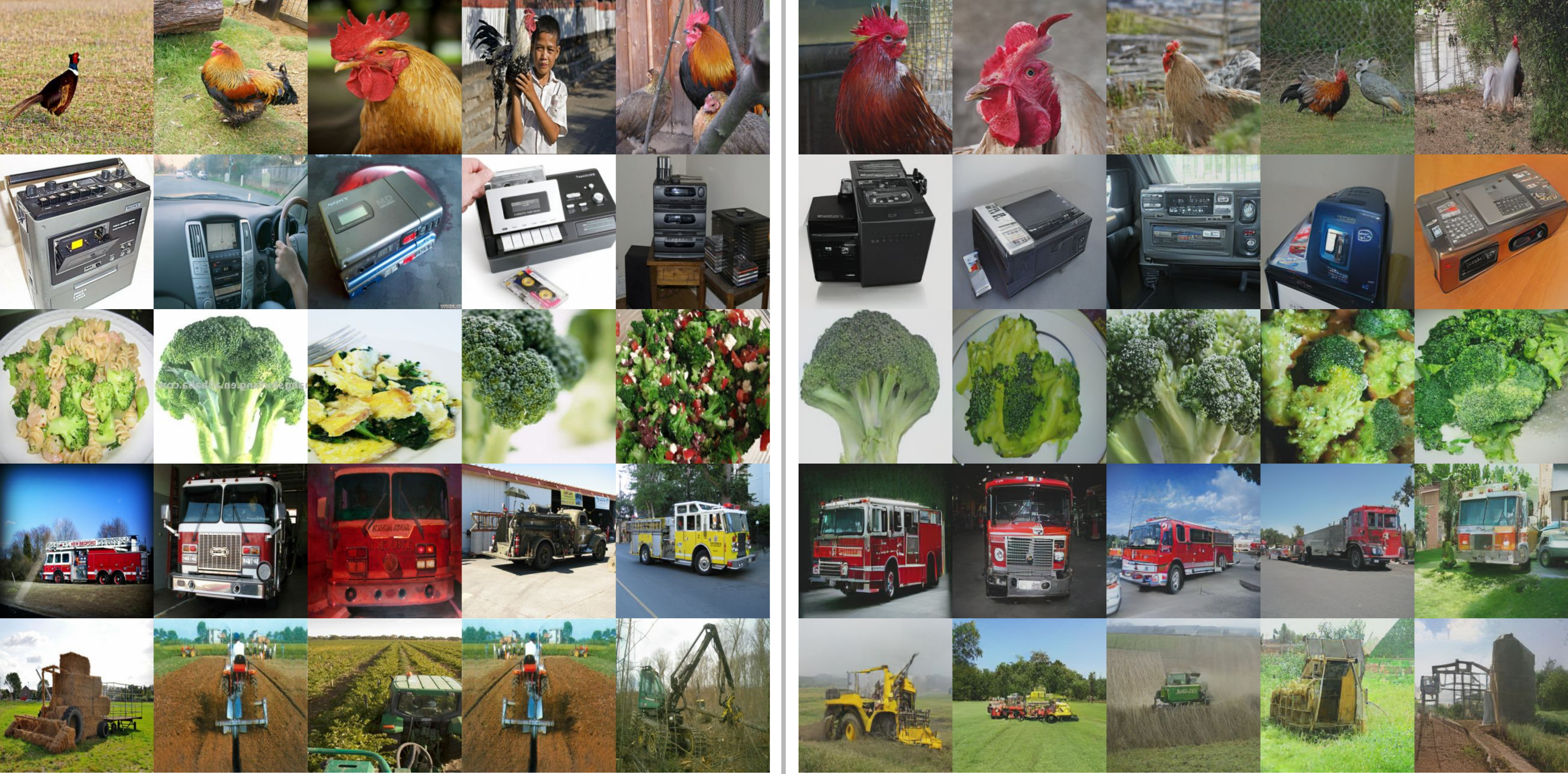}
        \caption{ImageNet, 5-shot, Left: real, Right: generation.}
        \label{fig:fewshot_imagenet_5shot_more}
    \end{subfigure}
    \begin{subfigure}[b]{\textwidth}
        \centering
        \includegraphics[width=0.95\linewidth]{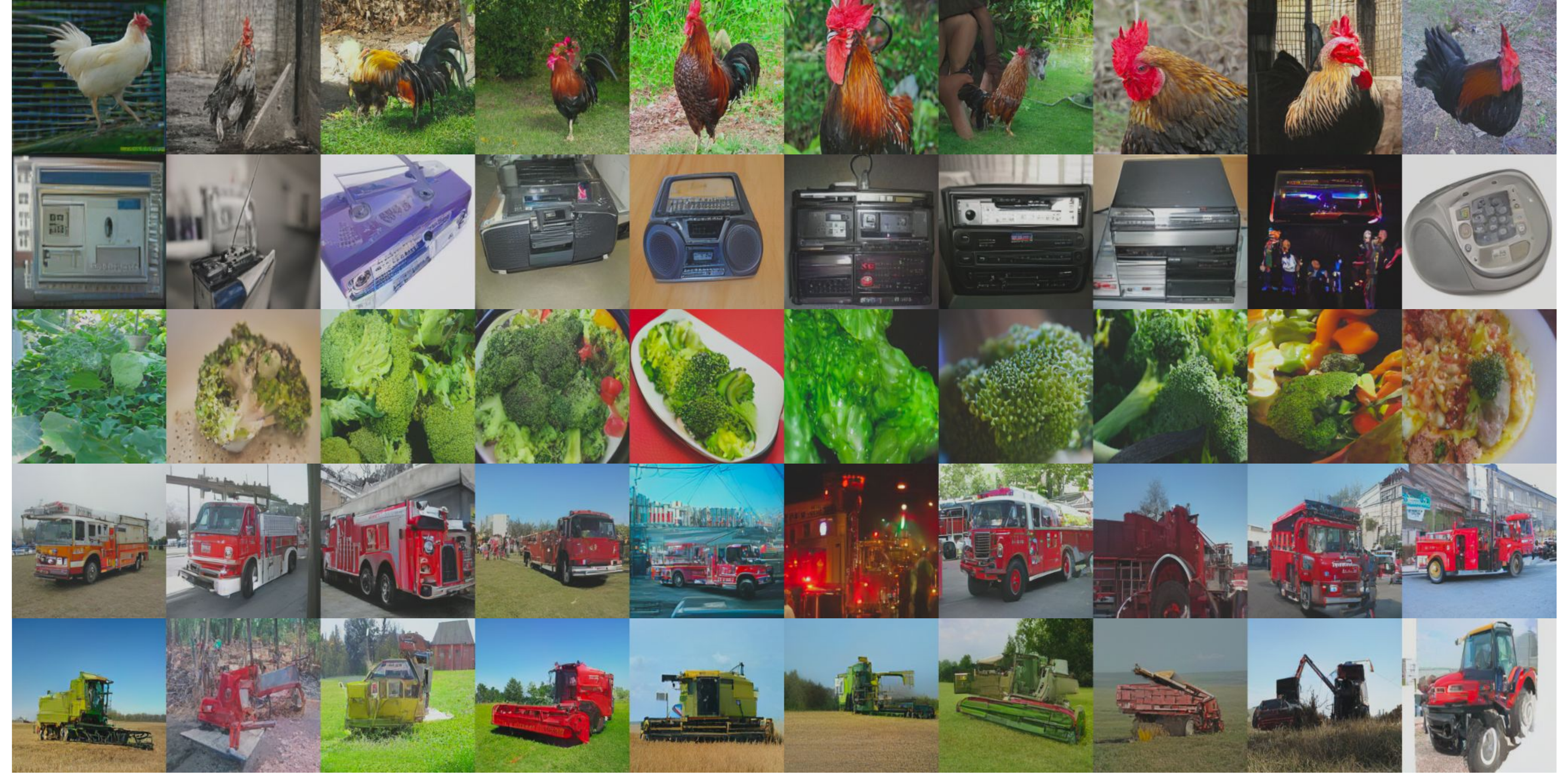}
        \caption{ImageNet, 500-shot, All generation, without cherry-picking.}
        \label{fig:fewshot_imagenet_500shot_more}
    \end{subfigure}
    \caption{Fewshot generation on ImageNet.}
    \label{fig:fewshot_imagenet_more}
\end{figure}

\begin{figure}
    \centering
    \begin{subfigure}[b]{\textwidth}
        \centering
        \includegraphics[width=0.95\linewidth]{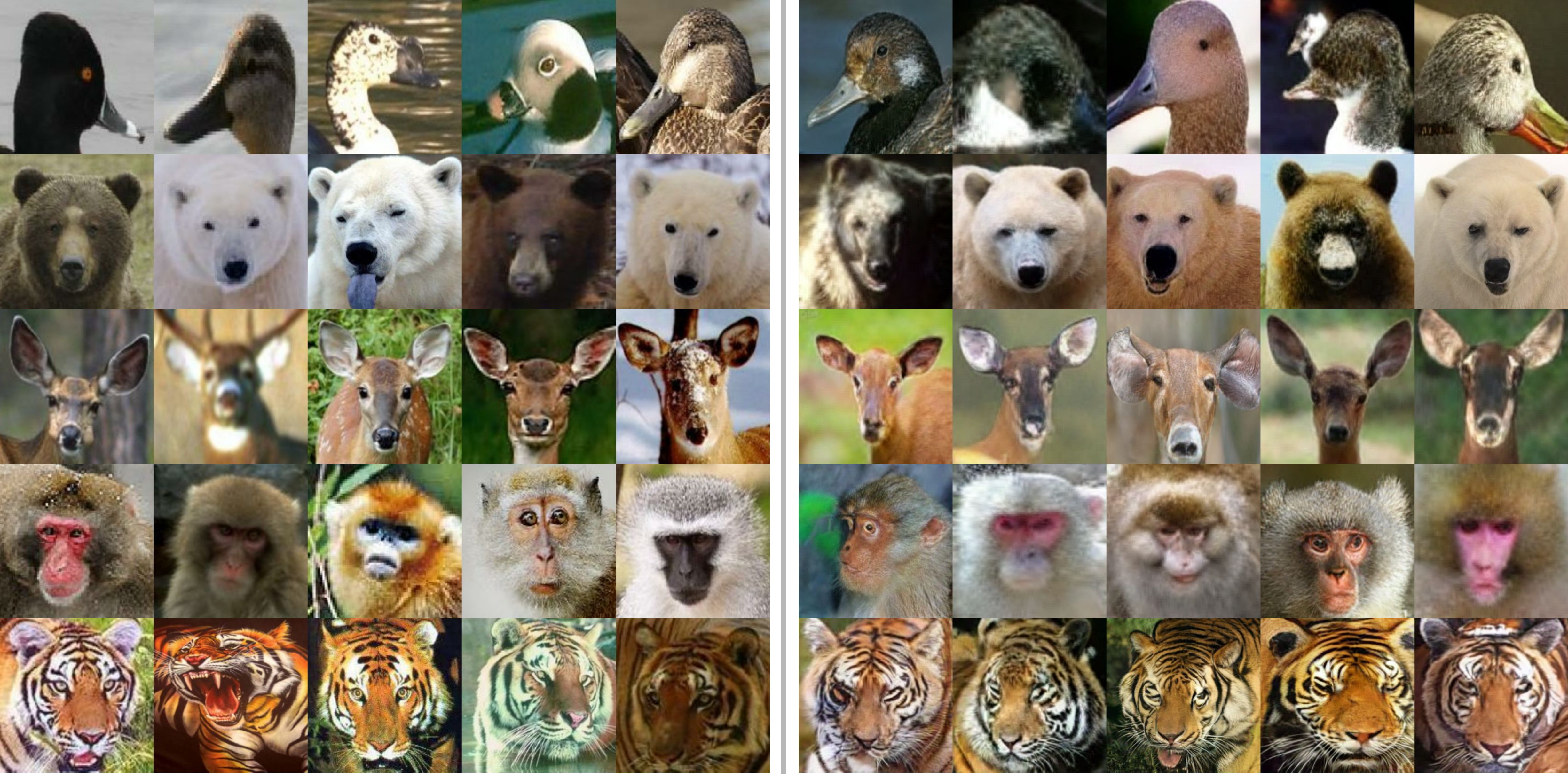}
        \caption{Animal Face, 5-shot, Left: real, Right: generation.}
        \label{fig:fewshot_animalface_5shot_more}
    \end{subfigure}
    \begin{subfigure}[b]{\textwidth}
        \centering
        \includegraphics[width=0.95\linewidth]{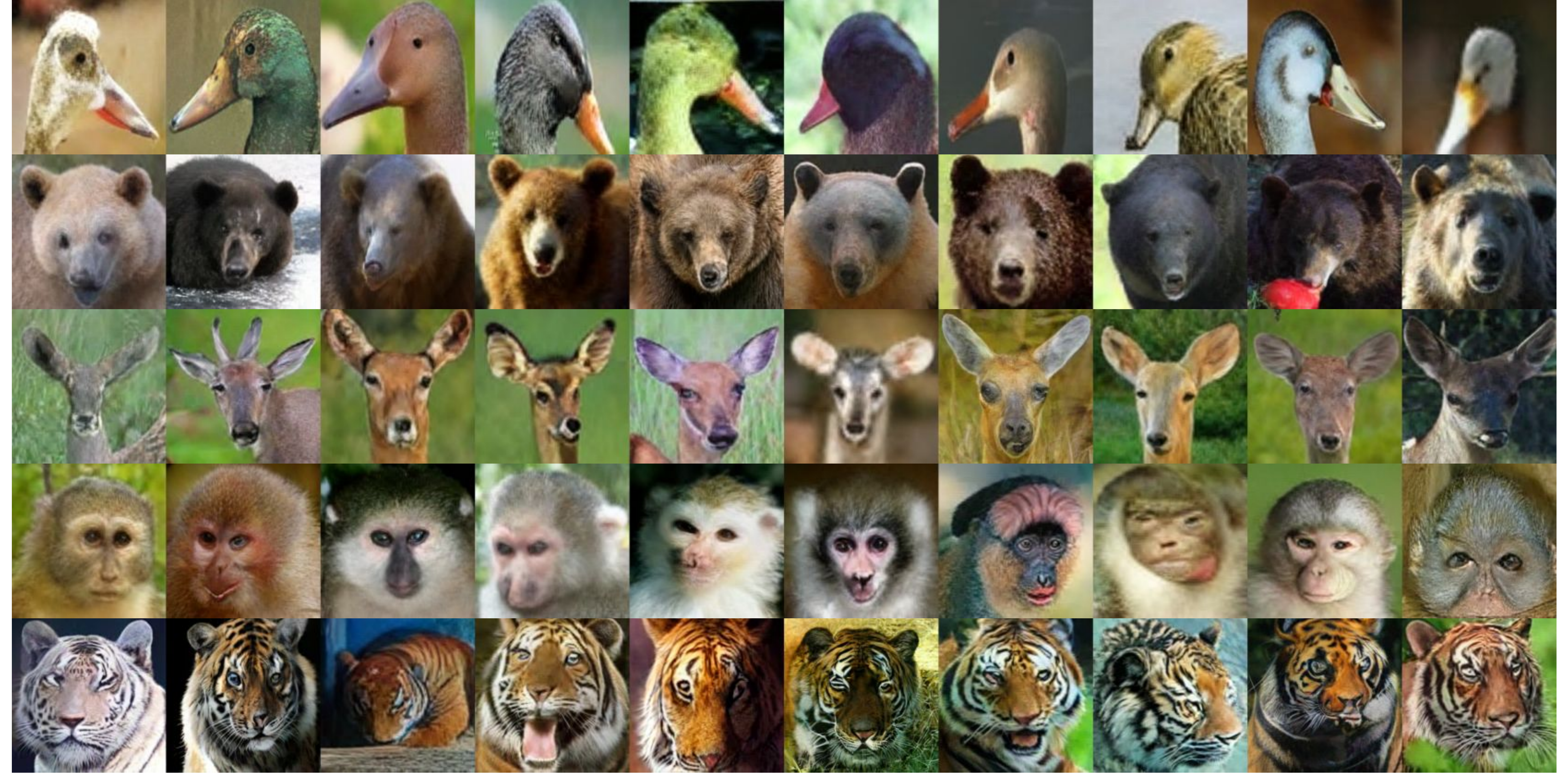}
        \caption{Animal Face, 100-shot, All generation, without cherry-picking.}
        \label{fig:fewshot_animalface_100shot_more}
    \end{subfigure}
    \caption{Fewshot generation on Animal Face.}
    \label{fig:fewshot_animalface_more}
\end{figure}

\end{document}